\documentclass[final]{cvpr}

\usepackage{times}
\usepackage{epsfig}
\usepackage{graphicx}
\usepackage{amsmath}
\usepackage{amssymb}

\usepackage{enumitem}
\usepackage[nocompress]{cite}
\usepackage[ruled,vlined,noend]{algorithm2e}
\usepackage{subcaption}
\usepackage{multirow}
\usepackage{booktabs}

\def\eg{\emph{e.g}\onedot} 
\def\ie{\emph{i.e}\onedot} 
\def\cf{\emph{cf}\onedot} 
\def\etc{\emph{etc}\onedot} 
 
\def\etal{\emph{et al}\onedot}

\newcommand{\concat}[0]{\mathbin\Vert}
\newcommand{\slimparagraph}[1]{{\noindent\bf#1}}

\usepackage[pagebackref=true,breaklinks=true,colorlinks,bookmarks=false]{hyperref}

\def\cvprPaperID{xxxx} %
\def\confYear{CVPR 2021}

\begin{document}

\title{ViP-DeepLab: Learning Visual Perception with Depth-aware Video Panoptic Segmentation}

\author{
Siyuan Qiao\textsuperscript{1}\thanks{Work done while an intern at Google.}~~~
Yukun Zhu\textsuperscript{2}~~~
Hartwig Adam\textsuperscript{2}~~~
Alan Yuille\textsuperscript{1}~~~
Liang-Chieh Chen\textsuperscript{2}\\
\textsuperscript{1}Johns Hopkins University~~~
\textsuperscript{2}Google Research
}

\maketitle

\urlstyle{same}

\begin{abstract}
    In this paper, we present ViP-DeepLab, a unified model attempting to tackle the long-standing and challenging inverse projection problem in vision, which we model as restoring the point clouds from perspective image sequences while providing each point with instance-level semantic interpretations.
    Solving this problem requires the vision models to predict the spatial location, semantic class, and temporally consistent instance label for each 3D point.
    ViP-DeepLab approaches it by jointly performing monocular depth estimation and video panoptic segmentation.
    We name this joint task as Depth-aware Video Panoptic Segmentation, and propose a new evaluation metric along with two derived datasets for it, which will be made available to the public.
    On the individual sub-tasks, ViP-DeepLab also achieves state-of-the-art results, outperforming previous methods by 5.1\% VPQ on Cityscapes-VPS, ranking 1st on the KITTI monocular depth estimation benchmark, and 1st on KITTI MOTS pedestrian.
    The datasets and the evaluation codes are made publicly available\footnote{
    \url{https://github.com/joe-siyuan-qiao/ViP-DeepLab}}.
\end{abstract}
\section{Introduction}

The inverse projection problem, one of the most fundamental problems in vision, refers to the ambiguous mapping from the retinal images to the sources of retinal stimulation.
Such a mapping requires retrieving all the visual information about the 3D environment using the limited signals contained in the 2D images~\cite{palmer1999vision,pizlo2001perception}.
Humans are able to easily establish this mapping by identifying objects, determining their sizes, and reconstructing the 3D scene layout, \etc.
To endow machines with similar abilities to visually perceive the 3D world, we aim to develop a model to tackle the inverse projection problem.

\begin{figure}
    \centering
    \includegraphics[width=\linewidth]{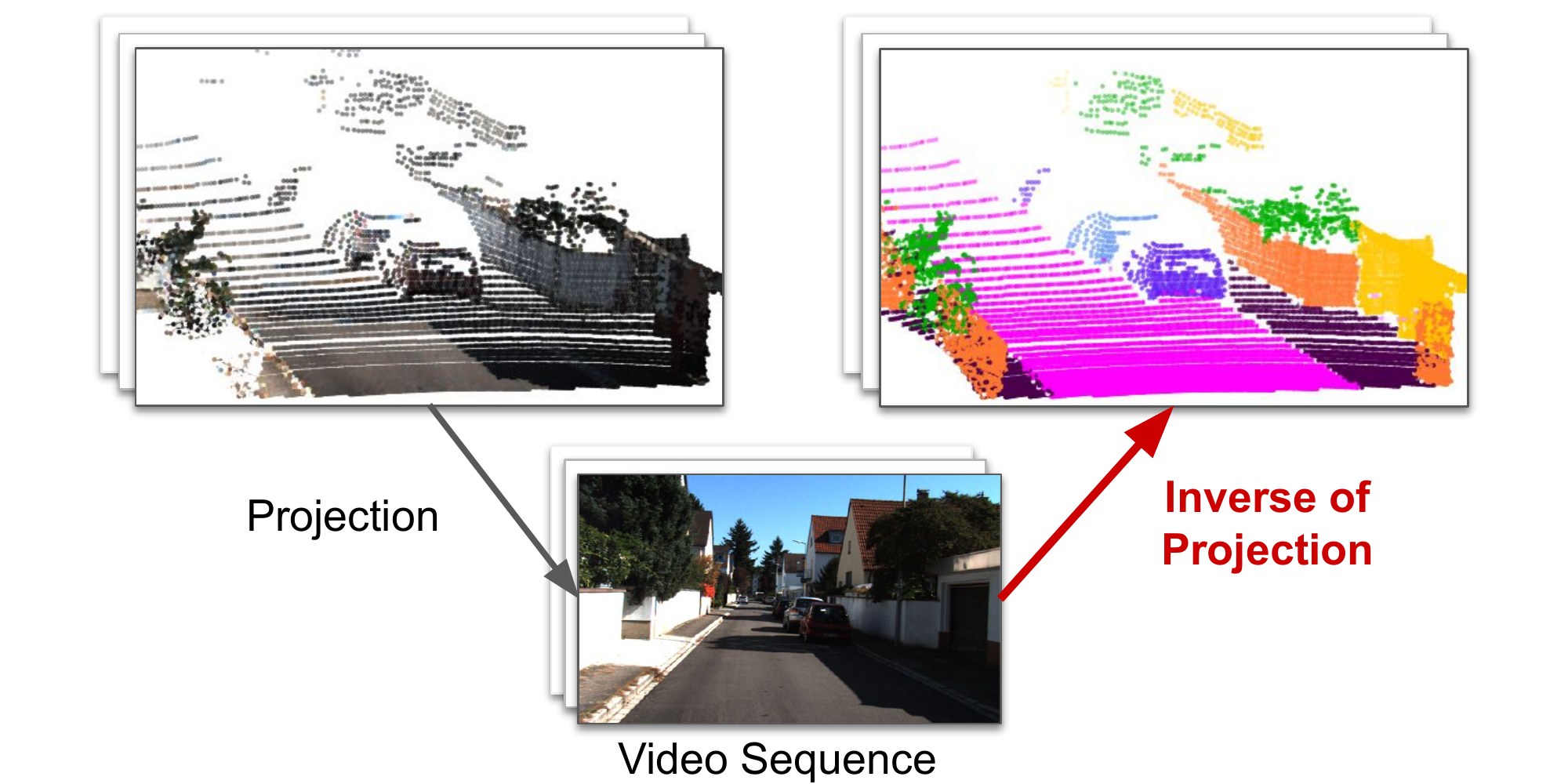}
    \caption{Projecting 3D points to the image plane results in 2D images.
    We study the inverse projection problem: how to restore the 3D points from 2D image sequences while providing temporally consistent instance-level semantic interpretations for the 3D points.
    }
    \label{fig:cover}
\end{figure}

As a step towards solving the inverse projection, the problem is simplified as restoring the 3D point clouds with semantic understandings from the perspective image sequences, which calls for vision models to predict the spatial location, semantic class, and temporally consistent instance label for each 3D point.
Fig.~\ref{fig:cover} shows an example of the inverse projection problem we study in this paper.
This simplified problem can be formulated as Depth-aware Video Panoptic Segmentation (DVPS) that contains two sub-tasks:
(i) monocular depth estimation~\cite{saxena2006learning}, which is used to estimate the spatial position of each 3D point that is projected to the image plane, and (ii) video panoptic segmentation~\cite{kim2020video}, which associates the 3D points with temporally consistent instance-level semantic predictions.

For the new task DVPS, we present two derived datasets accompanied by a new evaluation metric named Depth-aware Video Panoptic Quality (DVPQ).
DVPS datasets are hard to collect, as they need special depth sensors and a huge amount of labeling efforts.
Existing datasets usually lack some annotations or are not in the format for DVPS.
Our solution is to augment and convert existing datasets for DVPS, producing two new datasets, Cityscapes-DVPS and SemKITTI-DVPS.
Cityscapes-DVPS is derived from Cityscapes-VPS~\cite{kim2020video} by adding depth annotations from Cityscapes dataset~\cite{cordts2016cityscapes}, while
SemKITTI-DVPS is derived from SemanticKITTI~\cite{behley2019semantickitti} by projecting its annotated 3D point clouds to the image plane.
Additionally, the proposed metric DVPQ includes the metrics for depth estimation and video panoptic segmentation, requiring a vision model to simultaneously tackle the two sub-tasks.
To this end, we present ViP-DeepLab, a unified model that jointly performs video panoptic segmentation and monocular depth estimation for each pixel on the image plane.
In the following, we introduce how ViP-DeepLab tackles the two sub-tasks.

The first sub-task of DVPS is video panoptic segmentation~\cite{kim2020video}.
Panoptic segmentation~\cite{kirillov2019panoptic} unifies semantic segmentation~\cite{He2004CVPR} and instance segmentation~\cite{Hariharan2014ECCV} by assigning every pixel a semantic label and an instance ID.
It has been recently extended to the video domain, resulting in video panoptic segmentation~\cite{kim2020video}, which further demands each instance to have the same instance ID throughout the video sequence.
This poses additional challenges to panoptic segmentation as the model is now expected to be able to track objects in addiction to detecting and segmenting them.
Current approach VPSNet~\cite{kim2020video} adds a tracking head to learn the correspondence between the instances from different frames based on their regional feature similarity.
By contrast, our ViP-DeepLab takes a different approach to tracking objects.
Specifically, motivated by our finding that video panoptic segmentation can be modeled as concatenated image panoptic segmentation, we extend Panoptic-DeepLab~\cite{cheng2020panoptic} to perform center regression for {\it two} consecutive frames with respect to {\it only} the object centers that appear in the {\it first} frame.
During inference, this offset prediction allows ViP-DeepLab to group all the pixels in the two frames to the same object that appears in the first frame.
New instances emerge if they are not grouped to the previously detected instances.
This inference process continues for every two consecutive frames (with one overlapping frame) in a video sequence, stitching panoptic predictions together to form predictions with temporally consistent instance IDs.
Based on this simple design, our ViP-DeepLab outperforms VPSNet~\cite{kim2020video} by a large margin of 5.1\% VPQ, setting a new record on the Cityscapes-VPS dataset~\cite{kim2020video}.
Additionally, Multi-Object Tracking and Segmentation (MOTS)~\cite{voigtlaender2019mots} is a similar task to video panoptic segmentation, but only segments and tracks two classes: pedestrians and cars.
We therefore also apply our ViP-DeepLab to MOTS. As a result, ViP-DeepLab outperforms the current state-of-the-art PointTrack~\cite{xu2020segment} by 7.2\% and 2.5\% sMOTSA on pedestrians and cars, respectively, and ranks 1st on the leaderboard for KITTI MOTS pedestrian.

The second sub-task of DVPS is monocular depth estimation, which is challenging for both computers~\cite{saxena2006learning} and humans~\cite{howard2012perceiving}.
The state-of-the-art methods are mostly based on deep networks trained in a fully-supervised way~\cite{eigen2014depth,eigen2015predicting,fu2018deep,diaz2019soft}.
Following the same direction, our ViP-DeepLab appends another depth prediction head on top of Panoptic-DeepLab~\cite{cheng2020panoptic}.
Without using any additional {\it depth} training data, such a simple approach outperforms all the published and unpublished works on the KITTI benchmark~\cite{geiger2012we}.
Specifically, it outperforms DORN~\cite{fu2018deep} by 0.97 SILog, and even outperforms MPSD that uses extra planet-scale depth data~\cite{antequera2020mapillary}, breaking the long-standing record on the challenging KITTI depth estimation~\cite{uhrig2017sparsity}.
Notably, the differences between top-performing methods are all around 0.1 SILog, while our method significantly outperforms them.

To summarize, our contributions are listed as follows.
\setlist{nolistsep}
\begin{itemize}[itemsep=0pt]
    \item We propose a new task Depth-aware Video Panoptic Segmentation (DVPS), as a step towards solving the inverse projection problem by formulating it as joint video panoptic segmentation~\cite{kim2020video} and monocular depth estimation~\cite{saxena2006learning}.
    \item We present two DVPS datasets along with an evaluation metric Depth-aware Video Panoptic Quality (DVPQ).
    To facilitate future research, the datasets and the evaluation codes will be made publicly available.
    \item We develop ViP-DeepLab, a unified model for DVPS. On the individual sub-tasks, ViP-DeepLab ranks 1st on Cityscapes-VPS~\cite{kim2020video}, KITTI-MOTS pedestrian~\cite{voigtlaender2019mots}, and KITTI monocular depth estimation~\cite{geiger2012we}.
\end{itemize}

\begin{figure*}
    \centering
    \includegraphics[width=\linewidth]{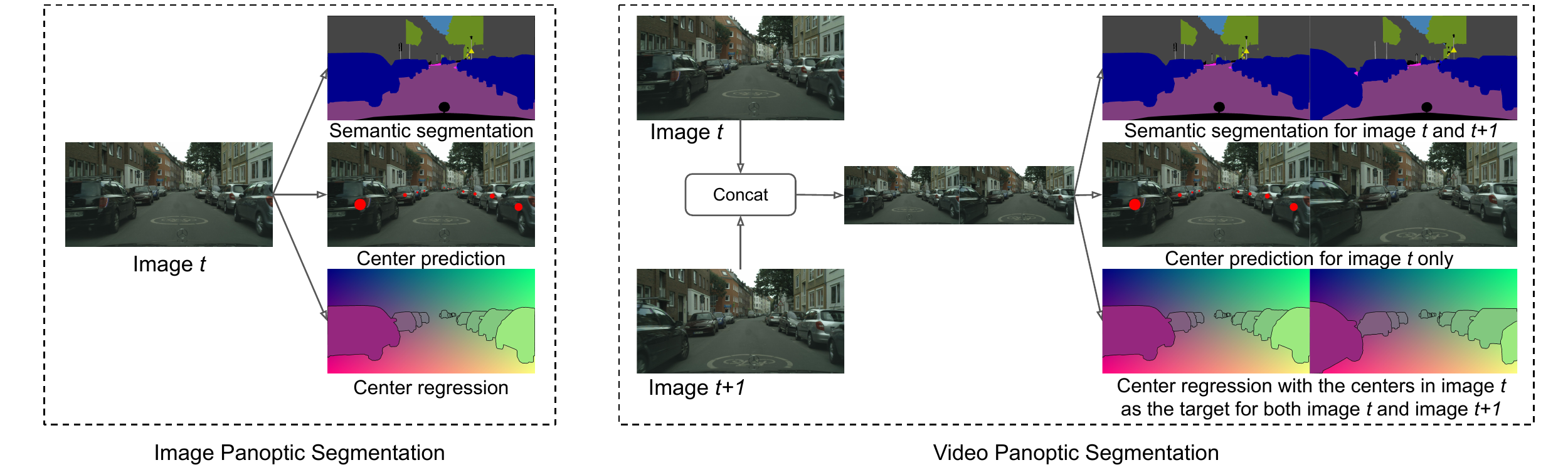}
    \caption{Comparing image panoptic segmentation and video panoptic segmentation.
    Our method is based on the finding that video panoptic segmentation can be modeled as concatenated image panoptic segmentation.
    Center regression is an offset map from each pixel to its object center.
    Here we draw the predicted centers instead of the offsets for clearer visualization.}
    \label{fig:vps}
\end{figure*}

\section{Related Work}

\slimparagraph{Panoptic Segmentation}
Recent methods for image panoptic segmentation can be grouped into two types: top-down (proposal-based) methods and bottom-up (box-free) methods.
Top-down methods employ a two-stage approach which generates object proposals followed by outputting panoptic predictions based on regional computations \cite{sofiiuk2019adaptis,li2020unifying,qiao2020detectors,wu2020bidirectional,lazarow2020learning,chen2020banet,li2018learning,li2019attention,xiong2019upsnet}.
For example, Panoptic FPN~\cite{kirillov2019panoptic} incorporates a semantic segmentation head into Mask R-CNN~\cite{he2017mask}.
Porzi \etal~\cite{porzi2019seamless} proposes a novel segmentation head to integrate FPN~\cite{lin2017feature} features by a lightweight DeepLab-like module~\cite{chen2017deeplab}.
Bottom-up panoptic segmentation methods group pixels to form instances on top of semantic segmentation prediction~\cite{yang2019deeperlab,wang2020axial,wang2020pixel}.
For example, SSAP~\cite{gao2019ssap} uses pixel-pair affinity pyramid~\cite{liu2018affinity} and a cascaded graph partition module~\cite{keuper2015efficient} to generate instances from coarse to fine.
BBFNet~\cite{bonde2020towards} uses Hough-voting~\cite{ballard1981generalizing,leibe2004combined} and Watershed transform~\cite{vincent1991watersheds,bai2017deep} to generate instance segmentation predictions.
Panoptic-DeepLab~\cite{cheng2020panoptic} employs class-agnostic instance center regression~\cite{kendall2018multi,uhrig2018box2pix,neven2019instance} on top of semantic segmentation outputs from DeepLab~\cite{chen2014semantic,chen2018encoder}.

\slimparagraph{Object Tracking}
One of the major tasks in video panoptic segmentation is object tracking.
Many trackers use tracking-by-detection,
which divides the task into two sub-tasks where an object detector (\eg \cite{felzenszwalb2009object,ren2015faster}) finds all objects and then an algorithm associates them~\cite{bewley2016simple,fang2018recurrent,leal2016learning,schulter2017deep,sharma2018beyond,tang2017multiple,wojke2017simple,xu2019spatial,zhu2018online,porzi2020learning}.
Another design is transforming object detectors to object trackers which detect and track objects at the same time~\cite{feichtenhofer2017detect,zhang2018integrated,bergmann2019tracking,peng2020chained,wang2020towards,zhang2020fair}.
For example, 
CenterTrack~\cite{zhou2020tracking} extends CenterNet~\cite{zhou2019objects} to predict offsets from the object center to its center in the previous frame.
STEm-Seg~\cite{athar2020stem} proposes to group all instance pixels in a video clip by learning a spatio-temporal embedding.
By contrast, our ViP-DeepLab implicitly performs object tracking by clustering all instance pixels in two consecutive video frames. Additionally, our method simply uses center
regression and achieves better results on MOTS~\cite{voigtlaender2019mots}.

\slimparagraph{Monocular Depth Estimation}
Monocular depth estimation predicts depth from a single image.
It can be learned in a supervised way~\cite{saxena2006learning,eigen2015predicting,li2015depth,cao2017estimating,fu2018deep,xu2017multi,gan2018monocular,yin2019enforcing,lee2019big},
by reconstructing images in the stereo setting~\cite{garg2016unsupervised,godard2017unsupervised,godard2019digging,xie2016deep3d,kuznietsov2017semi}, from videos~\cite{yin2018geonet,mahjourian2018unsupervised,wang2018learning}, or in relative order~\cite{chen2016single}.
ViP-DeepLab models monocular depth estimation as a dense regression problem, and we train it in a fully-supervised manner.
\begin{figure*}
    \centering
    \includegraphics[trim={100 0 100 0},clip,width=1.0\linewidth]{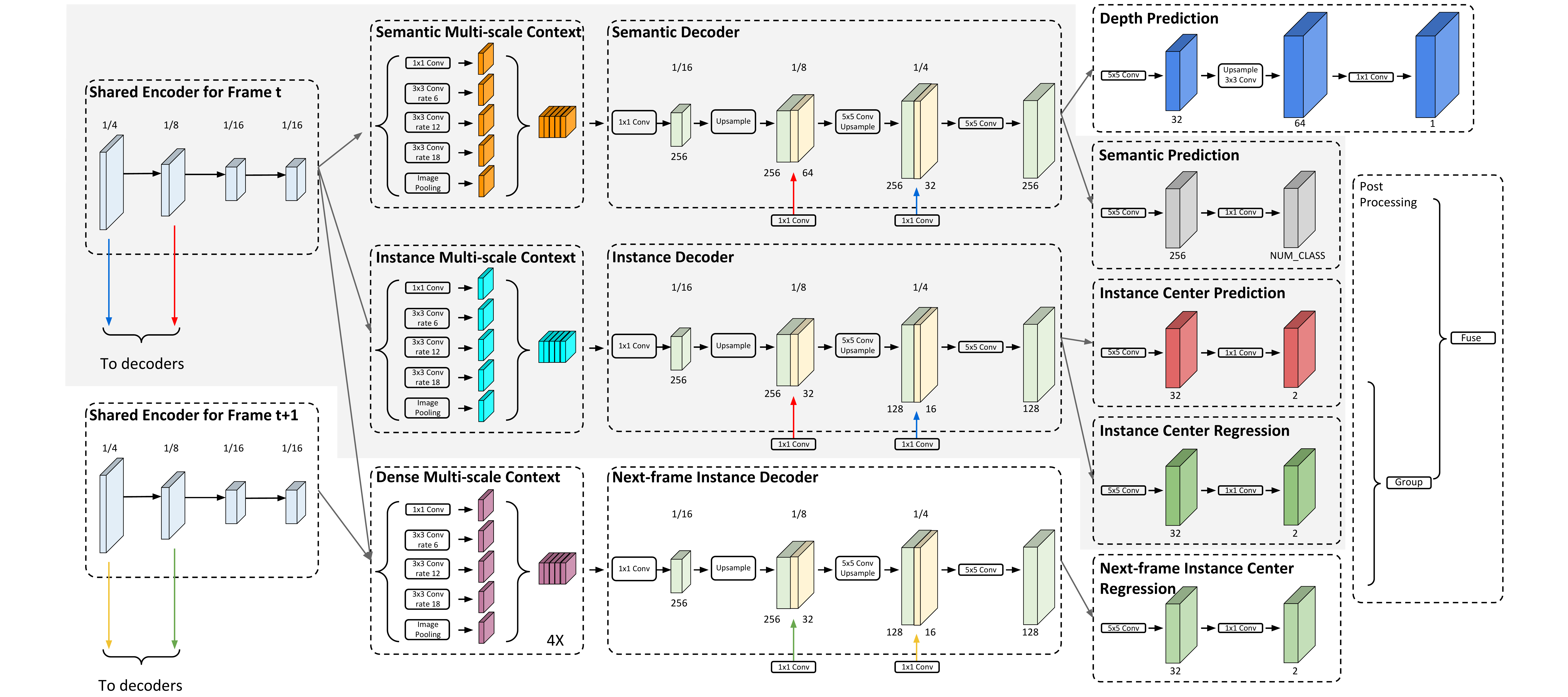}
    \caption{ViP-DeepLab expends Panoptic-DeepLab~\cite{cheng2020panoptic} (the gray part) by adding a depth prediction head to perform monocular depth estimation and a next-frame instance branch which regresses to the object centers in frame $t$ for frame $t+1$.
    }
    \label{fig:arch}
\end{figure*}

\section{ViP-DeepLab}

In this section, we present ViP-DeepLab, which extends Panoptic-DeepLab~\cite{cheng2020panoptic} to jointly perform video panoptic segmentation~\cite{kim2020video} and monocular depth estimation~\cite{saxena2006learning}.

\subsection{Video Panoptic Segmentation}

\slimparagraph{Rethinking Image and Video Panoptic Segmentation}
In the task of video panoptic segmentation, each instance is represented by a tube on the image plane and the time axis when the frames are stacked up.
Given a clip $I^{t:t+k}$ with time window $k$, true positive (TP) is defined by $\text{TP}=\{(u,\hat{u})\in U\times \hat{U}: \text{IoU}(u, \hat{u})>0.5\}$ where $U$ and $\hat{U}$ are the set of the ground-truth and predicted tubes, respectively.
False positives (FP) and false negatives (FN) are defined accordingly.
After accumulating the TP$_c$, FP$_c$, and FN$_c$ on all the clips with window size $k$ and class $c$, the evaluation metric Video Panoptic Quality (VPQ)~\cite{kim2020video} is defined by
\begin{equation}\label{eq:vpq}
    \text{VPQ}^{k} = \dfrac{1}{\text{N}_{\text{classes}}}\sum_c\dfrac{\sum_{(u,\hat{u})\in\text{TP}_c} \text{IoU}(u,\hat{u})}{|\text{TP}_c| + \frac{1}{2}|\text{FP}_c| + \frac{1}{2}|\text{FN}_c|}
\end{equation}
PQ~\cite{kirillov2019panoptic} is thus equal to $\text{VPQ}^{1}$ (\ie, $k=1$).

Our method is based on the connection between $\text{PQ}$ and $\text{VPQ}$.
For an image sequence $I_t$ ($t=1,...,T$), let $P_t$ denote the panoptic prediction and $Q_t$ be the ground-truth panoptic segmentation.
As VPQ$^k$ accumulates the PQ-related statistics from $P_t$ and $Q_t$ within a window of size $k$, we have
\begin{equation}\label{eq:pqvpq}
\small
    \text{VPQ}^k\Big(\Big[P_t, Q_t\Big]_{t=1}^{T}\Big)=\text{PQ}\Big( \Big[\concat_{i=t}^{t+k-1}P_i,\concat_{i=t}^{t+k-1}Q_i\Big]_{t=1}^{T-k+1} \Big)
\end{equation}
where $\concat_{i=t}^{t+k-1}P_i$ denotes the horizontal concatenation of $P_i$ from $t$ to $t+k-1$, and $\big[P_t, Q_t\big]_{t=1}^{T}$ denotes a list of pairs of $(P_t, Q_t)$ from $1$ to $T$ as the function input.

\equref{eq:pqvpq} reveals an interesting finding that video panoptic segmentation could be formulated as image panoptic segmentation with the images concatenated.
Such a finding motivates us to extend image panoptic segmentation models to video panoptic segmentation with extra modifications.

\paragraph{From Image to Video Panoptic Segmentation}
Panoptic-DeepLab~\cite{cheng2020panoptic} approaches the problem of image panoptic segmentation by solving three sub-tasks:
(1) semantic predictions for both `thing' and `stuff' classes,
(2) center prediction for each instance of `thing' classes,
and (3) center regression for each pixel of objects.
\figref{fig:vps} shows an example of the tasks on the left.
During inference, object centers with high confidence scores are kept, and each `thing' pixel is associated with the closest object center to form object instances.
Combining this `thing' prediction and the `stuff' prediction from semantic segmentation, Panoptic-DeepLab~\cite{cheng2020panoptic} generates the final panoptic prediction.

\begin{figure}
    \centering
    \includegraphics[trim={120 80 120 0},clip,width=\linewidth]{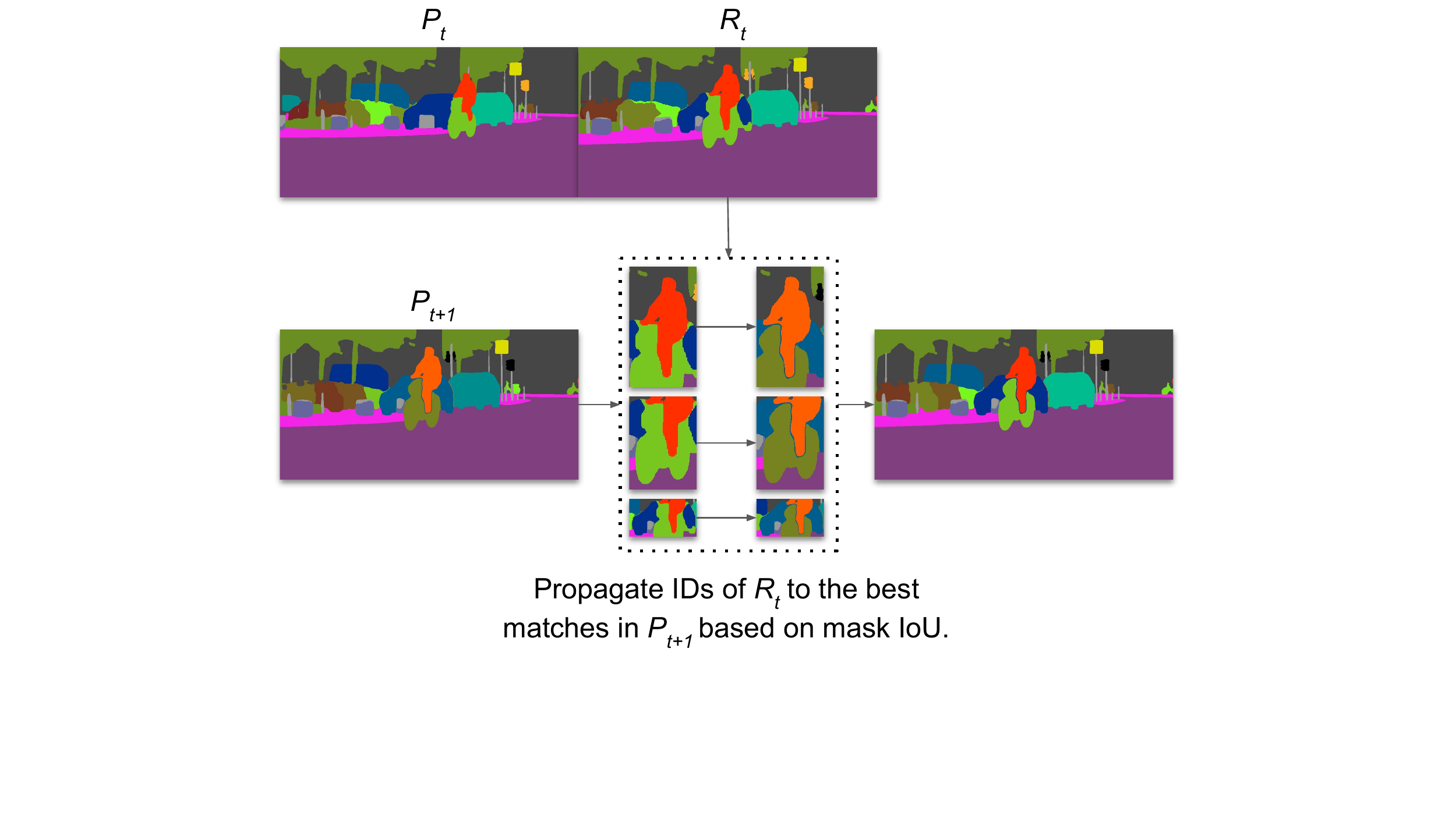}
    \caption{Visualization of stitching video panoptic predictions.
    It propagates IDs based on mask IoU between region pairs.
    ViP-DeepLab is capable of tracking objects with large movements, \eg, the cyclist in the image.
    Panoptic prediction of $R_t$ is of high quality, which is why a simple IoU-based stitching method works well in practice.
    }
    \label{fig:stitching}
\end{figure}

Our method extends Panoptic-DeepLab~\cite{cheng2020panoptic} to perform video panoptic segmentation.
As the right part of \figref{fig:vps} shows, it also breaks down the task of video panoptic segmentation into three sub-tasks: semantic segmentation, center prediction, and center regression.
During inference, our method takes image $t$ and $t+1$ concatenated horizontally as input, and only predict the centers in image $t$.
The center regression for both $t$ and $t+1$ will regress to the object centers in image $t$.
By doing so, our method detects the objects in the first frame, and finds all the pixels belonging to them in the first and the second frames.
Objects that appear only in the second frame are ignored here and will emerge again when the model works on the next image pair (\ie, ($t+1$, $t+2$)).
Our method models video panoptic segmentation as concatenated image panoptic segmentation, highly consistent with the definition of the metric VPQ.

\figref{fig:arch} shows the architecture of our method.
In order to perform the inference described above, we take image $t$ and $t+1$ as the input during training, and use the features of image $t$ to predict semantic segmentation, object centers and center offsets for image $t$.
In addition to that, we add a {\it next-frame instance branch} which predicts the center offsets for the pixels in image $t+1$ with respect to the centers in image $t$.
The backbone features of image $t$ and $t+1$ are concatenated along the feature axis before the {\it next-frame instance branch}.
As their backbone features are separated before concatenation, the {\it next-frame instance branch} needs a large receptive field to perform long-range center regression.
To address this, we use four ASPP modules in the branch, the output of which are densely-connected~\cite{huang2017densely,yang2018denseaspp} to dramatically increase the receptive field.
We name this densely-connected module as Cascade-ASPP.
Its architecture details are shown in Appendix.
Finally, the decoder in the {\it next-frame instance branch} uses the backbone features of image $t+1$ while the other branches use those of image $t$,
as indicated by the colored arrows in the figure.

\paragraph{Stitching Video Panoptic Predictions}
Our method outputs panoptic predictions with temporally consistent IDs for two consecutive frames.
To generate predictions for the entire sequence, we need to stitch the panoptic predictions.
Fig~\ref{fig:stitching} shows an example of our stitching method.
For each image pair $t$ and $t+1$, we split the panoptic prediction of the concatenated input in the middle, and use $P_t$ to denote the left prediction, and $R_t$ to denote the right one.
By doing so, $P_t$ becomes the panoptic prediction of image $t$, and $R_t$ becomes the panoptic prediction of image $t+1$ with instance IDs that are consistent with those of $P_t$.
The goal of stitching is to propagate IDs from $R_t$ to $P_{t+1}$ so that each object in $P_{t}$ and $P_{t+1}$ will have the same ID.

The ID propagation is based on mask IoU between region pairs.
For each region pair in $R_t$ and $P_{t+1}$, if they have the same class, and both find each other to have the largest mask IoU, then we propagate the ID between them.
Objects that do not receive IDs will become new instances.
A formal algorithm can be found in Appendix.

\subsection{Monocular Depth Estimation}
We model monocular depth estimation as a dense regression problem~\cite{eigen2014depth},
where each pixel will have an estimated depth.
As shown in \figref{fig:arch}, we add a depth prediction head on top of the decoded features of the semantic branch (\ie, Semantic Decoder), which upsamples the features by 2x and generates logits $f_d$ for depth regression:
\begin{equation}\label{eq:dep_reg}
    \text{Depth}=\text{MaxDepth} \times \text{Sigmoid}(f_d)
\end{equation}
MaxDepth controls the range of the predicted depth, which is set to $88$ for the range (about $0$ to $80$m) of KITTI~\cite{uhrig2017sparsity}.

Many metrics have been proposed to evaluate the quality of monocular depth prediction~\cite{geiger2012we}.
Among them, scale invariant logarithmic error~\cite{eigen2014depth} and relative squared error~\cite{geiger2012we}
are popular ones, which could also be directly optimized as training loss functions. We therefore combine them to train our depth prediction. 
Specifically, let $d$ and $\hat{d}$ denote the ground-truth and the predicted depth, respectively.
Our loss function for depth estimation is then defined by
\begin{align}\label{eq:depth_loss}
    \mathcal{L}_{\text{depth}}(d, \hat{d}) =& \frac{1}{n}\sum_i\Big(\log d_i - \log\hat{d}_i\Big)^2 -\frac{1}{n^2}\Big(\sum_i \log d_i \nonumber \\ & - \log\hat{d}_i\Big)^2 +\Big( \frac{1}{n} \sum_i \big(\frac{d_i - \hat{d}_i}{d_i}\big)^2  \Big)^{0.5}
\end{align}

\subsection{Depth-aware Video Panoptic Segmentation}
Motivated by solving the inverse projection problem, we introduce a challenging task, Depth-aware Video Panoptic Segmentation (DVPS), unifying the problems of monocular depth estimation and video panoptic segmentation.
In the task of DVPS, images are densely annotated with a tuple $(c, id, d)$ for each labeled pixel, where $c$, $id$ and $d$ denote its semantic class, instance ID and depth.
The model is expected to also generate a tuple $(\hat{c}, \hat{id}, \hat{d})$ for each pixel.

To evaluate methods for DVPS, we propose a metric called Depth-aware Video Panoptic Quality (DVPQ), which extends VPQ by additionally considering the depth prediction with the inlier metric.
Specifically, let $P$ and $Q$ be the prediction and ground-truth, respectively.
We use $P_i^c$, $P_i^{id}$ and $P_i^{d}$ to denote the predictions of example $i$
on the semantic class, instance ID, and depth.
The notations also apply to $Q$.
Let $k$ be the window size (as in \equref{eq:pqvpq}) and $\lambda$ be the depth threshold.
Then, $\text{DVPQ}^k_{\lambda}(P,Q)$ is defined by
\begin{equation}\label{eq:dvpq}
    \text{PQ}\Big( \Big[\concat_{i=t}^{t+k-1}\big(\hat{P_i^c}, P_i^{id}\big), \concat_{i=t}^{t+k-1}\big(Q^c_i, Q_i^{id}\big)\Big]_{t=1}^{T-k+1} \Big)
\end{equation}
where $\hat{P_i^c} = P_i^c$ for pixels that have {\it absolute relative} depth errors under $\lambda$ (\ie, $|P_i^d-Q_i^d|\leq \lambda Q_i^d$), 
and will be assigned a void label otherwise.
In other words, $\hat{P_i^c}$ filters out pixels that have large absolute relative depth errors.
As a result, the metrics VPQ~\cite{kim2020video} (also image PQ~\cite{kirillov2019panoptic}) and
depth inlier metric (\ie, $\max(P_i^d/Q_i^d, Q_i^d/P_i^d)=\delta<$ threshold)~\cite{eigen2014depth}
can be \textit{approximately} viewed as special cases for DVPQ.

Following~\cite{kim2020video}, we evaluate $\text{DVPQ}^k_{\lambda}$ for four different values of $k$ (depending on the dataset) and three values of $\lambda=\{0.1, 0.25, 0.5\}$.
Those values of $\lambda$ approximately correspond to the depth inlier metric $\delta < 1.1$, $\delta < 1.25$, and $\delta < 1.5$, respectively.
They are harder than the thresholds $1.25$, $1.25^2$ and $1.25^3$ that are commonly used in depth evaluation.
We choose harder thresholds as many methods are able to get $>99\%$ on the previous metrics~\cite{fu2018deep,lee2019big} .
Larger $k$ and smaller $\lambda$ correspond to a higher accuracy requirement for a long-term consistency of joint video panoptic segmentation and depth estimation.
The final number $\text{DVPQ}$ is obtained by averaging over all values of $k$ and $\lambda$.

\begin{figure*}
\centering
\begin{subfigure}{.33\textwidth}
  \centering
  \includegraphics[trim={0 200 0 200},clip,width=\linewidth]{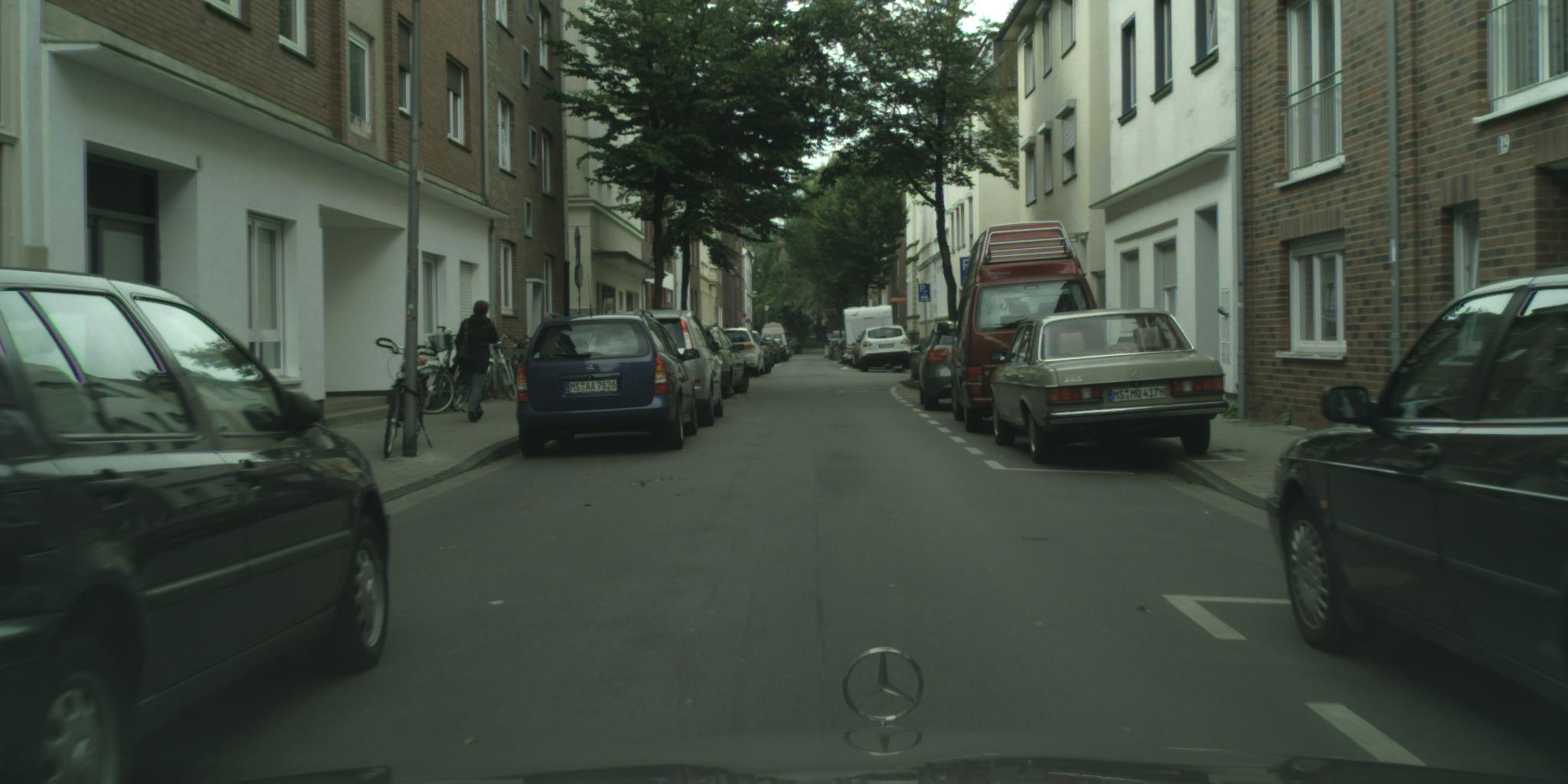}
\end{subfigure}
\begin{subfigure}{.33\textwidth}
  \centering
  \includegraphics[trim={0 200 0 200},clip,width=\linewidth]{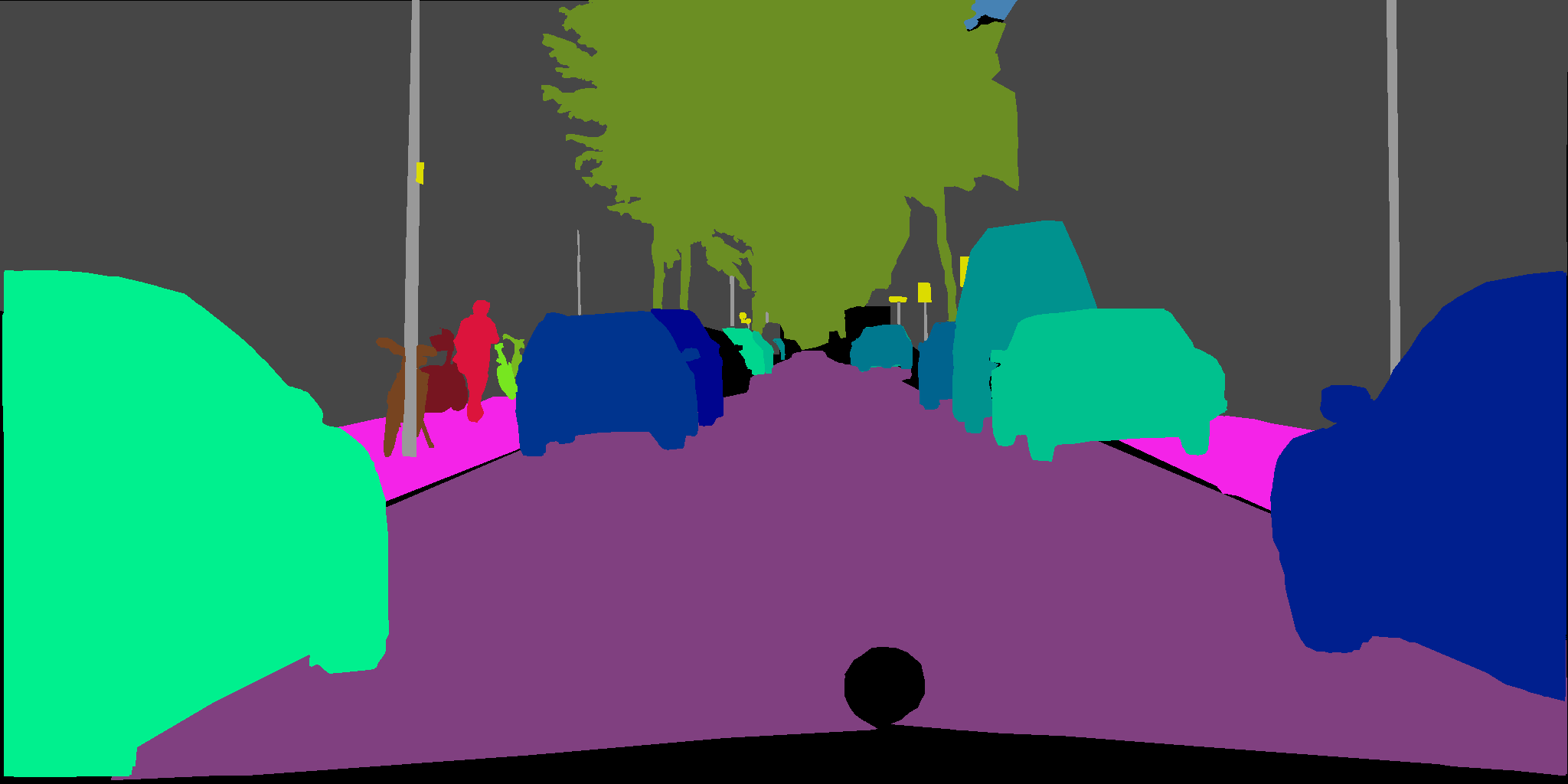}
\end{subfigure}
\begin{subfigure}{.33\textwidth}
  \centering
  \includegraphics[trim={0 200 0 200},clip,width=\linewidth]{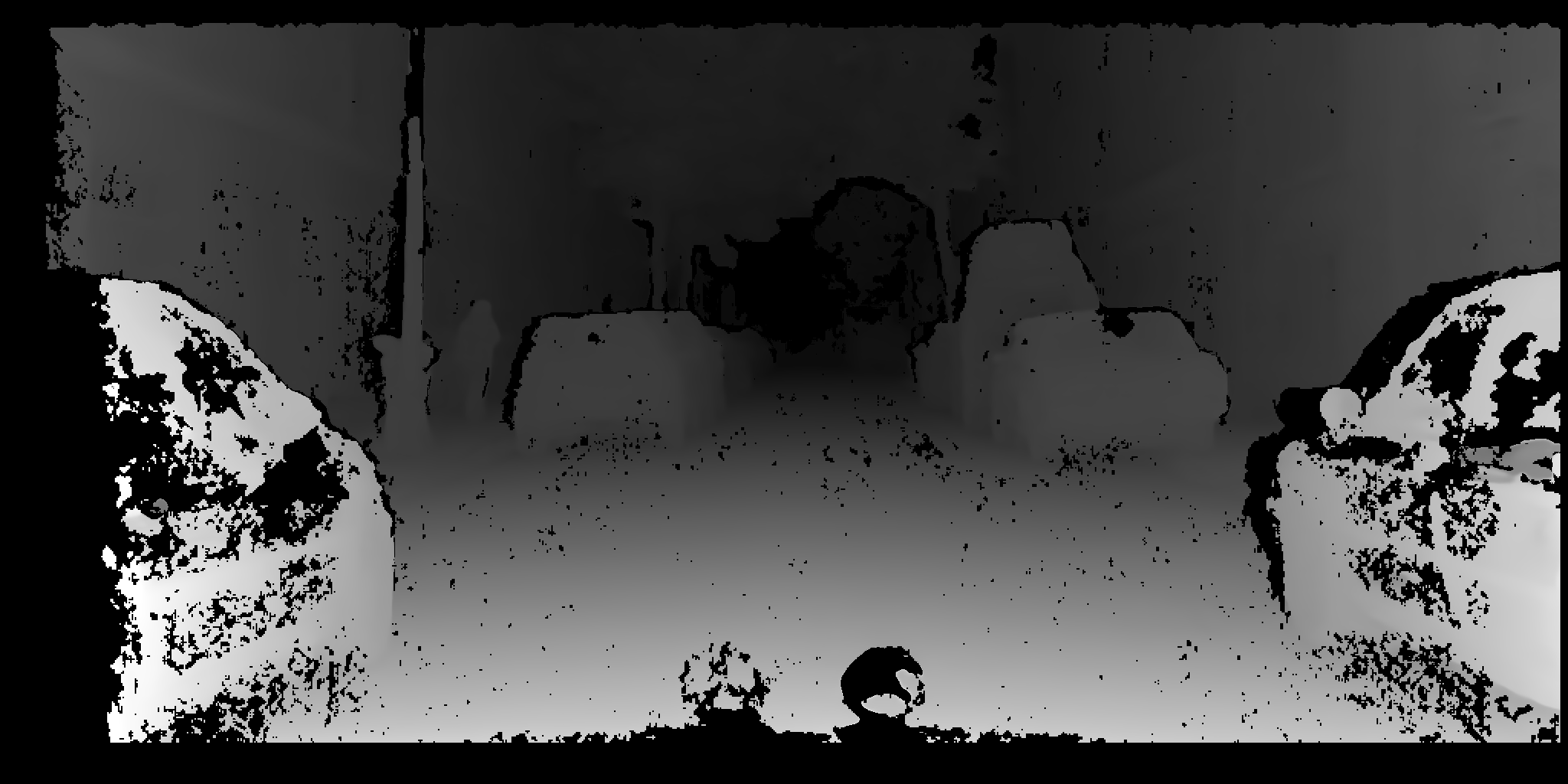}
\end{subfigure}\\[0.1em]
\begin{subfigure}{.33\textwidth}
  \centering
  \includegraphics[width=\linewidth]{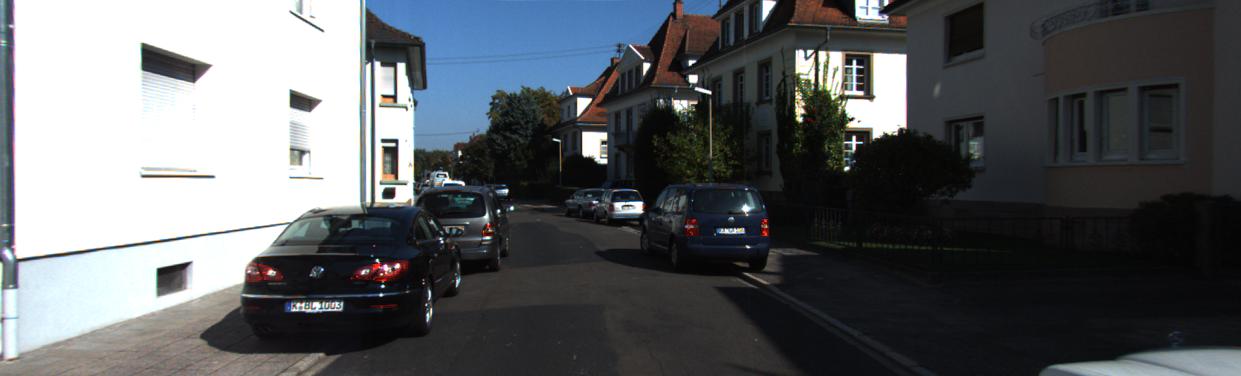}
\end{subfigure}
\begin{subfigure}{.33\textwidth}
  \centering
  \includegraphics[width=\linewidth]{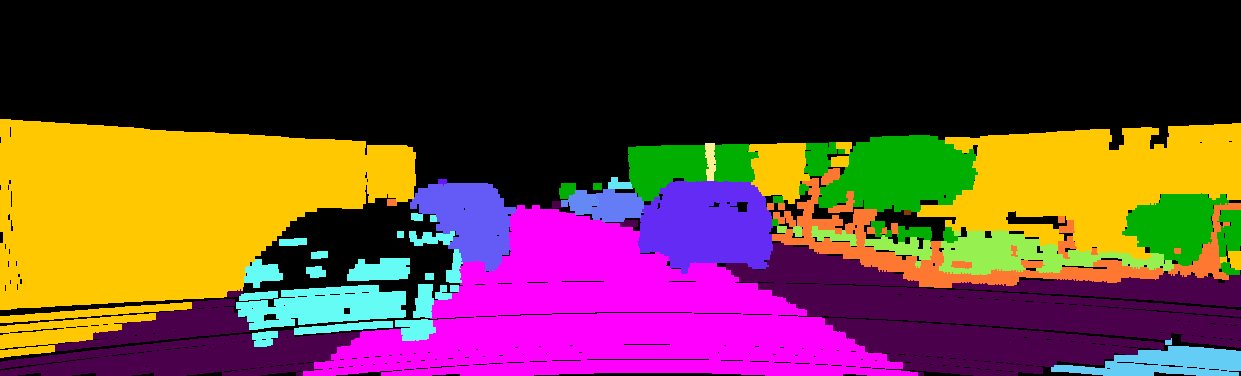}
\end{subfigure}
\begin{subfigure}{.33\textwidth}
  \centering
  \includegraphics[width=\linewidth]{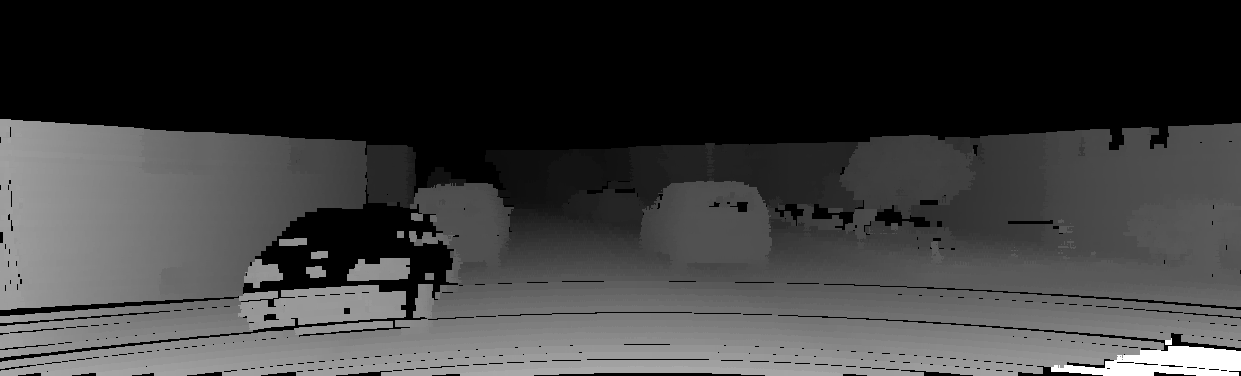}
\end{subfigure}
\caption{Dataset examples of Cityscapes-DVPS (top) and SemKITTI-DVPS (bottom).
From left to right: input image, video panoptic segmentation annotation, and depth map.
Regions are black if they are not covered by the velodyne data or they are removed by the data preprocessing step including disparity consistency check and non-foreground suppression.
}
\label{fig:datasets}
\end{figure*}

\section{Datasets}

To evaluate on the new task, Depth-aware Video Panoptic Segmentation, we create two new datasets, Cityscapes-DVPS and SemKITTI-DVPS.
\figref{fig:datasets} shows two examples, one for each dataset.
The details are elaborated below.

\subsection{Cityscapes-DVPS}
The original Cityscapes~\cite{cordts2016cityscapes} only contains image-level panoptic annotations.
Recently, Kim~\etal introduce a video panoptic segmentation dataset Cityscapes-VPS~\cite{kim2020video} by further annotating 6 frames out of each 30-frame video sequence (with a gap of 5 frames between each annotation), resulting in totally 3,000 annotated frames where the training, validation, and test sets have 2,400, 300, and 300 frames, respectively.
In the dataset, there are 19 semantic classes, including 8 `thing' and 11 `stuff' classes.

Even though Cityscapes-VPS contains video panoptic annotations, the depth annotations are missing.
We find that the depth annotations could be converted from the disparity maps via stereo images, provided by the original Cityscapes dataset~\cite{cordts2016cityscapes}.
However, the quality of the pre-computed disparity maps is not satisfactory.
To improve it, we select several modern disparity estimation methods~\cite{hirschmuller2007stereo,zhang2019ga,zhang2020adaptive,gu2020cascade} and follow the process similar to~\cite{cordts2016cityscapes}. 
Nevertheless, to discourage reproducing the depth generation process (so that one may game the benchmark), we do not disclose the details (\eg, the exact employed disparity method). The depth annotations will be made publicly available.

\subsection{SemKITTI-DVPS}

\begin{figure}
\centering
\begin{subfigure}{.49\linewidth}
  \centering
  \includegraphics[width=\linewidth]{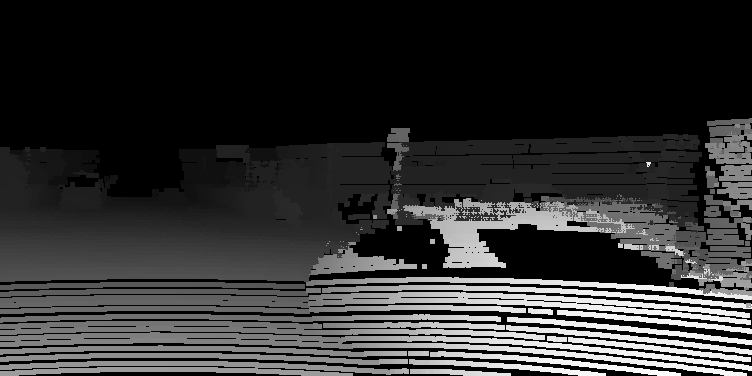}
\end{subfigure}\hfill
\begin{subfigure}{.49\linewidth}
  \centering
  \includegraphics[width=\linewidth]{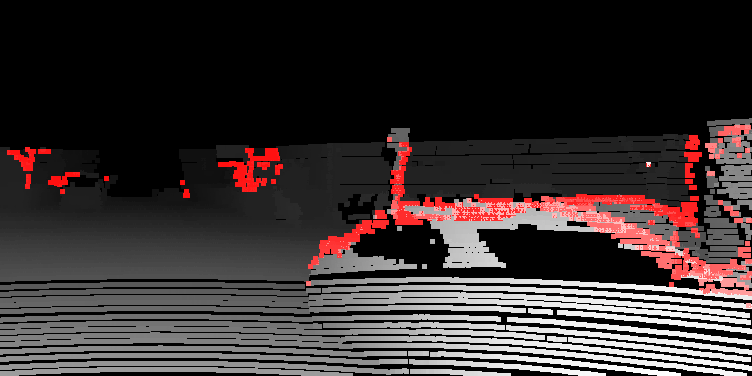}
\end{subfigure}\\[0.1em]
\begin{subfigure}{.49\linewidth}
  \centering
  \includegraphics[width=\linewidth]{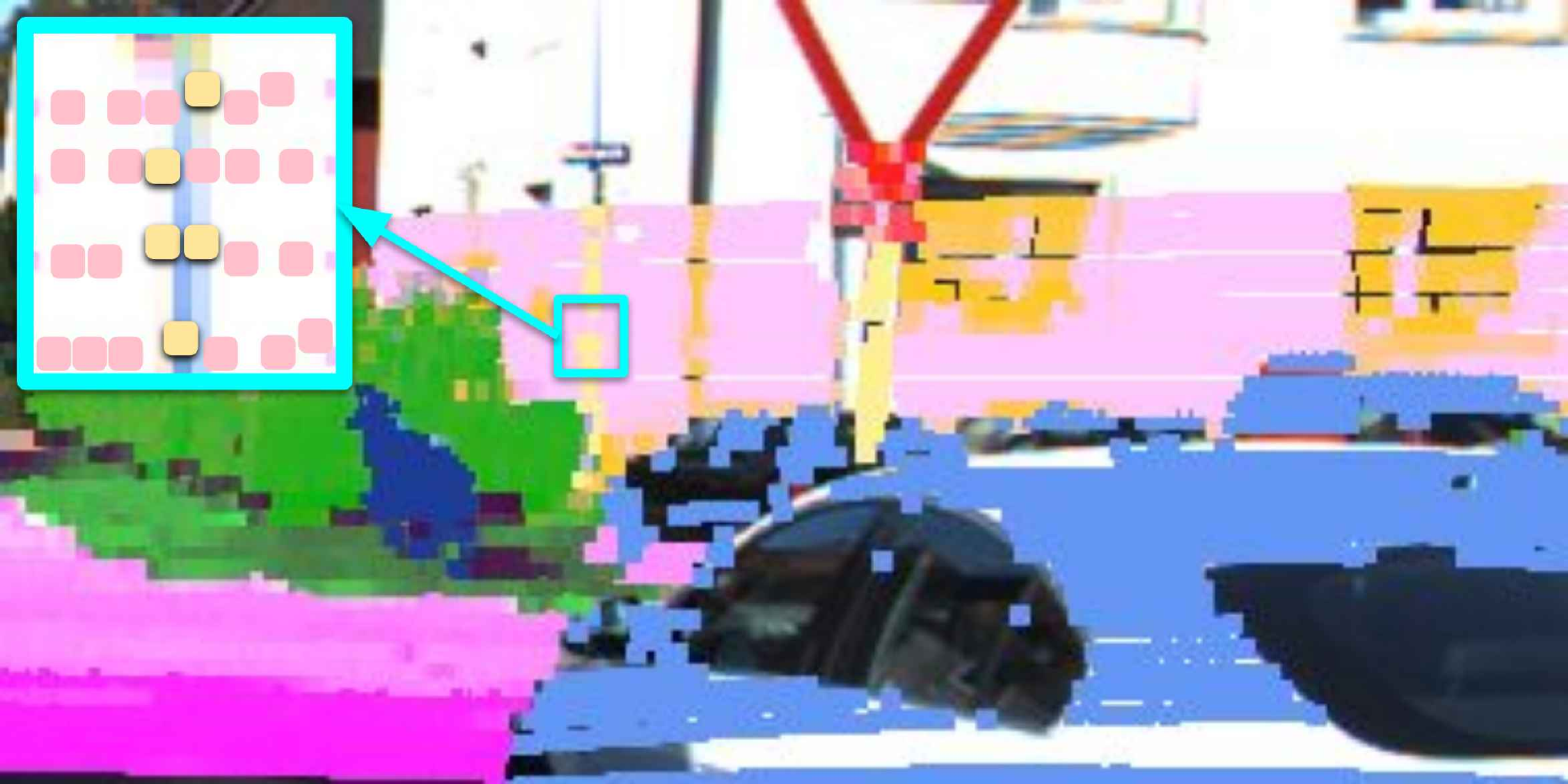}
\end{subfigure}\hfill
\begin{subfigure}{.49\linewidth}
  \centering
  \includegraphics[width=\linewidth]{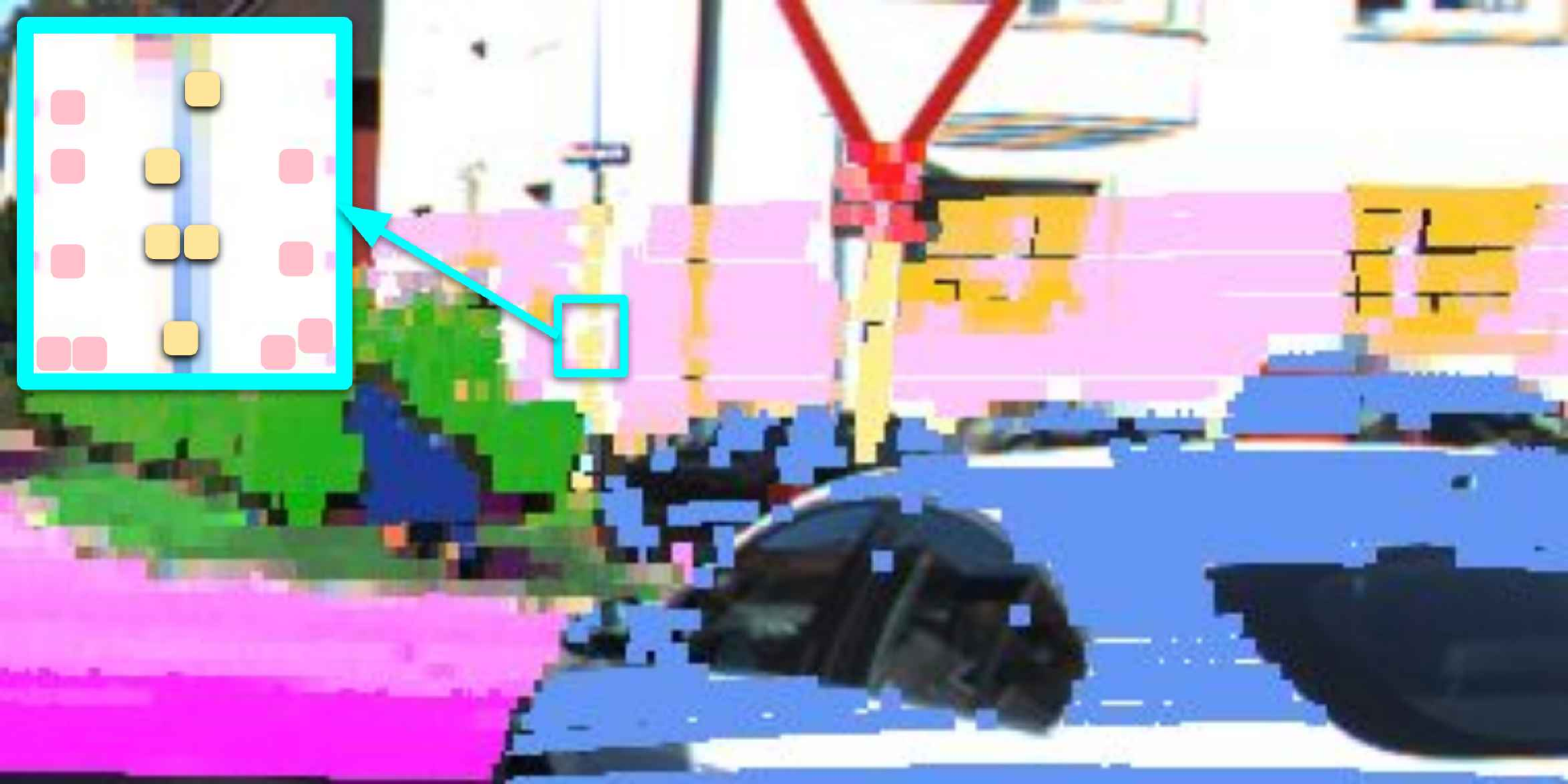}
\end{subfigure}
\caption{
Top: Removing occluded but falsely visible points highlighted in red by disparity consistency check.
Bottom: Removing the invading background points in pink for the thin object colored yellow by non-foreground suppression.
}
\label{fig:sk}
\end{figure}

SemanticKITTI dataset~\cite{behley2019semantickitti} is based on the odometry dataset of the KITTI Vision benchmark~\cite{geiger2012we}.
The dataset splits 22 sequences in to 11 training sequences and 11 test sequences.
The training sequence 08 is used for validation.
This dataset includes 8 `thing' and 11 `stuff' classes.

SemanticKITTI dataset provides perspective images and panoptic-labeled 3D point clouds (\ie, semantic class and instance ID are annotated).
To convert it for our use, we project the 3D point clouds into the image plane.
However, there are two challenges when converting the dataset, as presented in \figref{fig:sk}.
The first problem is that some point clouds are not visible to the camera but are recorded and labeled.
For example, the first row of \figref{fig:sk} shows that some regions behind the car become visible in the converted depth map due to the alignment of different sensors.
To address this issue, we follow Uhrig \etal~\cite{uhrig2017sparsity} and use the same disparity methods for Cityscapes-DVPS to remove the sampled points that exhibit large relative errors, which are highlighted in red in the right figure.
We refer to this processing as disparity consistency check.
The second problem is that the regions of thin objects (\eg, poles) are usually invaded by the far-away background point cloud after projection.
To alleviate this problem, for a small image patch, the projected background points are removed if there exits at least one foreground point that is closer to the camera.
We refer to this processing as non-foreground suppression.
In practice, we use a small $7\times7$ image patch. 
Doing so leaves clear boundaries for thin objects so they can be identified without confusion as shown in the second row of \figref{fig:sk}.

\begin{table*}[!t]
\renewcommand{\arraystretch}{0.9}
\small
\centering
    \begin{tabular}{l|c|c|c|c|c}
    \toprule[0.2em]
    $\text{DVPQ}_\lambda^k$ on Cityscapes-DVPS & k = 1 & k = 2 & k = 3 & k = 4 & Average\\
    \toprule[0.2em]
    $\lambda$ = 0.50 & 68.7 $\vert$ 61.4 $\vert$ 74.0 & 61.7 $\vert$ 48.5 $\vert$ 71.3 & 58.4 $\vert$ 42.1 $\vert$ 70.2 & 56.3 $\vert$ 38.0 $\vert$ 69.5 & 61.3 $\vert$ 47.5 $\vert$ 71.2 \\
    $\lambda$ = 0.25 & 66.5 $\vert$ 60.4 $\vert$ 71.0 & 59.5 $\vert$ 47.6 $\vert$ 68.2 & 56.2 $\vert$ 41.3 $\vert$ 67.1 & 54.2 $\vert$ 37.3 $\vert$ 66.5 & 59.1 $\vert$ 46.7 $\vert$ 68.2 \\
    $\lambda$ = 0.10 & 50.5 $\vert$ 45.8 $\vert$ 53.9 & 45.6 $\vert$ 36.9 $\vert$ 51.9 & 42.6 $\vert$ 31.7 $\vert$ 50.6 & 40.8 $\vert$ 28.4 $\vert$ 49.8 & 44.9 $\vert$ 35.7 $\vert$ 51.5 \\
    \midrule
    Average & 61.9 $\vert$ 55.9 $\vert$ 66.3 & 55.6 $\vert$ 44.3 $\vert$ 63.8 & 52.4 $\vert$ 38.4 $\vert$ 62.6 & 50.4 $\vert$ 34.6 $\vert$ 61.9 & \textbf{55.1 $\vert$ 43.3 $\vert$ 63.6} \\
    \bottomrule[0.1em]
    \toprule[0.2em]
    $\text{DVPQ}_\lambda^k$ on SemKITTI-DVPS & k = 1 & k = 5 & k = 10 & k = 20 & Average\\
    \toprule[0.2em]
    $\lambda$ = 0.50 & 54.7 $\vert$ 46.4 $\vert$ 60.6 & 51.5 $\vert$ 41.0 $\vert$ 59.1 & 50.1 $\vert$ 38.5 $\vert$ 58.5 & 49.2 $\vert$ 36.9 $\vert$ 58.2 & 51.4 $\vert$ 40.7 $\vert$ 59.1 \\
    $\lambda$ = 0.25 & 52.0 $\vert$ 44.8 $\vert$ 57.3 & 48.8 $\vert$ 39.4 $\vert$ 55.7 & 47.4 $\vert$ 37.0 $\vert$ 55.1 & 46.6 $\vert$ 35.6 $\vert$ 54.7 & 48.7 $\vert$ 39.2 $\vert$ 55.7\\
    $\lambda$ = 0.10 & 40.0 $\vert$ 34.7 $\vert$ 43.8 & 37.1 $\vert$ 30.3 $\vert$ 42.0 & 35.8 $\vert$ 28.3 $\vert$ 41.2 & 34.5 $\vert$ 26.5 $\vert$ 40.4 & 36.8 $\vert$ 30.0 $\vert$ 41.9 \\
    \midrule
    Average & 48.9 $\vert$ 42.0 $\vert$ 53.9 & 45.8 $\vert$ 36.9 $\vert$ 52.3 & 44.4 $\vert$ 34.6 $\vert$ 51.6 & 43.4 $\vert$ 33.0 $\vert$ 51.1 & \textbf{45.6 $\vert$ 36.6 $\vert$ 52.2} \\
    \bottomrule[0.1em]
    \end{tabular}
    \caption{ViP-DeepLab performance for the task of {\it Depth-aware Video Panoptic Segmentation} (DVPS) evaluated on Cityscapes-DVPS and SemKITTI-DVPS.
    Each cell shows $\text{DVPQ}_{\lambda}^{k} ~\vert~ \text{DVPQ}_{\lambda}^{k}\text{-Thing} ~\vert~ \text{DVPQ}_{\lambda}^{k}\text{-Stuff}$ where $\lambda$ is the threshold of relative depth error, and $k$ is the number of frames. Smaller $\lambda$ and larger $k$ correspond to a higher accuracy requirement.
    }
    \label{tab:dvps}
\end{table*}

\section{Experiments}\label{sec:exp}

In this section, we first present our major results on the new task Depth-aware Video Panoptic Segmentation.
Then, we show our method applied to three sub-tasks, including video panoptic segmentation~\cite{kim2020video}, monocular depth estimation~\cite{geiger2012we}, and multi-object tracking and segmentation~\cite{voigtlaender2019mots}.

\subsection{Depth-aware Video Panoptic Segmentation}

\tabref{tab:dvps} shows our results on Depth-aware Video Panoptic Segmentation.
We evaluate our method on the datasets Cityscapes-DVPS and SemKITTI-DVPS that we are going to make publicly available, so that the research community can compare their methods with it.
The evaluation is based on our proposed $\text{DVPQ}_{\lambda}^{k}$ metric (\equref{eq:dvpq}), where $\lambda$ is the threshold of relative depth error, and $k$ denotes the length of the short video clip used in evaluation. 

\begin{figure*}[!t]
\centering
\begin{subfigure}{.245\linewidth}
  \centering
  \includegraphics[trim={0 256 0 256},clip,width=\linewidth]{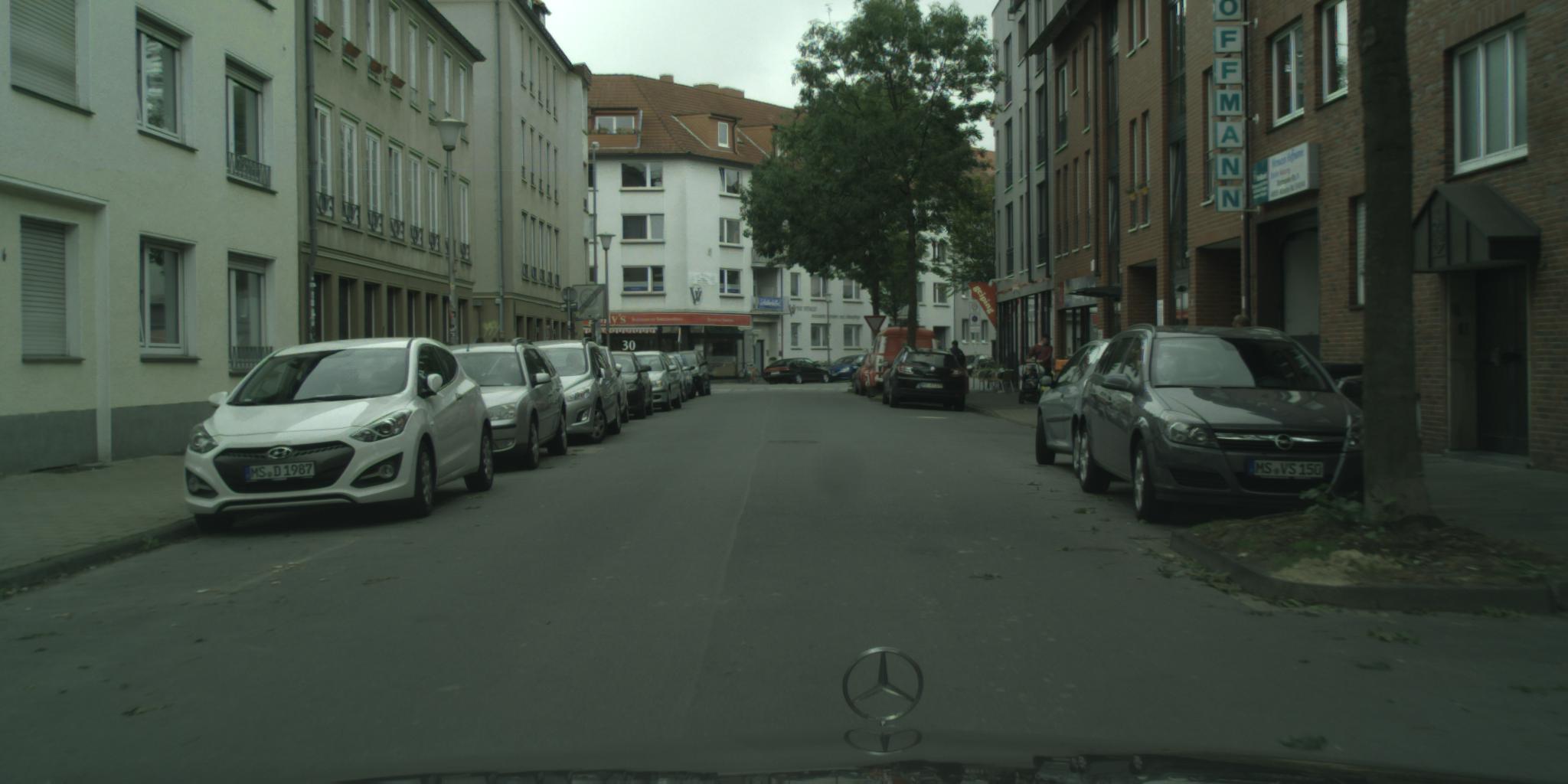}
\end{subfigure}\hfill
\begin{subfigure}{.245\linewidth}
  \centering
  \includegraphics[trim={0 256 0 256},clip,width=\linewidth]{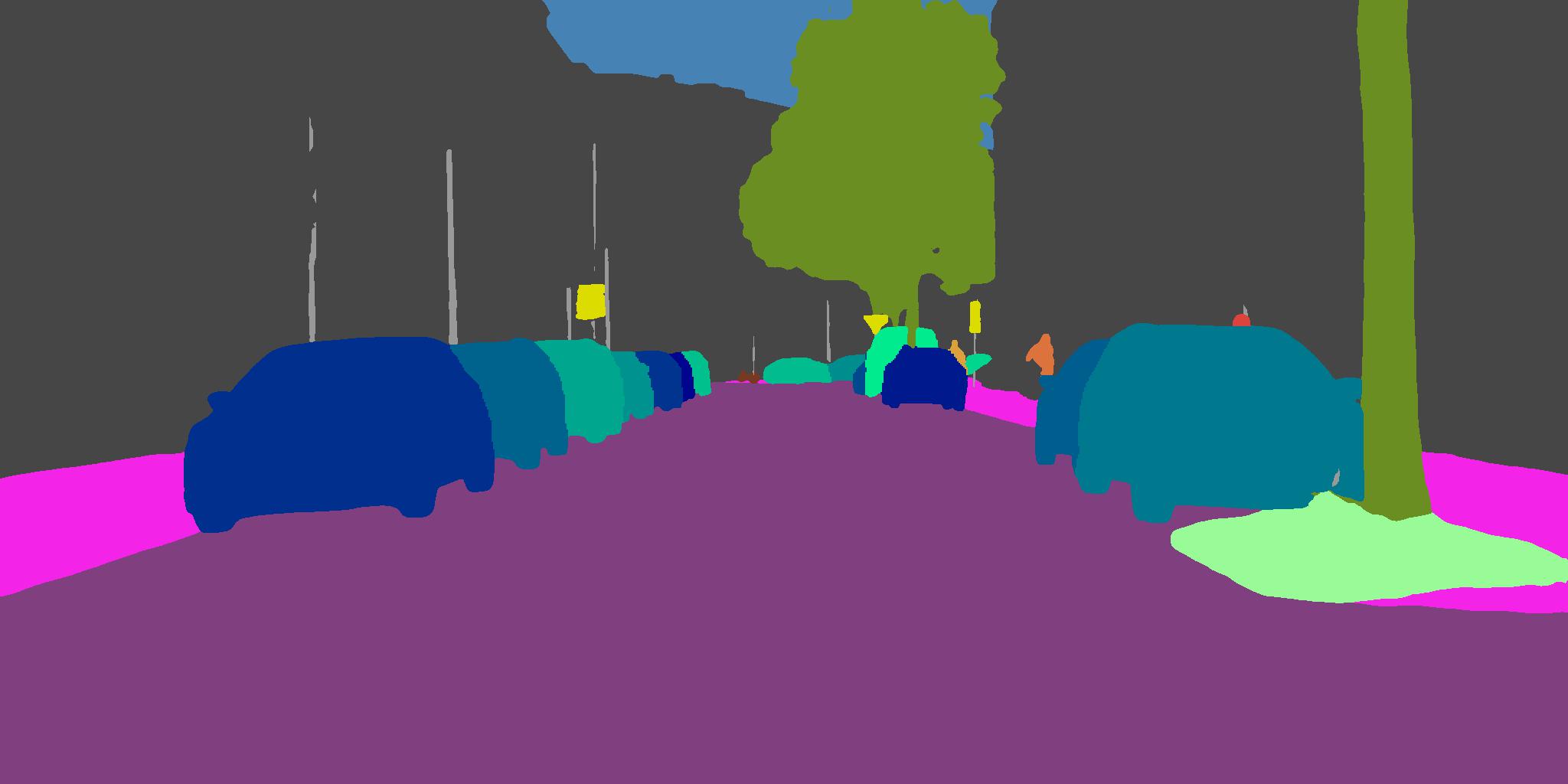}
\end{subfigure}\hfill
\begin{subfigure}{.245\linewidth}
  \centering
  \includegraphics[trim={0 256 0 256},clip,width=\linewidth]{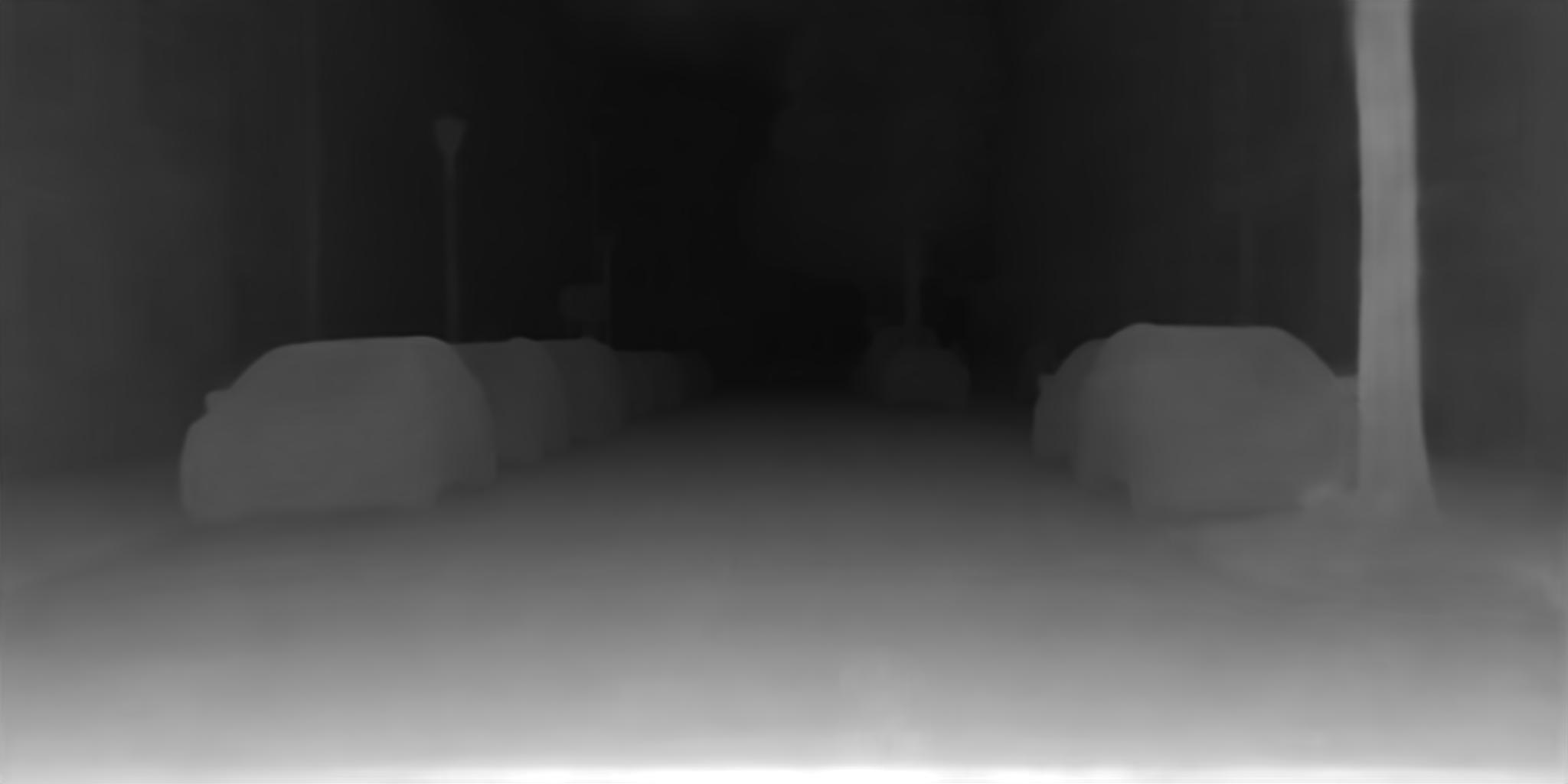}
\end{subfigure}\hfill
\begin{subfigure}{.245\linewidth}
  \centering
  \includegraphics[width=\linewidth]{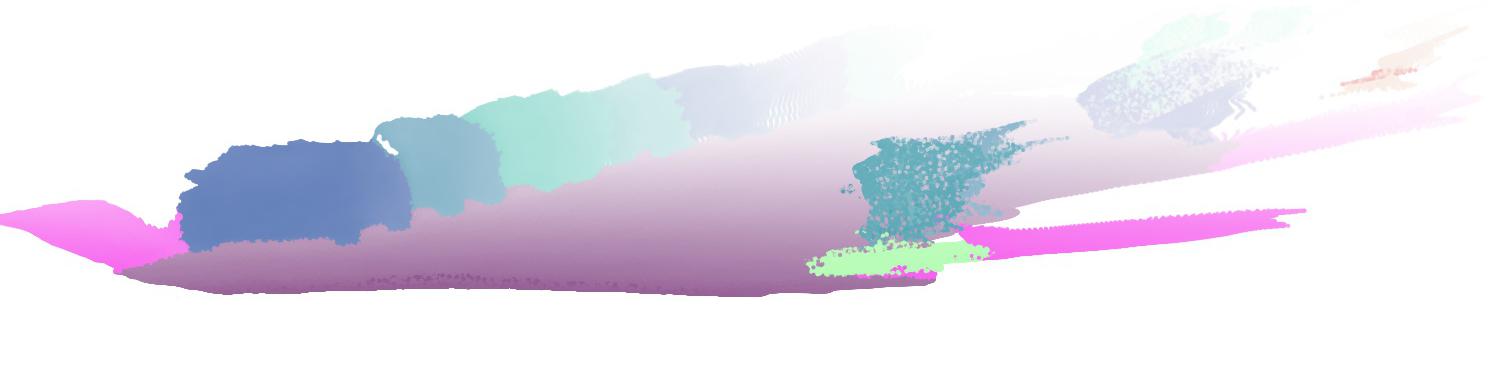}
\end{subfigure}\\
\begin{subfigure}{.245\linewidth}
  \centering
  \includegraphics[trim={0 256 0 256},clip,width=\linewidth]{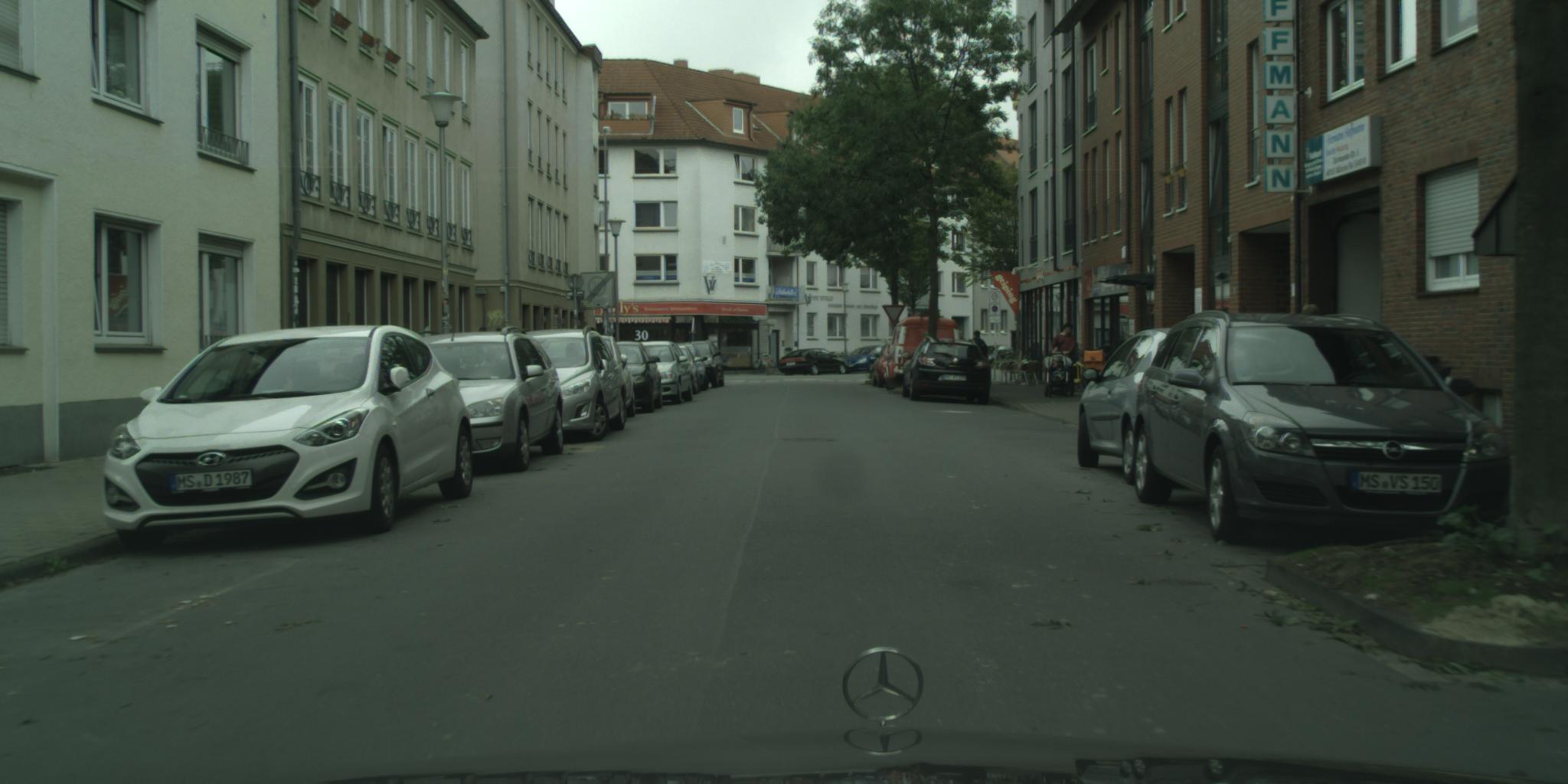}
\end{subfigure}\hfill
\begin{subfigure}{.245\linewidth}
  \centering
  \includegraphics[trim={0 256 0 256},clip,width=\linewidth]{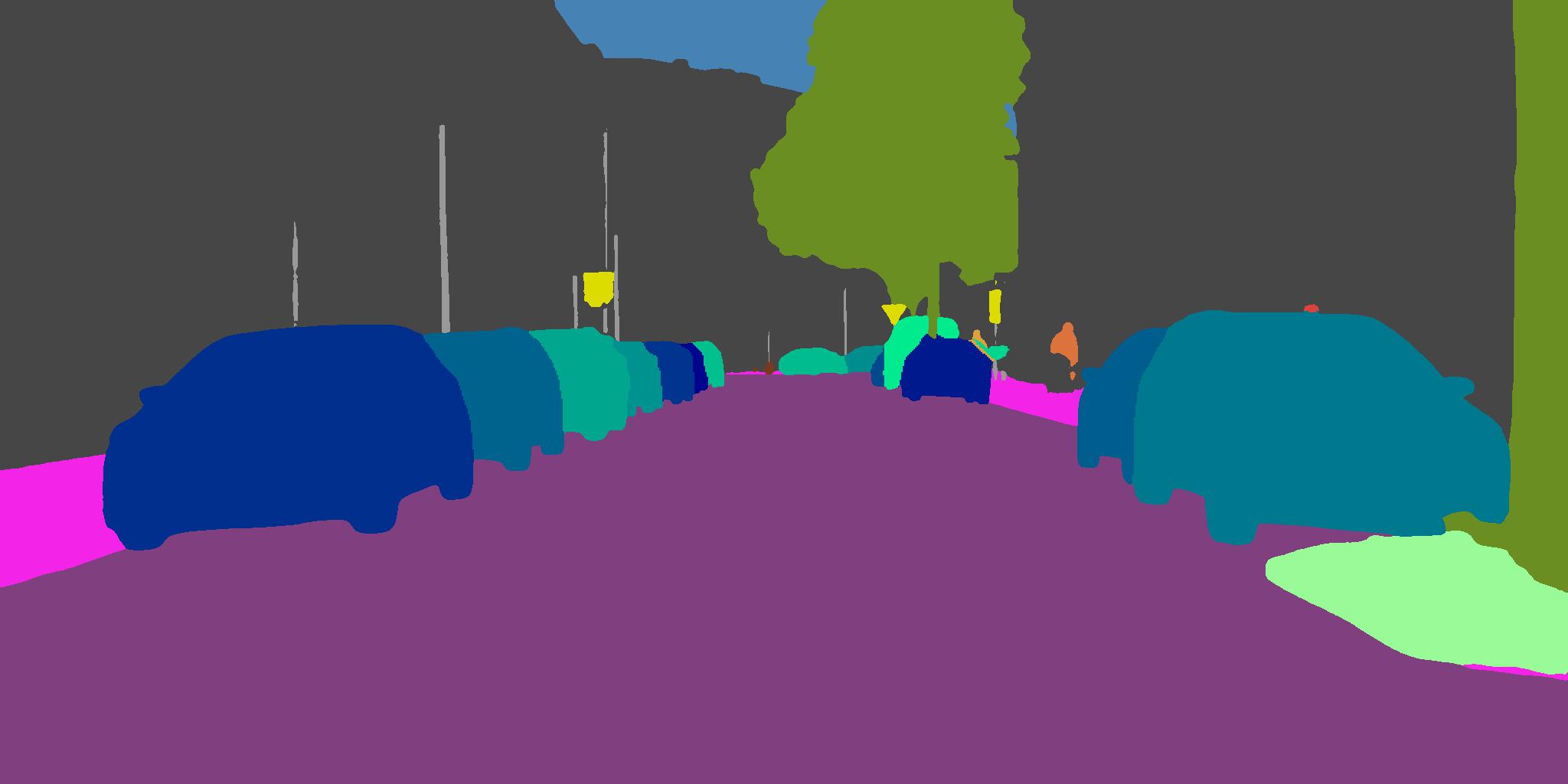}
\end{subfigure}\hfill
\begin{subfigure}{.245\linewidth}
  \centering
  \includegraphics[trim={0 256 0 256},clip,width=\linewidth]{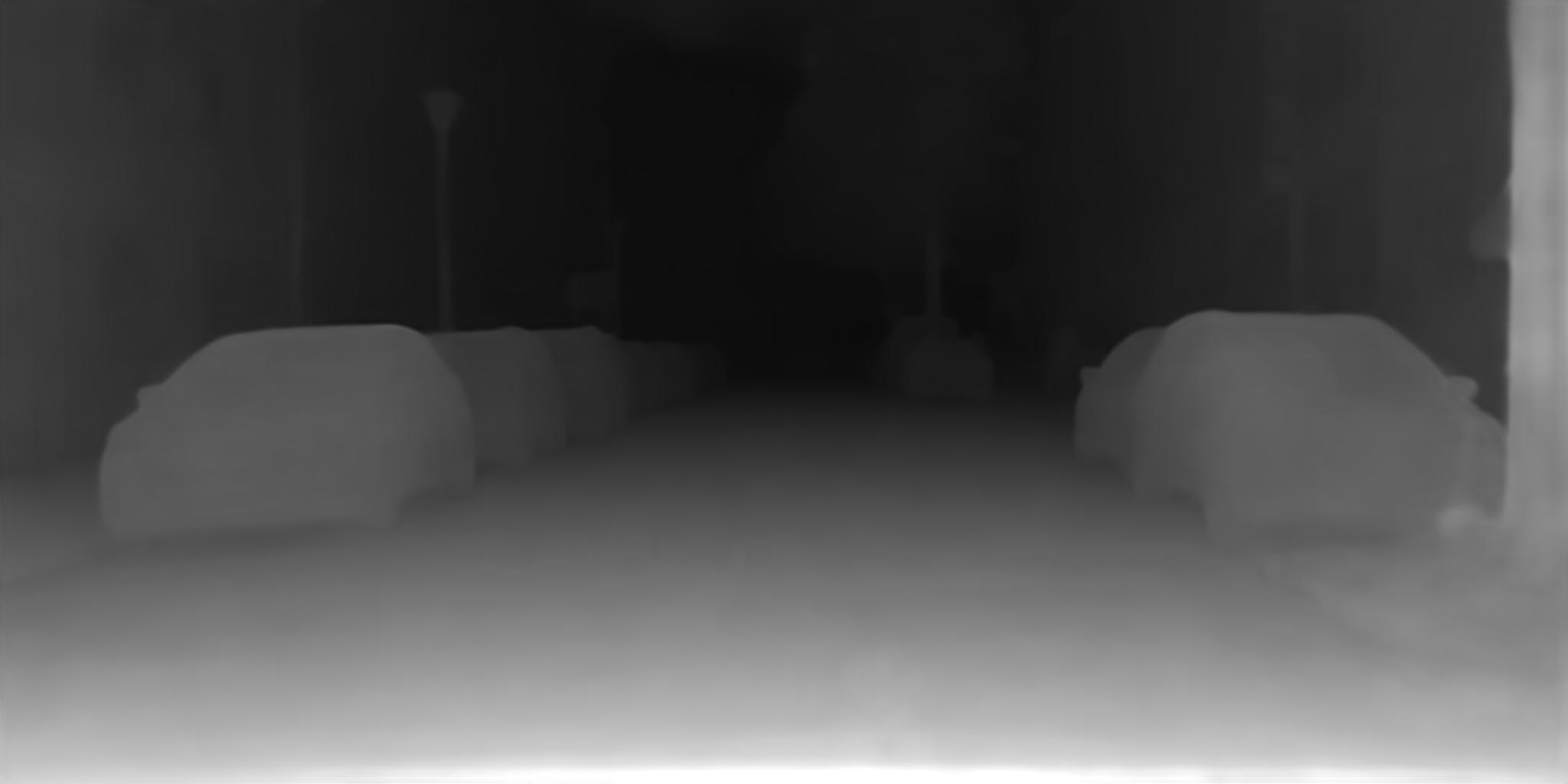}
\end{subfigure}\hfill
\begin{subfigure}{.245\linewidth}
  \centering
  \includegraphics[width=\linewidth]{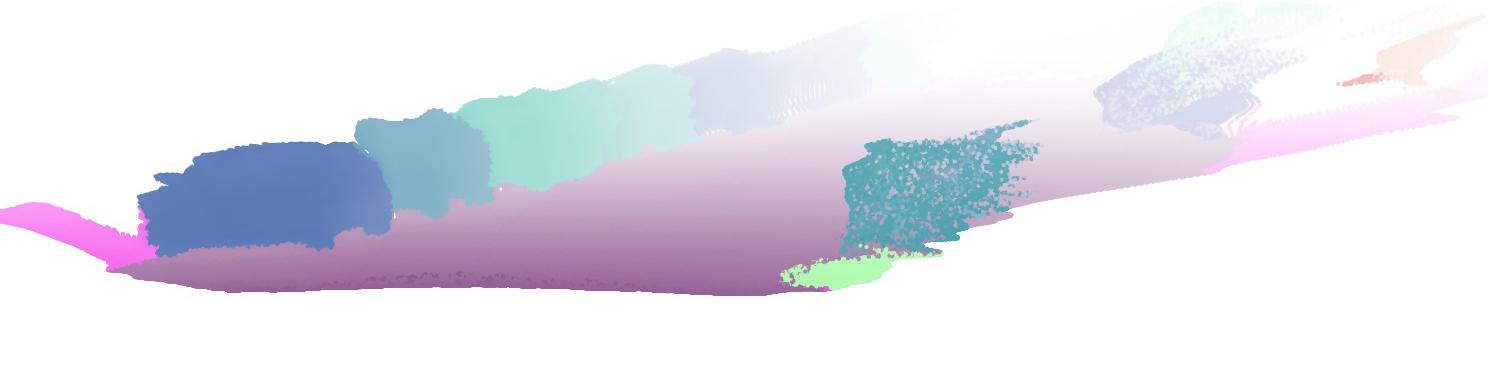}
\end{subfigure}\\[0.25em]
\begin{subfigure}{.245\linewidth}
  \centering
  \includegraphics[width=\linewidth]{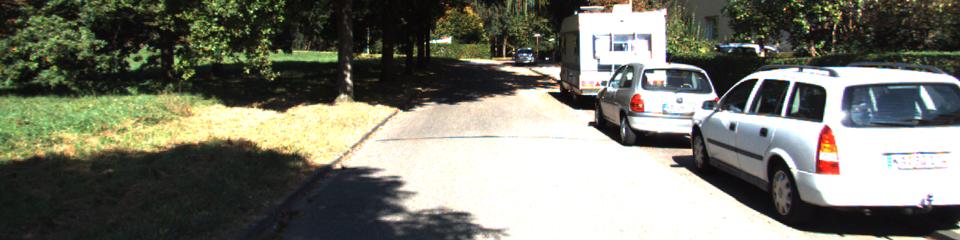}
\end{subfigure}\hfill
\begin{subfigure}{.245\linewidth}
  \centering
  \includegraphics[width=\linewidth]{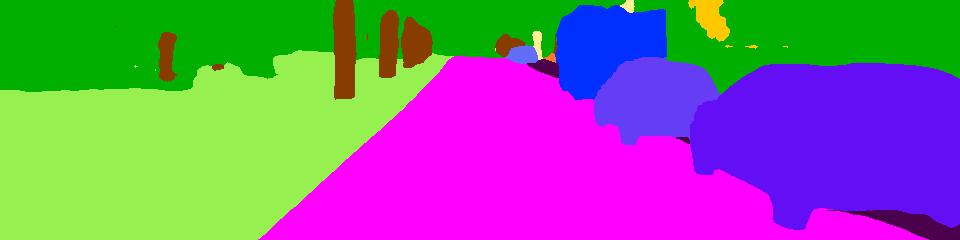}
\end{subfigure}\hfill
\begin{subfigure}{.245\linewidth}
  \centering
  \includegraphics[width=\linewidth]{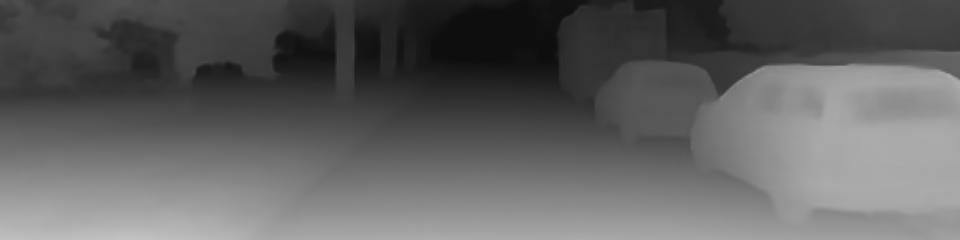}
\end{subfigure}\hfill
\begin{subfigure}{.245\linewidth}
  \centering
  \includegraphics[width=\linewidth]{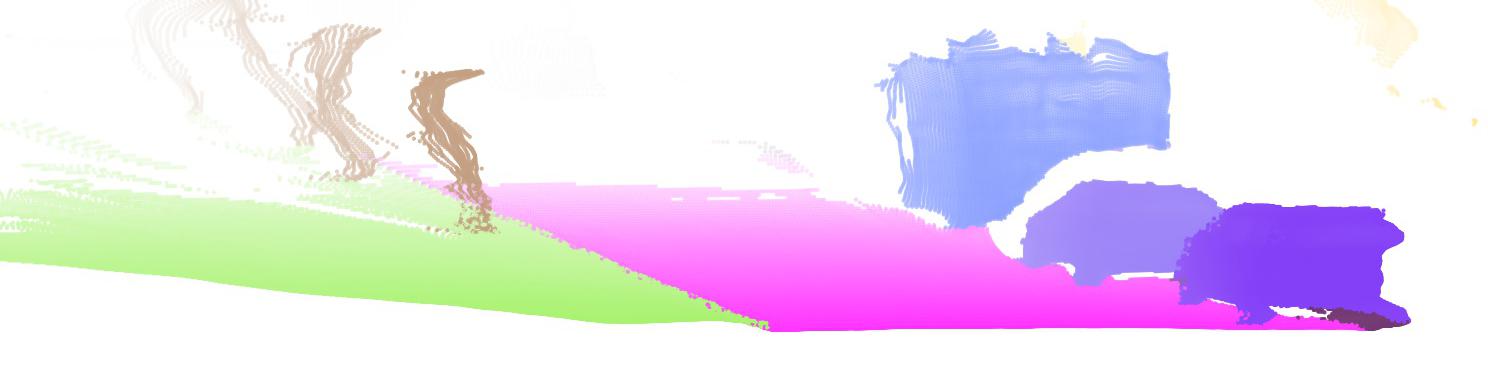}
\end{subfigure}\\
\begin{subfigure}{.245\linewidth}
  \centering
  \includegraphics[width=\linewidth]{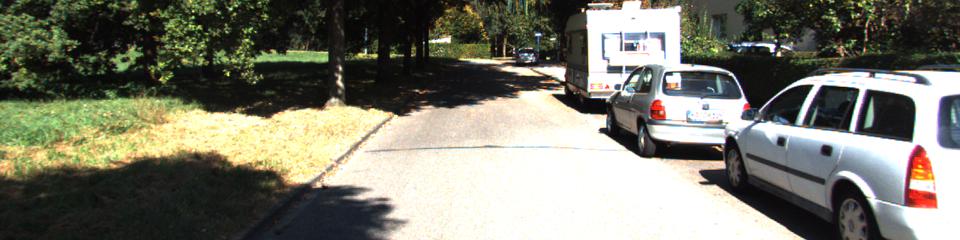}
\end{subfigure}\hfill
\begin{subfigure}{.245\linewidth}
  \centering
  \includegraphics[width=\linewidth]{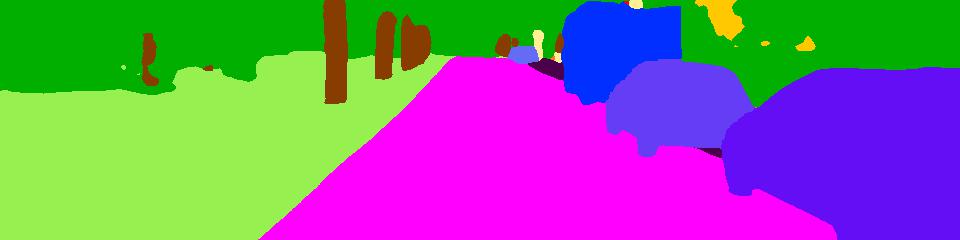}
\end{subfigure}\hfill
\begin{subfigure}{.245\linewidth}
  \centering
  \includegraphics[width=\linewidth]{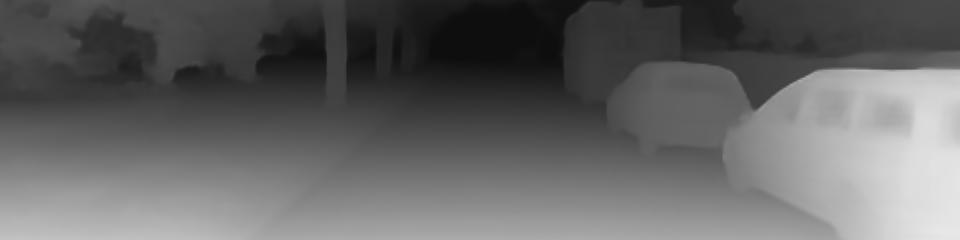}
\end{subfigure}\hfill
\begin{subfigure}{.245\linewidth}
  \centering
  \includegraphics[width=\linewidth]{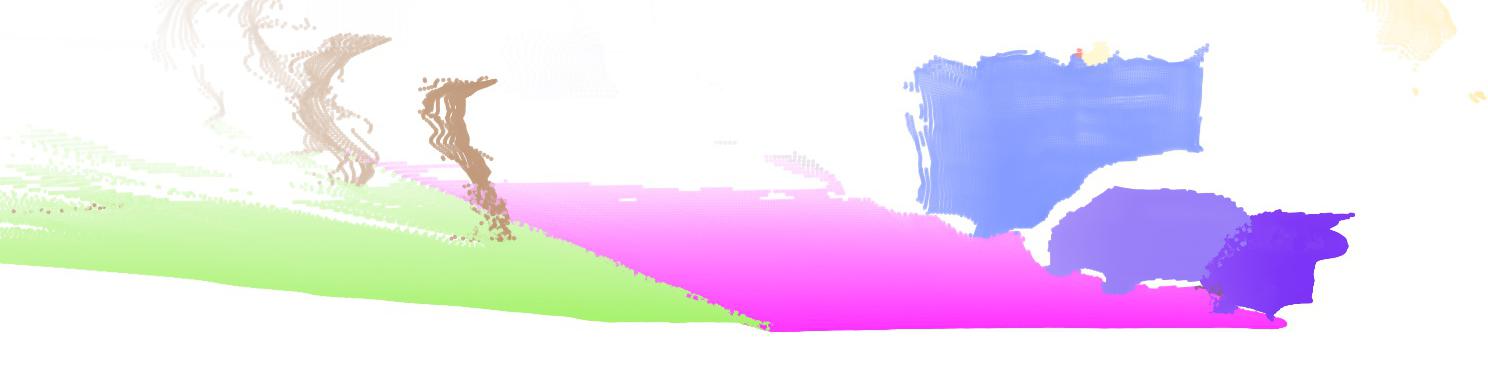}
\end{subfigure}
\caption{Prediction visualizations on Cityscapes-DVPS (top) and SemKITTI-DVPS (bottom). From left to right: input image, temporally consistent panoptic segmentation prediction, monocular depth prediction, and point cloud visualization.}
\label{fig:visualization}
\end{figure*}

Following~\cite{kim2020video}, we set $k=\{1, 2, 3, 4\}$ out of the total 6 frames per video sequence for Cityscapes-DVPS. By contrast, we set $k=\{1, 5, 10, 20\}$ for SemKITTI-DVPS which contains much longer video sequences, and we aim to evaluate a longer temporal consistency.
We study the drops of $\text{DVPQ}_{\lambda}^{k}$ as the number of frames $k$ increases, where smaller performance drops indicate higher temporal consistency.
Interestingly, as the number of frames $k$ increases, the performance drops on SemKITTI-DVPS are smaller than that on Cityscapes-DVPS.
For example, DVPQ$_{0.5}^1$ - DVPQ$_{0.5}^2$ on Cityscapes-DVPS is $7\%$, while DVPQ$_{0.5}^1$ - DVPQ$_{0.5}^5$ on SemKITTI-DVPS is $3.2\%$.
We speculate that this is because the annotation frame rate is higher on SemKITTI-DVPS (\cf only every 5th frame is annotated on Cityscapes-DVPS), making our ViP-DeepLab's offsets prediction easier for the following frames, despite the evaluation clip length $k$ is larger.
At last, we use the mean of $\text{DVPS}_{\lambda}^k$ with different ${\lambda}$ and $k$ as the final performance score.
\figref{fig:visualization} visualizes the predictions of our method on the validation set of Cityscapes-DVPS (top) and SemKITTI-DVPS (bottom), where the second column shows $P_t$ and $R_t$ defined in prediction stitching.
Although the training samples of SemKITTI-DVPS are sparse points, our method is able to predict smooth and sharp predictions, as the points are evenly distributed in the regions covered by the velodyne data.
Please see the supplementary material for more visualization results.
After experimenting on DVPS, we compare ViP-DeepLab with the previous state-of-the-arts on the sub-tasks to showcase its strong performance.

\subsection{Video Panoptic Segmentation}

\begin{table}[!t]
\renewcommand{\arraystretch}{0.9}
\small
    \centering
    \begin{tabular}{l|c|c}
        \toprule[0.2em]
        Val set & VPSNet~\cite{kim2020video} & ViP-DeepLab \\
        \toprule[0.2em]
        k = 1~~~~ & 65.0 $\vert$ 59.0 $\vert$ 69.4 & 70.4 $\vert$ 63.2 $\vert$ 75.7 \\
        k = 2 & 57.6 $\vert$ 45.1 $\vert$ 66.7 & 63.6 $\vert$ 50.7 $\vert$ 73.0 \\
        k = 3 & 54.4 $\vert$ 39.2 $\vert$ 65.6 & 60.1 $\vert$ 44.0 $\vert$ 71.9 \\
        k = 4 & 52.8 $\vert$ 35.8 $\vert$ 65.3 & 58.1 $\vert$ 40.2 $\vert$ 71.2 \\
        \midrule
        VPQ & 57.5 $\vert$ 44.8 $\vert$ 66.7 & \textbf{63.1 $\vert$ 49.5 $\vert$ 73.0} \\
        \bottomrule[0.1em]
        \toprule[0.2em]
        Test set & VPSNet~\cite{kim2020video} & ViP-DeepLab \\
        \toprule[0.2em]
        k = 1 & 64.2 $\vert$ 59.0 $\vert$ 67.7 & 68.9 $\vert$ 61.6 $\vert$ 73.5 \\
        k = 2 & 57.9 $\vert$ 46.5 $\vert$ 65.1 & 62.9 $\vert$ 51.0 $\vert$ 70.5 \\
        k = 3 & 54.8 $\vert$ 41.1 $\vert$ 63.4 & 59.9 $\vert$ 46.0 $\vert$ 68.8 \\
        k = 4 & 52.6 $\vert$ 36.5 $\vert$ 62.9 & 58.2 $\vert$ 42.1 $\vert$ 68.4 \\
        \midrule
        VPQ & 57.4 $\vert$ 45.8 $\vert$ 64.8 & \textbf{62.5 $\vert$ 50.2 $\vert$ 70.3} \\
        \bottomrule[0.1em]
    \end{tabular}
    \caption{VPQ on Cityscapes-VPS.
    Each cell shows $\text{VPQ}^{k} ~\vert~ \text{VPQ}^{k}\text{-Thing} ~\vert~ \text{VPQ}^{k}\text{-Stuff}$. VPQ is averaged over $k=\{1, 2, 3, 4\}$.
    $k=\{0, 5, 10, 15\}$ in~\cite{kim2020video} correspond to $k=\{1, 2, 3, 4\}$ in this paper as we use different notations.
    }
    \label{tab:vps}
\end{table}

\begin{table}[!t]
\renewcommand{\arraystretch}{0.9}
\small
    \centering
    \begin{tabular}{l|ccccc}
    \toprule[0.2em]
    Method &  k = 1 & k = 2 & k = 3 & k = 4 & VPQ \\
    \toprule[0.2em]
    Baseline & 65.7 & 58.9 & 55.8 & 53.6 & 58.5 \\
    + MV & 66.7 & 59.3 & 56.1 & 54.1 & 59.0 \\
    + CS & 67.9 & 60.4 & 56.8 & 54.7 & 59.9 \\
    + DenseContext & 68.2 & 61.3 & 58.2 & 56.1 & 60.9\\
    + AutoAug~\cite{cubuk2018autoaugment} & 68.6 & 61.6 & 58.6 & 56.3 & 61.3 \\
    + RFP~\cite{qiao2020detectors} & 69.2 & 62.3 & 59.2 & 57.0 & 61.9 \\
    + TTA & 70.3 & 63.2 & 59.9 & 57.5 & 62.7 \\
    + SSL & 70.4 & 63.6 & 60.1 & 58.1 & 63.1 \\
    \bottomrule[0.1em]
    \end{tabular}
    \caption{Ablation Study on Cityscapes-VPS.}
    \label{tab:abl}
\end{table}

The first sub-task of DVPS is Video Panoptic Segmentation (VPS).
We conduct experiments on Cityscapes-VPS following the setting of \cite{kim2020video}.
\tabref{tab:vps} shows our major results on their validation set (top) and the test set where the test set annotations are not available to the public (bottom).
As the table shows, our method outperforms VPSNet~\cite{kim2020video} by 5.6\% VPQ on the validation set and 5.1\% VPQ on the test set.

\tabref{tab:abl} shows the ablation study on Cityscapes-VPS.
The baseline is our method with backbone WR-41~\cite{zagoruyko2016wide,wu2019wider,chen2020semi} pre-trained on ImageNet~\cite{russakovsky2015imagenet}.
Next, `MV' initializes the model with a checkpoint pretrained on Mapillary Vistas~\cite{neuhold2017mapillary}.
`CS' uses a model further pretrained on Cityscapes videos with pseudo labels~\cite{chen2020semi} on the \textit{train} sequence.
Both `MV' and `CS' only involve image panoptic segmentation pretraining.
Hence, they mainly improves image PQ (\ie $k=1$)
but increases the gaps between VPQ$^k$ (\eg, VPQ$^1$ - VPQ$^2$), showing the temporal consistency benefits less from the pretrained models.
Then, `DenseContext' increases the number of the context modules (from 1 to 4) for the next-frame instance branch, which narrows down the gaps between VPQ$^k$.
`AutoAug' uses AutoAugment~\cite{cubuk2018autoaugment} to augment the data.
`RFP' adds Recursive Feature Pyramid (RFP)~\cite{qiao2020detectors} to enhance the backbone.
`TTA' stands for test-time augmentation, which includes multi-scale inference at scales 0.5:1.75:0.25 and horizontal flipping.
In `SSL', we follow Naive-Student~\cite{chen2020semi} to generate temporally consistent pseudo labels on the unlabeled \textit{train} sequence in Cityscapes videos~\cite{cordts2016cityscapes}, which adds more training samples for temporal consistency, as demonstrated by +0.1\% on VPQ$^1$ and +0.6\% on VPQ$^{4}$.

\subsection{Monocular Depth Estimation}

\begin{table}[!t]
\renewcommand{\arraystretch}{0.9}
\small
    \centering
    \begin{tabular}{l|c|cccc}
    \toprule[0.2em]
    Method & Rank & SILog & sqRel & absRel & iRMSE \\
    \toprule[0.2em]
    DORN~\cite{fu2018deep} & 10 & 11.77 & 2.23 & 8.78 & 12.98 \\
    BTS~\cite{lee2019big} & 9 & 11.67 & 2.21 & 9.04 & 12.23 \\
    BANet~\cite{aich2020bidirectional} & 8 & 11.61 & 2.29 & 9.38 & 12.23 \\
    MPSD~\cite{antequera2020mapillary} & 2 & 11.12 & 2.07 & 8.99 & 11.56 \\
    \midrule
    ViP-DeepLab & 1 & 10.80 & 2.19 & 8.94 & 11.77 \\
    \bottomrule[0.1em]
    \end{tabular}
    \caption{KITTI Depth Prediction Leaderboard. Ranking includes published and unpublished methods.}
    \label{tab:kitti_depth}
\end{table}

The second sub-task of DVPS is monocular depth estimation.
We test our method on the KITTI depth benchmark~\cite{uhrig2017sparsity}.
\tabref{tab:kitti_depth} shows the results on the leaderboard.
Our model is pretrained on Mapillary Vistas~\cite{neuhold2017mapillary} and Cityscapes videos with pseudo labels~\cite{chen2020semi} (\ie, the same pretrained checkpoint we used in the previous experiments).
Then the model is fine-tuned with the training and validation set provided by KITTI depth benchmark~\cite{uhrig2017sparsity}.
However, the model is slightly different from the previous ones in the following aspects.
It does not use RFP~\cite{qiao2020detectors}.
In TTA, it only has horizontal flipping.
We use $\pm5$ degrees of random rotation during training,
which improves SILog by 0.27.
The previous models use a decoder with stride 8 and 4.
Here, we find it useful to further exploit decoder stride 2,
which improves SILog by 0.17.
After the above changes, our method achieves the best results on KITTI depth benchmark.

\begin{table}[!t]
\small
\setlength{\tabcolsep}{0.16em}
    \centering
    \scalebox{0.92}{
    \begin{tabular}{l|c|cc|c|cc}
    \toprule[0.2em]
    & \multicolumn{3}{c|}{Pedestrians} & \multicolumn{3}{c}{Cars}  \\
    \cline{2-7}
    Method & {\footnotesize Rank} & {\footnotesize sMOTSA} & {\footnotesize MOTSA} &  {\footnotesize Rank} & {\footnotesize sMOTSA} & {\footnotesize MOTSA} \\
    \toprule[0.2em]
    TrackR-CNN~\cite{voigtlaender2019mots} & 20 & 47.3 & 66.1 & 19 & 67.0 & 79.6 \\
    MOTSFusion~\cite{luiten2020track} & 13 & 58.7 & 72.9 & 12 & 75.0 & 84.1 \\
    PointTrack~\cite{xu2020segment} & 11 & 61.5 & 76.5 & 5 & 78.5 & 90.9 \\
    ReMOTS~\cite{yangremots} & 6 & 66.0 & 81.3 & 9 & 75.9 & 86.7 \\
    \midrule
    ViP-DeepLab &  & 67.7 & 83.4 &  & 80.6 & 90.3 \\
    ViP-DeepLab + KF & 1 & 68.7 & 84.5 & 3 & 81.0 & 90.7 \\
    \bottomrule[0.1em]
    \end{tabular}
    }
    \caption{KITTI MOTS Leaderboard. Ranking includes published and unpublished methods.
    }
    \label{tab:kitti_mots}
\end{table}

\subsection{Multi-Object Tracking and Segmentation}
Finally, we evaluate our method on the KITTI MOTS benchmark~\cite{voigtlaender2019mots}.
\tabref{tab:kitti_mots} shows the leaderboard results.
Different from the previous experiments, this benchmark only tracks pedestrians and cars.
Adopting the same strategy as we used for Cityscapes-VPS, ViP-DeepLab outperforms all the published methods and achieves 67.7\% and 80.6\% sMOTSA for pedestrians and cars, respectively.
To further improve our results, we use Kalman filter (KF)~\cite{wang2020towards}
to re-localize missing objects that are occluded or detection failures.
This mechanism improves the sMOTSA by 1.0\% and 0.4\% for pedestrians and cars, respectively.
\section{Conclusion}

In this paper, we propose a new challenging task Depth-aware Video Panoptic Segmentation, which combines monocular depth estimation and video panoptic segmentation, as a step towards solving the inverse projection problem in vision.
For this task, we propose Depth-aware Video Panoptic Quality as the evaluation metric along with two derived datasets.
We present ViP-DeepLab as a strong baseline for this task.
Additionally, our ViP-DeepLab also achieves state-of-the-art performances on several sub-tasks, including monocular depth estimation, video panoptic segmentation, and multi-object tracking and segmentation.

\subsubsection*{Acknowledgments}
We would like to thank Maxwell Collins for the feedbacks, and the support from Google Mobile Vision team.

{\small
\bibliographystyle{ieee_fullname}
\bibliography{egbib}
}

\clearpage

\appendix

\section{Stitching Algorithm}

\begin{algorithm}[t]
\SetAlgoLined
\SetKwInOut{Input}{Input}
\SetKwInOut{Output}{Output}
\Input{Panoptic prediction $P_t$ for image $t$,
and panoptic prediction $R_t$ for image $t+1$ when concatenated with image $t$.
}
\Output{In-place stitched panoptic predictions $P_t$.}
 \For{$t=1, 2, ..., T-1$}{
  Increase all instance IDs of $P_{t+1}$ by the maximum of the instance IDs of  $P_{1}$ to $P_{t}$\;
  // \textit{Find all overlapping region pairs $\mathbb{C}$ between} \\
  // \textit{$R_{t}$ and $P_{t+1}$ with the same classes.} \\
  Let $\mathbb{S}=[(r, p)~|~r\in R_{t}\land p\in P_{t+1}\land r\cap p > 0]$\;
  Let $\mathbb{C}=[(r, p)~|~(r,p)\in\mathbb{S}\land \text{cls}(r) = \text{cls}(p)]$ where $\text{cls}(\cdot)$ denotes the class\;
  Sort $\mathbb{C}=[(r, p)]$ with respect to $\text{IoU}(r, p)$ in ascending order\;
  // \textit{For each region in $R_{t}$, $\mathbb{M}$ stores the region in} \\
  // \textit{$P_{t+1}$ having the largest IoU with it. $\mathbb{N}$ stores} \\
  // \textit{the opposite direction.} \\
  Let $\mathbb{M}$, $\mathbb{N}$ be empty dictionaries\;
  \For{$(r,p) \in \mathbb{C}$}{
    $\mathbb{M}[r] = p$ and $\mathbb{N}[p] = r$\;
  }
  \For{$r\rightarrow p \in \mathbb{M}$}{
    // \textit{Propagate IDs from $R_t$ to $P_{t+1}$ and $R_{t+1}$} \\
    // \textit{if $r$ and $p$ map to each other in $\mathbb{M}$ and $\mathbb{N}$}.\\
    \If{$r=\mathbb{N}[\mathbb{M}[r]]$}{
        \If{$t<T-1$}{
            Assign the region in $R_{t+1}$ that has the ID of $p$ with the ID of $r$\;
        }
        Assign the region of $p$ in $P_{t+1}$ with the instance ID of $r$ in $R_t$\;
    }
  }
 }
 \caption{Stitching Video Panoptic Predictions}
 \label{algo:stitch}
\end{algorithm}

Alg.~\ref{algo:stitch} shows the details of the algorithm to stitch video panoptic predictions to form predictions with consistent IDs throughout the entire sequence.
We split the panoptic prediction of the concatenated image pair $t$ and $t+1$ in the middle, and use $P_t$ and $R_t$ to denote the left and the right prediction, respectively.
This makes $P_t$ the panoptic prediction of image $t$, and $R_t$ the panoptic prediction of image $t+1$ with instance IDs that are consistent with those of $P_t$.
The objective of the algorithm is to propagate IDs from $R_t$ to $P_{t+1}$ so that each object in $P_{t}$ and $P_{t+1}$ will have the same ID.
The ID propagation is based on mask IoU between region pairs.
For each region $r$ in $R_t$, we find the region $p$ in $P_{t+1}$ that has the same class and the largest IoU with it.
We use $\mathbb{M}$ to store this mapping.
Similarly, for each region $p$ in $P_{t+1}$, we also find the region $r$ in $R_{t}$ that has the same class and the largest IoU with it.
We use $\mathbb{N}$ to store this mapping.
If a region $r$ of $R_{t}$ and a region $p$ of $P_{t+1}$ are matched to each other (\ie $\mathbb{M}(r)=p$ and $\mathbb{N}(p)=r$), then we propagate the ID from $r$ to $p$.

\begin{table}[!t]
\small
\setlength{\tabcolsep}{0.16em}
    \centering
    \scalebox{0.92}{
    \begin{tabular}{l|ccc|ccc}
    \toprule[0.2em]
    & \multicolumn{3}{c|}{Pedestrians} & \multicolumn{3}{c}{Cars}  \\
    \cline{2-7}
    Method & {\footnotesize sMOTSA} & {\footnotesize MOTSA} & {\footnotesize IDS} &  {\footnotesize sMOTSA} & {\footnotesize MOTSA} & {\footnotesize IDS} \\
    \toprule[0.2em]
    TrackR-CNN~\cite{voigtlaender2019mots} & 46.8 & 65.1 & 78 & 76.2 & 87.8 & 93 \\
    MOTSNet~\cite{porzi2020learning} & 54.6 & 69.3 & - & 78.1 & 87.2 & - \\
    MOTSFusion~\cite{luiten2020track} & 58.9 & 71.9 & 36 & 82.6 & 90.2 & 51 \\
    PointTrack~\cite{xu2020segment} & 62.4 & 77.3 & 19 & 85.5 & 94.9 & 22  \\
    \midrule
    ViP-DeepLab + KF & 68.3 & 83.2 & 15 & 86.0 & 94.7 & 52 \\
    \bottomrule[0.1em]
    \end{tabular}
    }
    \caption{Results on KITTI MOTS validation set.}
    \label{tab:kitti_mots_val}
\end{table}

\begin{table}[]
\small
\setlength{\tabcolsep}{0.16em}
    \centering
    \begin{tabular}{c|cccc|c|c|c}
    \toprule[0.2em]
    $\mathcal{L}_{\text{depth}}$ weight & k = 1 & k = 2 & k = 3 & k = 4 & VPQ & absRel & DVPQ \\
    \toprule[0.2em]
    0.1 & 68.9 & 61.9 & 58.8 & 56.5 & 61.5 & 9.51 & 51.3 \\
    1.0 & 69.0 & 62.0 & 58.7 & 56.5 & 61.6 & 7.21 & 55.1 \\
    10 & 67.8 & 61.1 & 57.5 & 55.5 & 60.5 & 6.54 & 54.3 \\
    \bottomrule[0.1em]
    \end{tabular}
    \caption{
    ViP-DeepLab trained with different training weights for $\mathcal{L}_{\text{depth}}$ on Cityscapes-DVPS.
    }
    \label{tab:loss_weight}
\end{table}

\begin{table*}[!t]
    \centering
    \small
    \begin{tabular}{l|cccccccc}
    \toprule[0.2em]
    Method & {\footnotesize $\delta < 1.25\uparrow$} & {\footnotesize$\delta < 1.25^2\uparrow$} & {\footnotesize$\delta < 1.25^3\uparrow$} & {\footnotesize absRel$\downarrow$}  & {\footnotesize sqRel$\downarrow$} & {\footnotesize RMSE$\downarrow$} & {\footnotesize RMSElog$\downarrow$} & {\footnotesize SILog$\downarrow$} \\
    \toprule[0.2em]
    \cite{diaz2019soft} & 95.77 & 99.21 & 99.75 & 6.99 & 1.27 & 2.86 & 0.104 & 9.73 \\
    Ours &  96.27 & 99.41 & 99.81 & 5.72 & 0.96 & 2.58 & 0.092 & 8.47 \\
    \bottomrule[0.1em]
    \end{tabular}
    \caption{Results on the official KITTI depth validation set. $\uparrow$: The higher the better. $\downarrow$: The lower the better.
    }
    \label{tab:kitti_depth_val}
\end{table*}

\begin{figure*}
    \centering
    \includegraphics[trim={200 320 150 320}, clip, width=\linewidth]{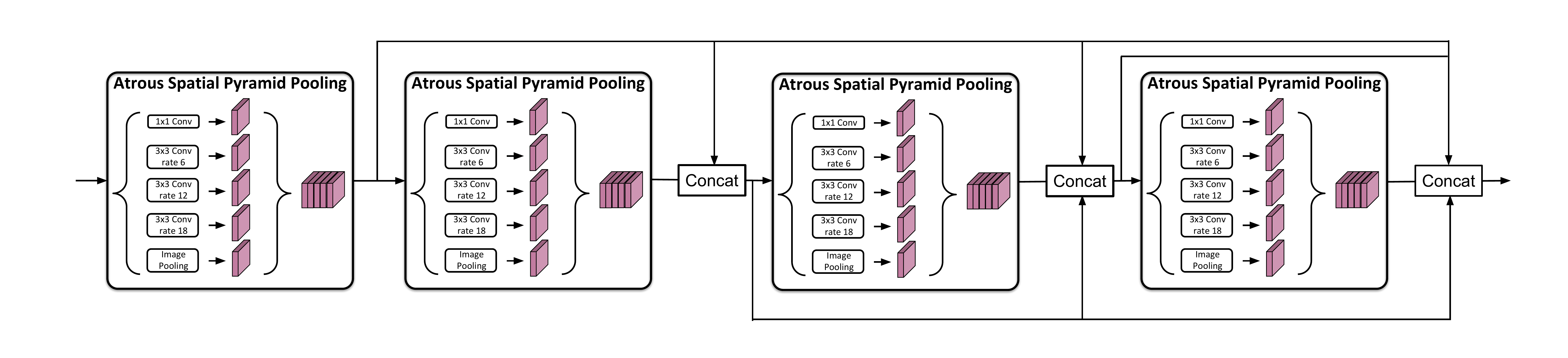}
    \caption{The architecture of Cascade-ASPP, which is employed as \textit{Dense Multi-scale Context} in the next-frame instance branch.
    It cascades four ASPP modules with the outputs densely connected.
    }
    \label{fig:dense}
\end{figure*}

\section{More Experiments}

\paragraph{KITTI MOTS Validation Set}
We first evaluate ViP-DeepLab on the validation set of KITTI MOTS benchmark~\cite{voigtlaender2019mots}.
\tabref{tab:kitti_mots_val} shows the comparisons between ViP-DeepLab and the previous methods.
We adopt the same strategy as we used for training models for KITTI MOTS Leaderboard except that the training data used in here does not include the validation set.
As shown in the table, our method equipped with Kalman filter outperforms the previous methods by a large margin.

\paragraph{Effects of Depth Loss Weight}
Next, we study the effects of different training weights for the depth loss $\mathcal{L}_{\text{depth}}$.
In the previous experiments on Cityscapes-DVPS and SemKITTI-DVPS, we use the depth loss defined by \equref{eq:depth_loss}, which has a loss weight of 1.0.
For the purpose of ablation study, we change the training weight from 1.0 to 10 and 0.1.
The results are shown in \tabref{tab:loss_weight}.
From the table we can see that as the $\mathcal{L}_{\text{depth}}$ weight increases, ViP-DeepLab performs better on the sub-task monocular depth estimation (\ie absRel becomes lower), but worse on the sub-task video panoptic segmentation (\ie VPQ becomes lower).
This is consistent with our intuition that a larger $\mathcal{L}_{\text{depth}}$ weight makes the model focus more on the task of monocular depth estimation.
The metric that matters most here is DVPQ, which unifies the metrics of both sub-tasks.
In order to get a high DVPQ score, the predictions must be accurate on both tasks.
Therefore, finding a balanced $\mathcal{L}_{\text{depth}}$ weight is critical to get a high DVPQ.
As the table shows, setting $\mathcal{L}_{\text{depth}}$ weight to 1.0 achieves the best results among the three choices.

\paragraph{KITTI Depth Validation Set}
Finally, we show the performance of ViP-DeepLab on the official validation set of KITTI depth benchmark~\cite{uhrig2017sparsity}.
The validation set has 1,000 cropped images.
\tabref{tab:kitti_depth_val} compares our method with previous methods that report their performances on it.
Our method outperforms the previous methods by a large margin on all the metrics.

\section{Cascade-ASPP}
\figref{fig:dense} shows the architecture of Cascade-ASPP.
It is used as the module \textit{Dense Multi-scale Context} in the next-frame instance branch shown in \figref{fig:arch}.
It cascades four ASPP modules with their outputs densely connected.
The motivation of Cascade-ASPP is to dramatically increase the receptive field of the next-frame instance branch.
As demonstrated in \tabref{tab:abl}, Cascade-ASPP (\ie DenseContext) improves the performances of video panoptic segmentation on Cityscapes-VPS compared with the single ASPP variant.

\section{More Visualizations}
\begin{figure*}[!t]
\centering
\begin{subfigure}{.245\linewidth}
  \centering
  \includegraphics[trim={0 100 0 100},clip,width=\linewidth]{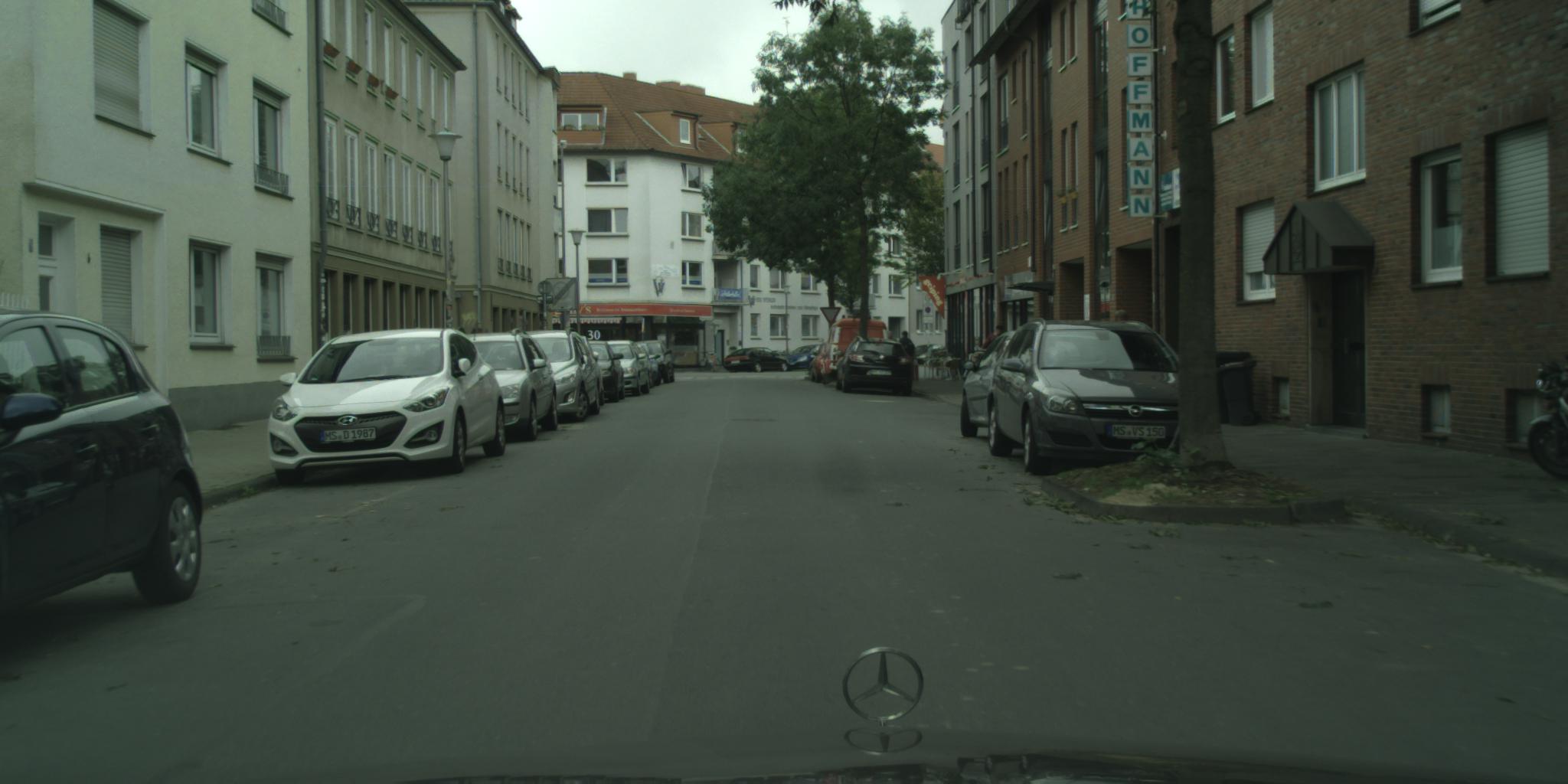}
\end{subfigure}\hfill\begin{subfigure}{.245\linewidth}
  \centering
  \includegraphics[trim={0 100 0 100},clip,width=\linewidth]{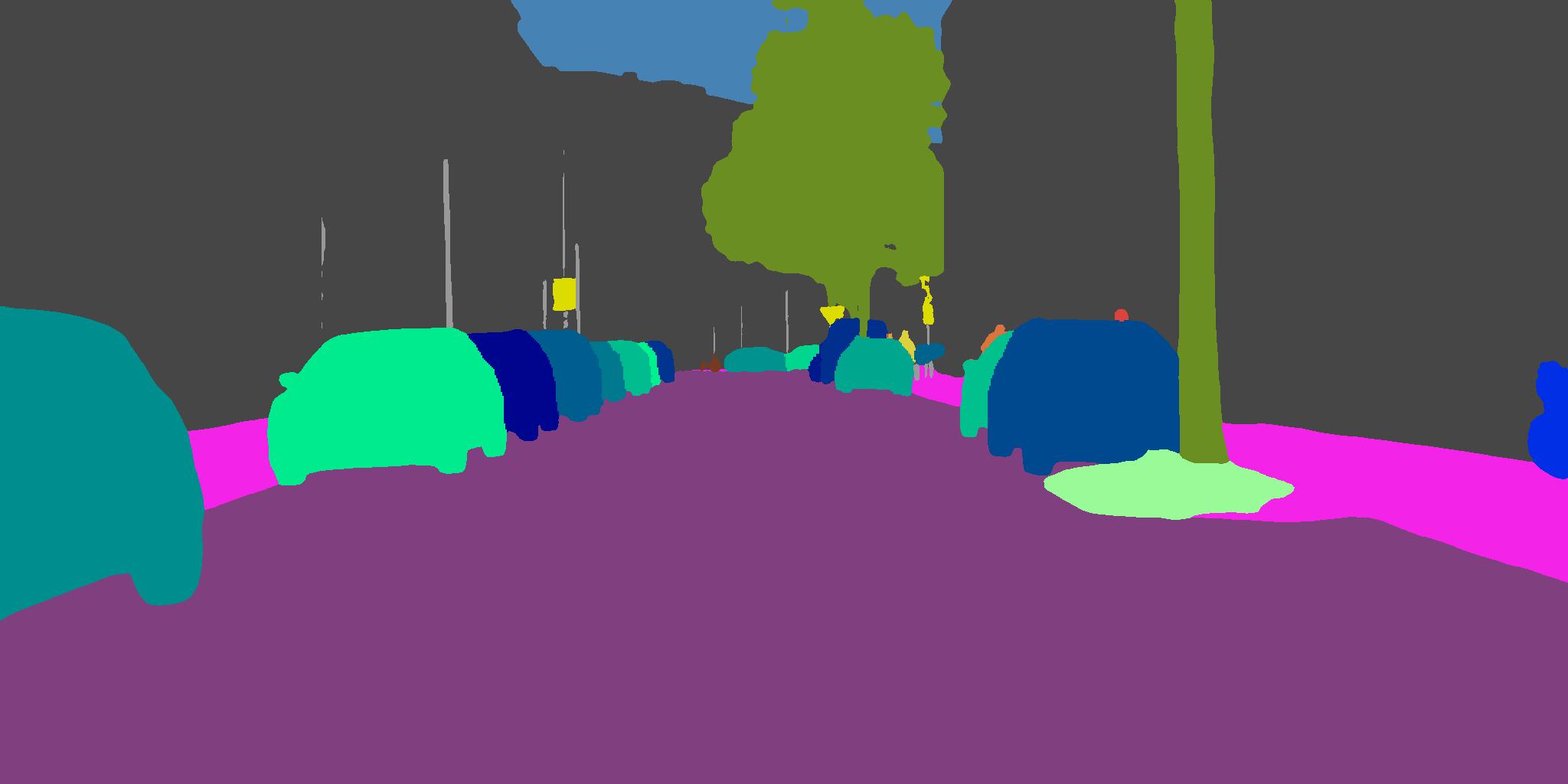}
\end{subfigure}\hfill\begin{subfigure}{.245\linewidth}
  \centering
  \includegraphics[trim={0 100 0 100},clip,width=\linewidth]{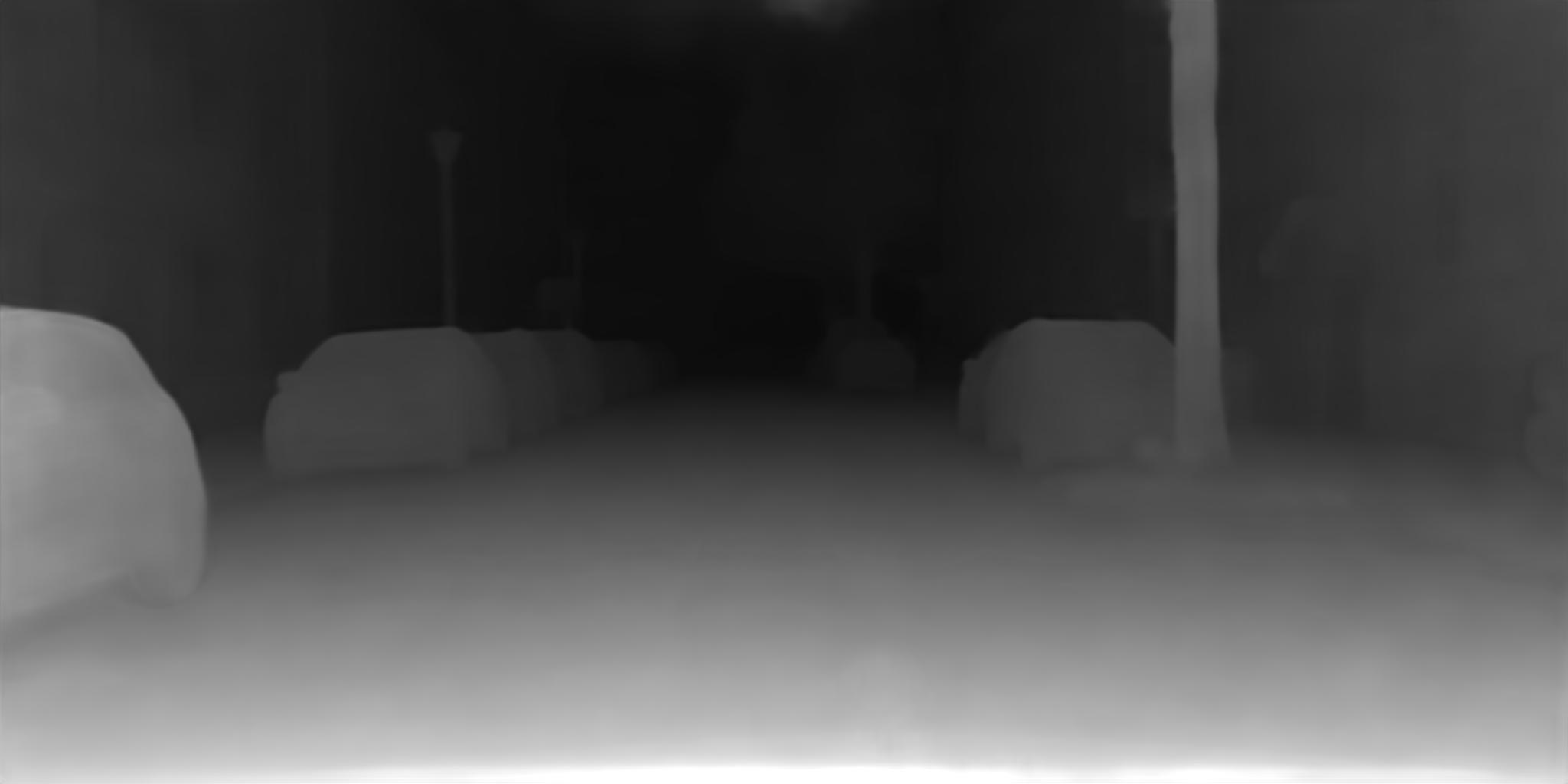}
\end{subfigure}\hfill\begin{subfigure}{.245\linewidth}
  \centering
  \includegraphics[trim={0 100 0 100},clip,width=\linewidth]{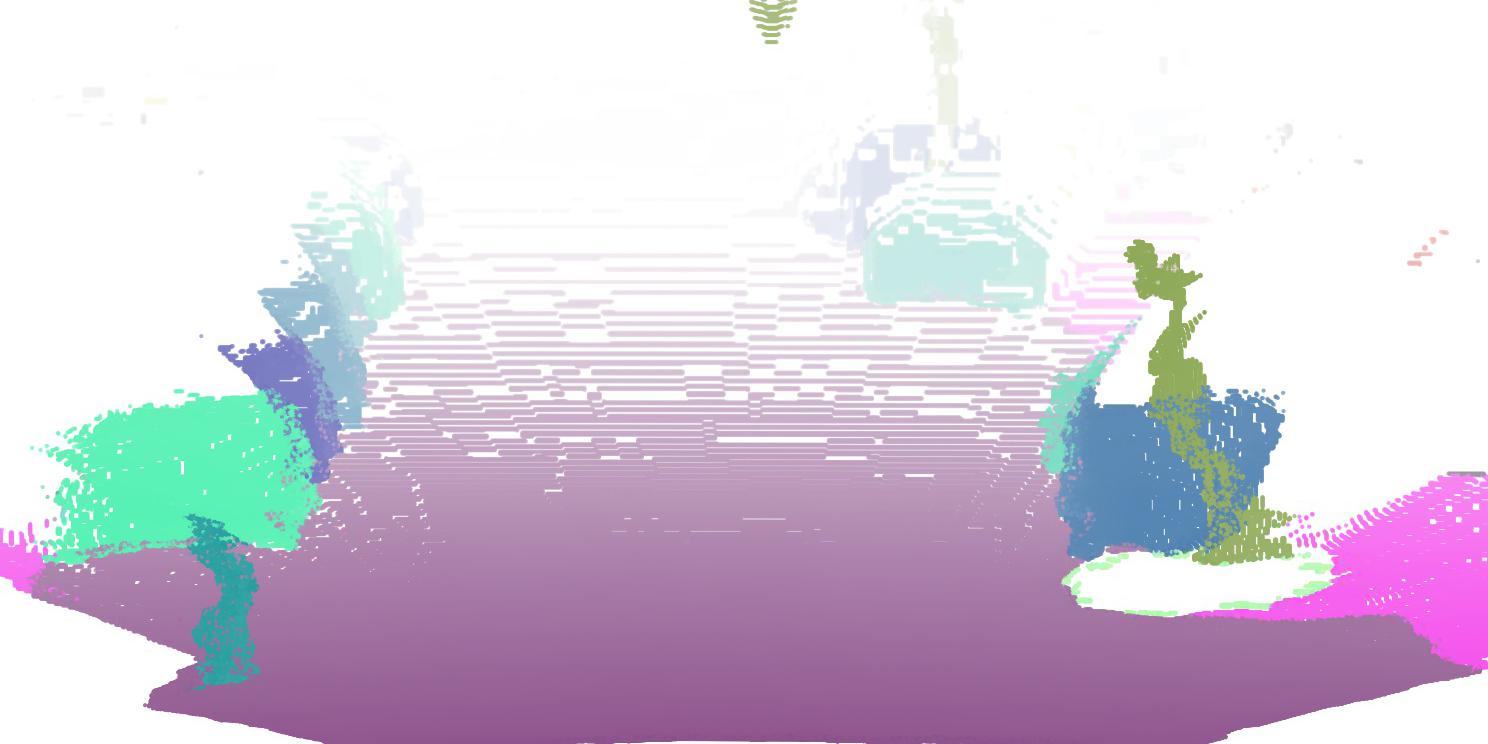}
\end{subfigure}\\\begin{subfigure}{.245\linewidth}
  \centering
  \includegraphics[trim={0 100 0 100},clip,width=\linewidth]{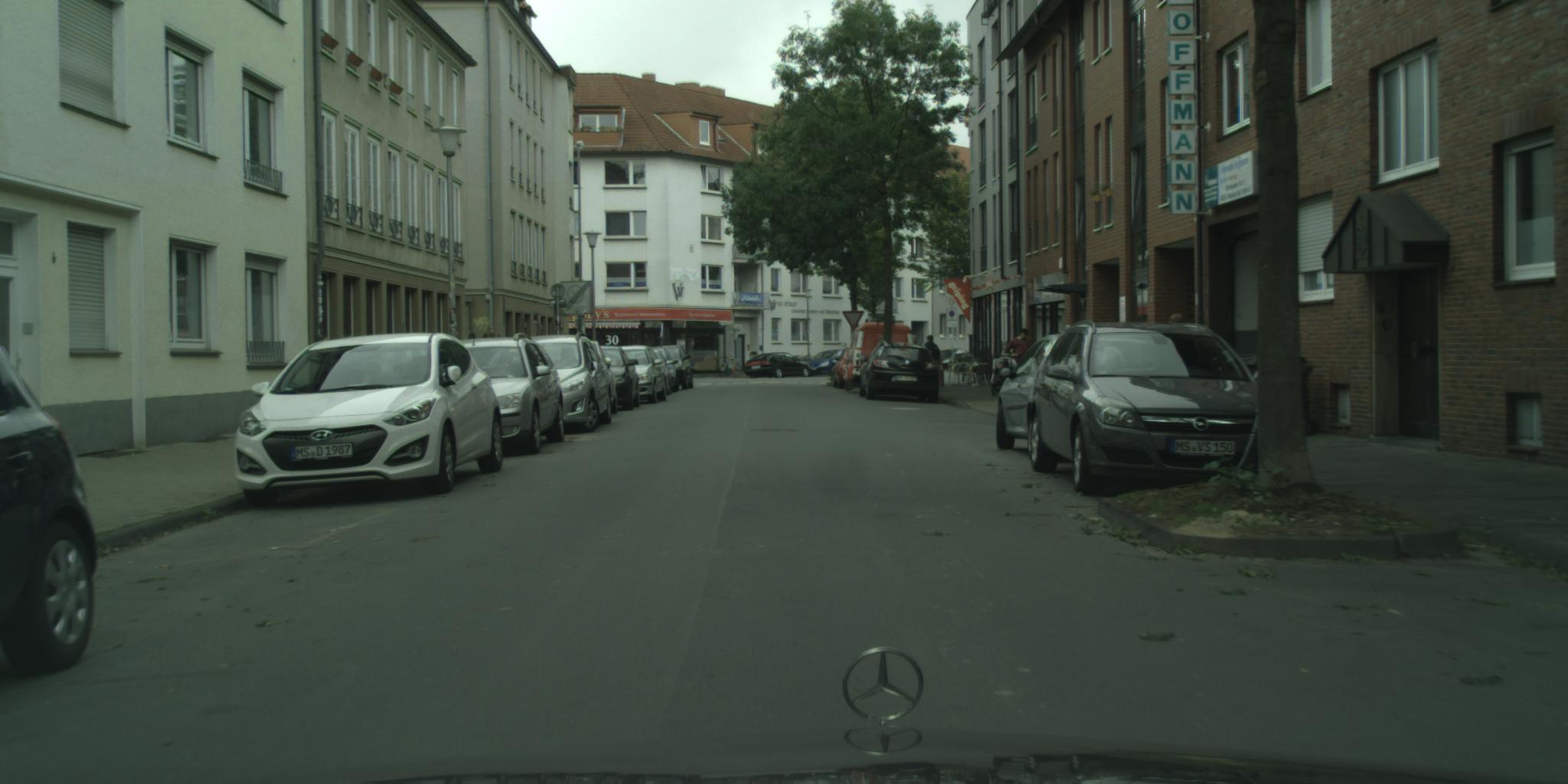}
\end{subfigure}\hfill\begin{subfigure}{.245\linewidth}
  \centering
  \includegraphics[trim={0 100 0 100},clip,width=\linewidth]{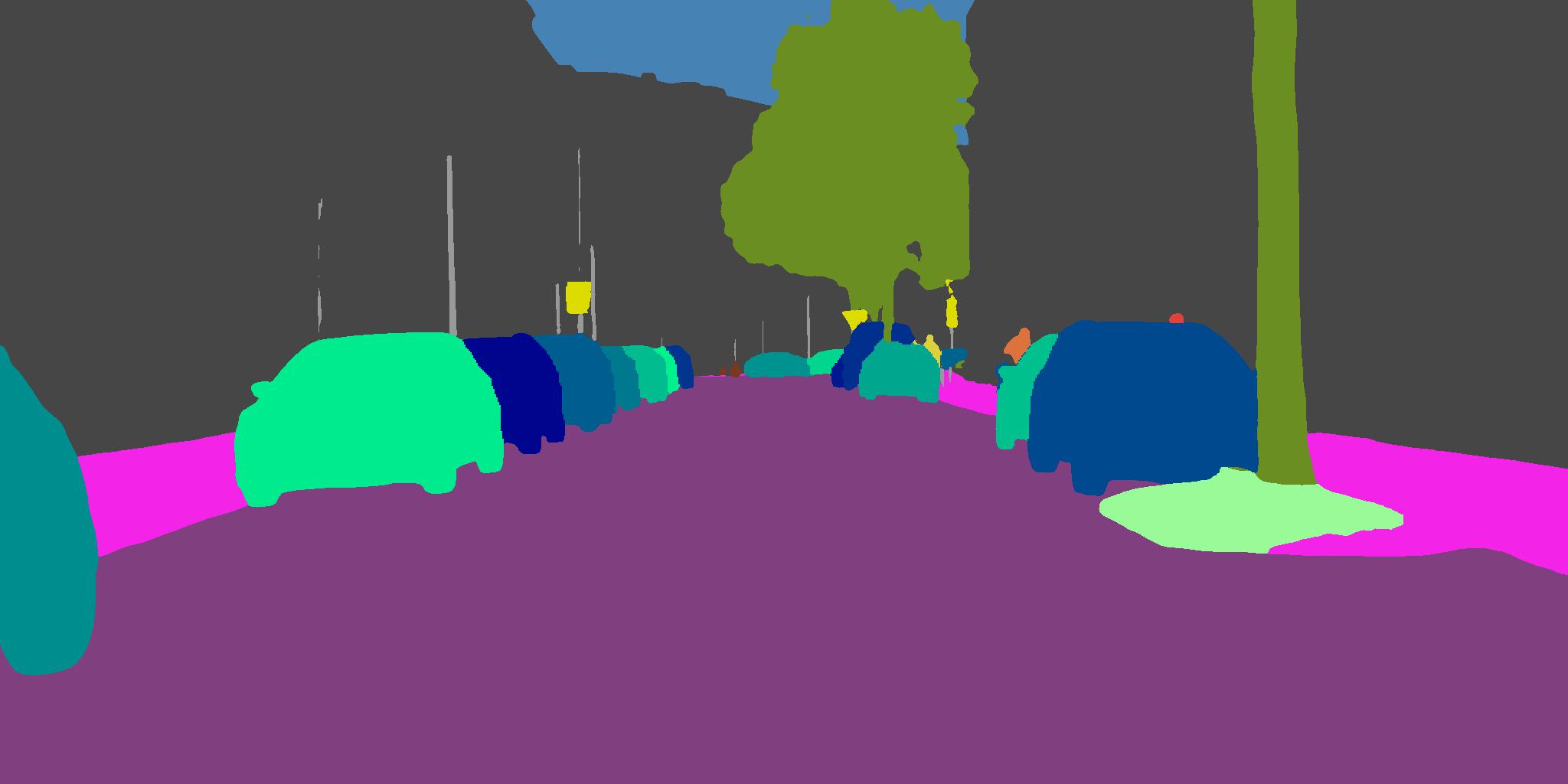}
\end{subfigure}\hfill\begin{subfigure}{.245\linewidth}
  \centering
  \includegraphics[trim={0 100 0 100},clip,width=\linewidth]{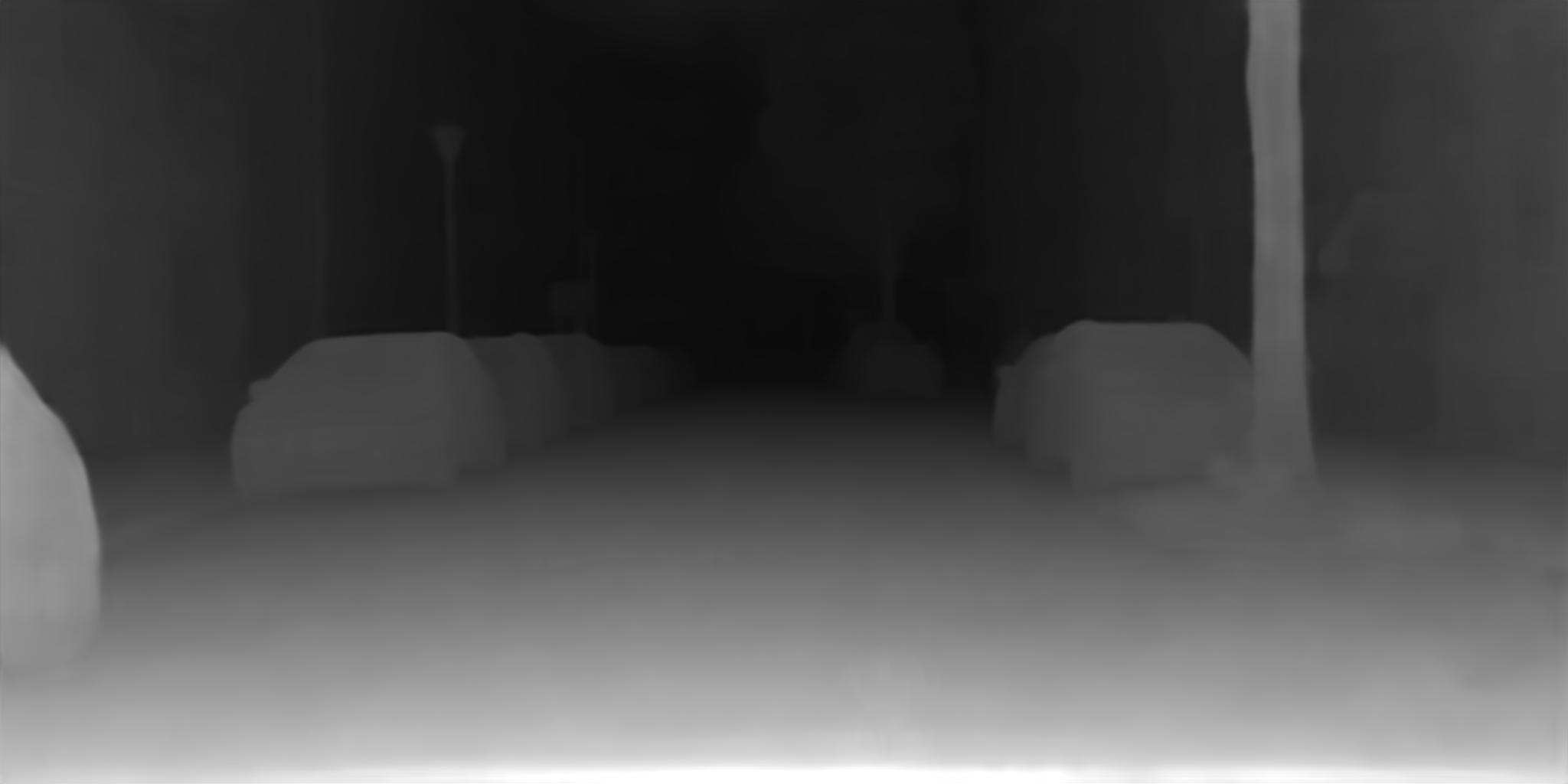}
\end{subfigure}\hfill\begin{subfigure}{.245\linewidth}
  \centering
  \includegraphics[trim={0 100 0 100},clip,width=\linewidth]{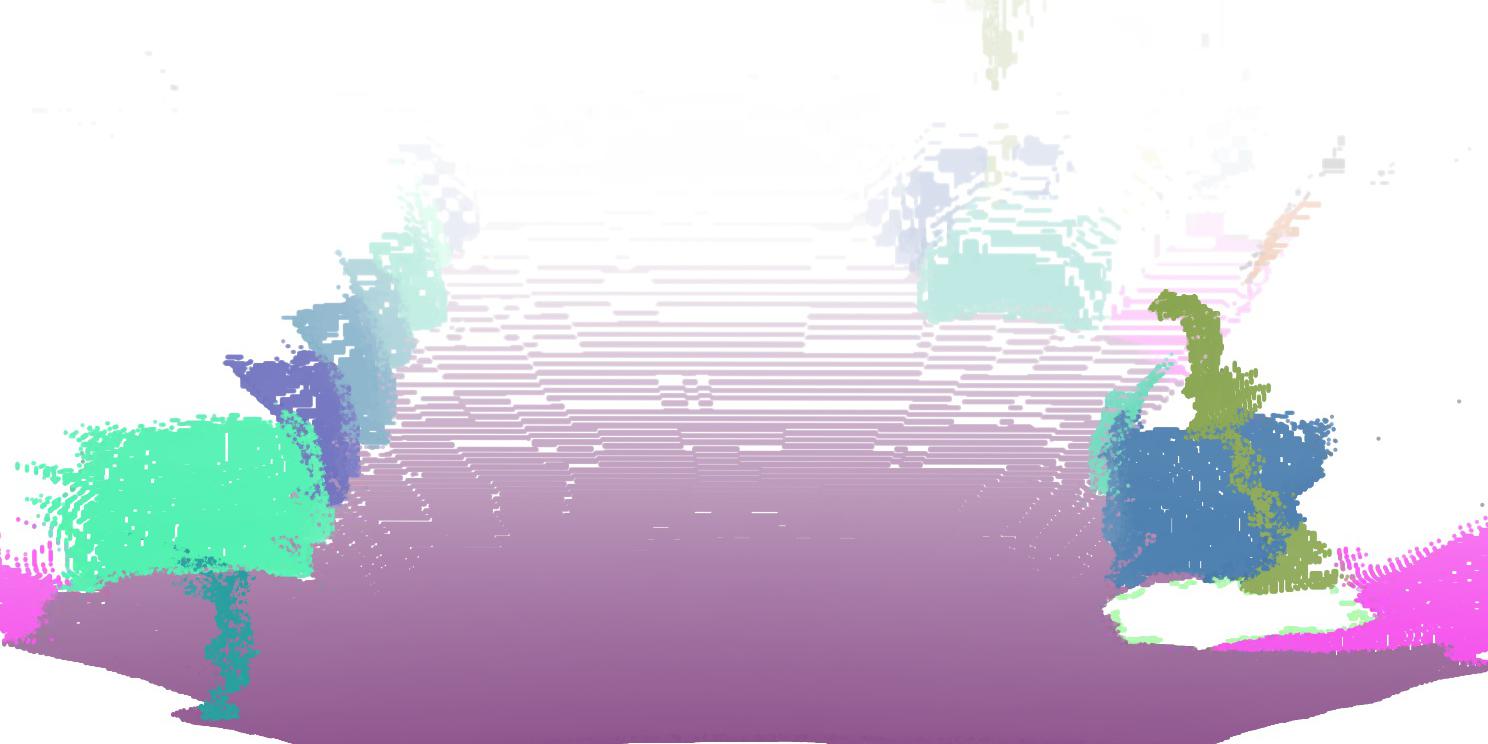}
\end{subfigure}\\\begin{subfigure}{.245\linewidth}
  \centering
  \includegraphics[trim={0 100 0 100},clip,width=\linewidth]{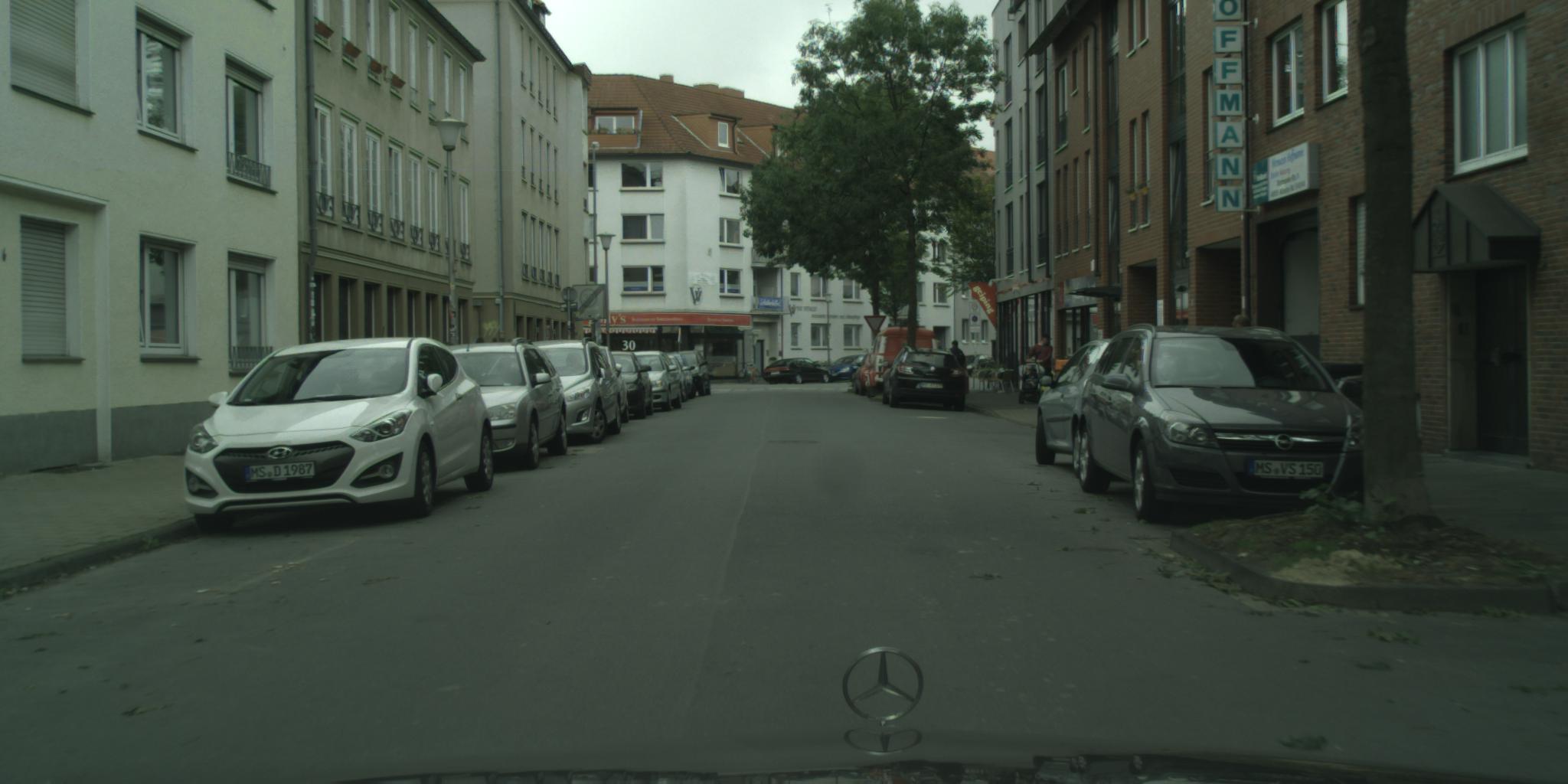}
\end{subfigure}\hfill\begin{subfigure}{.245\linewidth}
  \centering
  \includegraphics[trim={0 100 0 100},clip,width=\linewidth]{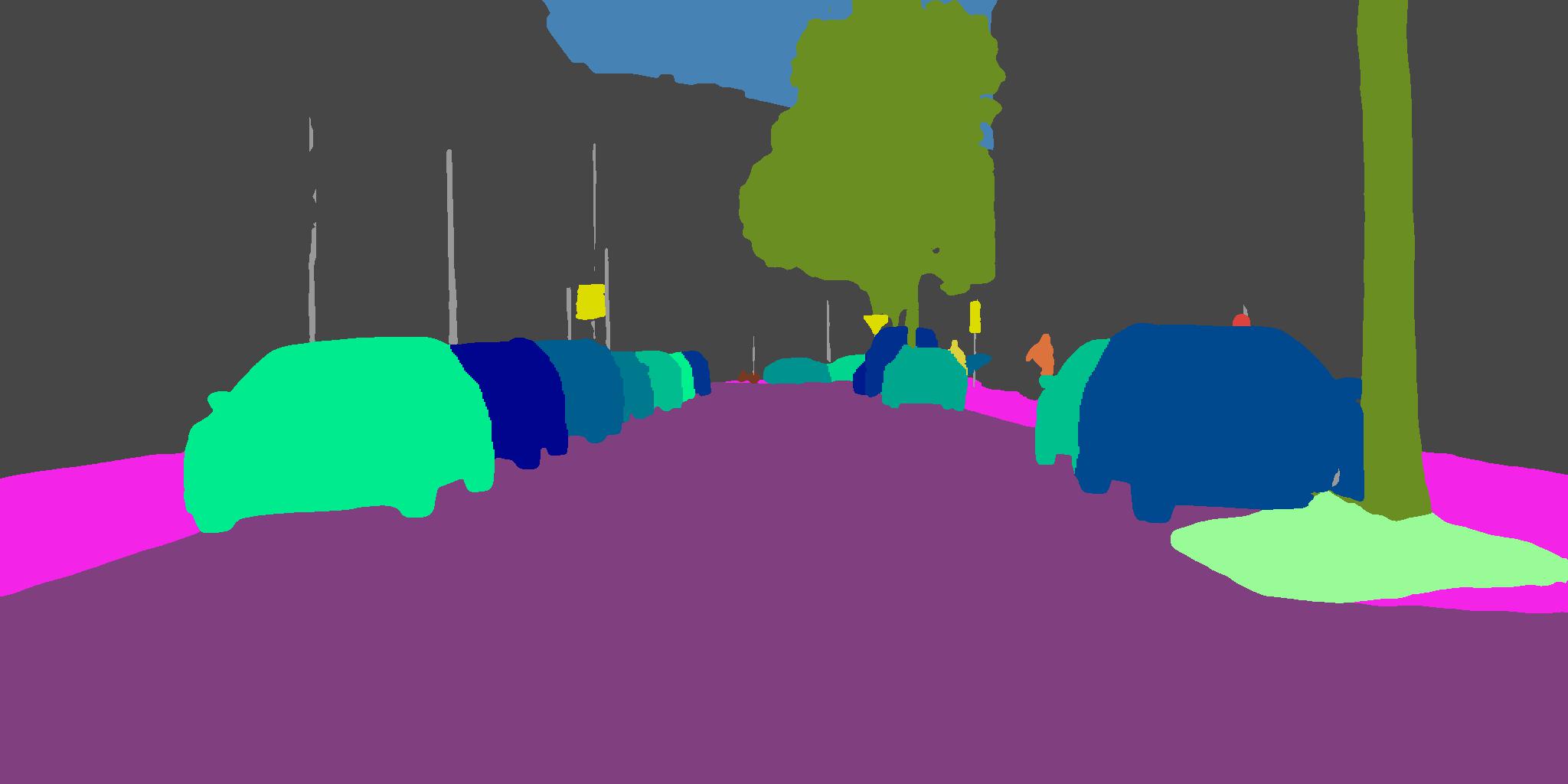}
\end{subfigure}\hfill\begin{subfigure}{.245\linewidth}
  \centering
  \includegraphics[trim={0 100 0 100},clip,width=\linewidth]{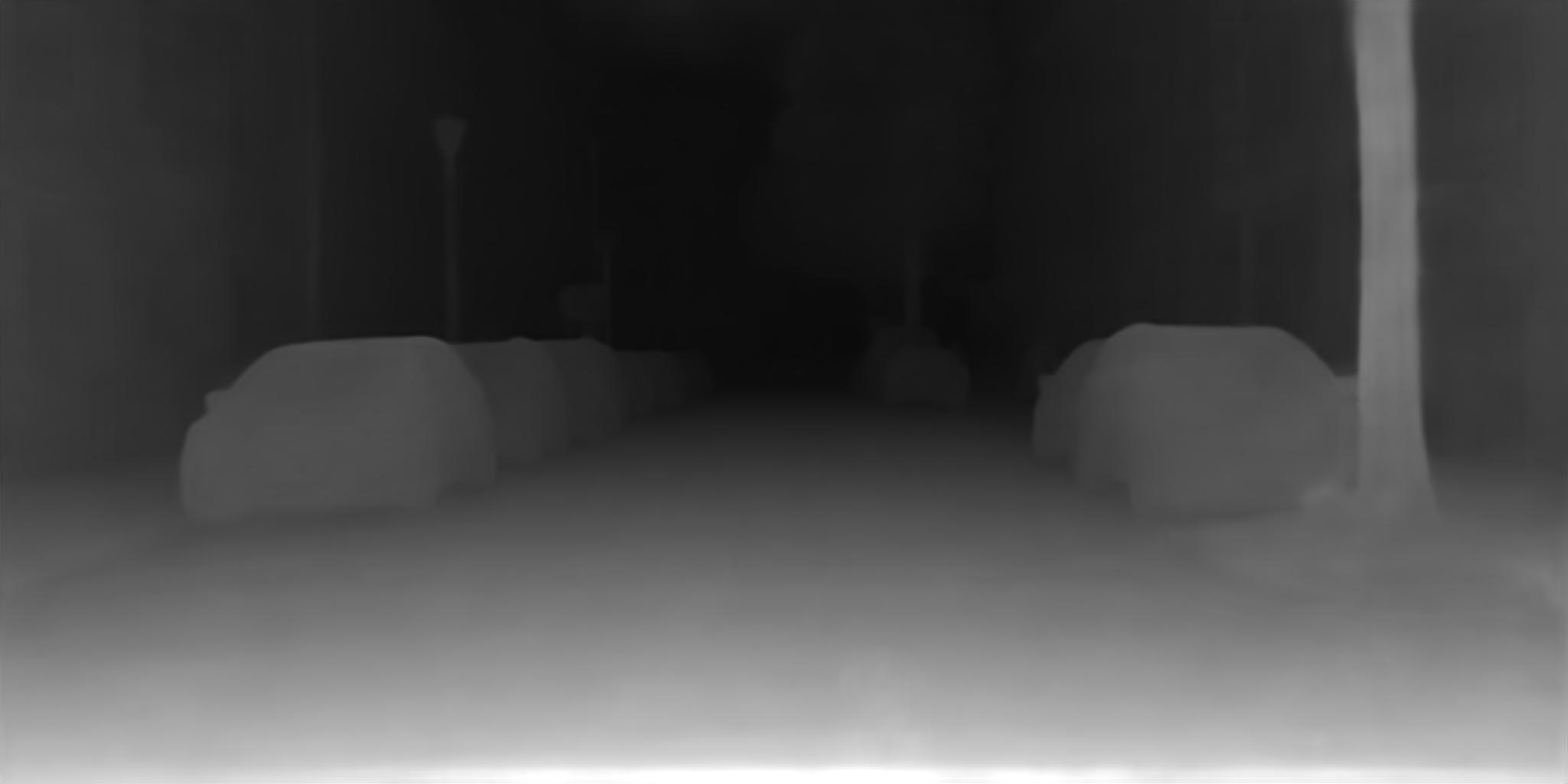}
\end{subfigure}\hfill\begin{subfigure}{.245\linewidth}
  \centering
  \includegraphics[trim={0 100 0 100},clip,width=\linewidth]{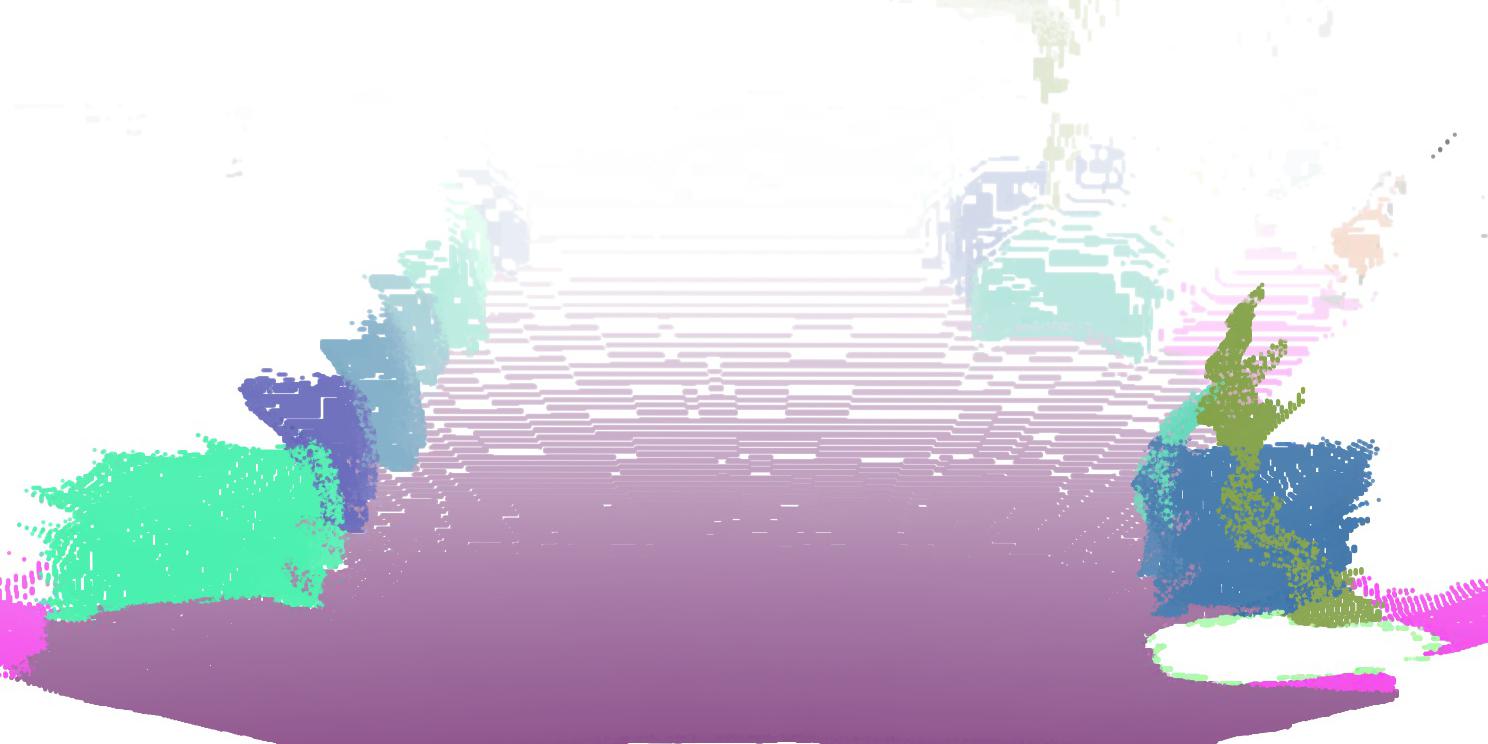}
\end{subfigure}\\\begin{subfigure}{.245\linewidth}
  \centering
  \includegraphics[trim={0 100 0 100},clip,width=\linewidth]{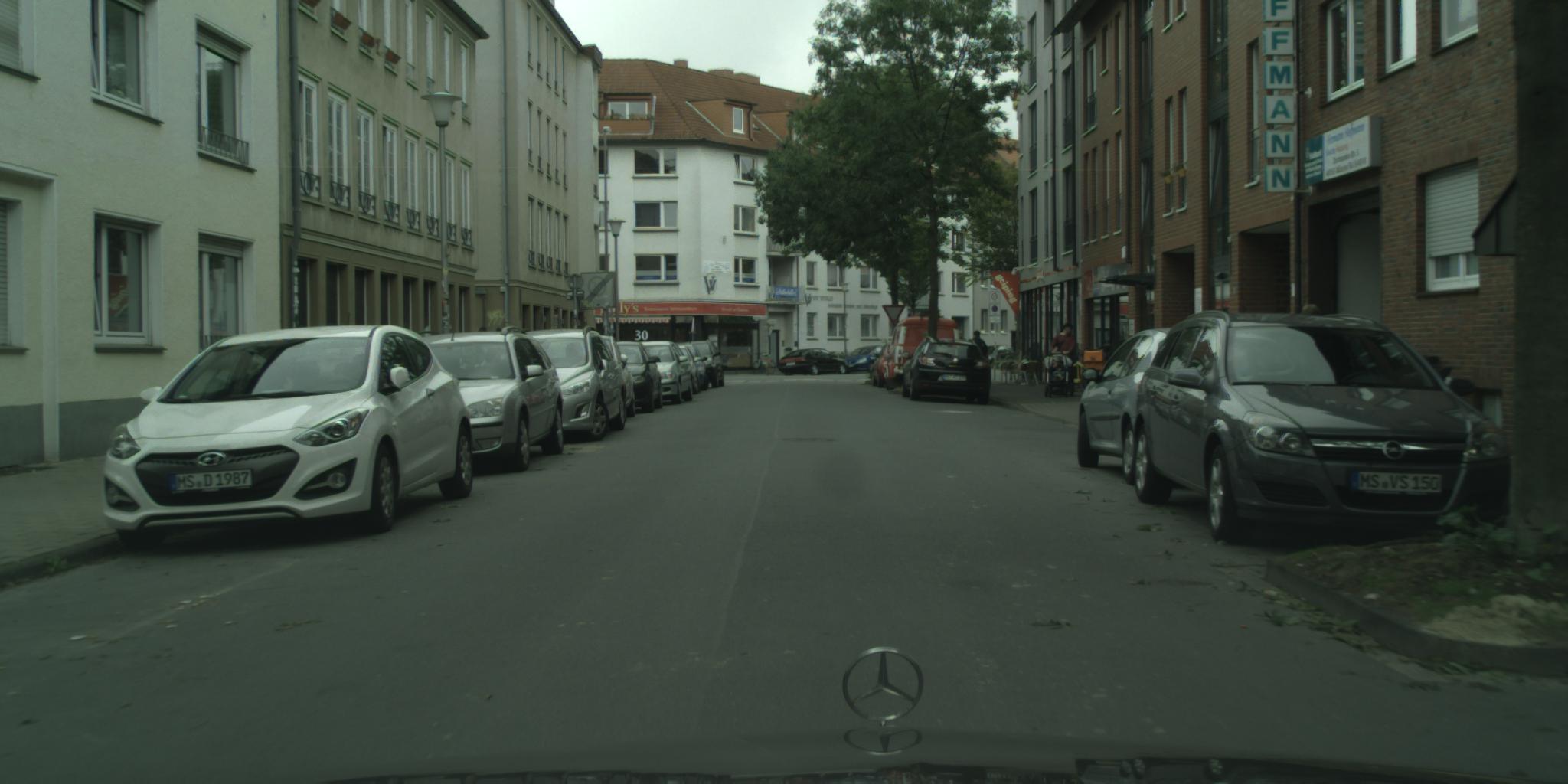}
\end{subfigure}\hfill\begin{subfigure}{.245\linewidth}
  \centering
  \includegraphics[trim={0 100 0 100},clip,width=\linewidth]{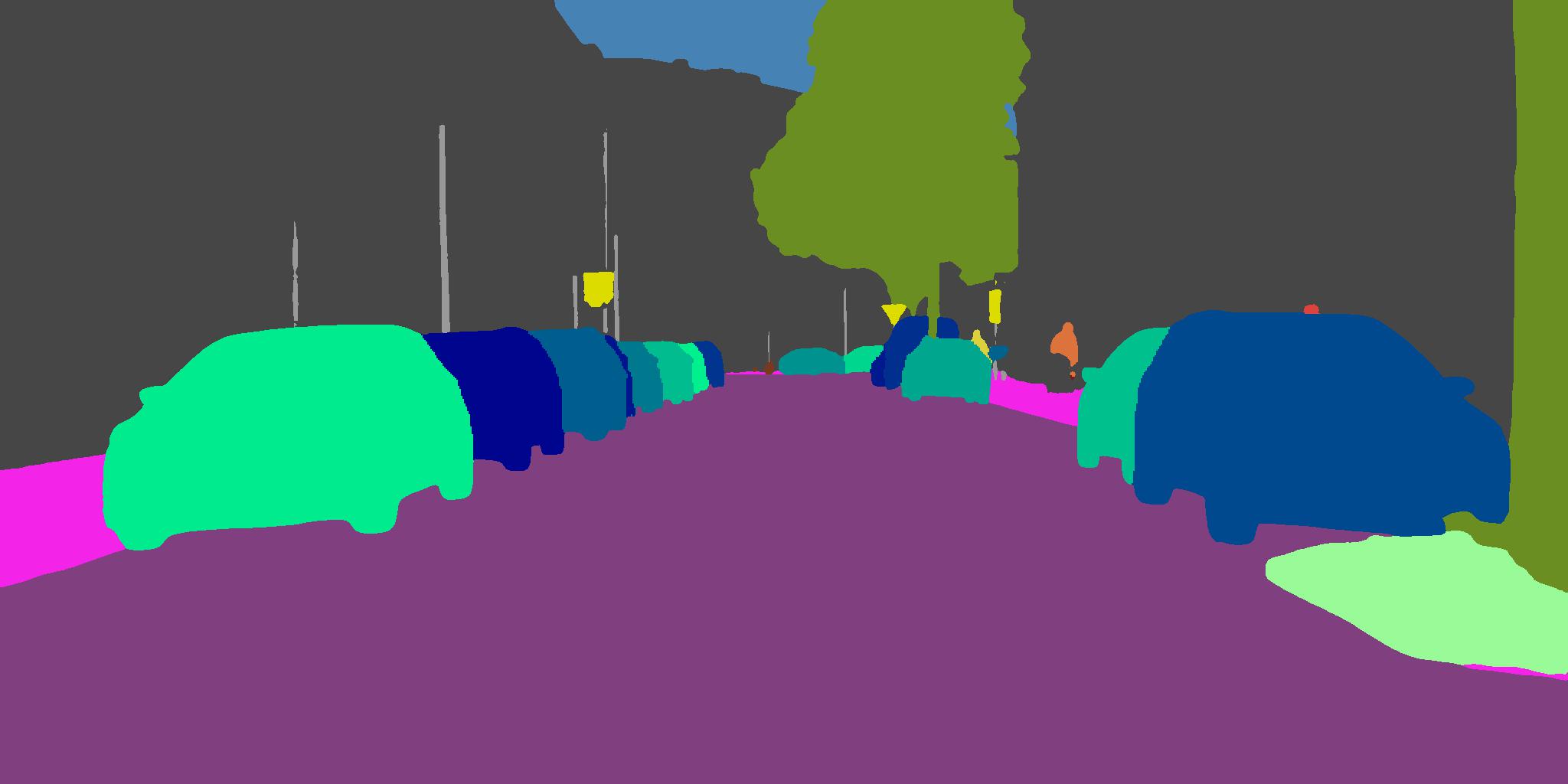}
\end{subfigure}\hfill\begin{subfigure}{.245\linewidth}
  \centering
  \includegraphics[trim={0 100 0 100},clip,width=\linewidth]{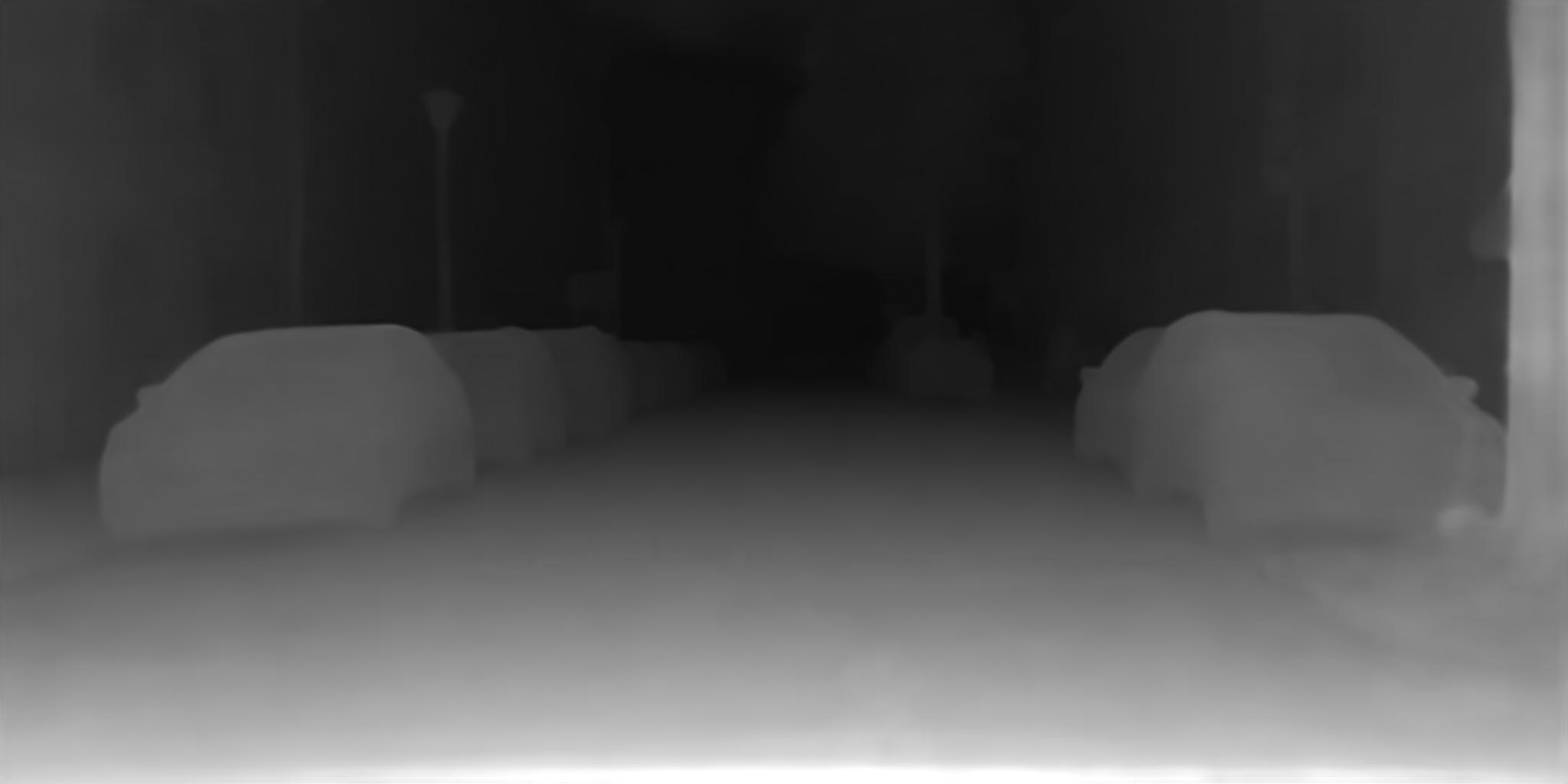}
\end{subfigure}\hfill\begin{subfigure}{.245\linewidth}
  \centering
  \includegraphics[trim={0 100 0 100},clip,width=\linewidth]{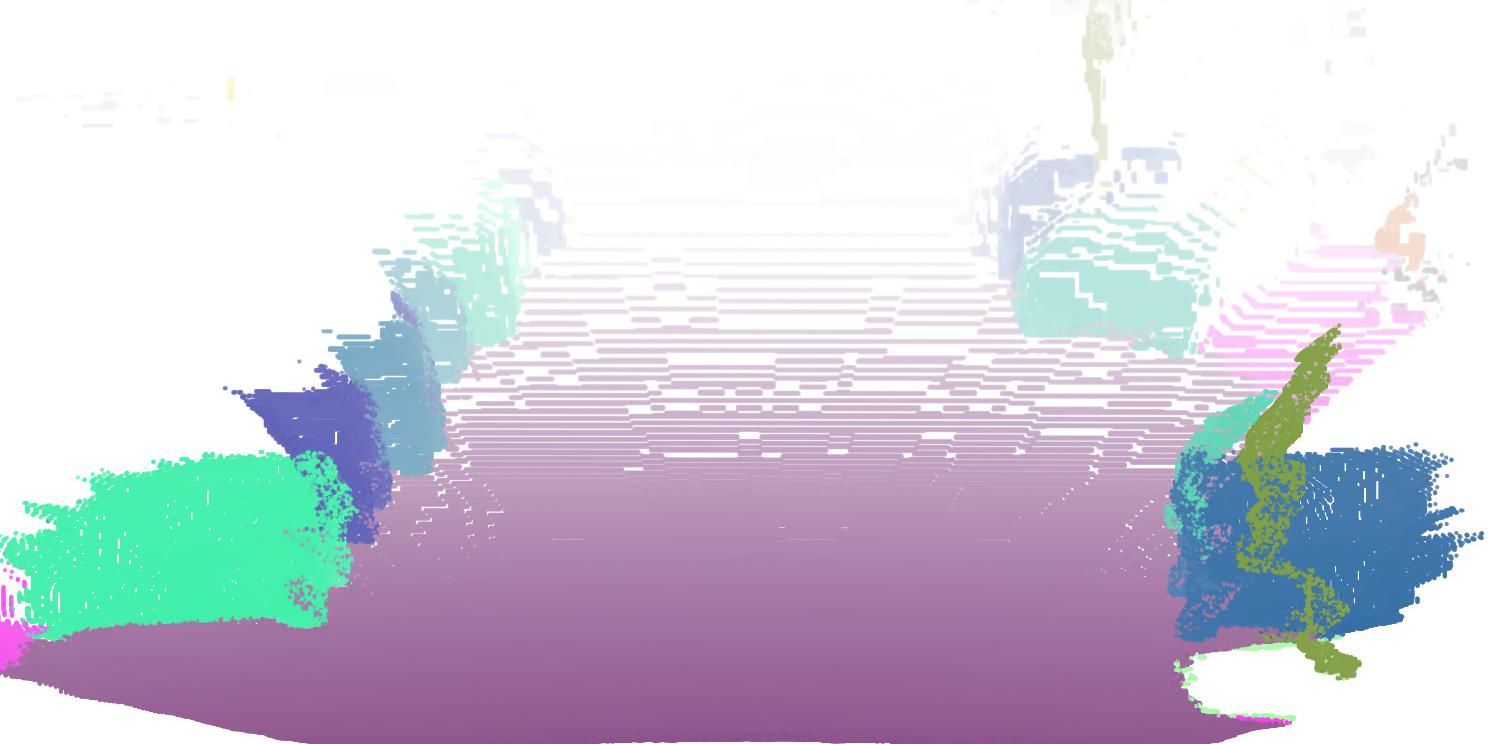}
\end{subfigure}\\\begin{subfigure}{.245\linewidth}
  \centering
  \includegraphics[trim={0 100 0 100},clip,width=\linewidth]{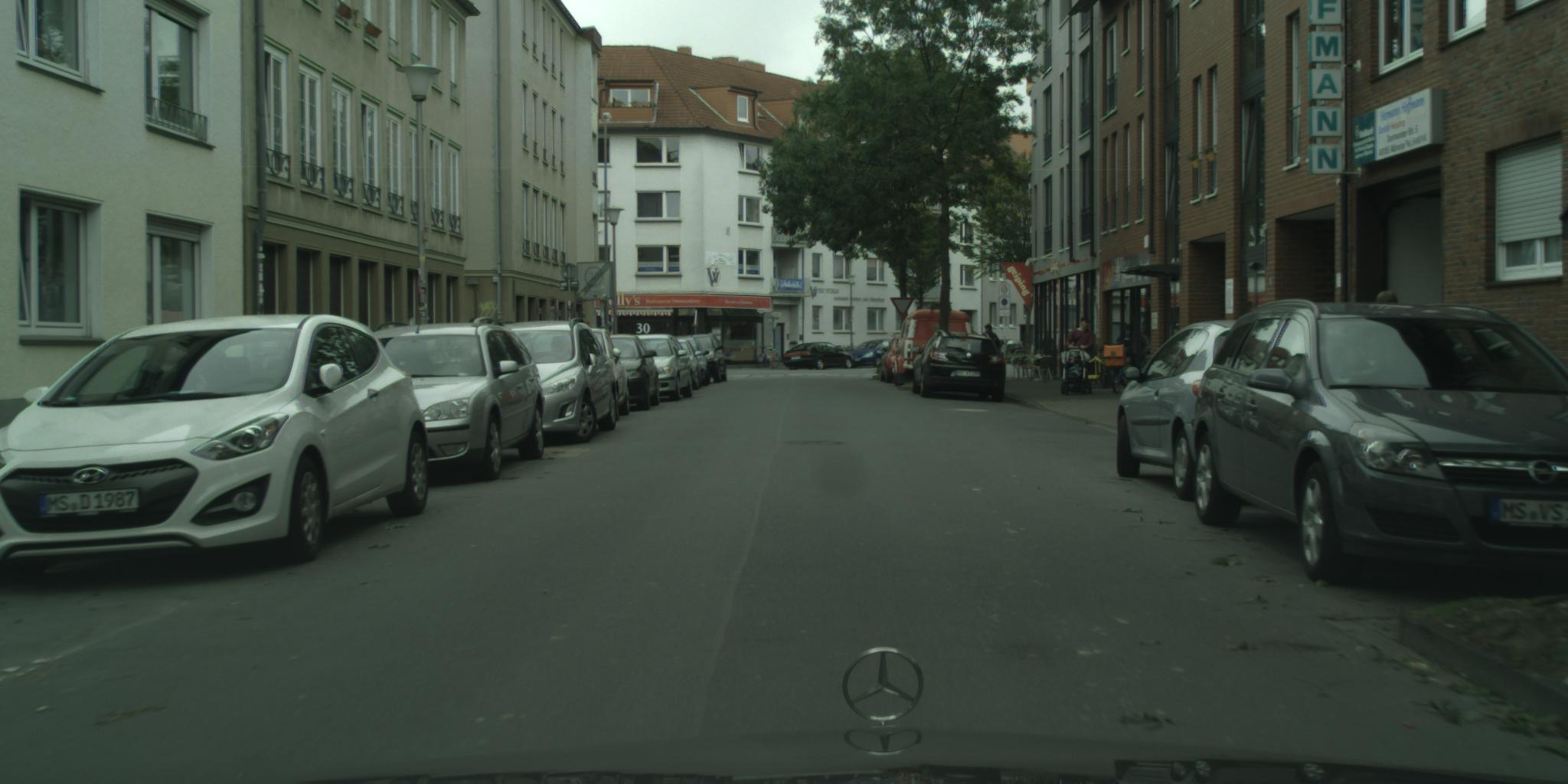}
\end{subfigure}\hfill\begin{subfigure}{.245\linewidth}
  \centering
  \includegraphics[trim={0 100 0 100},clip,width=\linewidth]{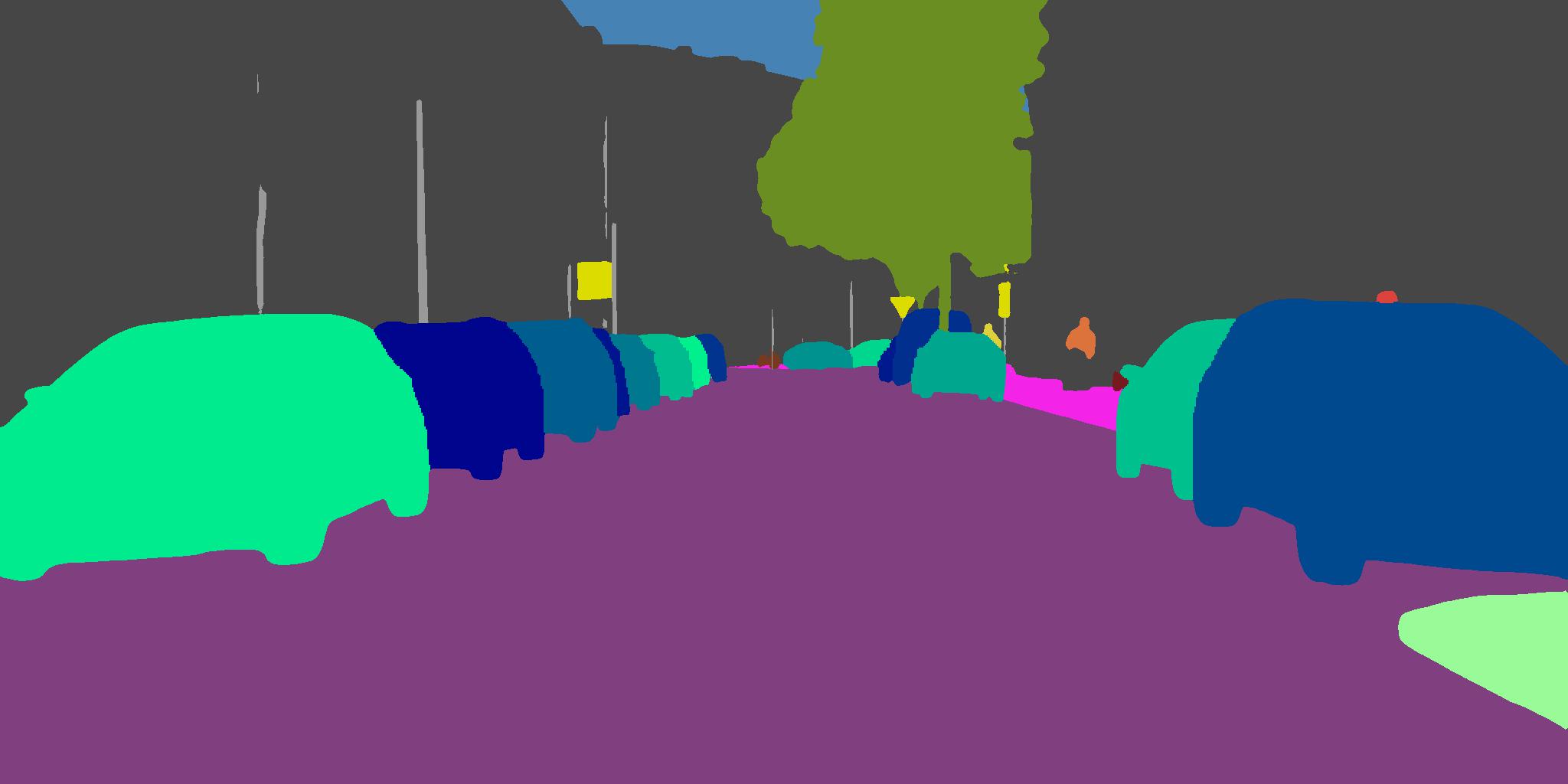}
\end{subfigure}\hfill\begin{subfigure}{.245\linewidth}
  \centering
  \includegraphics[trim={0 100 0 100},clip,width=\linewidth]{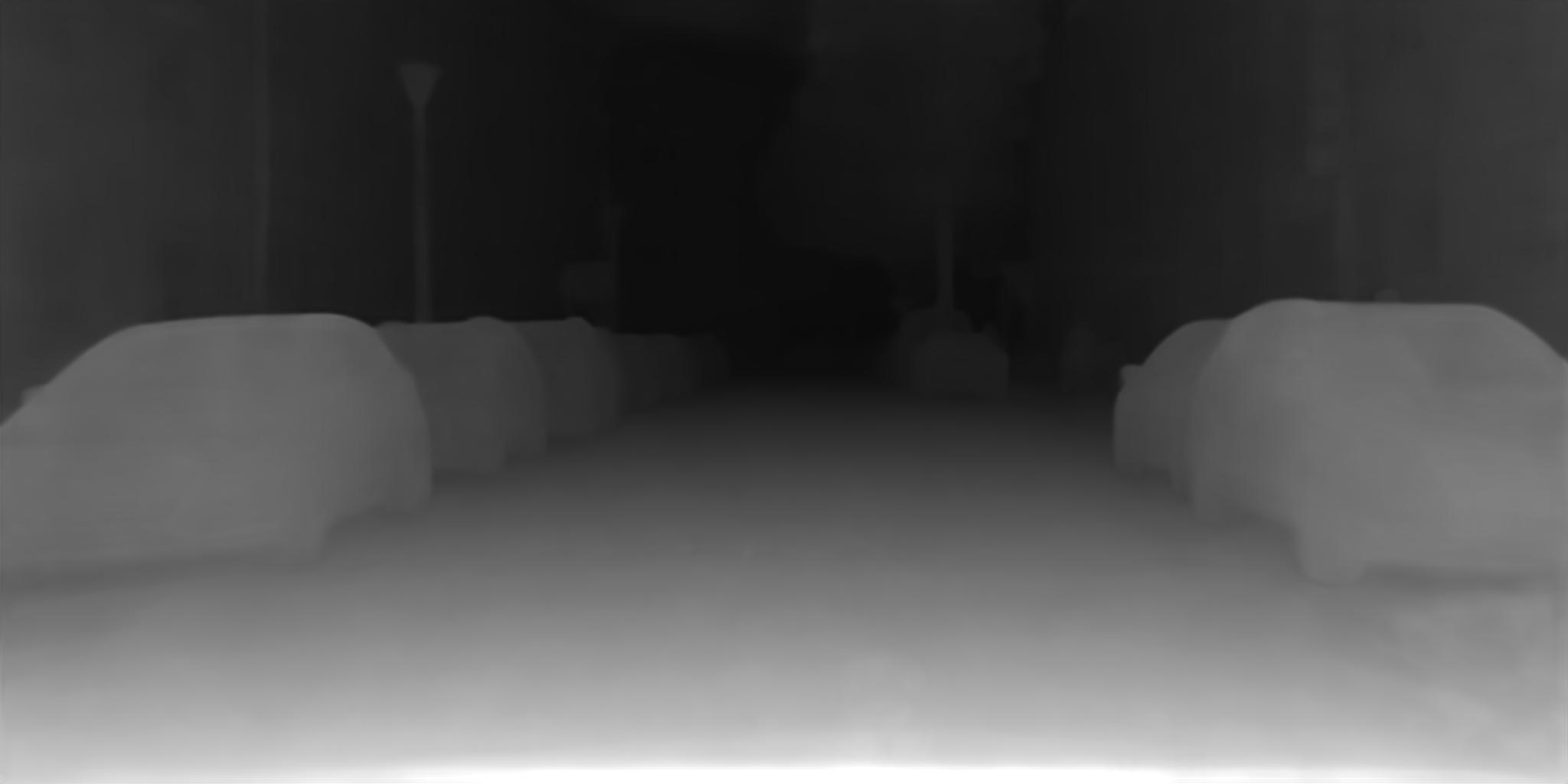}
\end{subfigure}\hfill\begin{subfigure}{.245\linewidth}
  \centering
  \includegraphics[trim={0 100 0 100},clip,width=\linewidth]{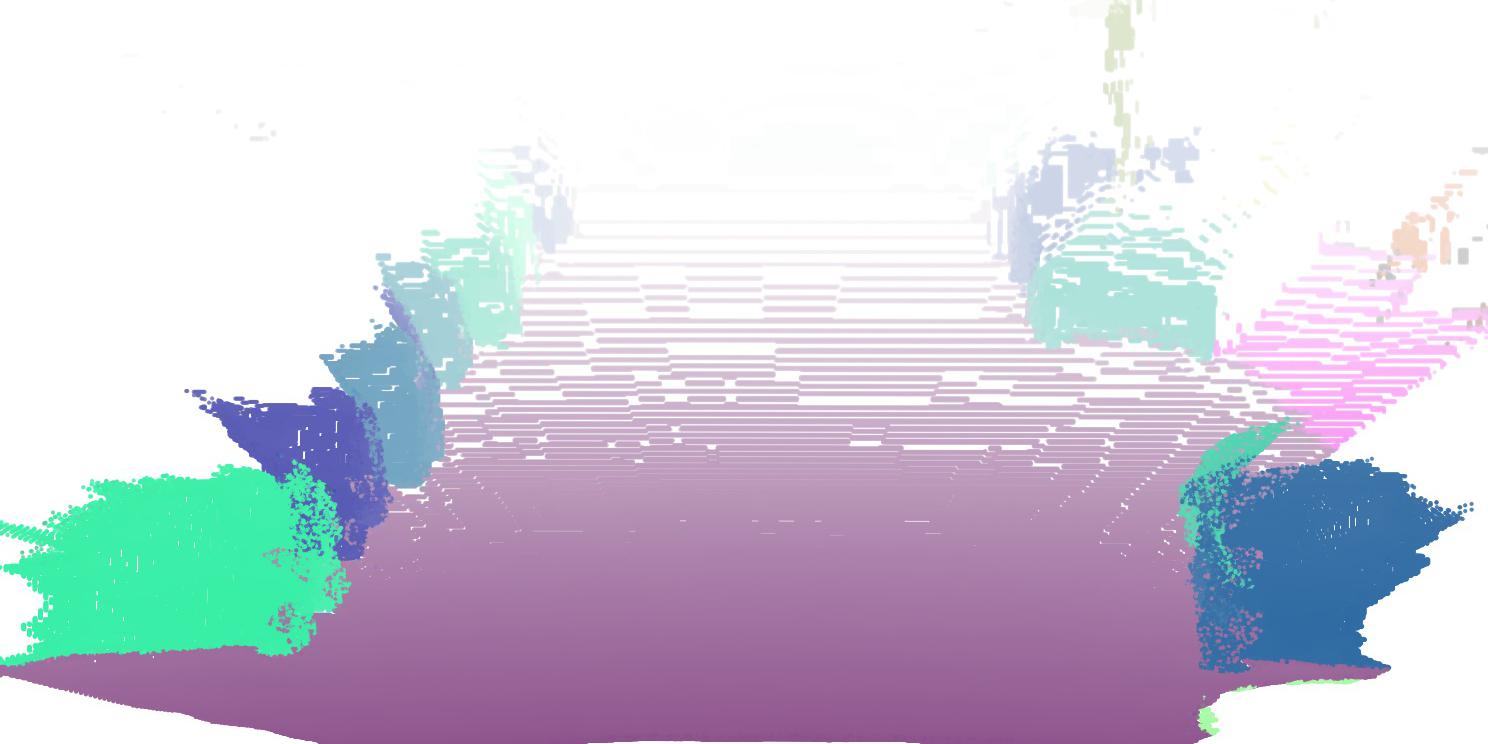}
\end{subfigure}\\\begin{subfigure}{.245\linewidth}
  \centering
  \includegraphics[trim={0 100 0 100},clip,width=\linewidth]{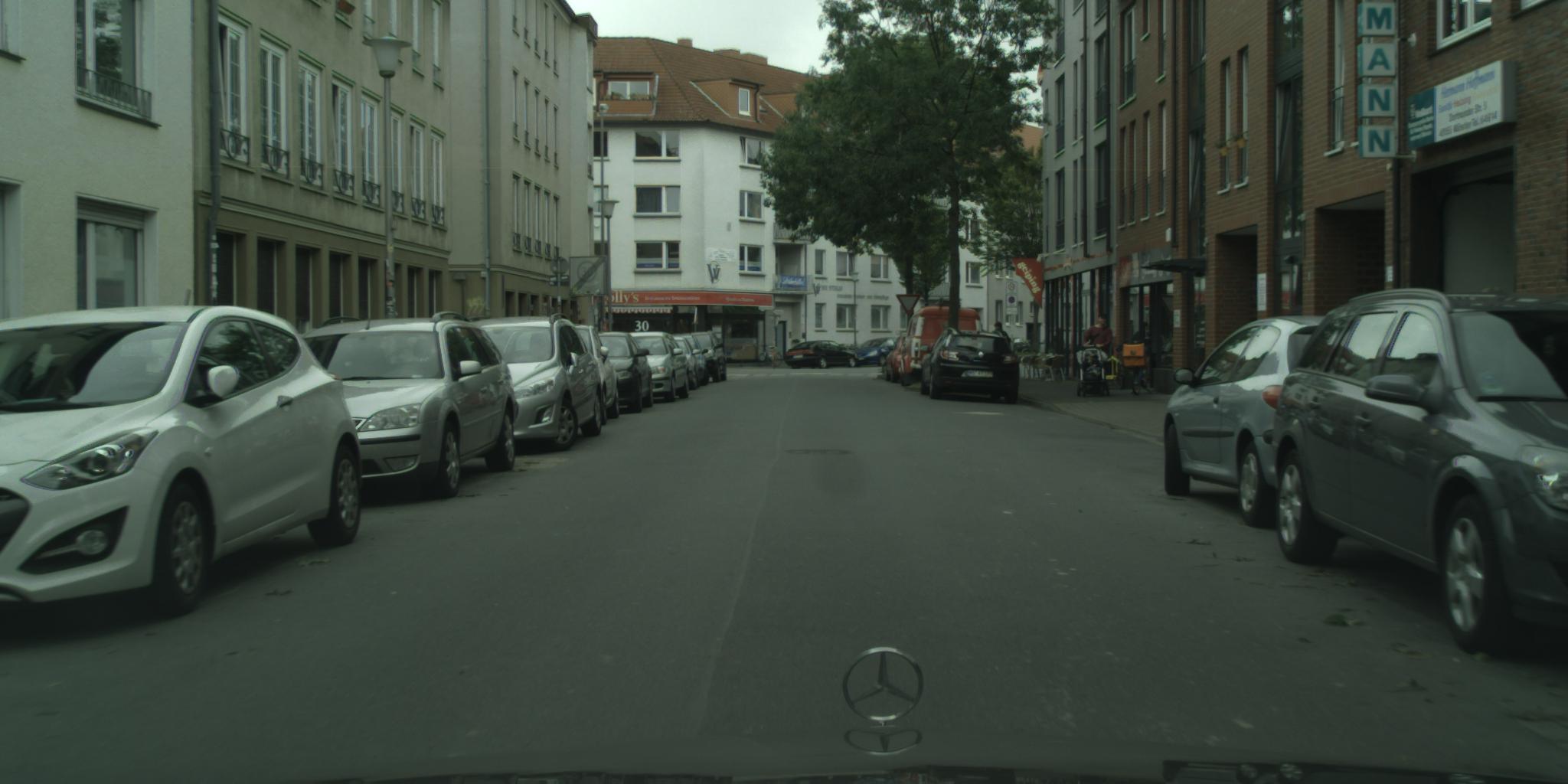}
\end{subfigure}\hfill\begin{subfigure}{.245\linewidth}
  \centering
  \includegraphics[trim={0 100 0 100},clip,width=\linewidth]{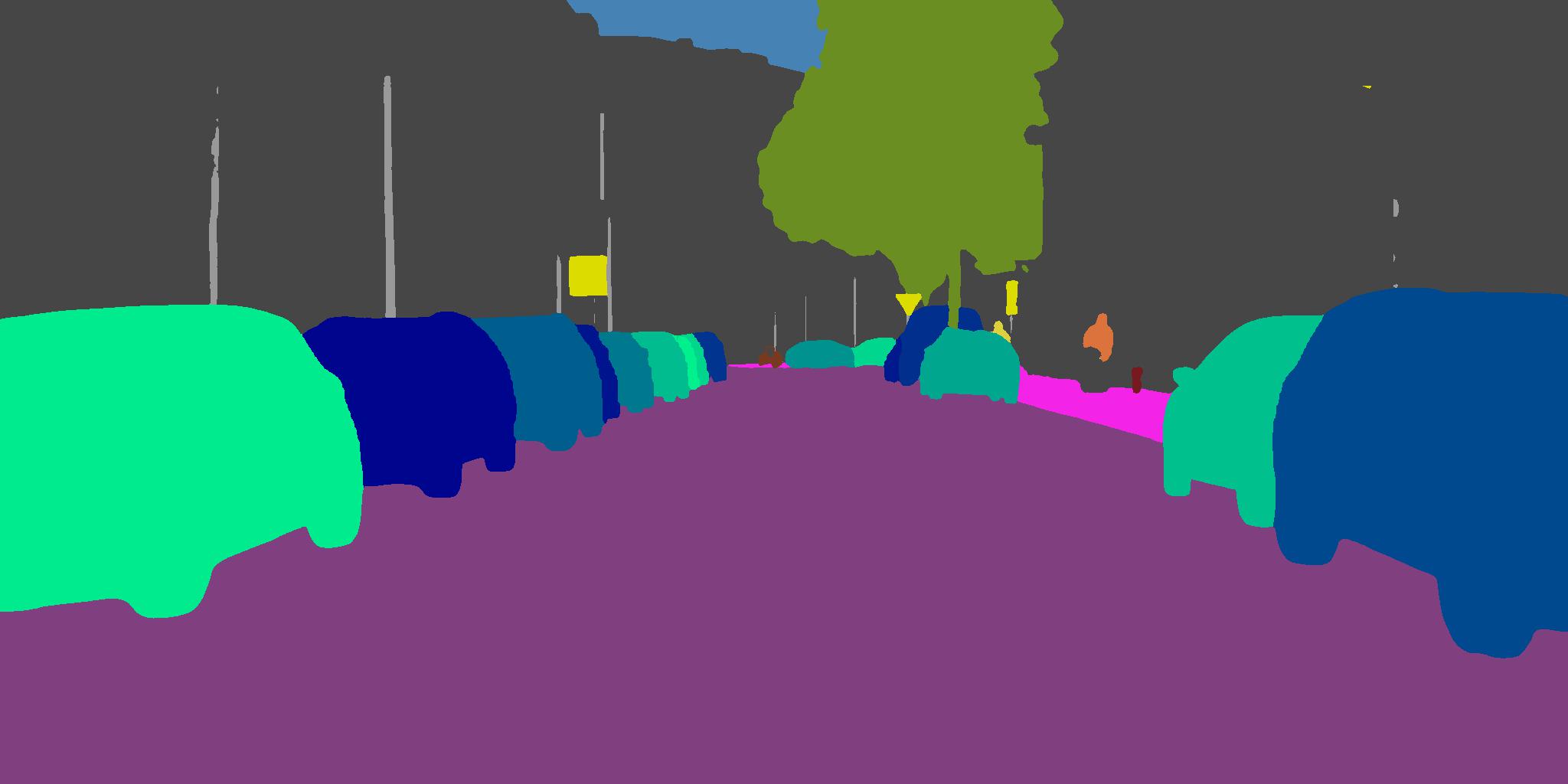}
\end{subfigure}\hfill\begin{subfigure}{.245\linewidth}
  \centering
  \includegraphics[trim={0 100 0 100},clip,width=\linewidth]{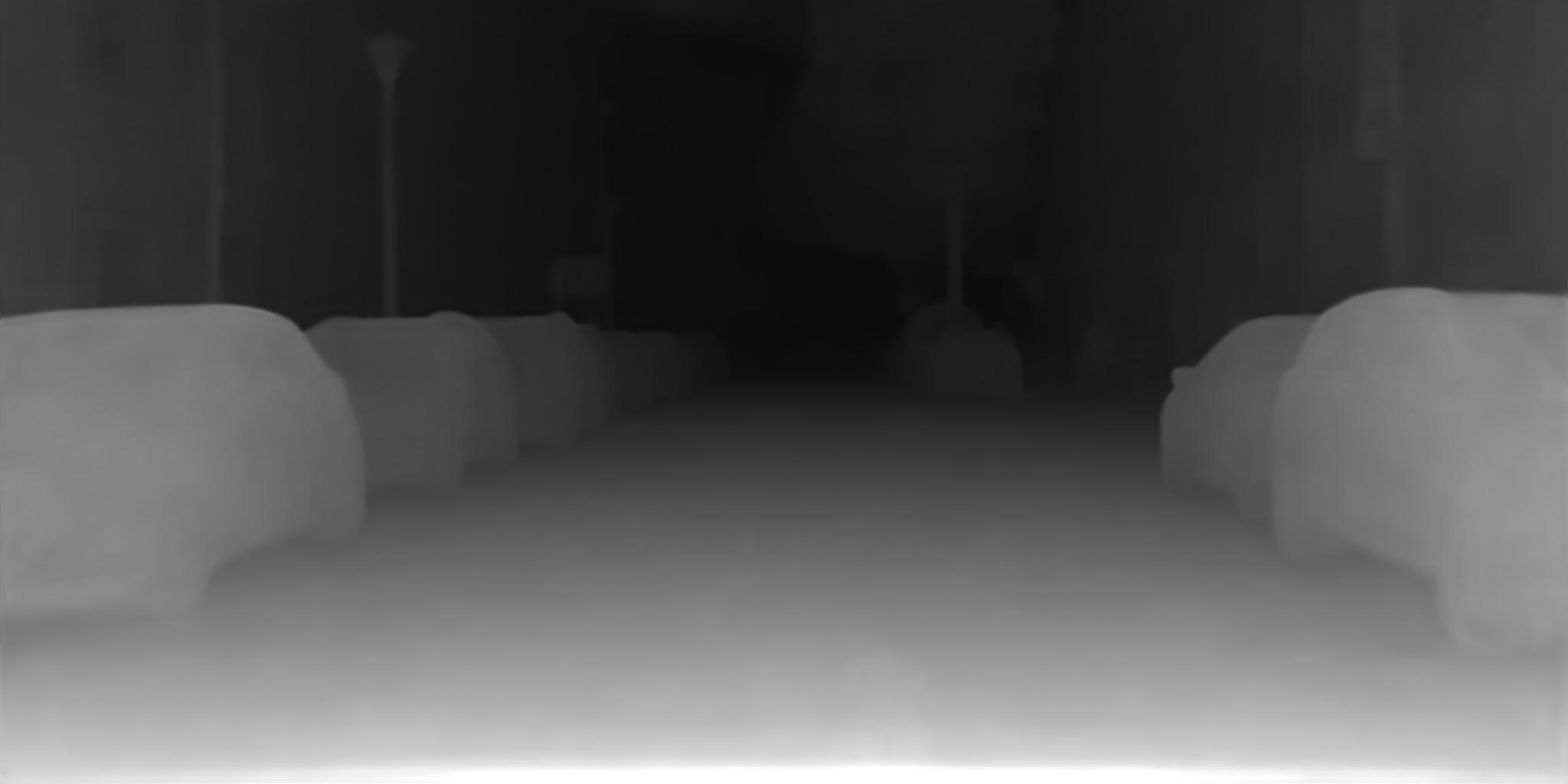}
\end{subfigure}\hfill\begin{subfigure}{.245\linewidth}
  \centering
  \includegraphics[trim={0 100 0 100},clip,width=\linewidth]{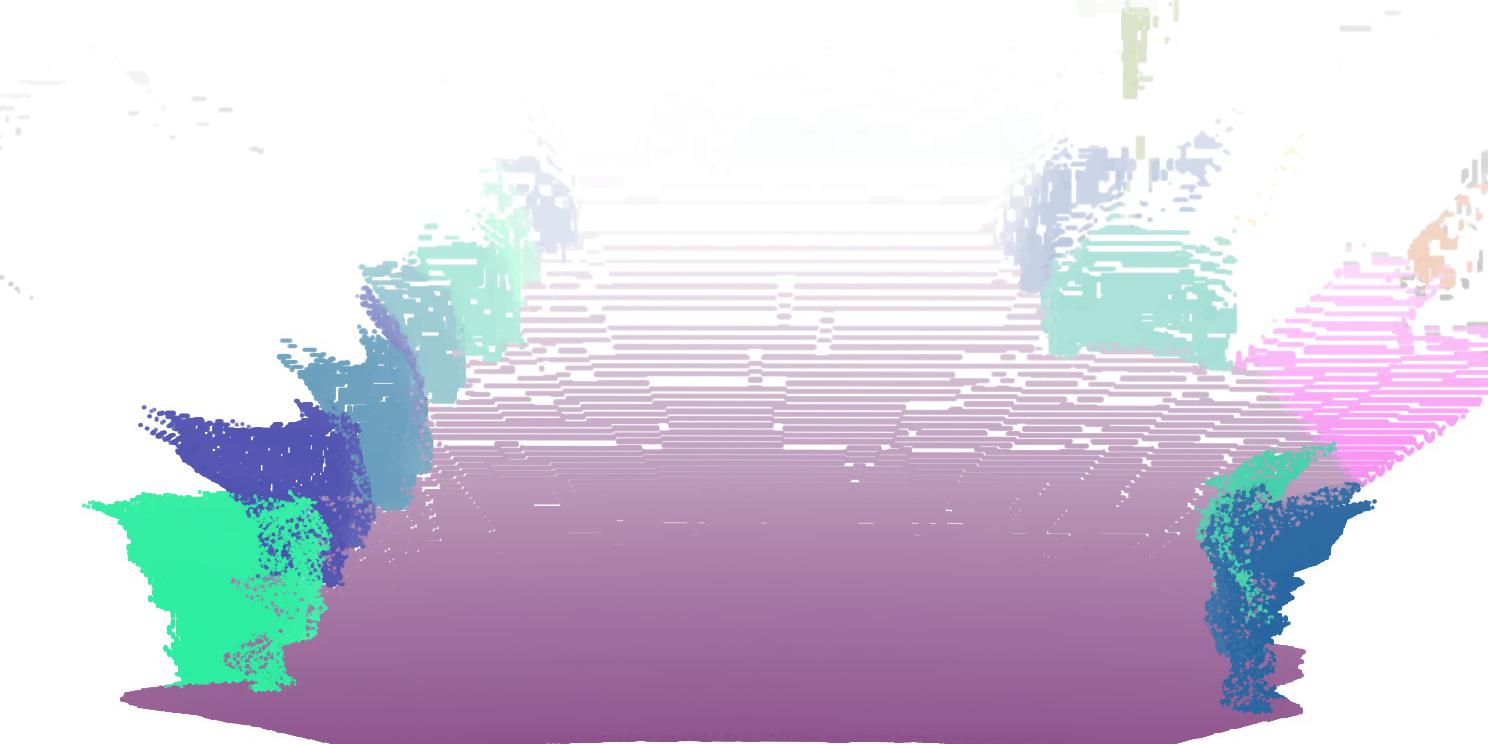}
\end{subfigure}\\\begin{subfigure}{.245\linewidth}
  \centering
  \includegraphics[trim={0 100 0 100},clip,width=\linewidth]{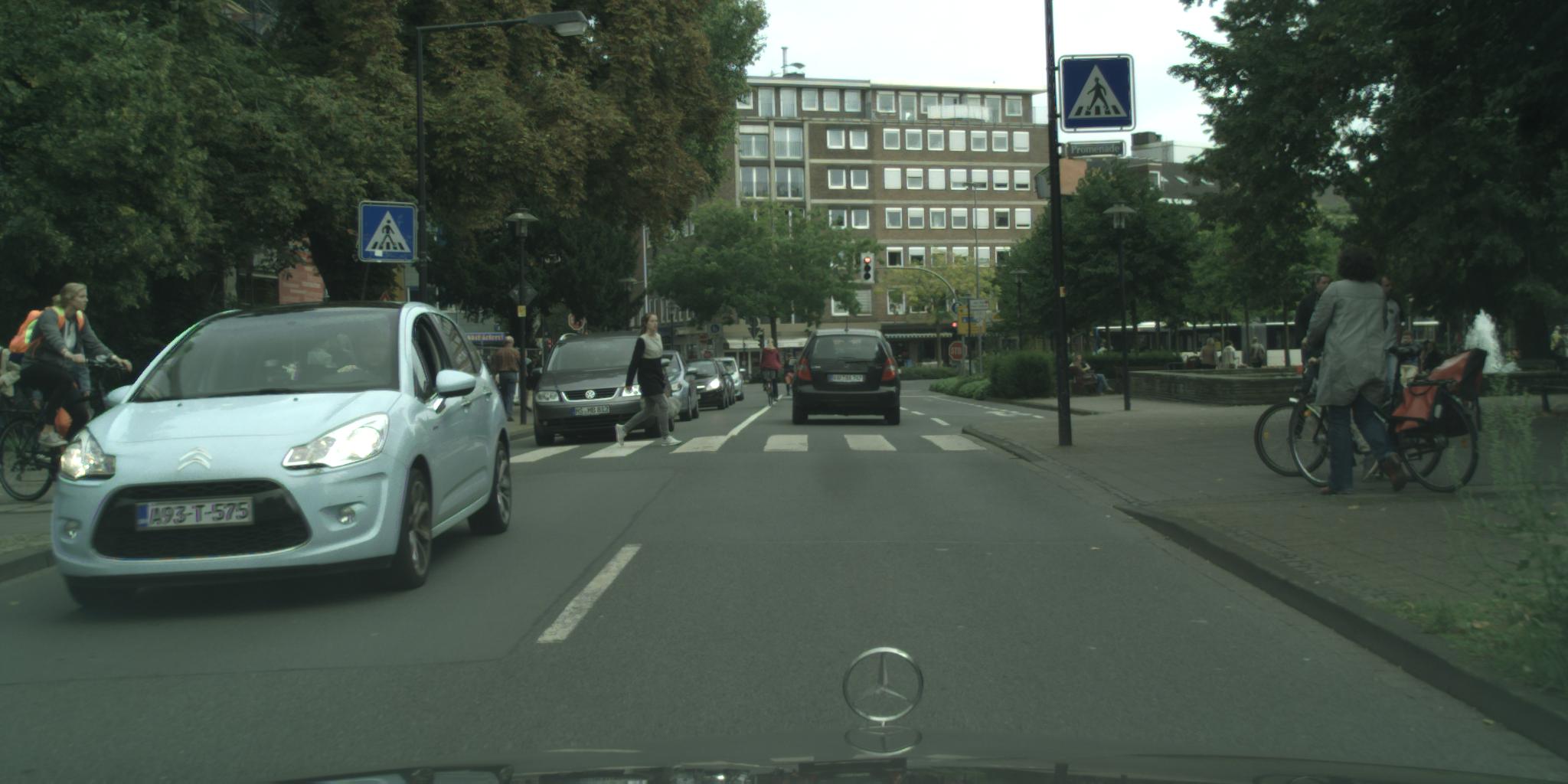}
\end{subfigure}\hfill\begin{subfigure}{.245\linewidth}
  \centering
  \includegraphics[trim={0 100 0 100},clip,width=\linewidth]{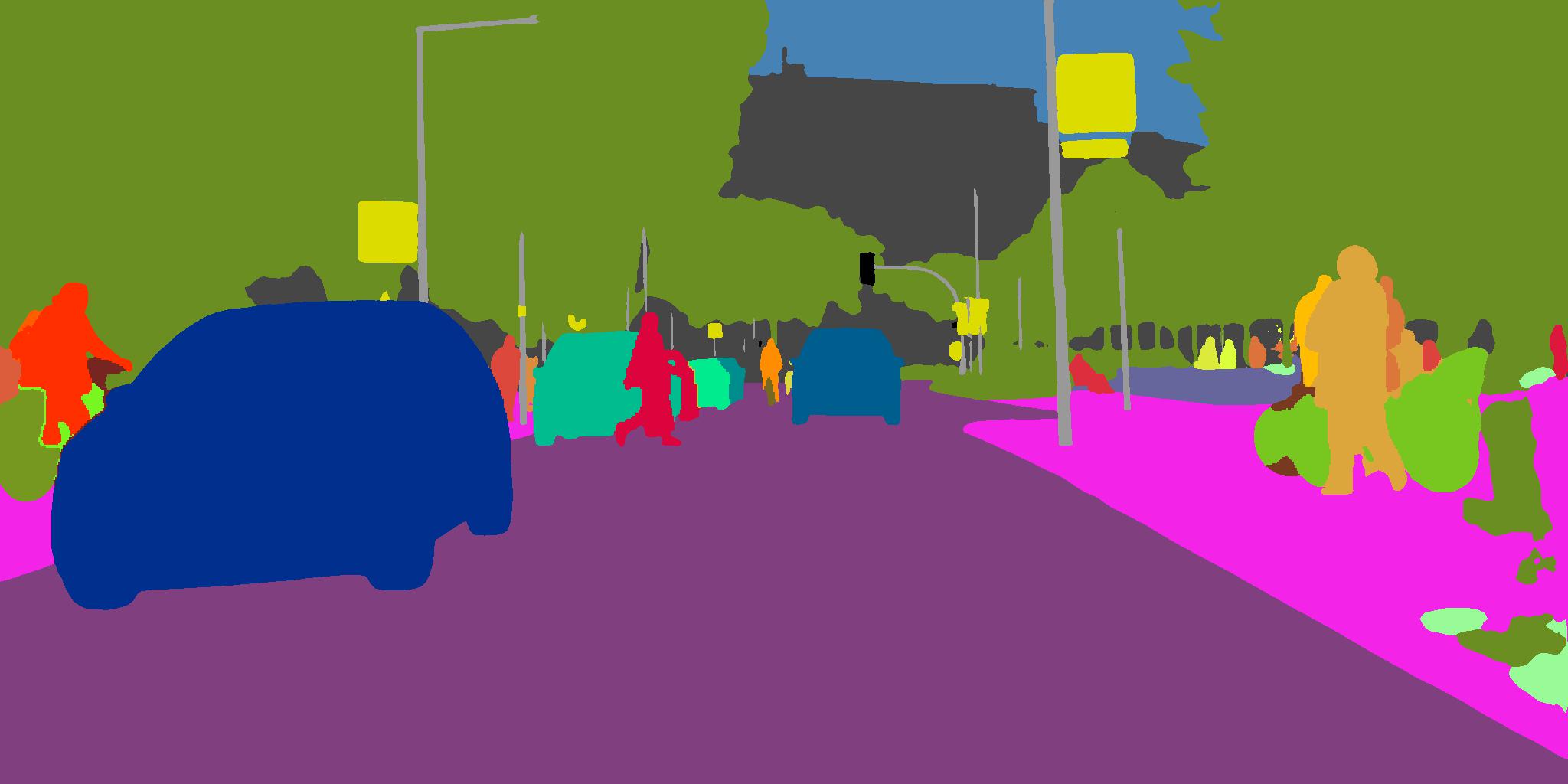}
\end{subfigure}\hfill\begin{subfigure}{.245\linewidth}
  \centering
  \includegraphics[trim={0 100 0 100},clip,width=\linewidth]{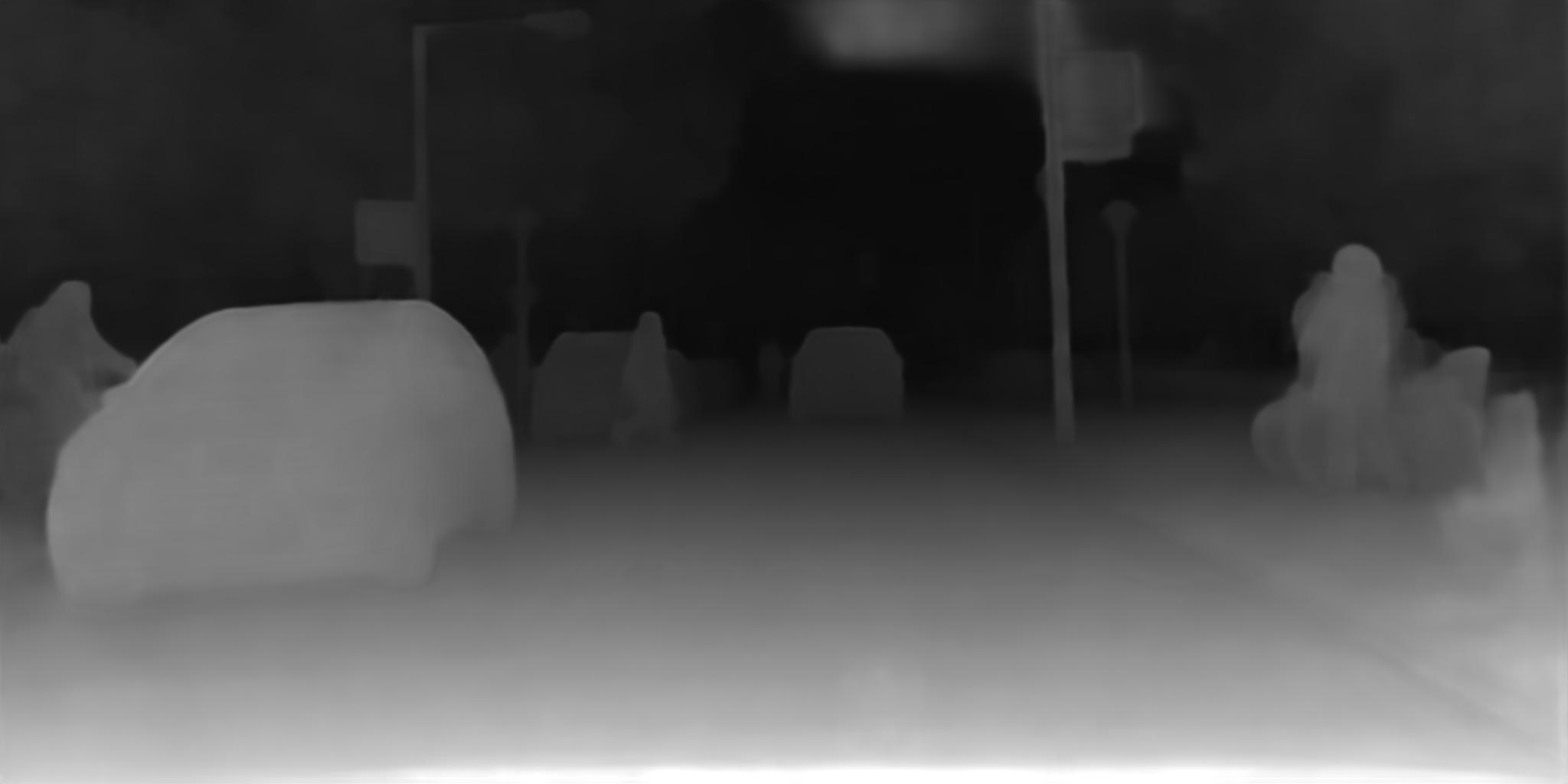}
\end{subfigure}\hfill\begin{subfigure}{.245\linewidth}
  \centering
  \includegraphics[trim={0 100 0 100},clip,width=\linewidth]{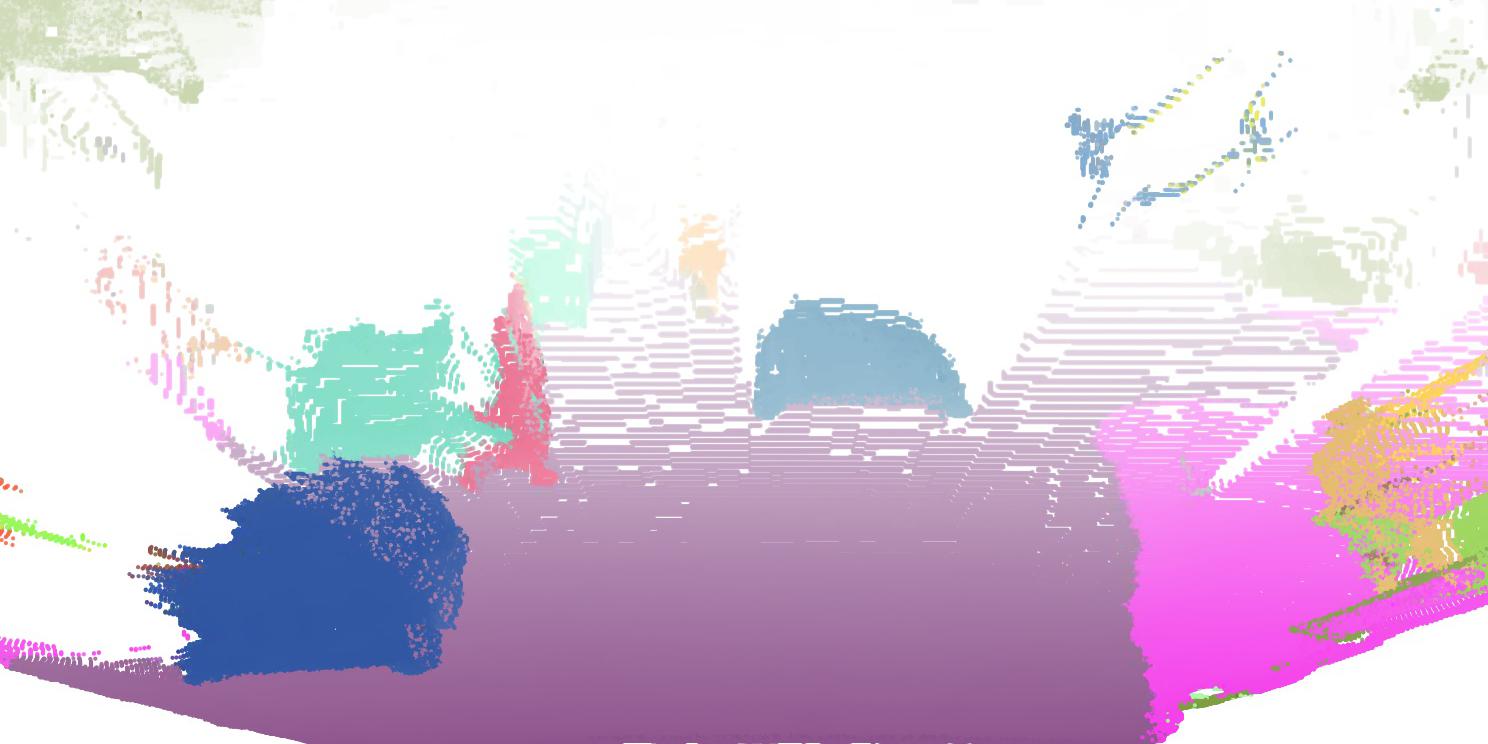}
\end{subfigure}\\\begin{subfigure}{.245\linewidth}
  \centering
  \includegraphics[trim={0 100 0 100},clip,width=\linewidth]{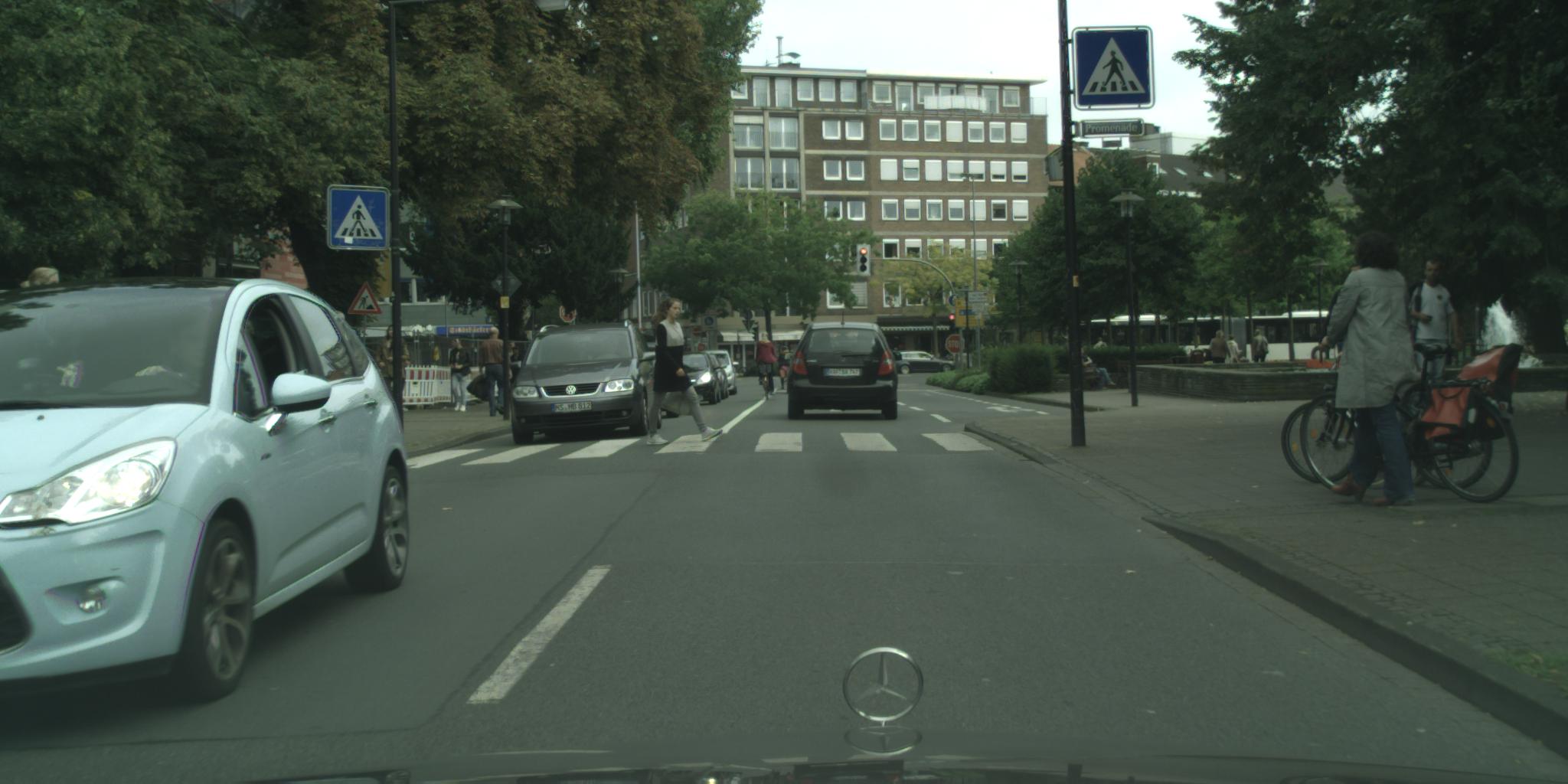}
\end{subfigure}\hfill\begin{subfigure}{.245\linewidth}
  \centering
  \includegraphics[trim={0 100 0 100},clip,width=\linewidth]{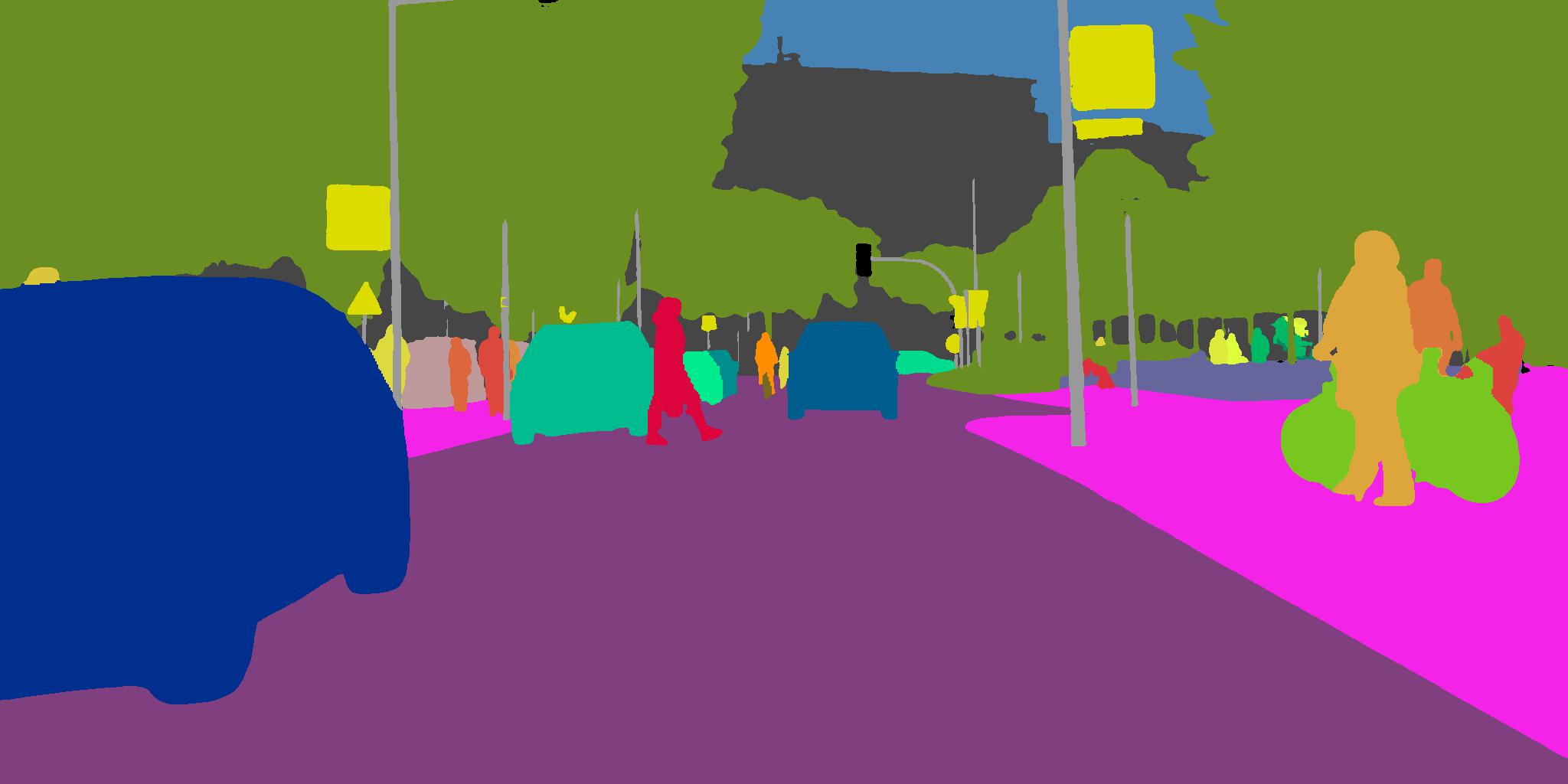}
\end{subfigure}\hfill\begin{subfigure}{.245\linewidth}
  \centering
  \includegraphics[trim={0 100 0 100},clip,width=\linewidth]{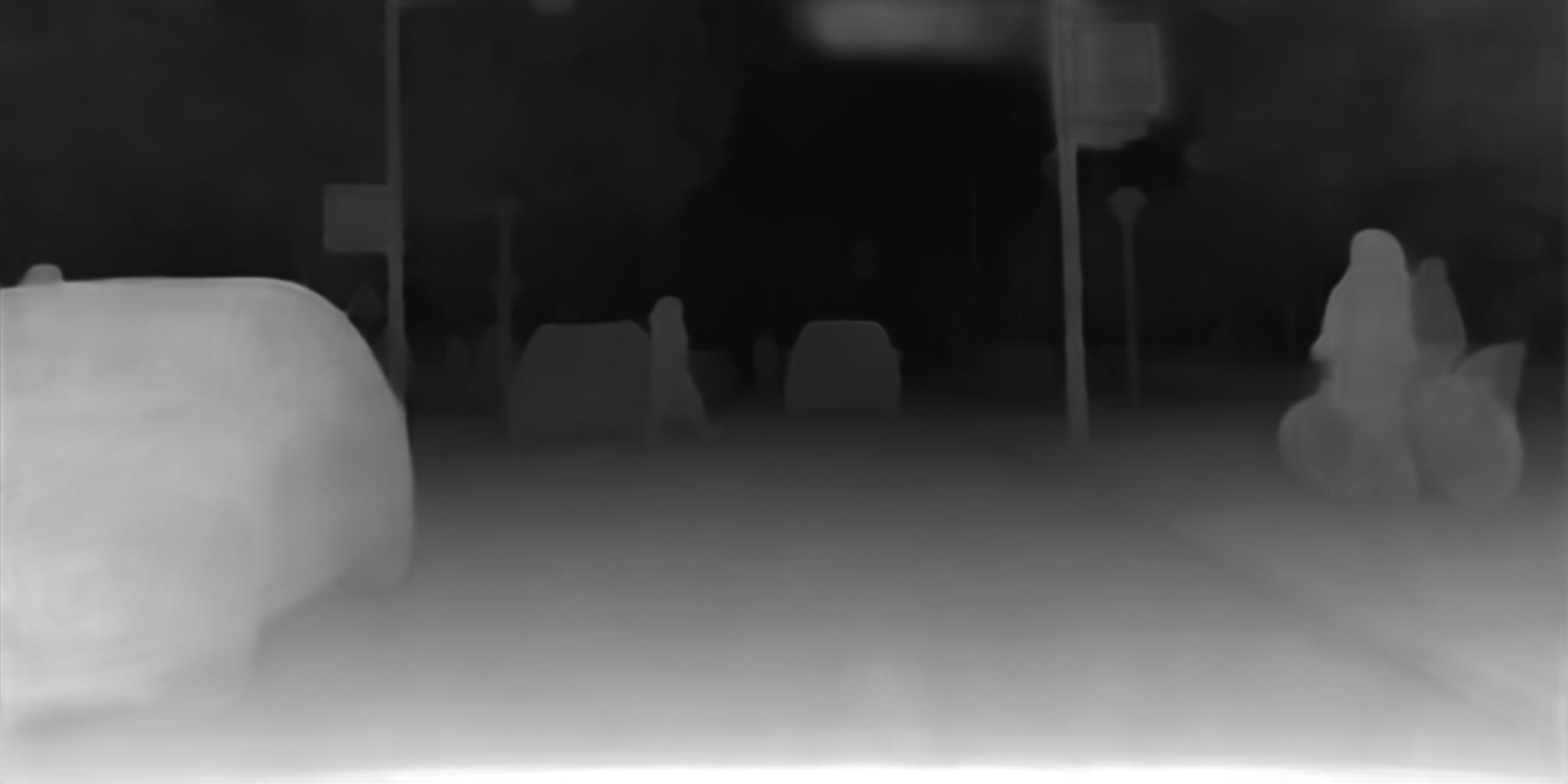}
\end{subfigure}\hfill\begin{subfigure}{.245\linewidth}
  \centering
  \includegraphics[trim={0 100 0 100},clip,width=\linewidth]{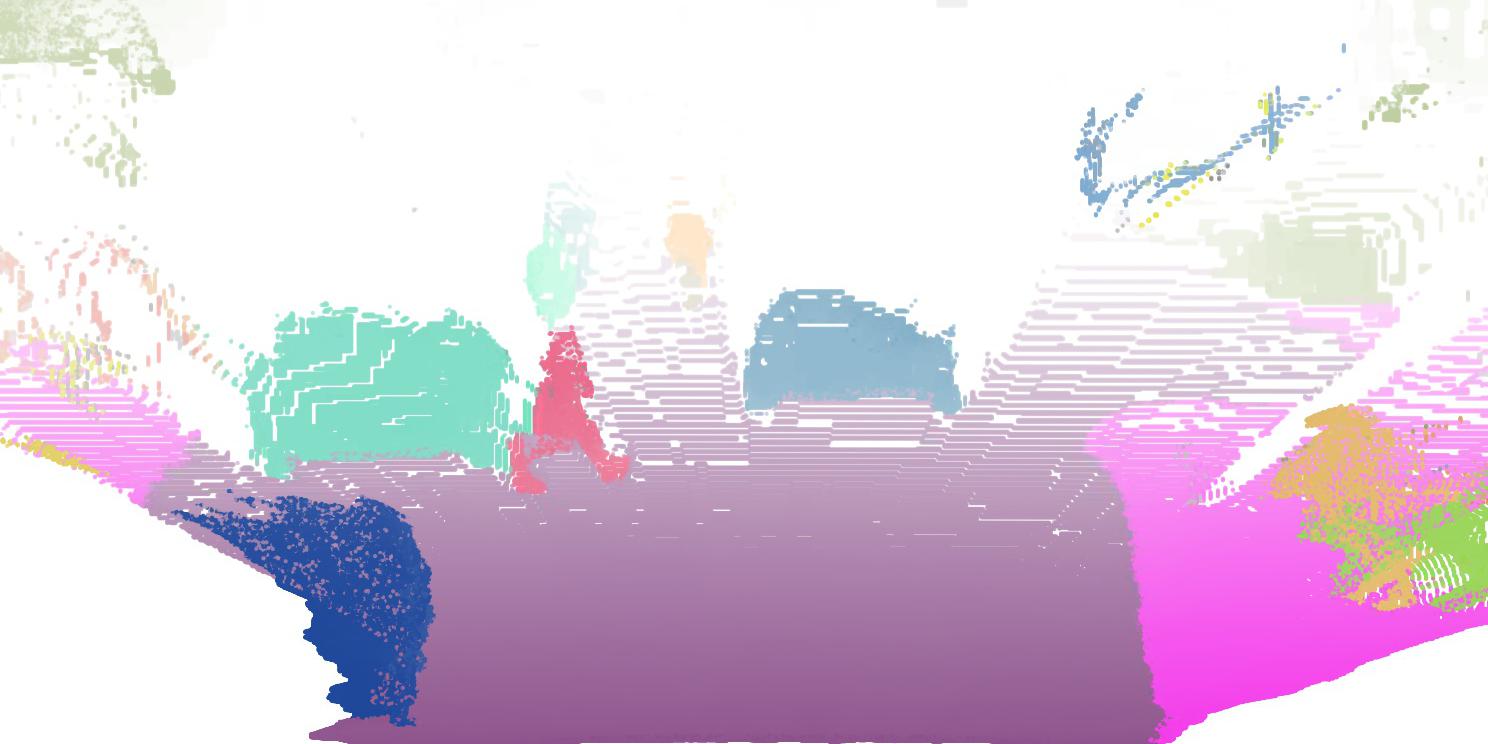}
\end{subfigure}\\\begin{subfigure}{.245\linewidth}
  \centering
  \includegraphics[trim={0 100 0 100},clip,width=\linewidth]{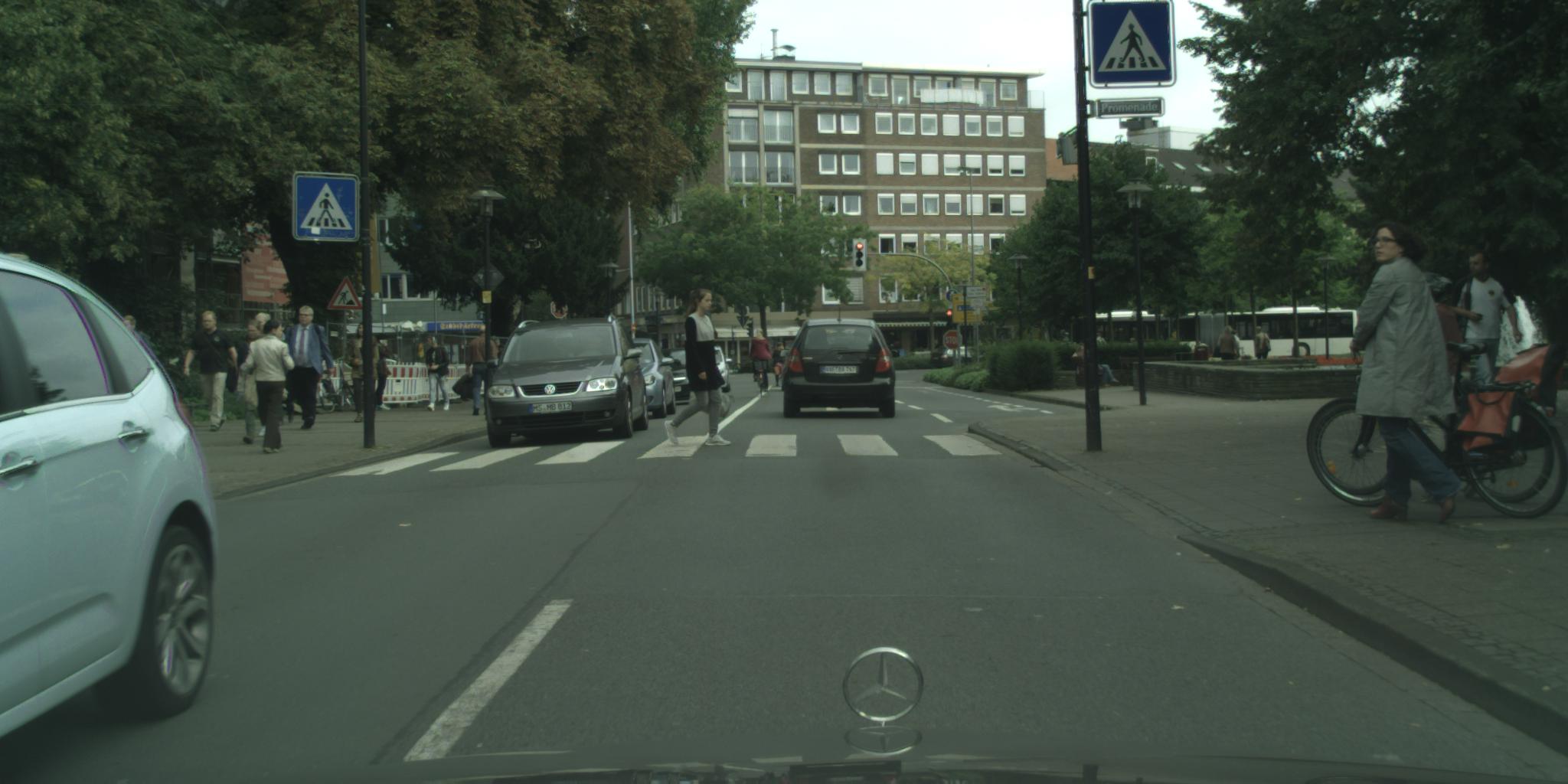}
\end{subfigure}\hfill\begin{subfigure}{.245\linewidth}
  \centering
  \includegraphics[trim={0 100 0 100},clip,width=\linewidth]{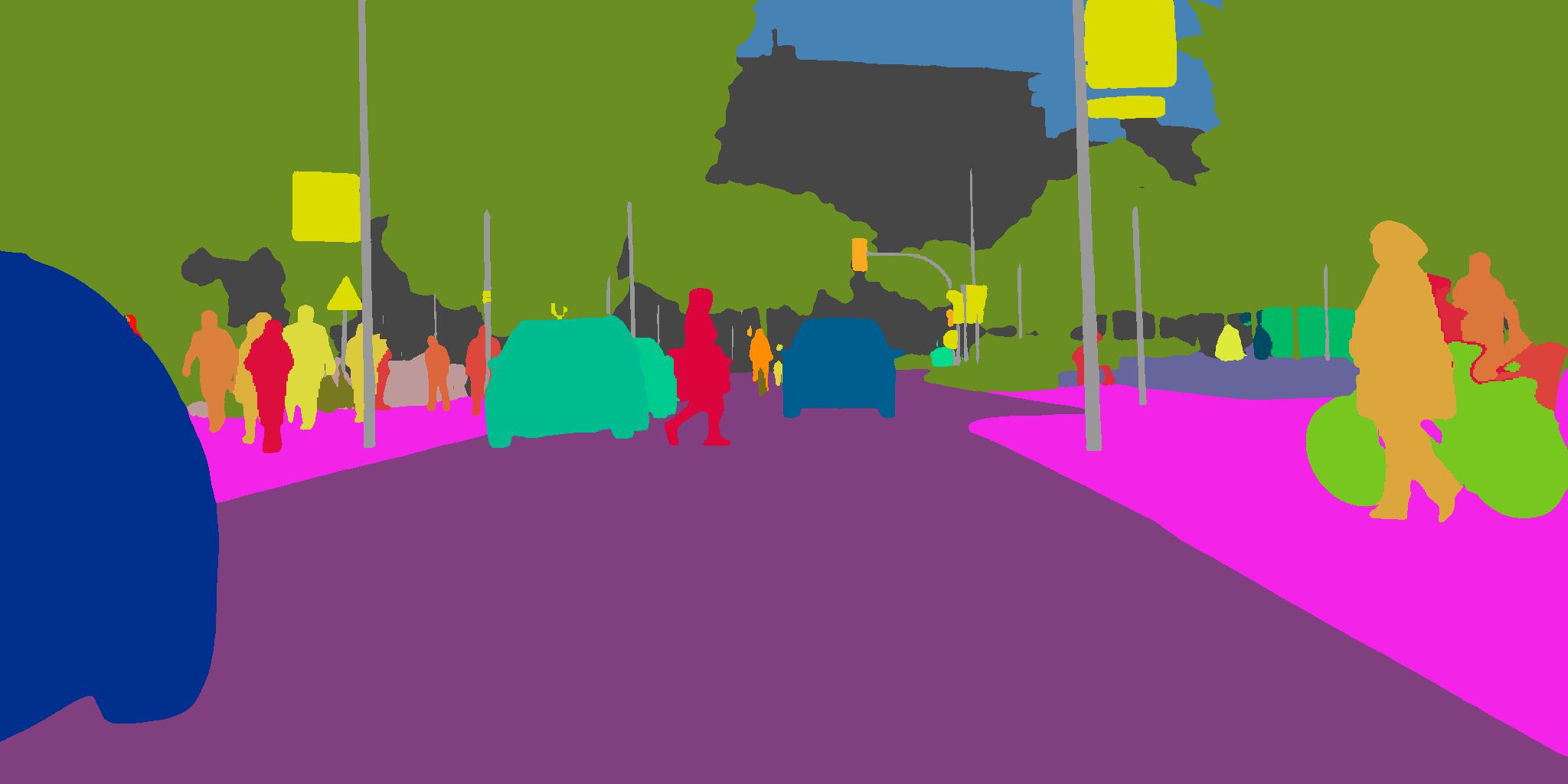}
\end{subfigure}\hfill\begin{subfigure}{.245\linewidth}
  \centering
  \includegraphics[trim={0 100 0 100},clip,width=\linewidth]{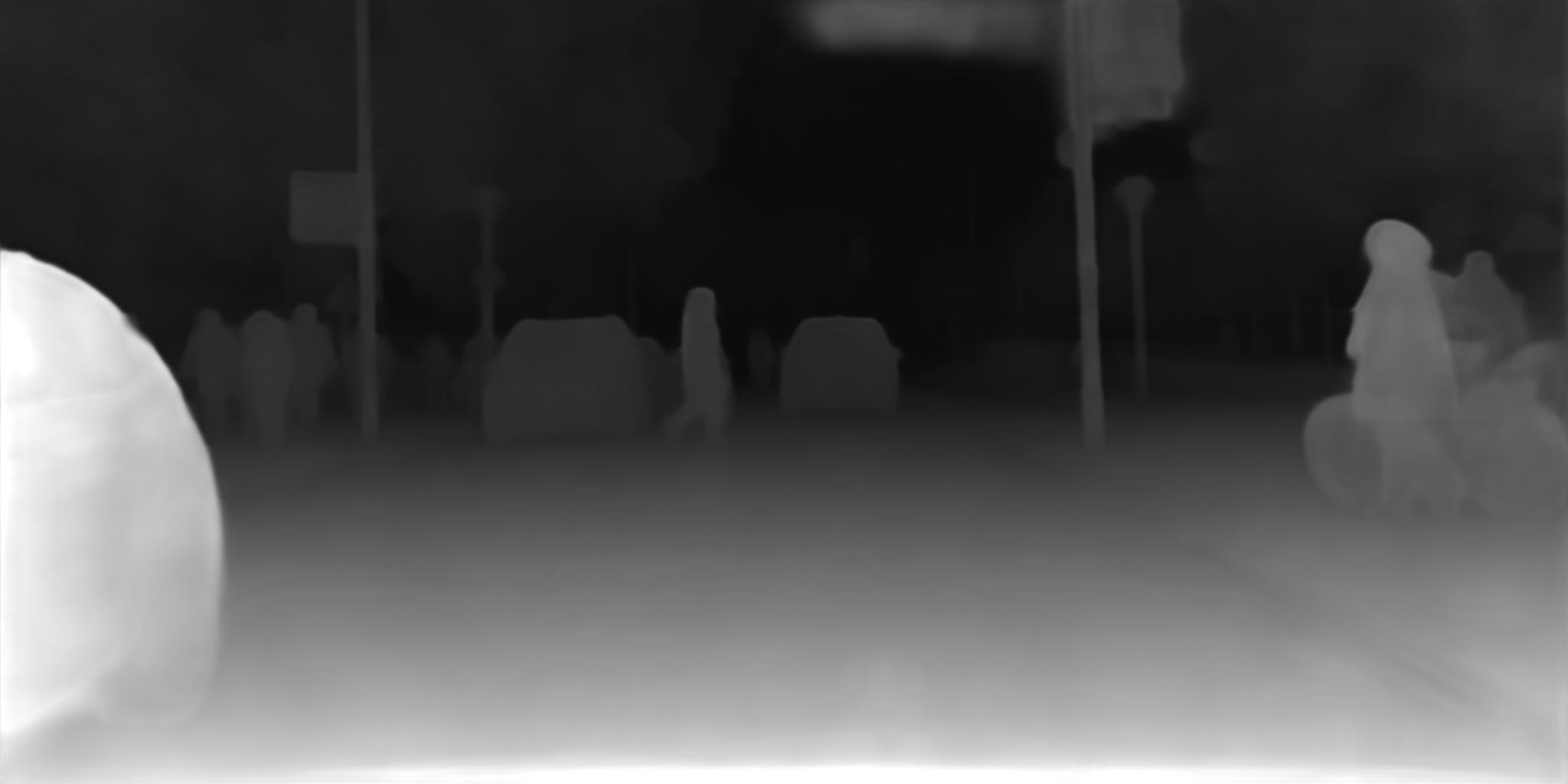}
\end{subfigure}\hfill\begin{subfigure}{.245\linewidth}
  \centering
  \includegraphics[trim={0 100 0 100},clip,width=\linewidth]{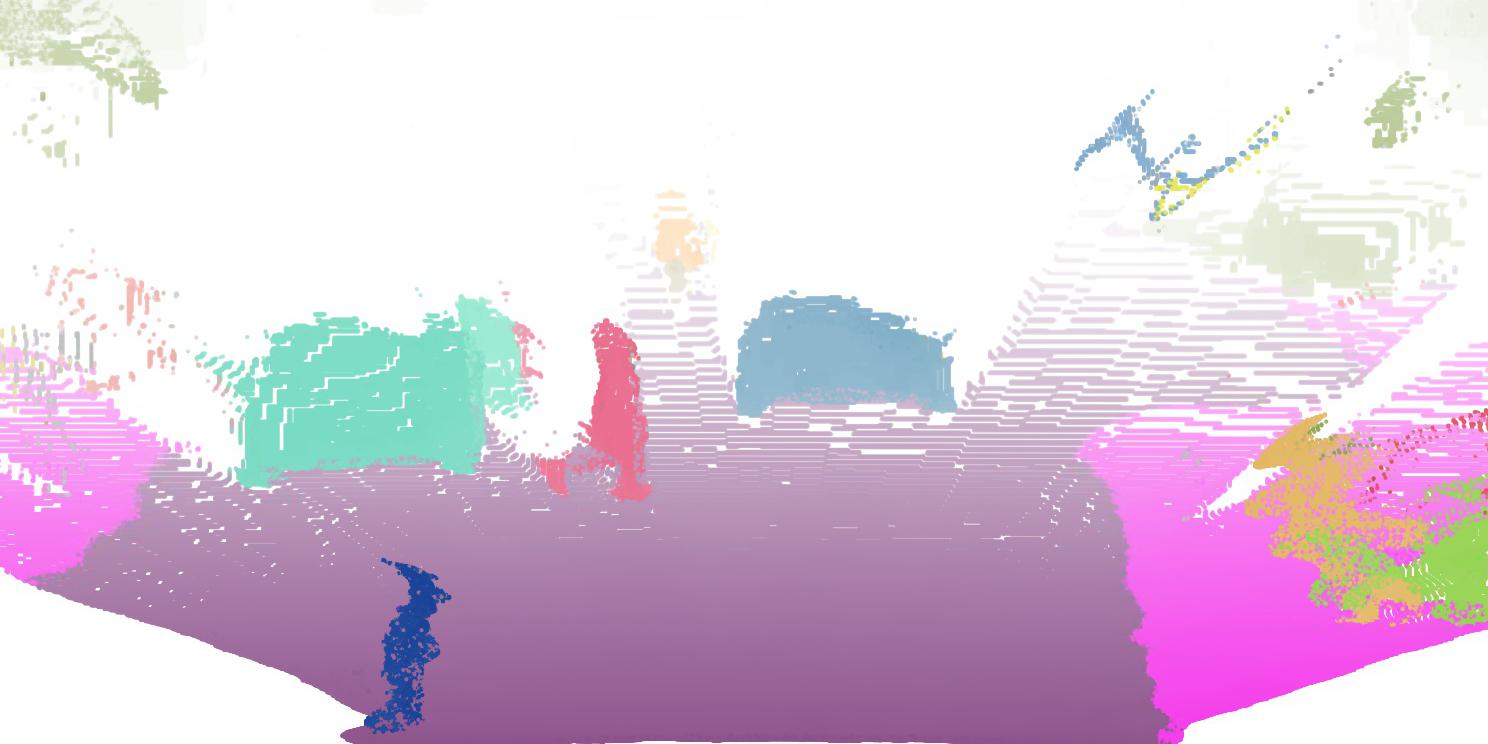}
\end{subfigure}\\\begin{subfigure}{.245\linewidth}
  \centering
  \includegraphics[trim={0 100 0 100},clip,width=\linewidth]{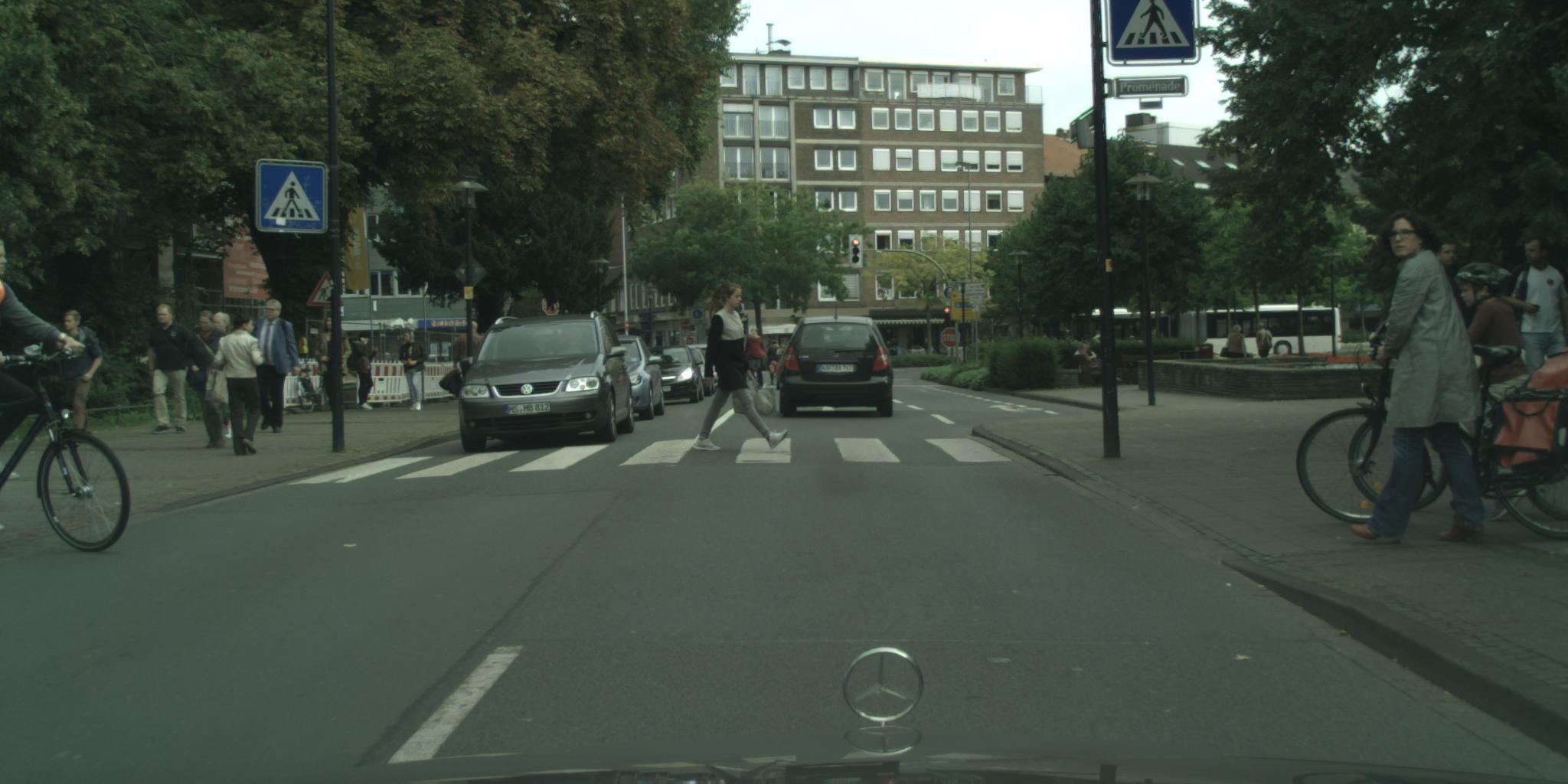}
\end{subfigure}\hfill\begin{subfigure}{.245\linewidth}
  \centering
  \includegraphics[trim={0 100 0 100},clip,width=\linewidth]{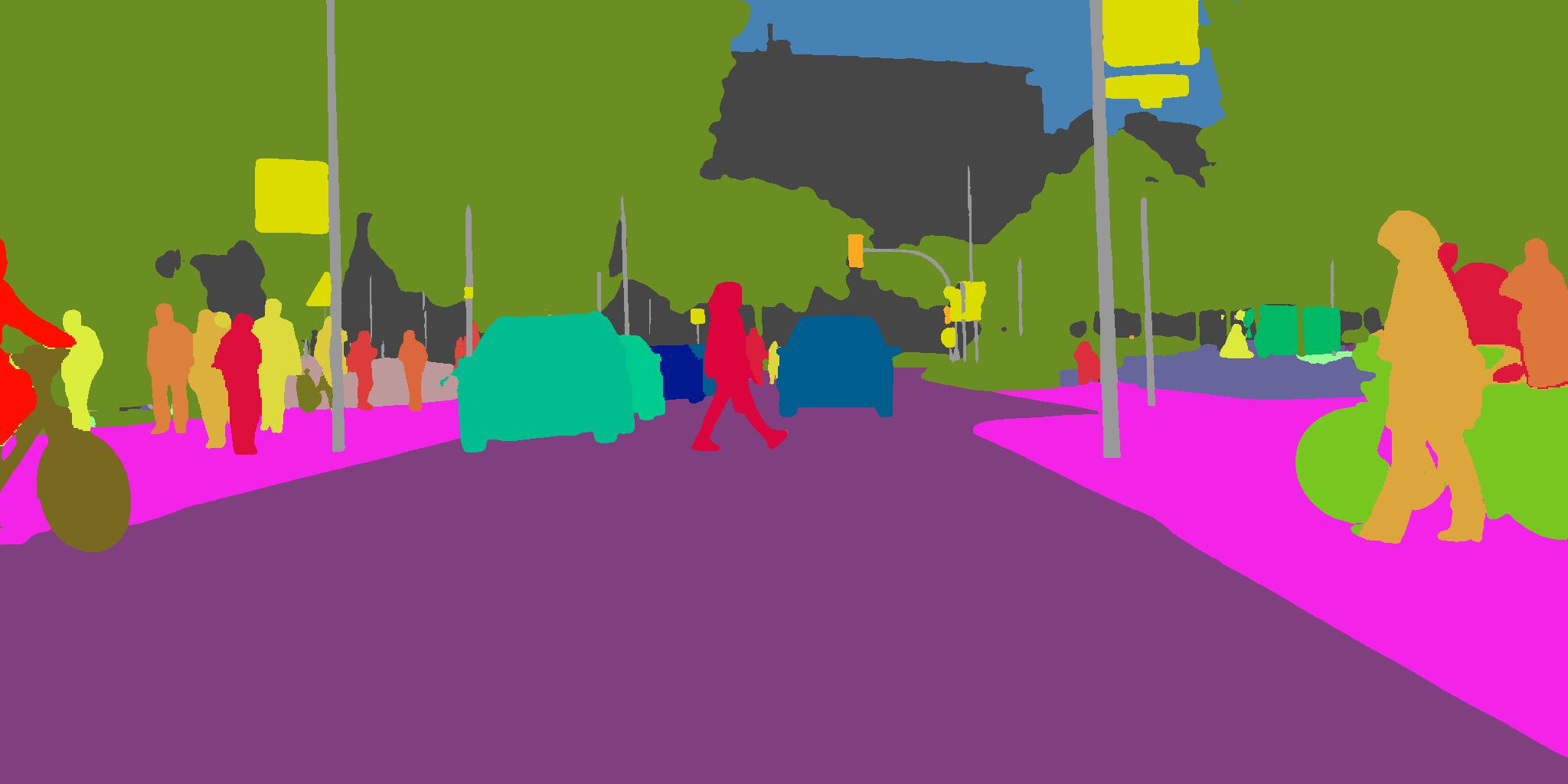}
\end{subfigure}\hfill\begin{subfigure}{.245\linewidth}
  \centering
  \includegraphics[trim={0 100 0 100},clip,width=\linewidth]{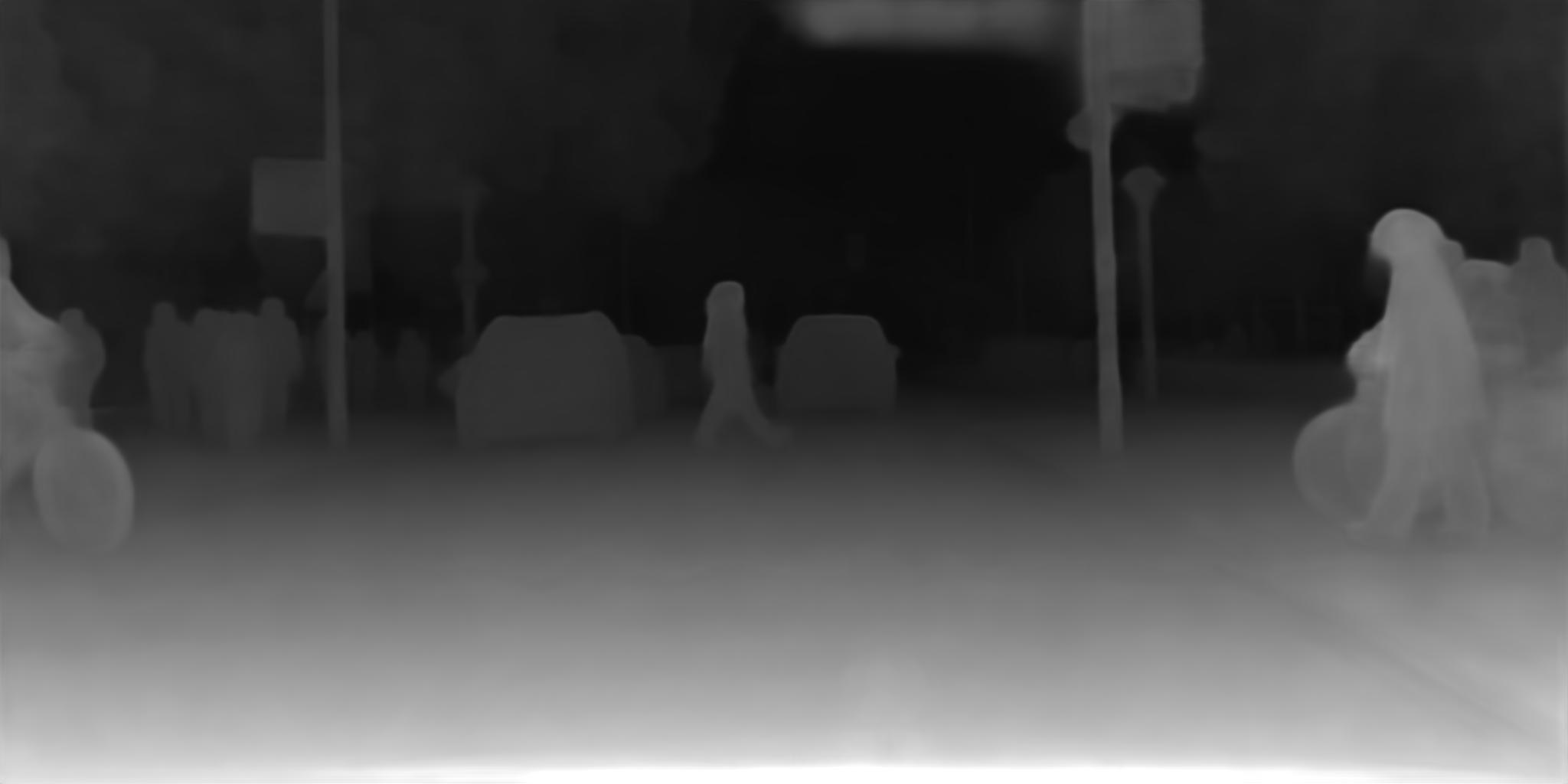}
\end{subfigure}\hfill\begin{subfigure}{.245\linewidth}
  \centering
  \includegraphics[trim={0 100 0 100},clip,width=\linewidth]{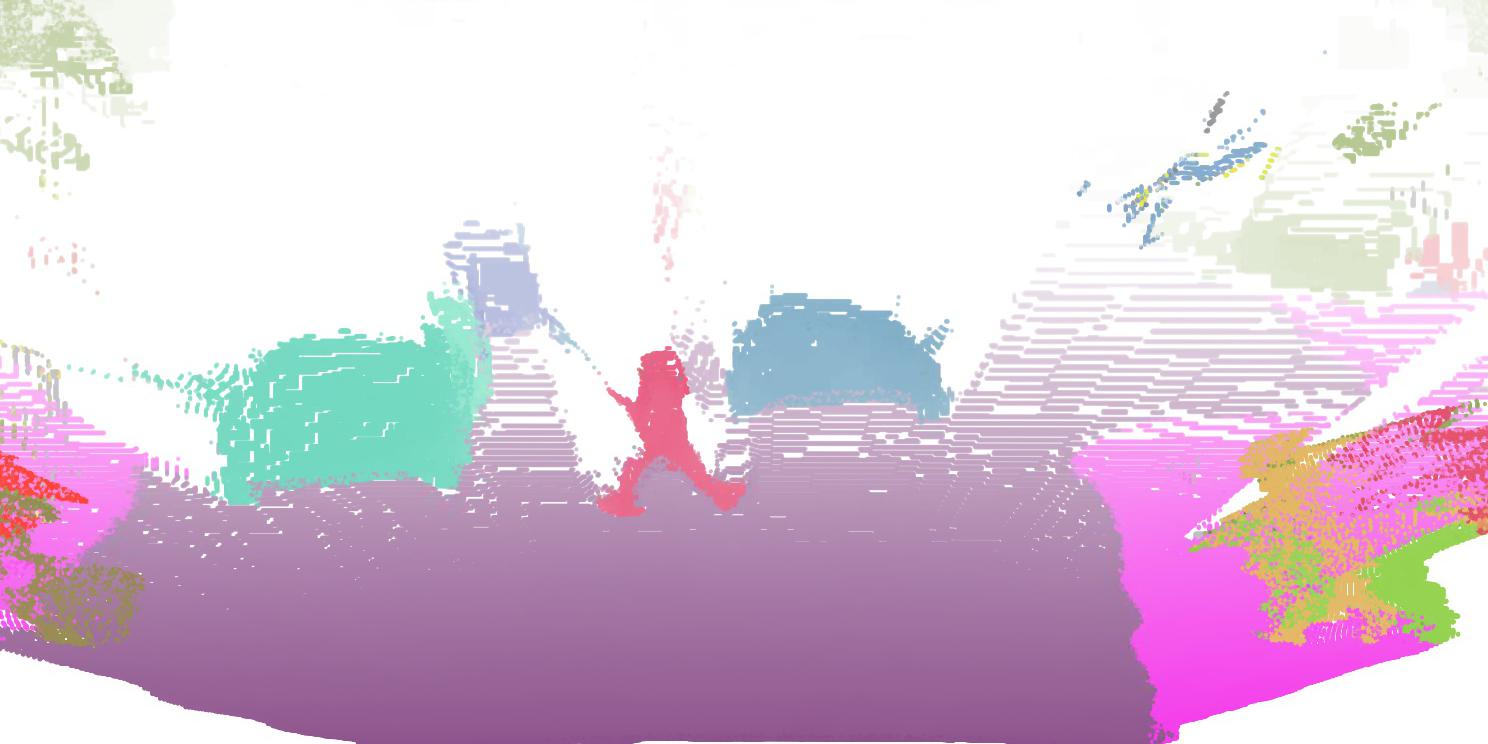}
\end{subfigure}\\\begin{subfigure}{.245\linewidth}
  \centering
  \includegraphics[trim={0 100 0 100},clip,width=\linewidth]{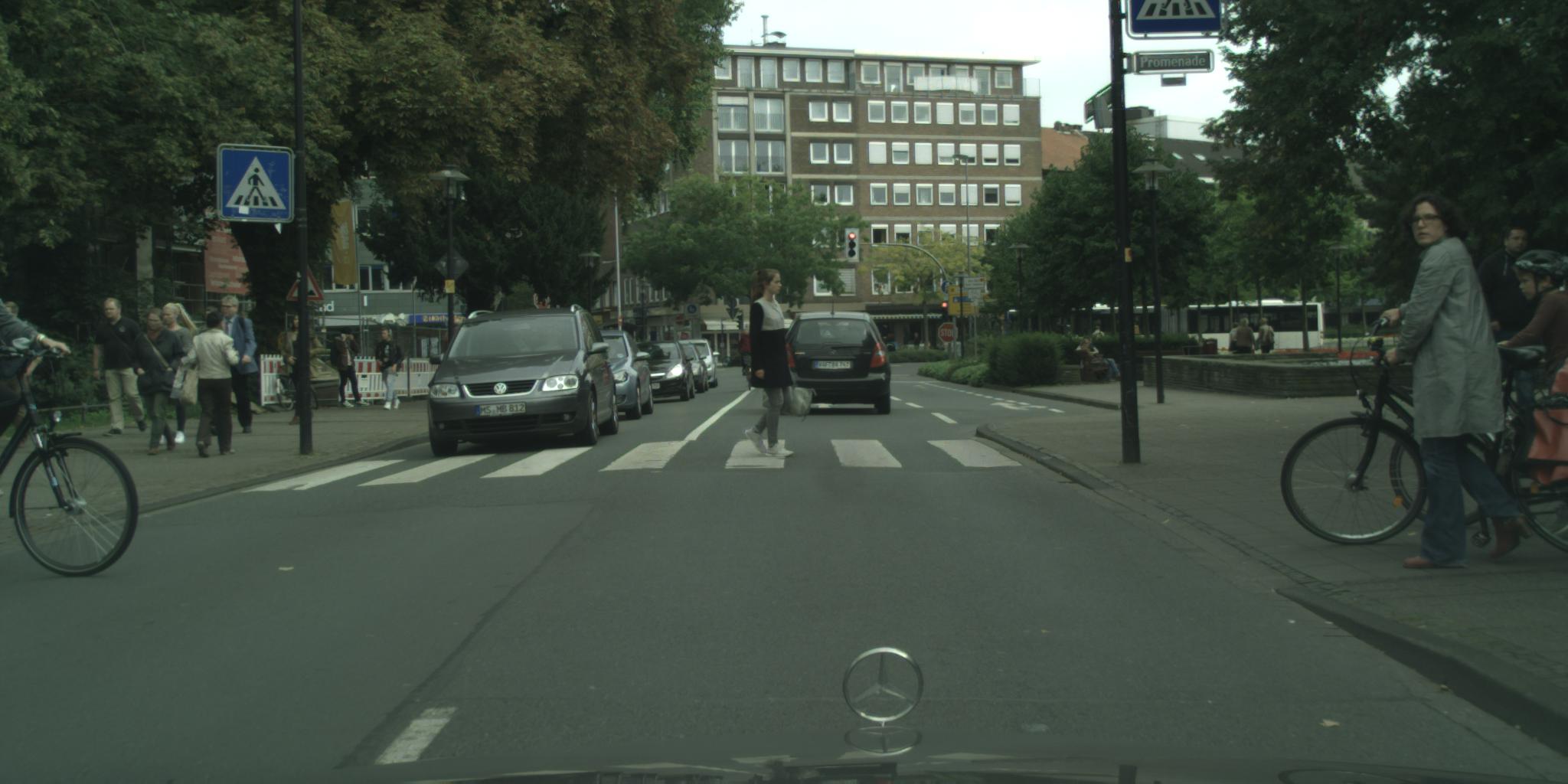}
\end{subfigure}\hfill\begin{subfigure}{.245\linewidth}
  \centering
  \includegraphics[trim={0 100 0 100},clip,width=\linewidth]{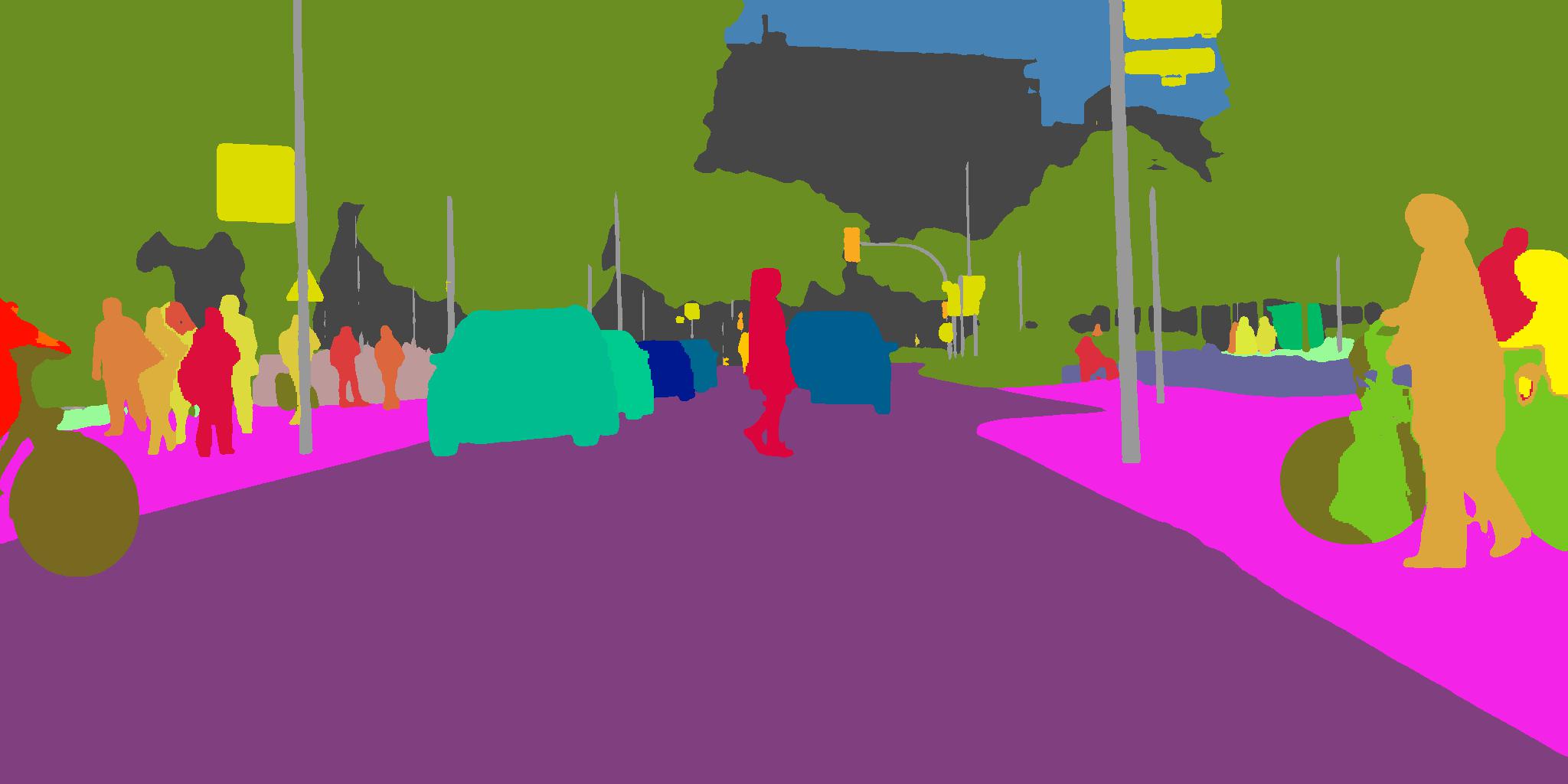}
\end{subfigure}\hfill\begin{subfigure}{.245\linewidth}
  \centering
  \includegraphics[trim={0 100 0 100},clip,width=\linewidth]{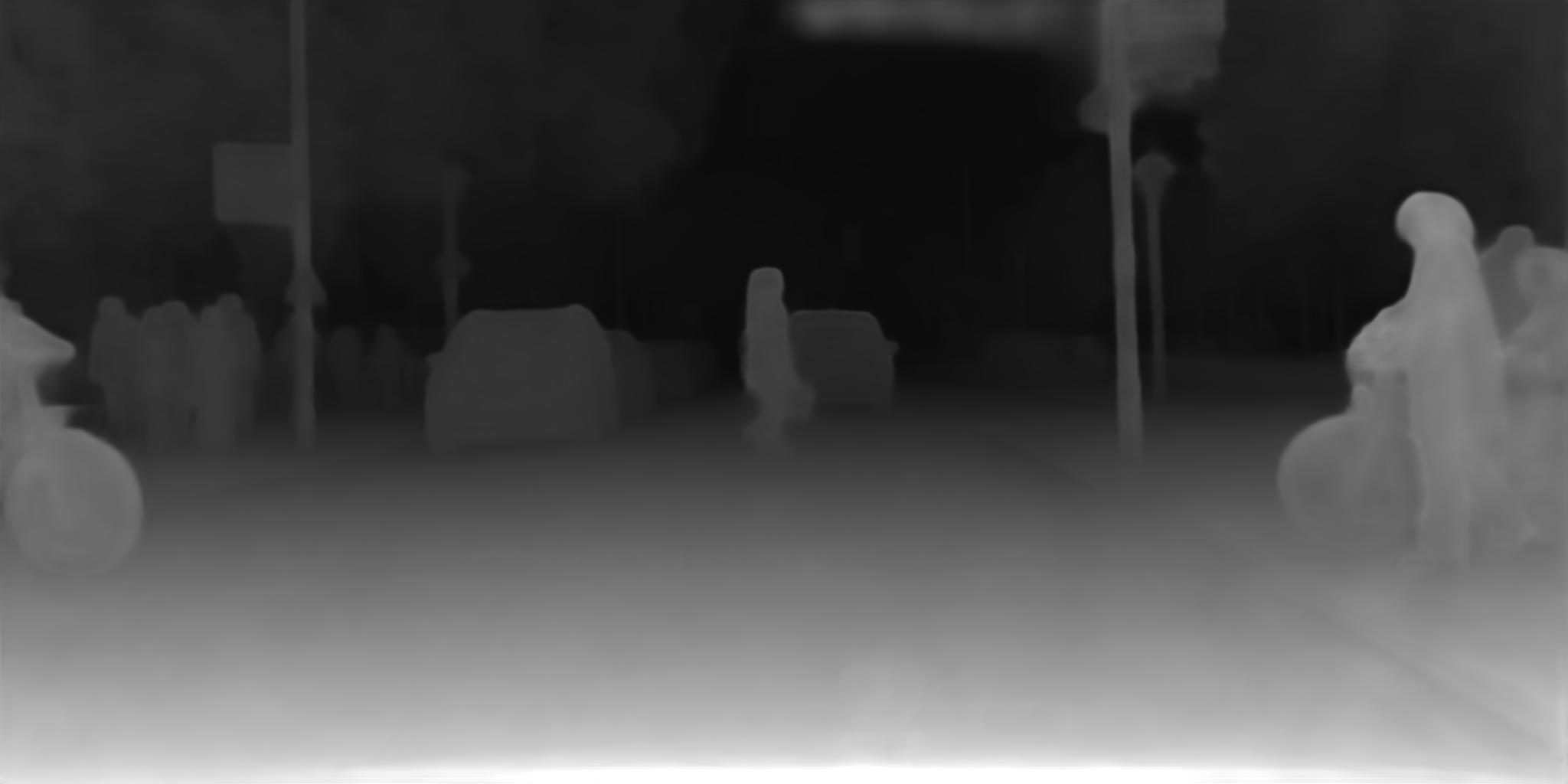}
\end{subfigure}\hfill\begin{subfigure}{.245\linewidth}
  \centering
  \includegraphics[trim={0 100 0 100},clip,width=\linewidth]{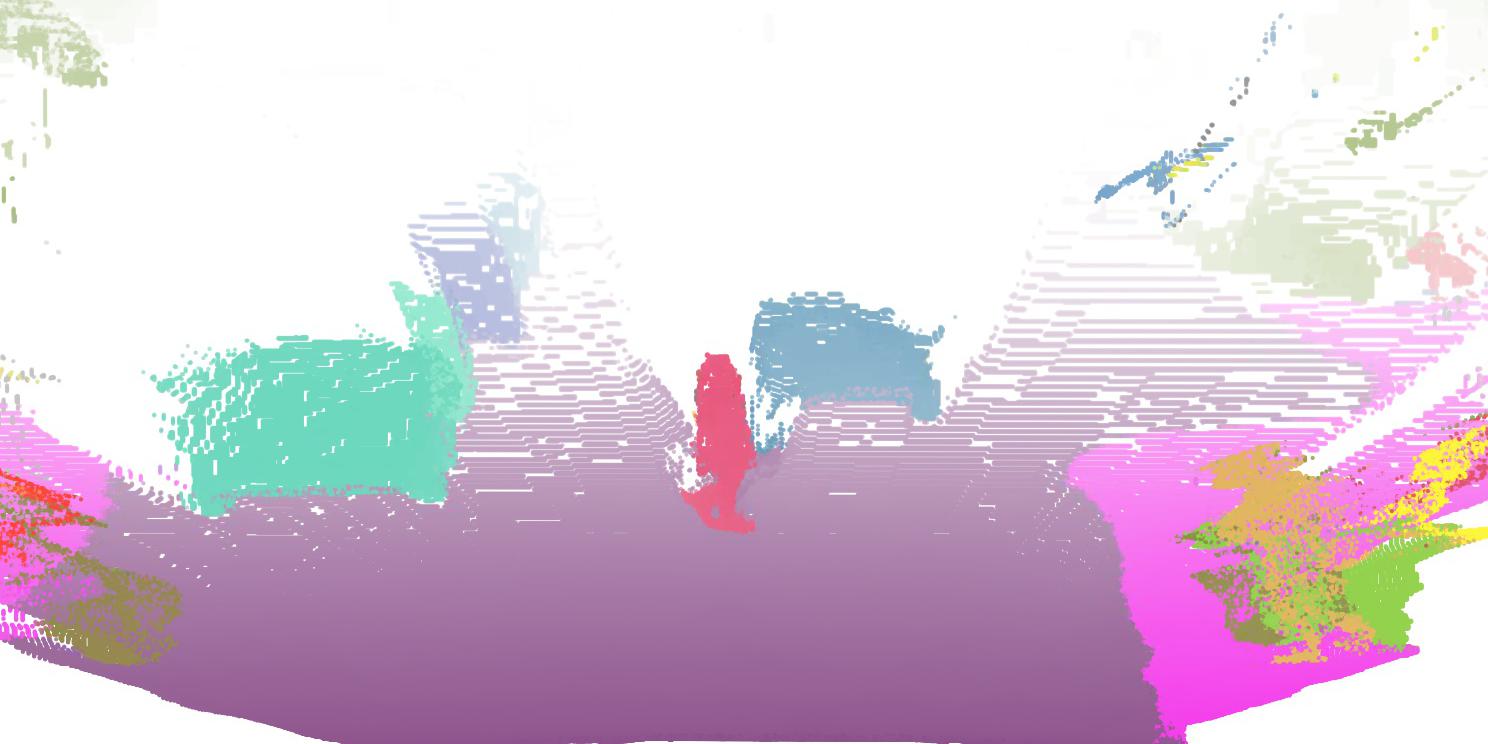}
\end{subfigure}\\\begin{subfigure}{.245\linewidth}
  \centering
  \includegraphics[trim={0 100 0 100},clip,width=\linewidth]{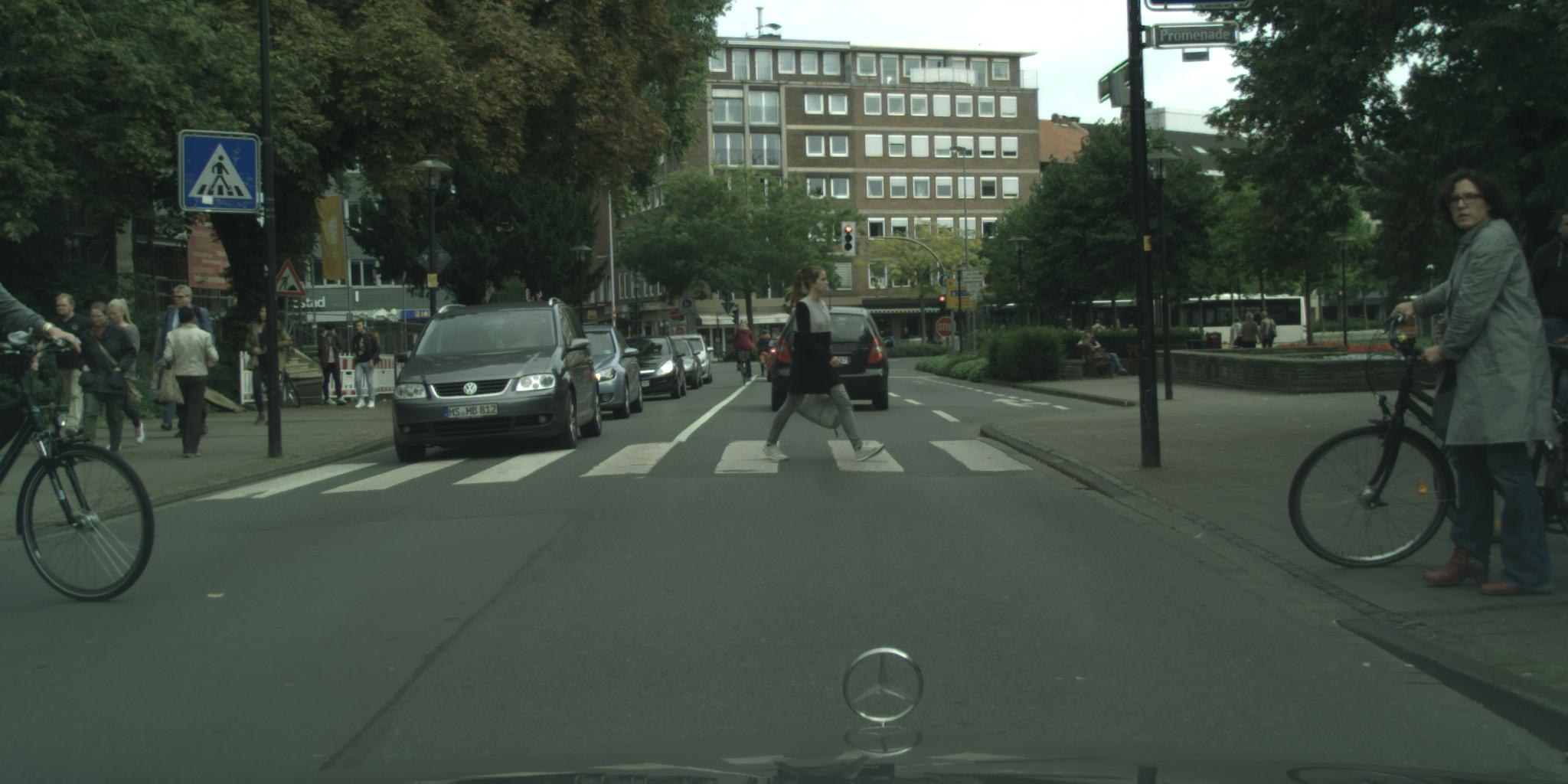}
\end{subfigure}\hfill\begin{subfigure}{.245\linewidth}
  \centering
  \includegraphics[trim={0 100 0 100},clip,width=\linewidth]{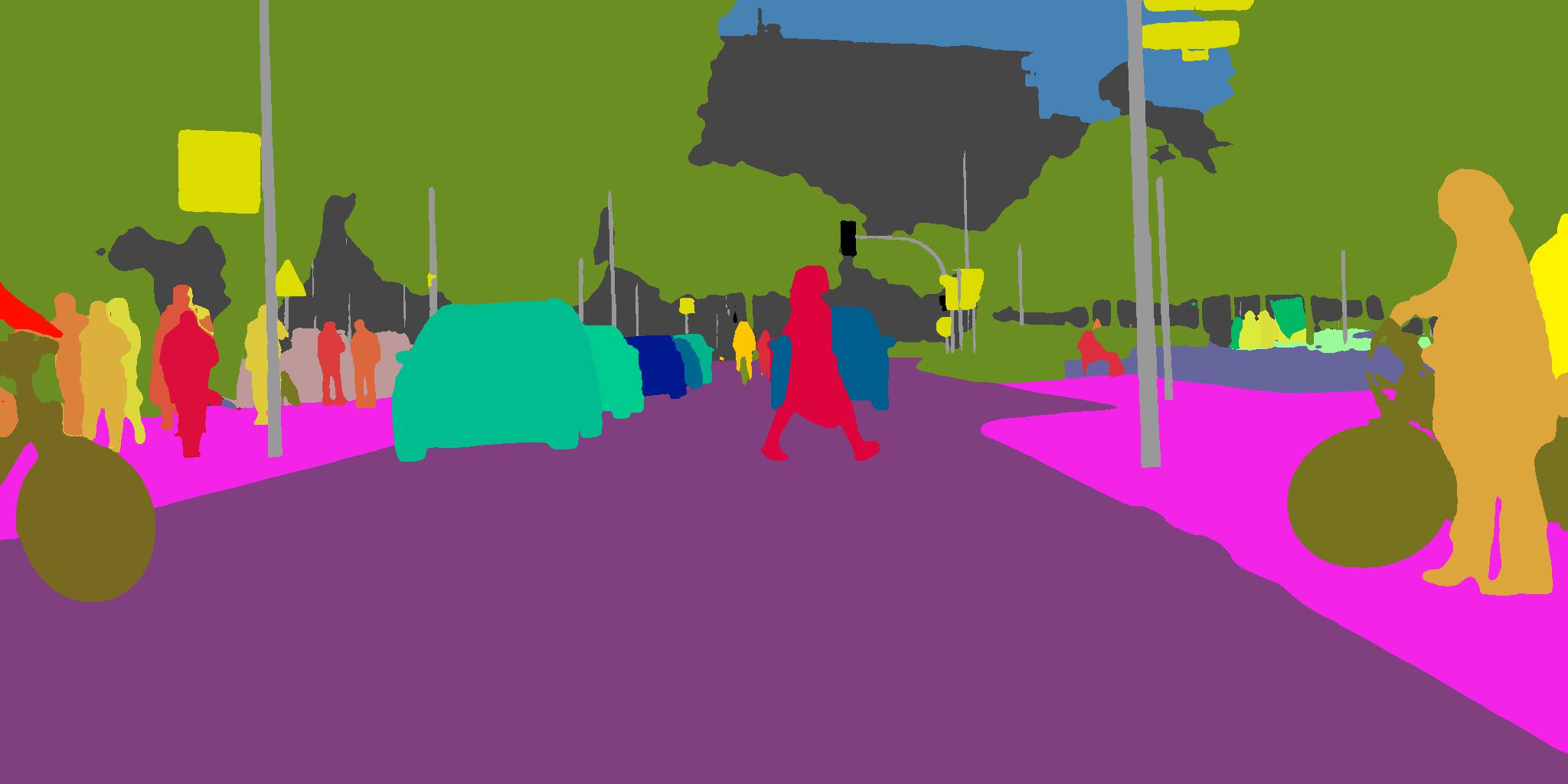}
\end{subfigure}\hfill\begin{subfigure}{.245\linewidth}
  \centering
  \includegraphics[trim={0 100 0 100},clip,width=\linewidth]{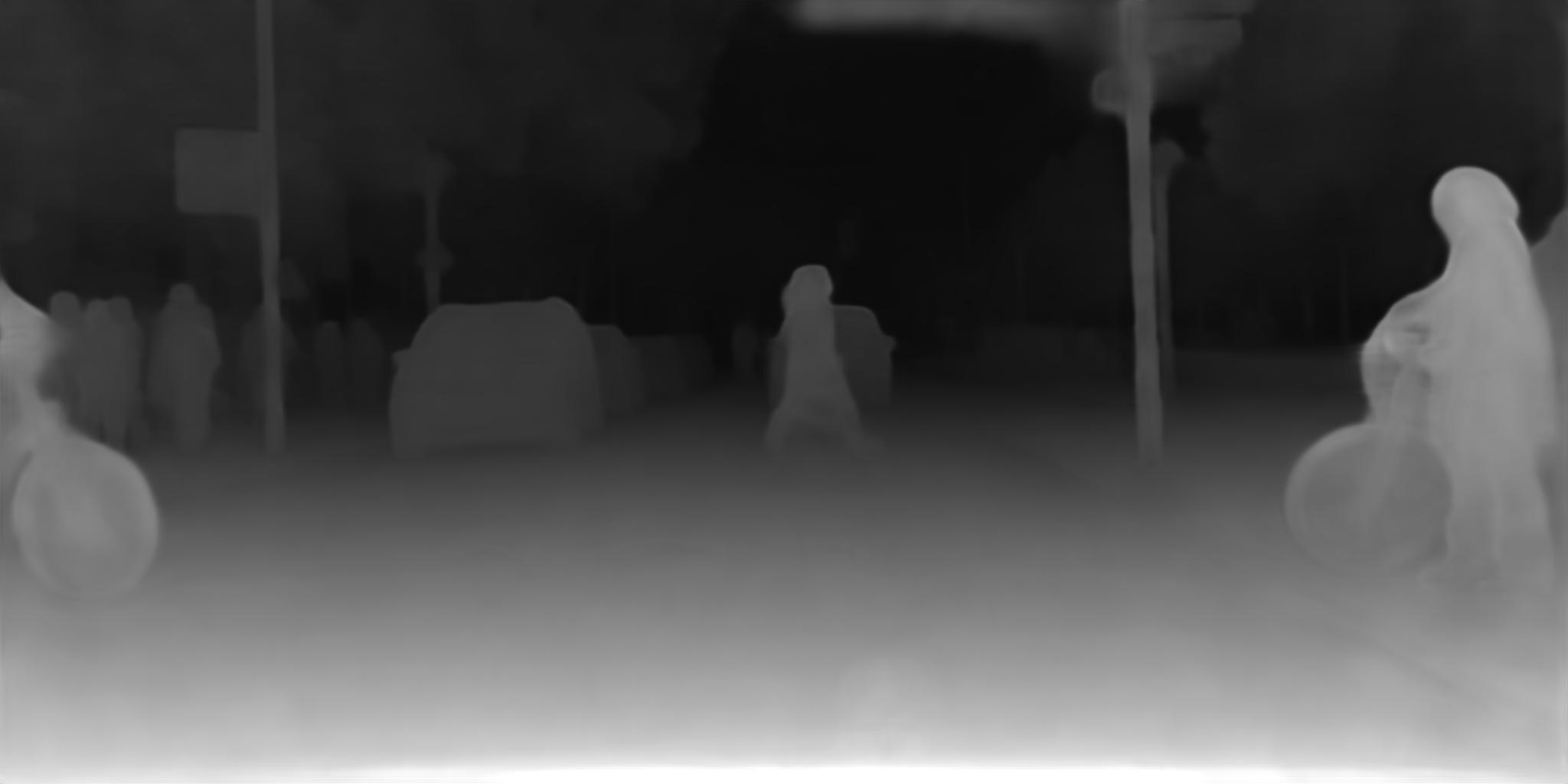}
\end{subfigure}\hfill\begin{subfigure}{.245\linewidth}
  \centering
  \includegraphics[trim={0 100 0 100},clip,width=\linewidth]{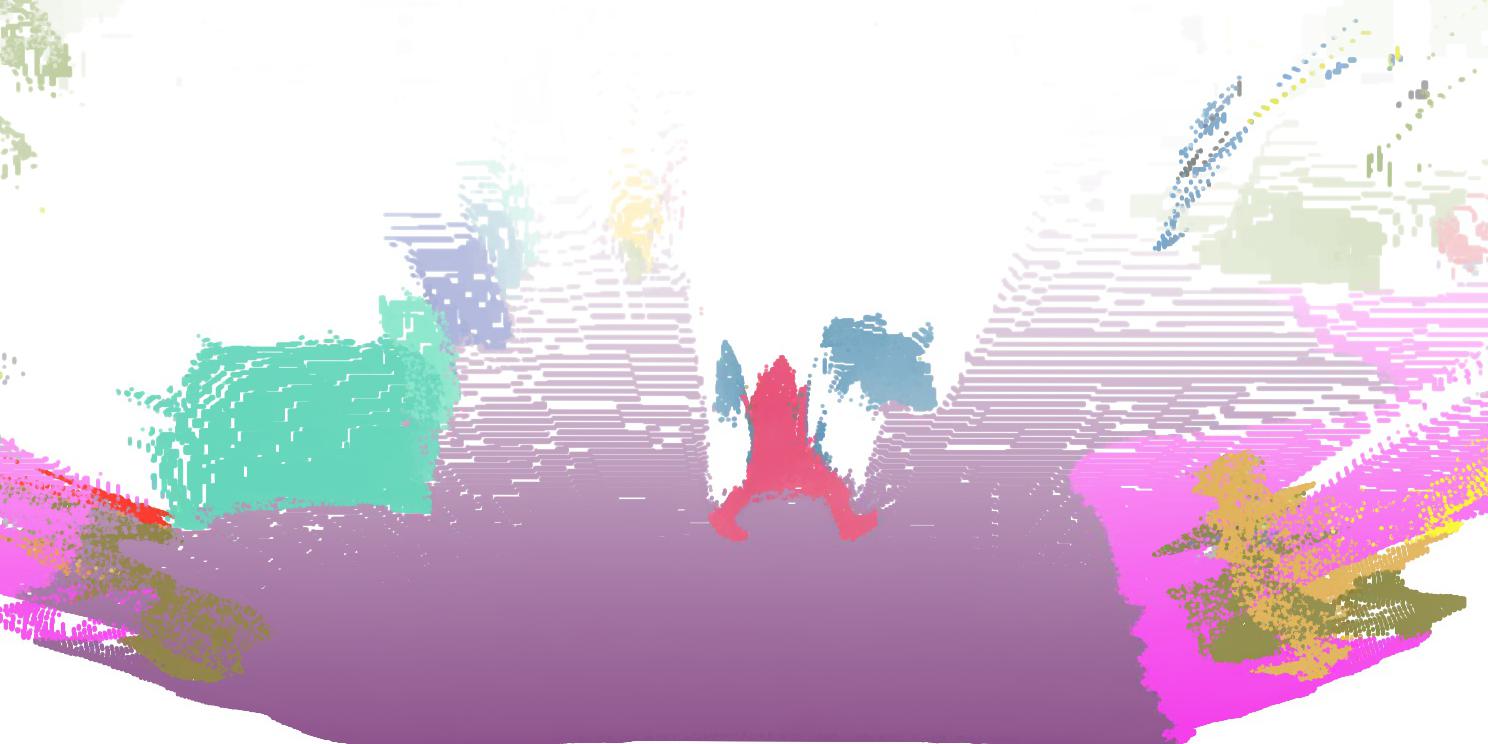}
\end{subfigure}\\
\caption{Prediction visualizations on Cityscapes-DVPS. From left to right: input image, temporally consistent panoptic segmentation prediction, monocular depth prediction, and point cloud visualization.}
\label{fig:cs_1}
\end{figure*}

\begin{figure*}[!t]
\centering
\begin{subfigure}{.245\linewidth}
  \centering
  \includegraphics[trim={0 100 0 100},clip,width=\linewidth]{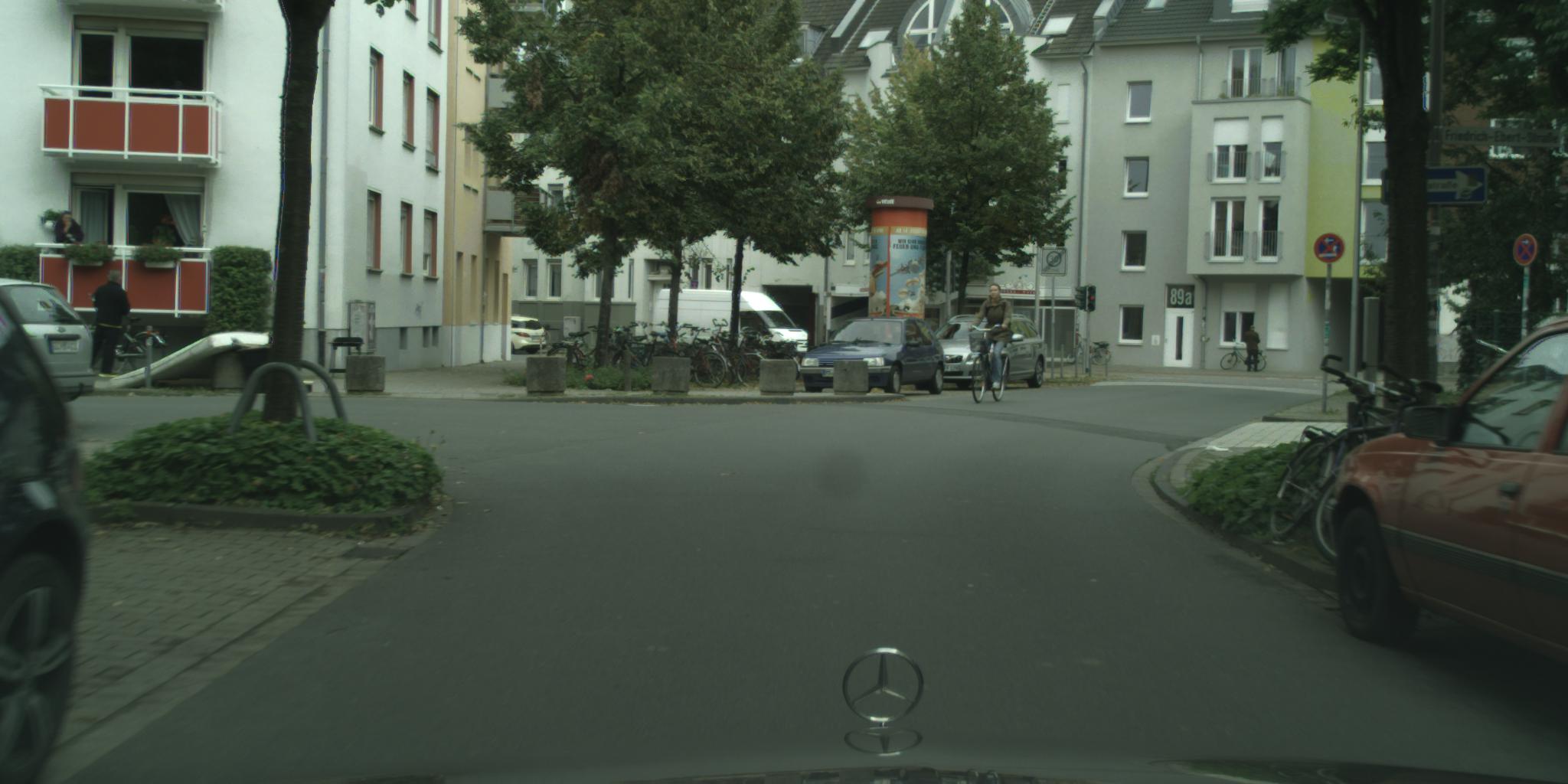}
\end{subfigure}\hfill\begin{subfigure}{.245\linewidth}
  \centering
  \includegraphics[trim={0 100 0 100},clip,width=\linewidth]{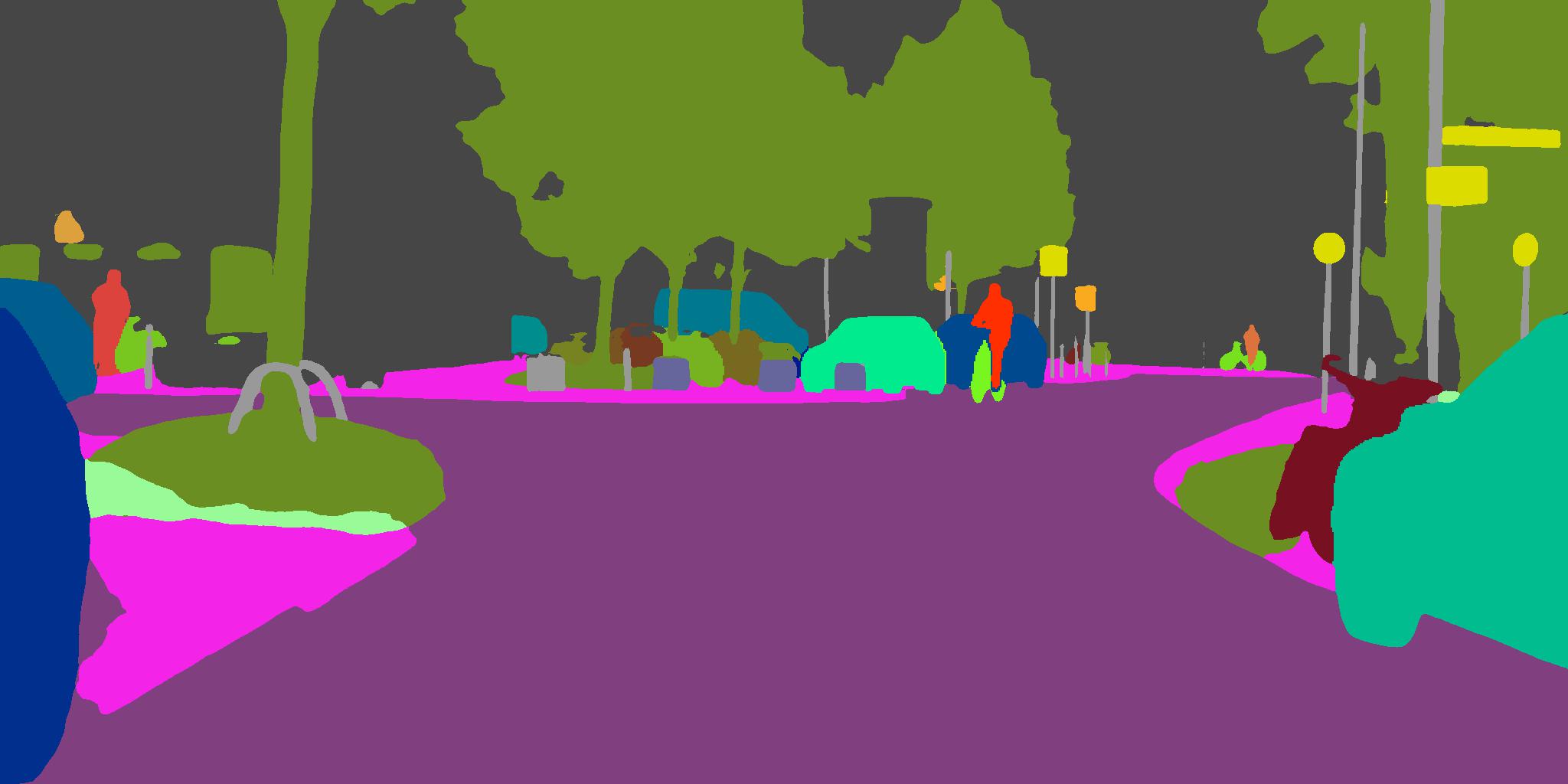}
\end{subfigure}\hfill\begin{subfigure}{.245\linewidth}
  \centering
  \includegraphics[trim={0 100 0 100},clip,width=\linewidth]{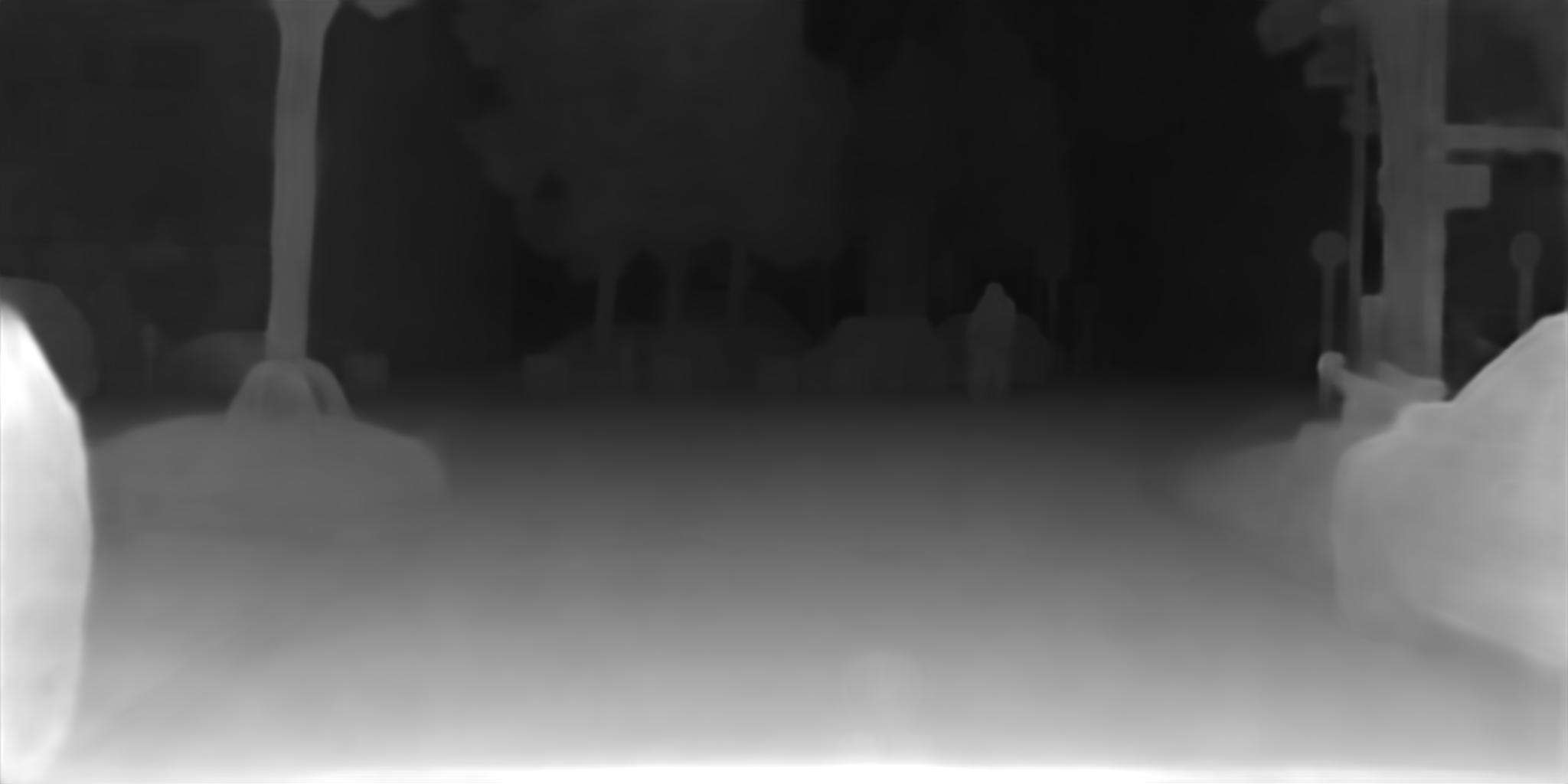}
\end{subfigure}\hfill\begin{subfigure}{.245\linewidth}
  \centering
  \includegraphics[trim={400 130 0 200},clip,width=\linewidth]{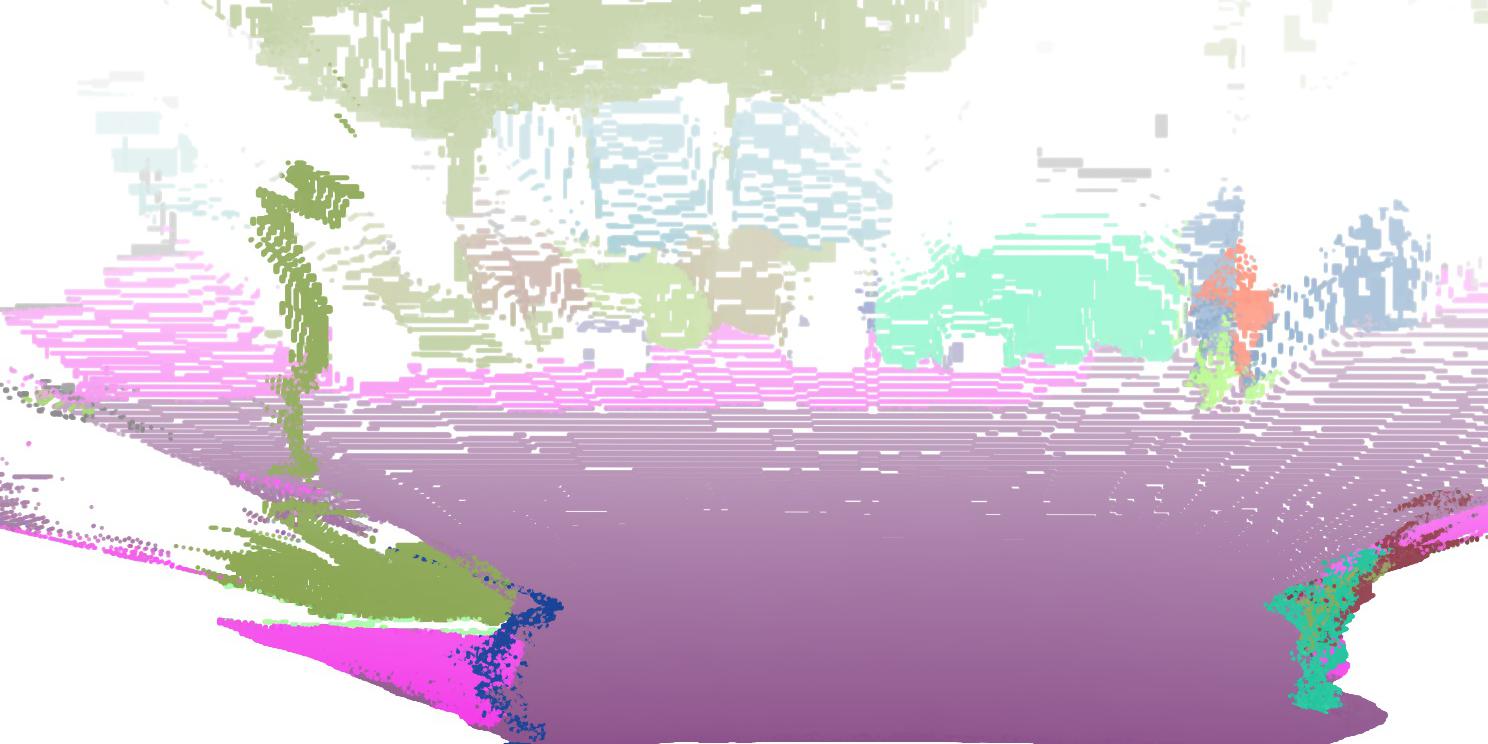}
\end{subfigure}\\\begin{subfigure}{.245\linewidth}
  \centering
  \includegraphics[trim={0 100 0 100},clip,width=\linewidth]{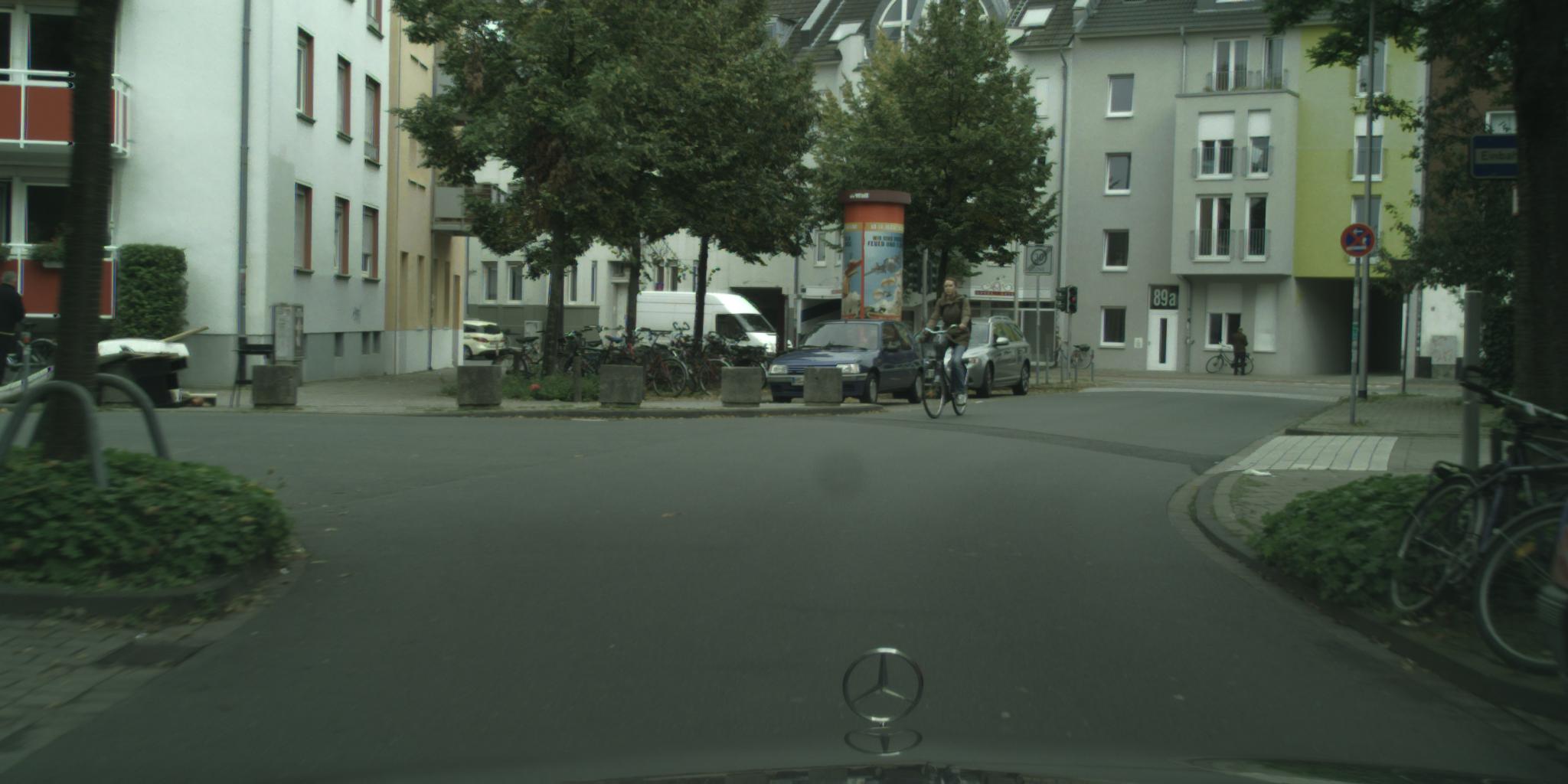}
\end{subfigure}\hfill\begin{subfigure}{.245\linewidth}
  \centering
  \includegraphics[trim={0 100 0 100},clip,width=\linewidth]{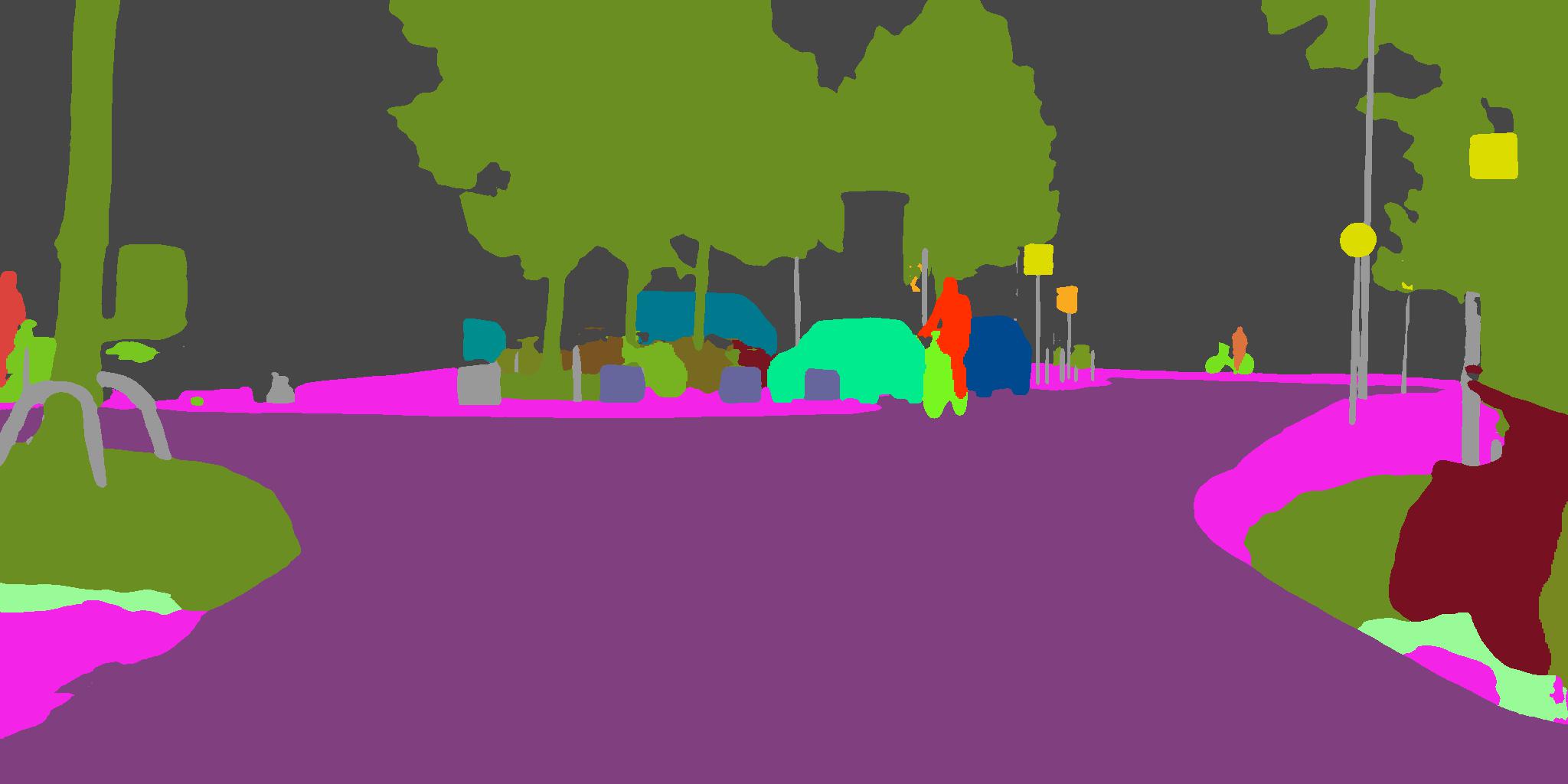}
\end{subfigure}\hfill\begin{subfigure}{.245\linewidth}
  \centering
  \includegraphics[trim={0 100 0 100},clip,width=\linewidth]{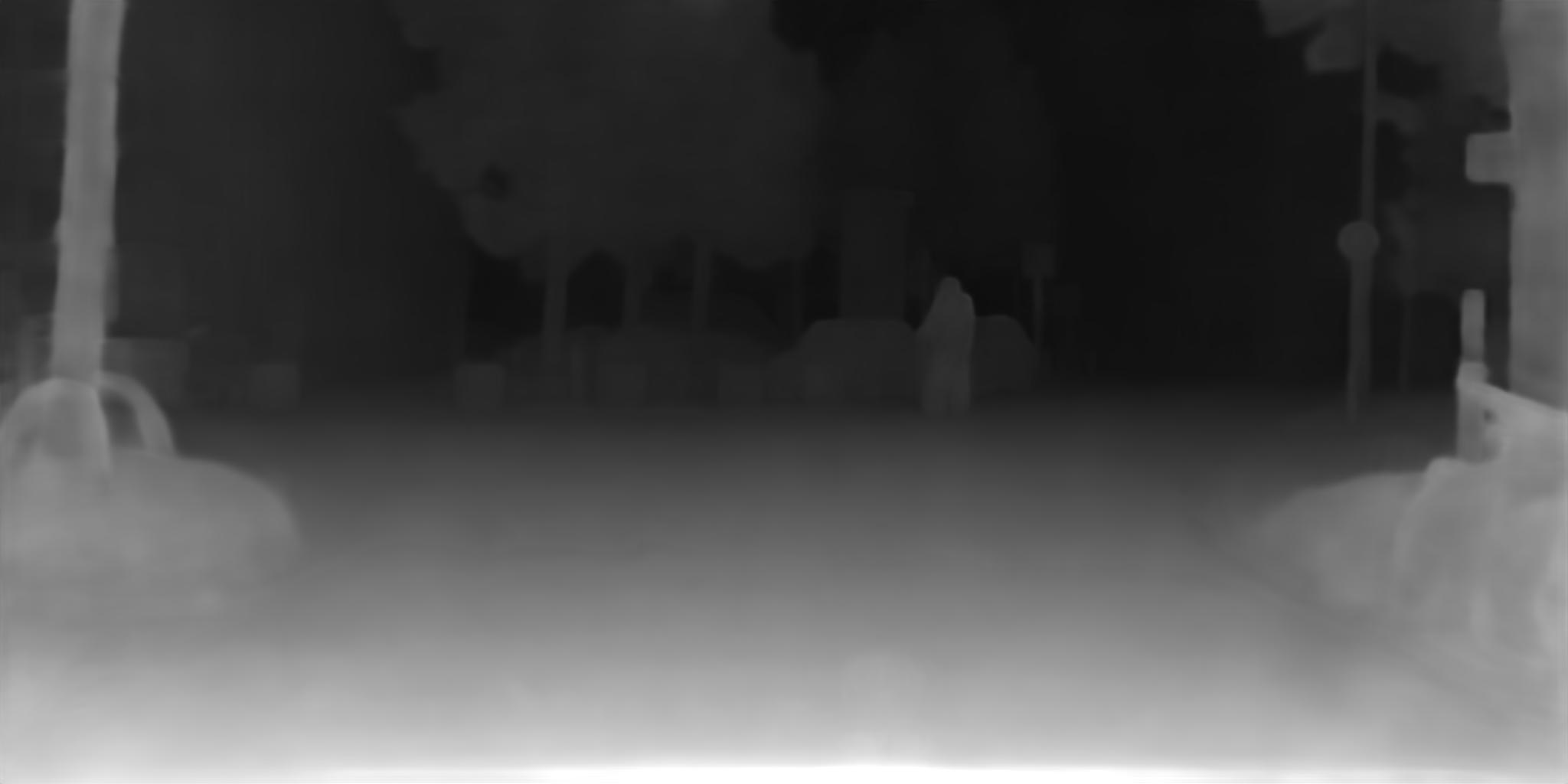}
\end{subfigure}\hfill\begin{subfigure}{.245\linewidth}
  \centering
  \includegraphics[trim={400 130 0 200},clip,width=\linewidth]{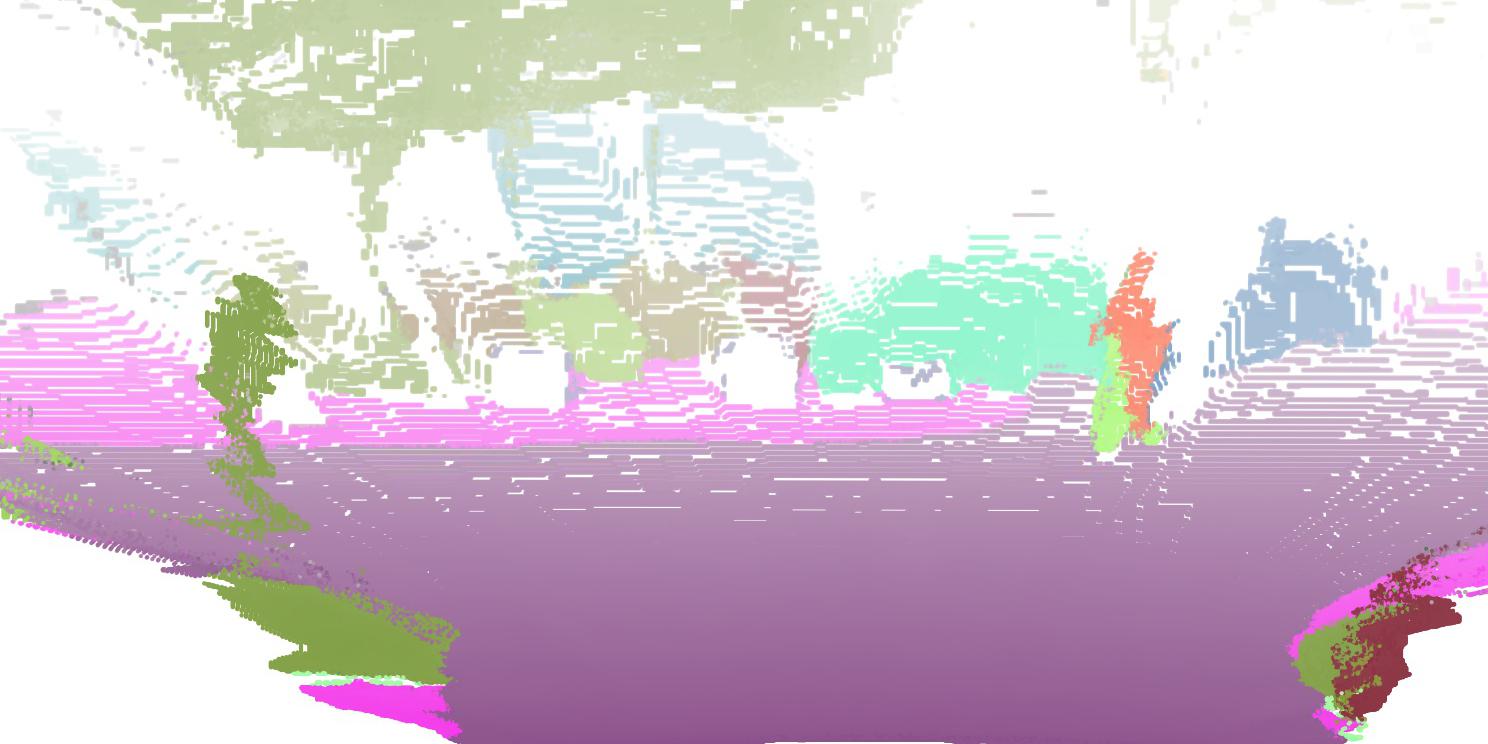}
\end{subfigure}\\\begin{subfigure}{.245\linewidth}
  \centering
  \includegraphics[trim={0 100 0 100},clip,width=\linewidth]{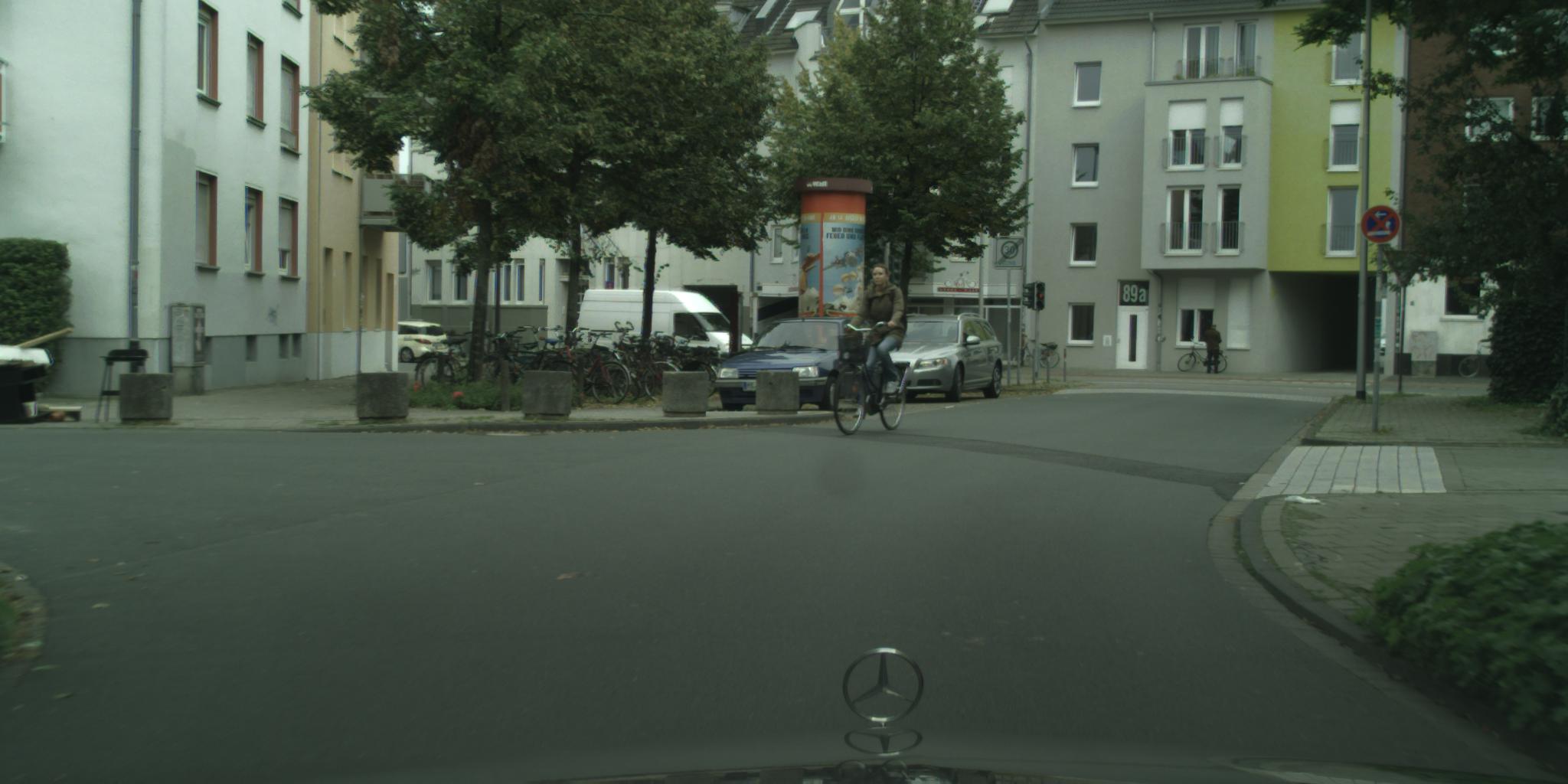}
\end{subfigure}\hfill\begin{subfigure}{.245\linewidth}
  \centering
  \includegraphics[trim={0 100 0 100},clip,width=\linewidth]{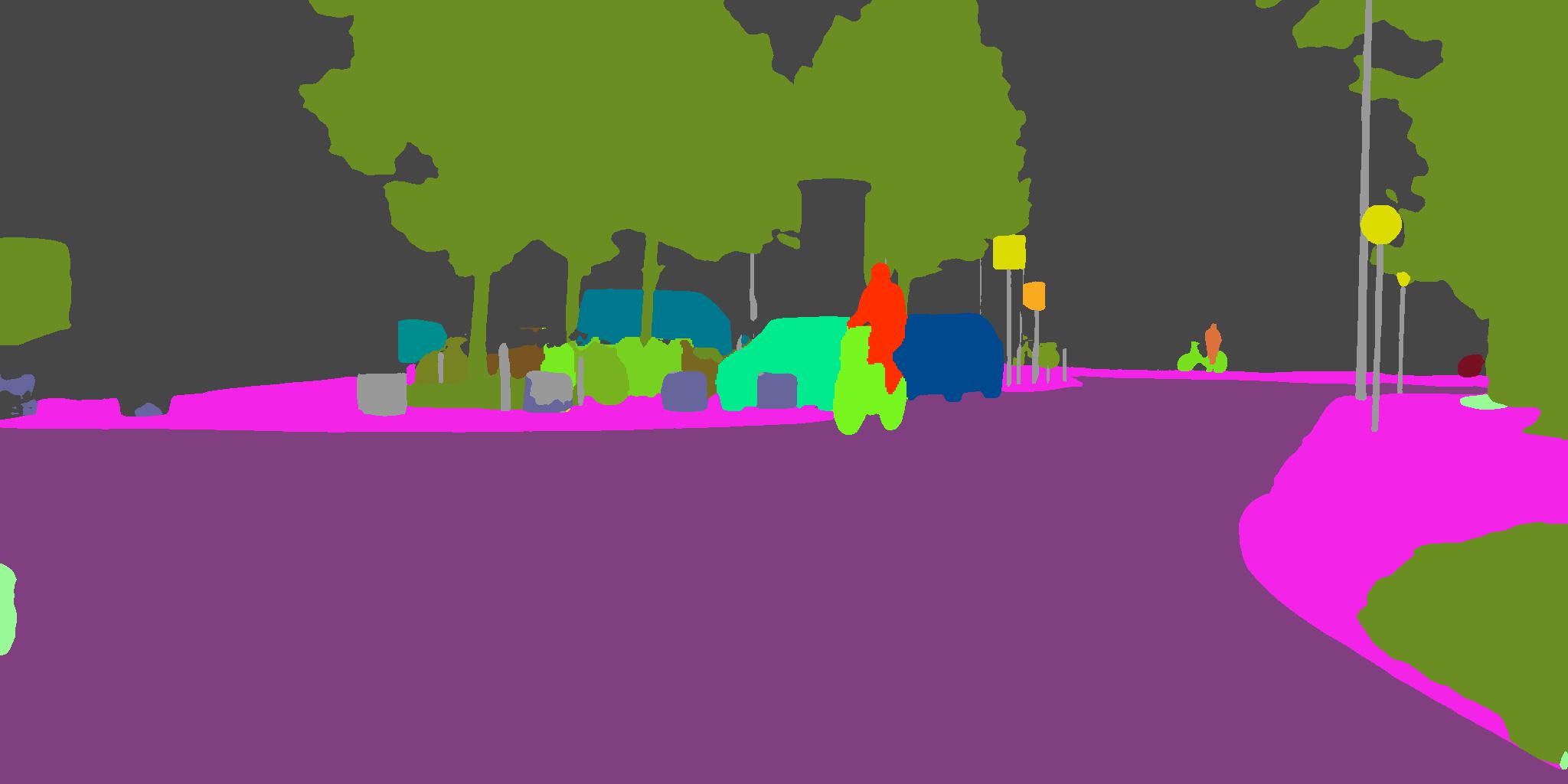}
\end{subfigure}\hfill\begin{subfigure}{.245\linewidth}
  \centering
  \includegraphics[trim={0 100 0 100},clip,width=\linewidth]{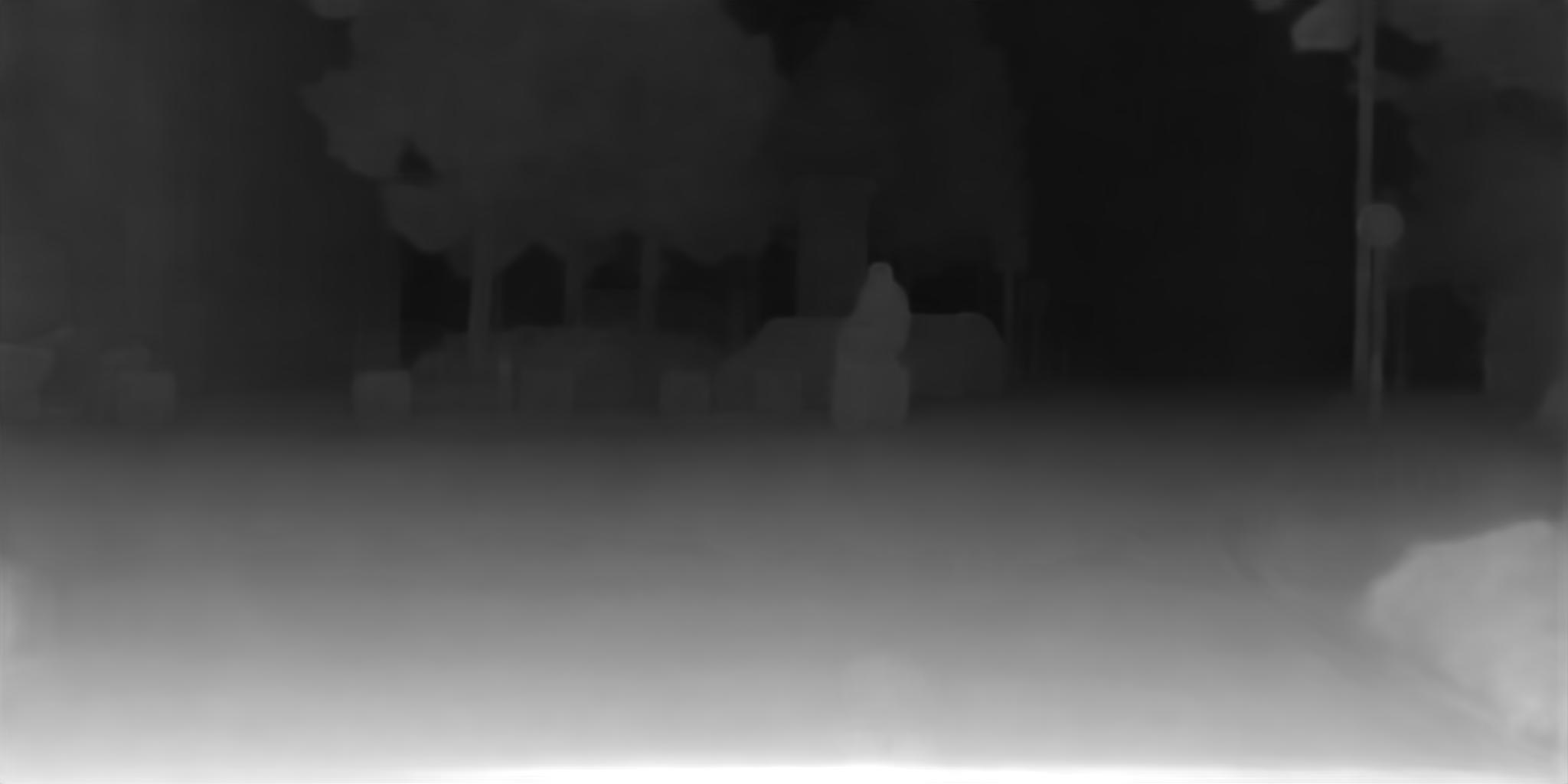}
\end{subfigure}\hfill\begin{subfigure}{.245\linewidth}
  \centering
  \includegraphics[trim={400 130 0 200},clip,width=\linewidth]{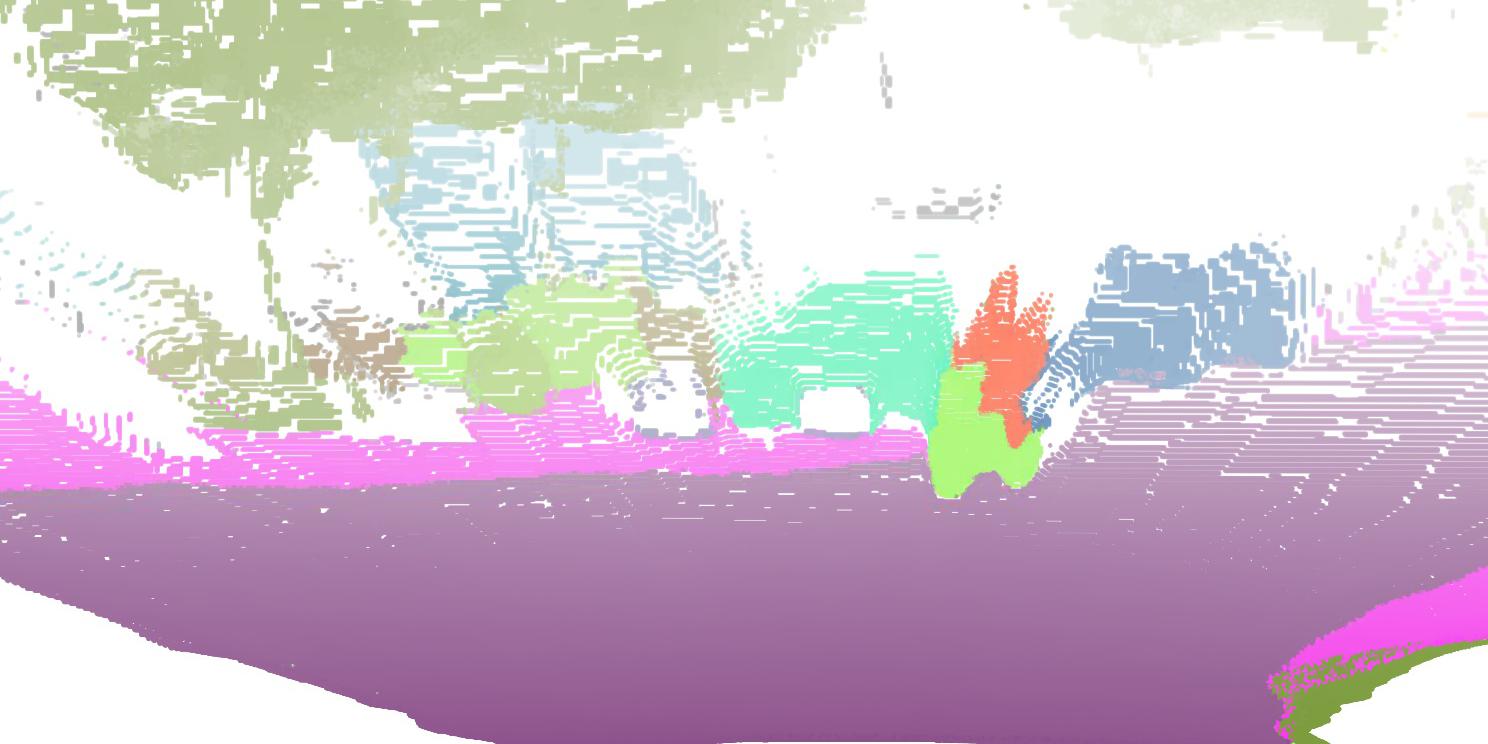}
\end{subfigure}\\\begin{subfigure}{.245\linewidth}
  \centering
  \includegraphics[trim={0 100 0 100},clip,width=\linewidth]{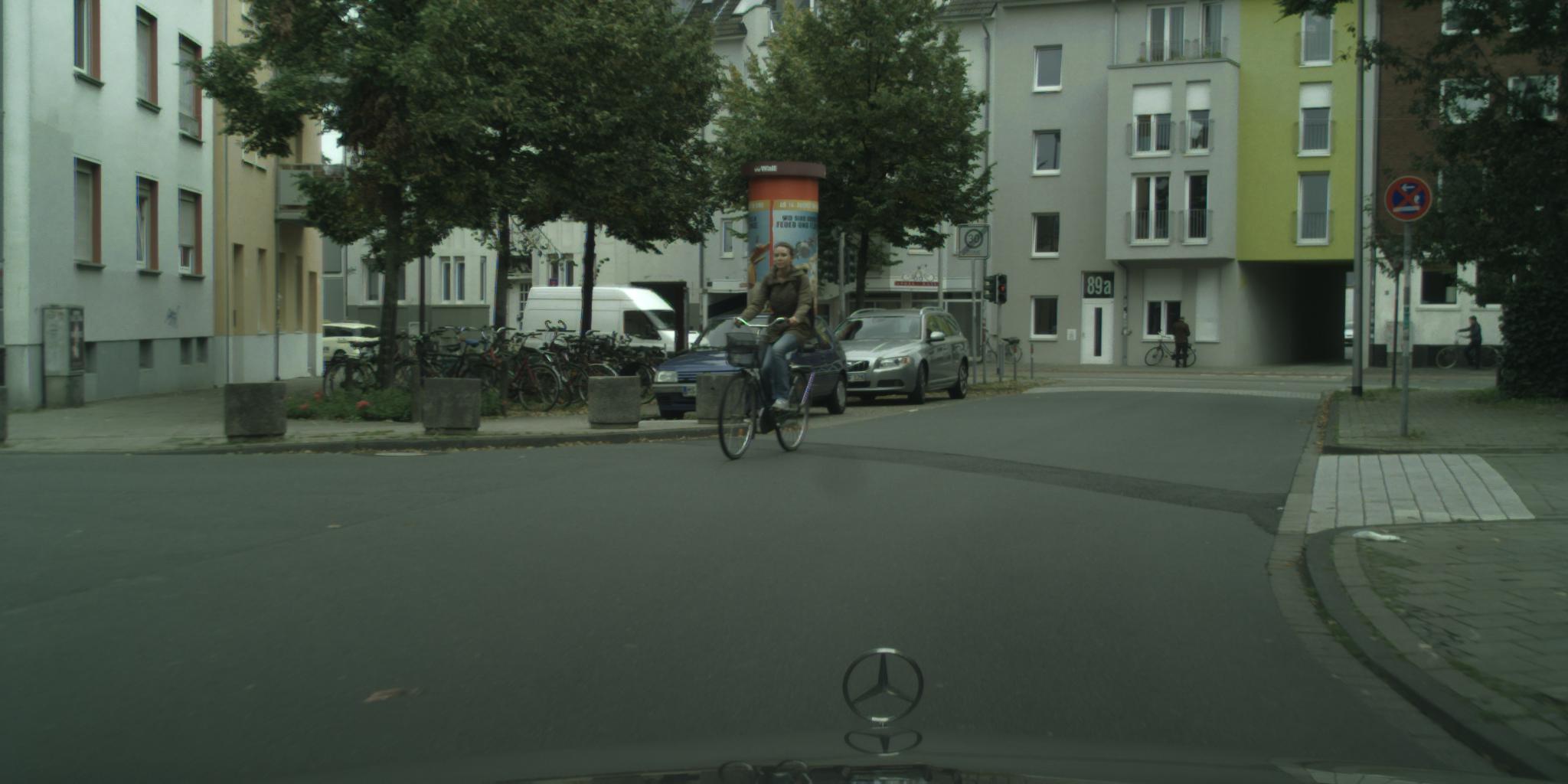}
\end{subfigure}\hfill\begin{subfigure}{.245\linewidth}
  \centering
  \includegraphics[trim={0 100 0 100},clip,width=\linewidth]{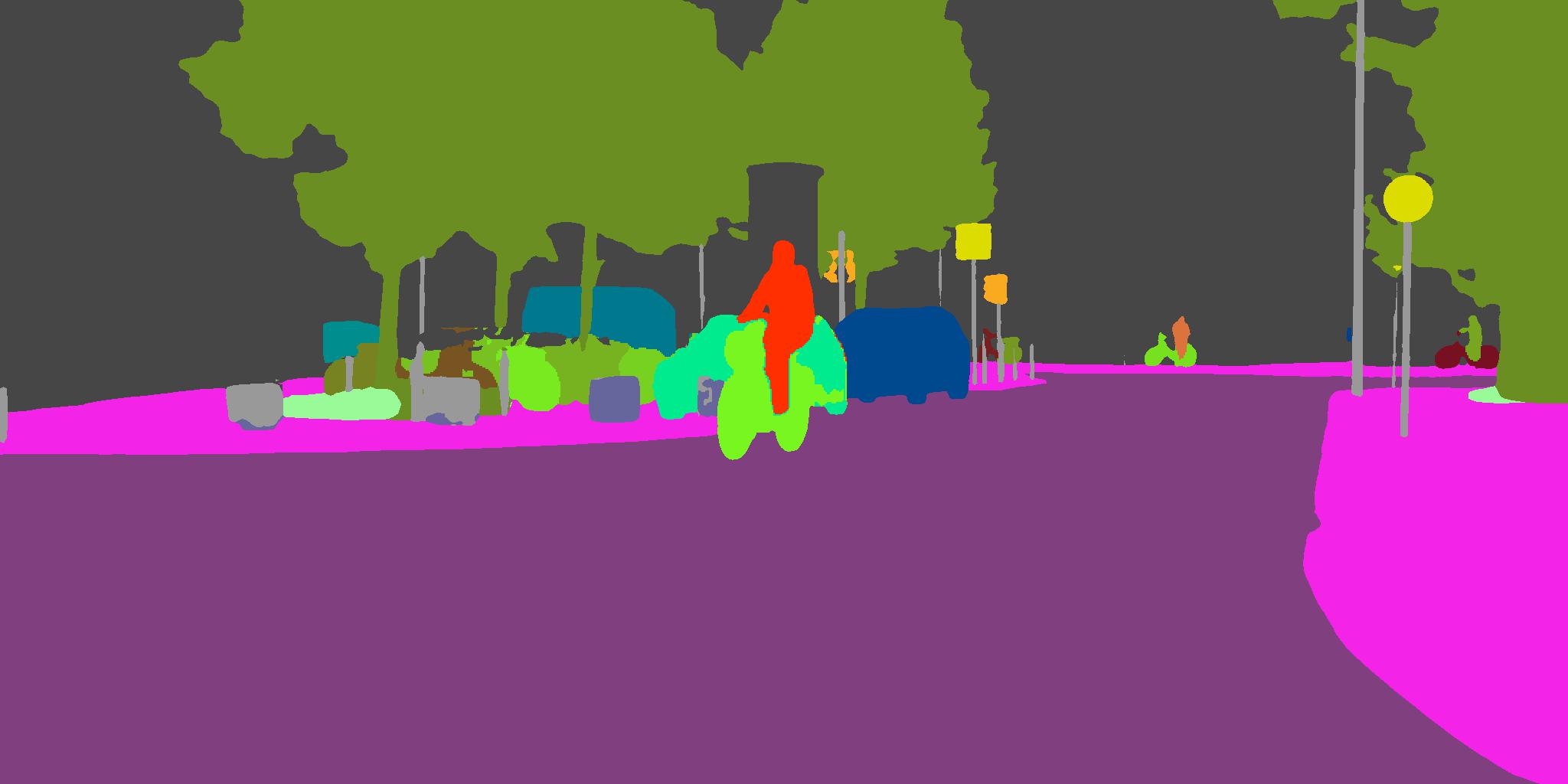}
\end{subfigure}\hfill\begin{subfigure}{.245\linewidth}
  \centering
  \includegraphics[trim={0 100 0 100},clip,width=\linewidth]{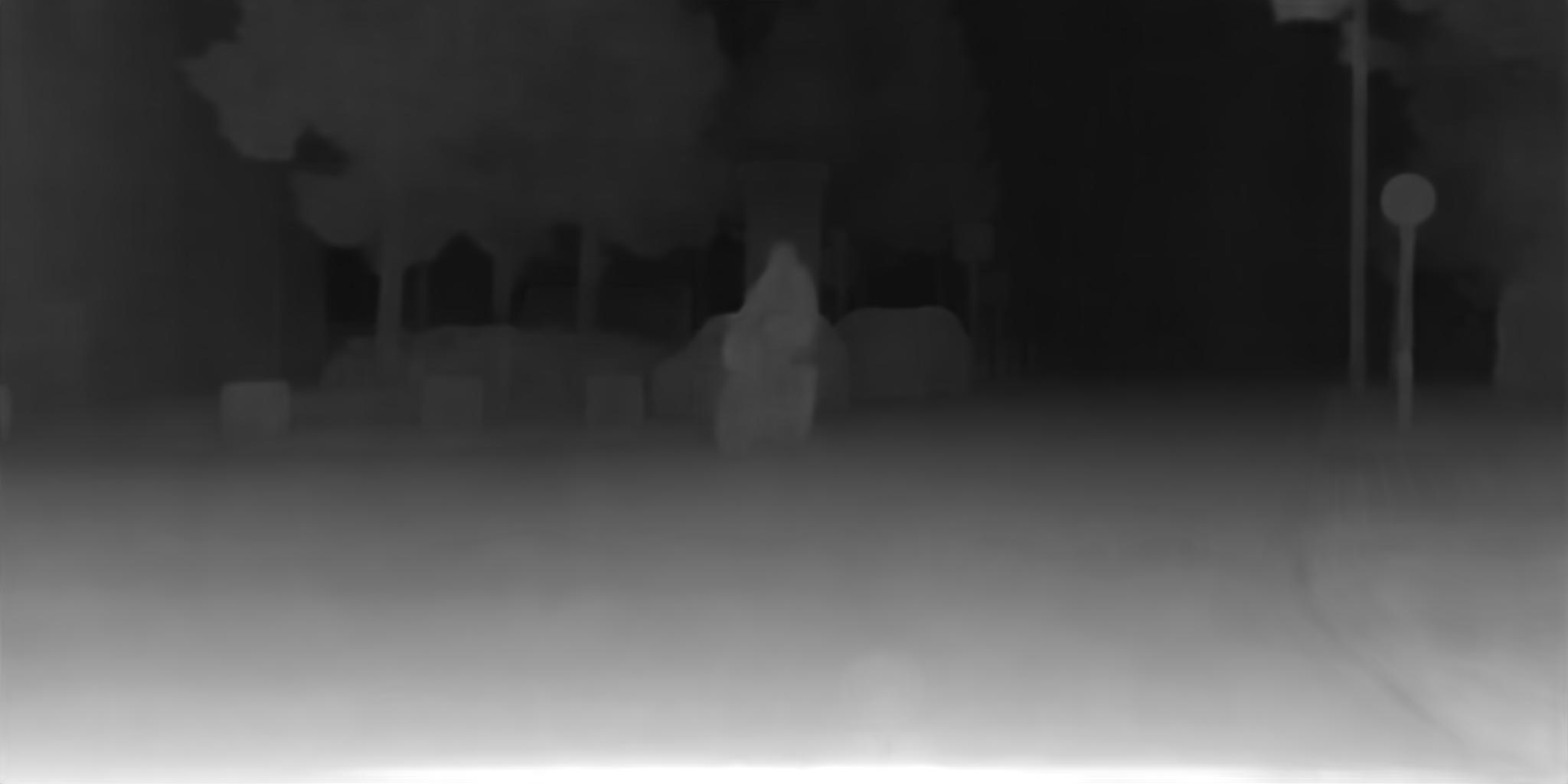}
\end{subfigure}\hfill\begin{subfigure}{.245\linewidth}
  \centering
  \includegraphics[trim={400 130 0 200},clip,width=\linewidth]{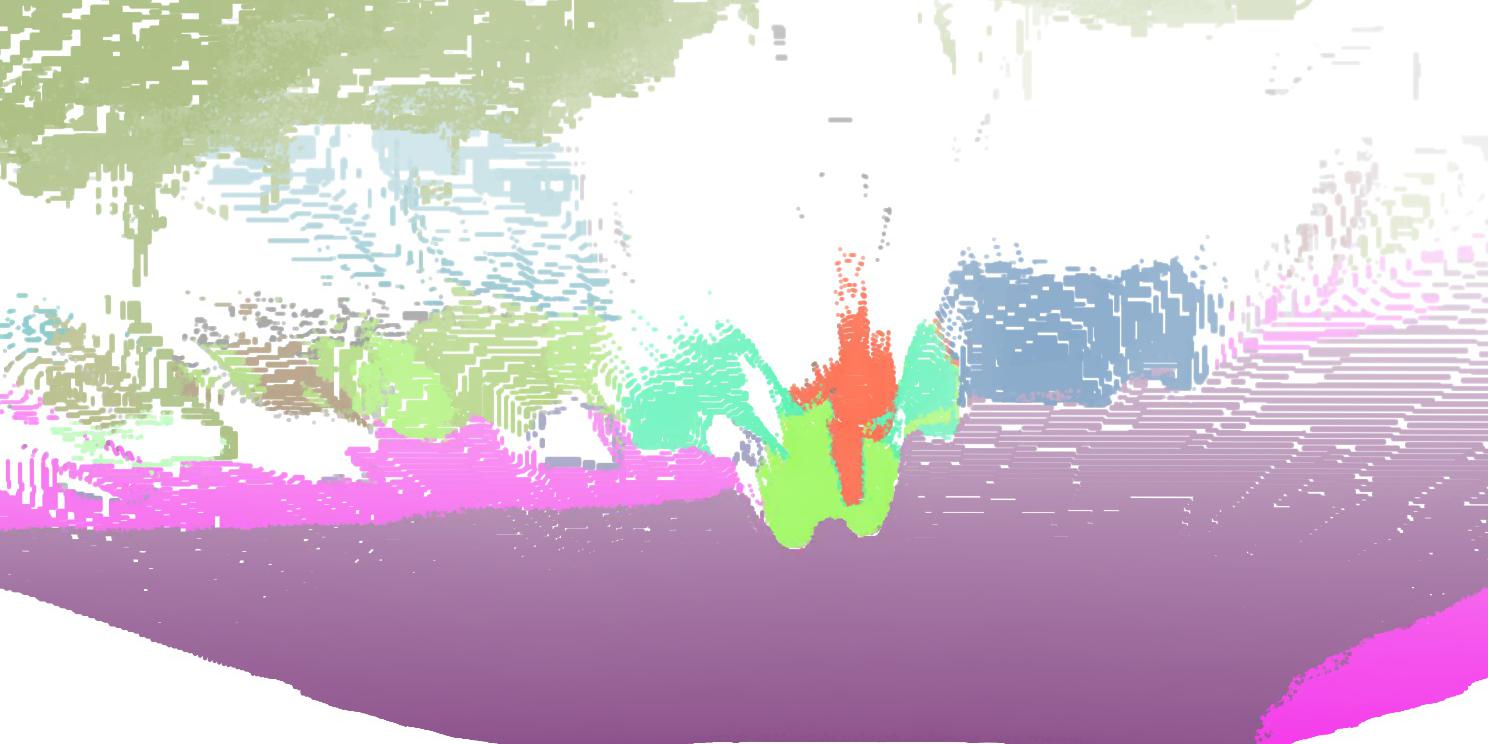}
\end{subfigure}\\\begin{subfigure}{.245\linewidth}
  \centering
  \includegraphics[trim={0 100 0 100},clip,width=\linewidth]{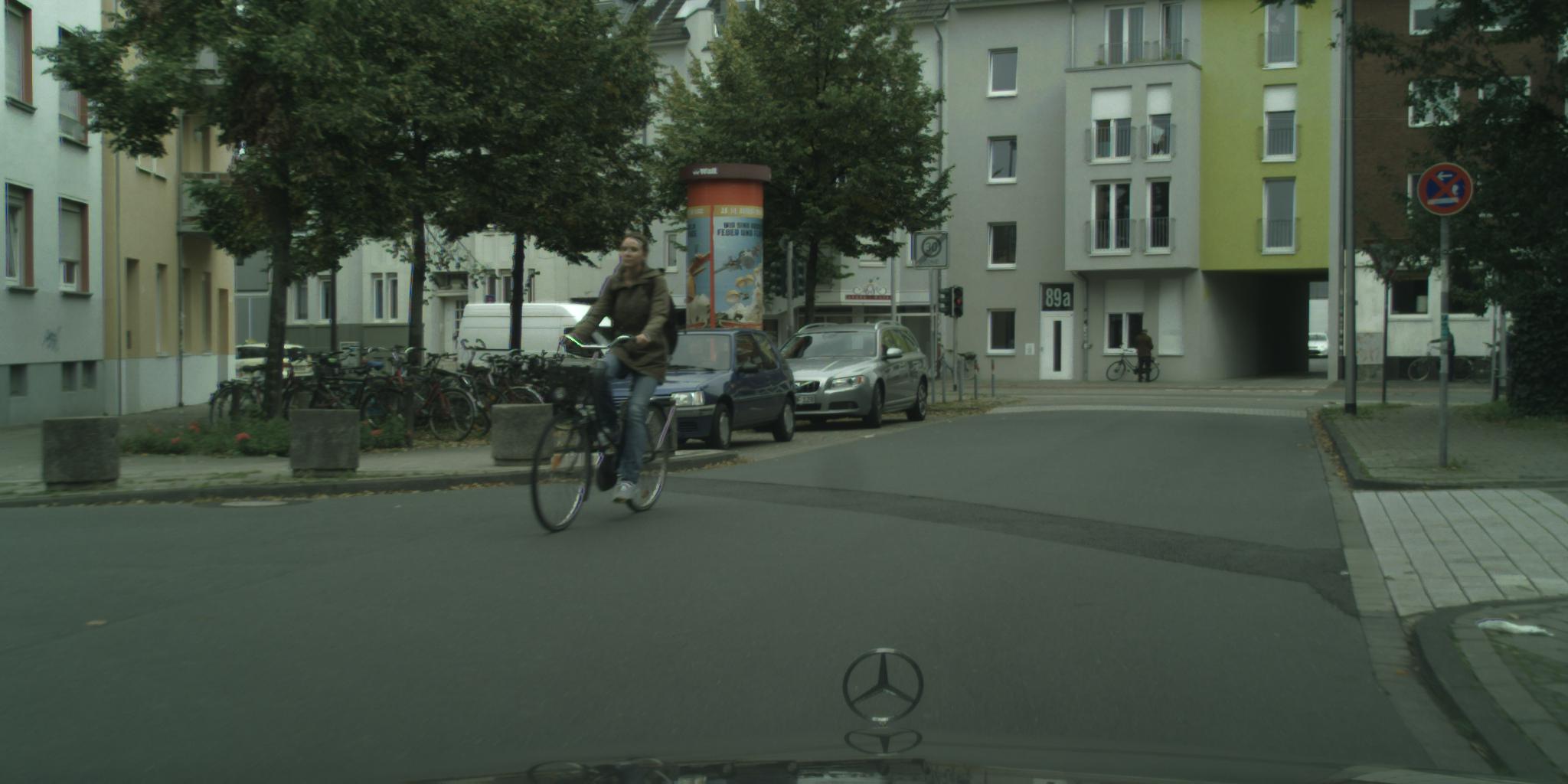}
\end{subfigure}\hfill\begin{subfigure}{.245\linewidth}
  \centering
  \includegraphics[trim={0 100 0 100},clip,width=\linewidth]{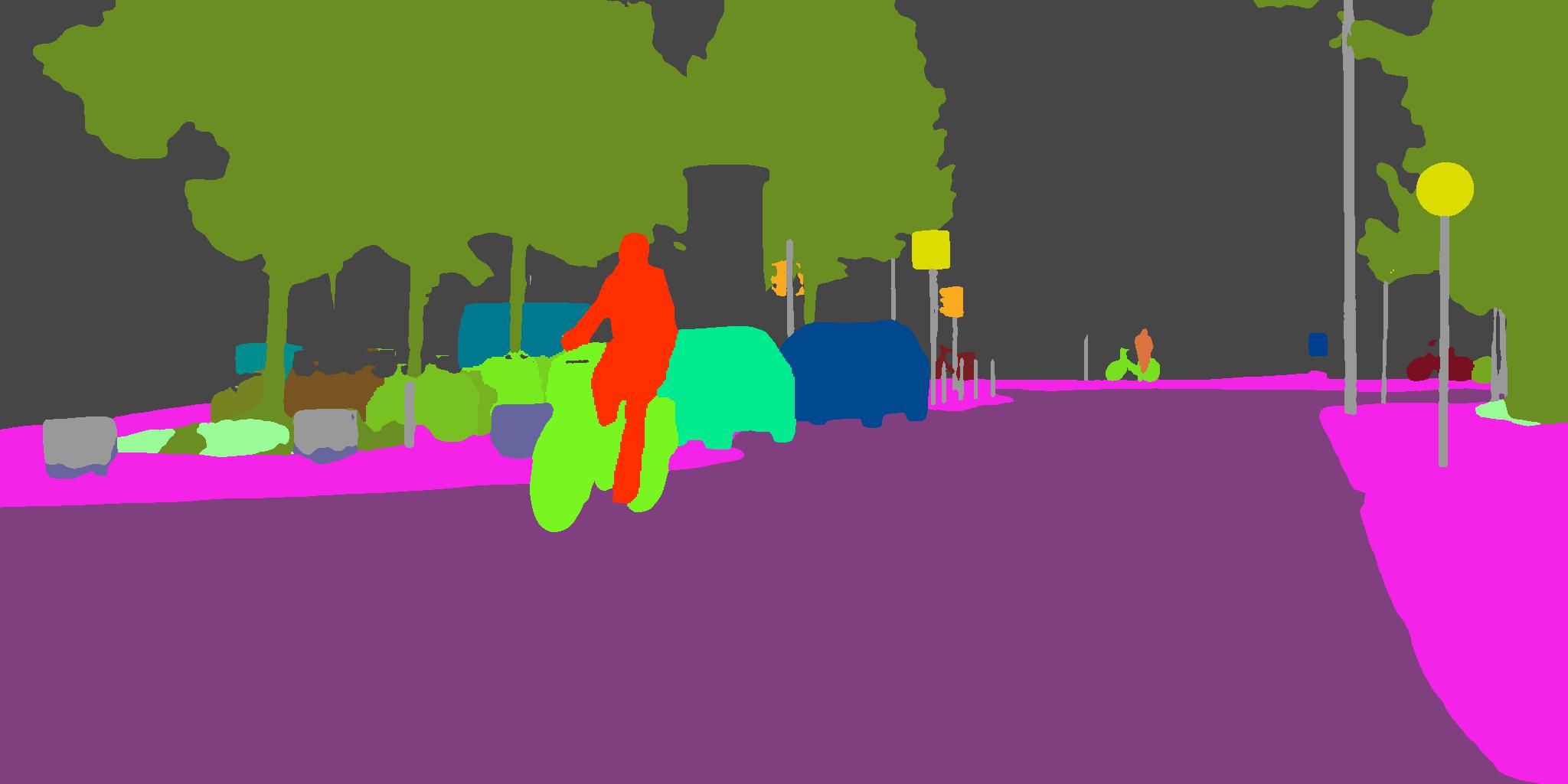}
\end{subfigure}\hfill\begin{subfigure}{.245\linewidth}
  \centering
  \includegraphics[trim={0 100 0 100},clip,width=\linewidth]{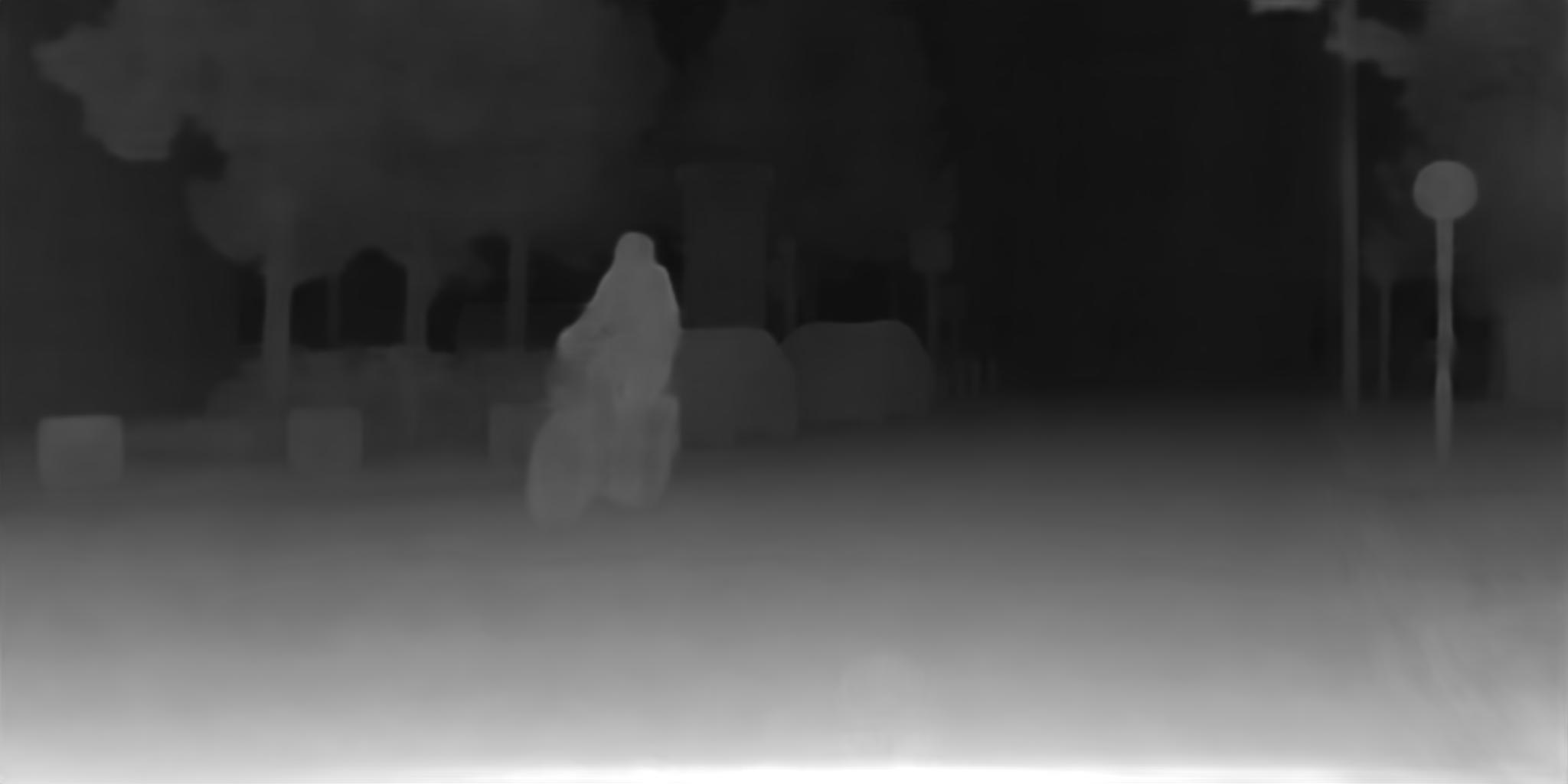}
\end{subfigure}\hfill\begin{subfigure}{.245\linewidth}
  \centering
  \includegraphics[trim={400 130 0 200},clip,width=\linewidth]{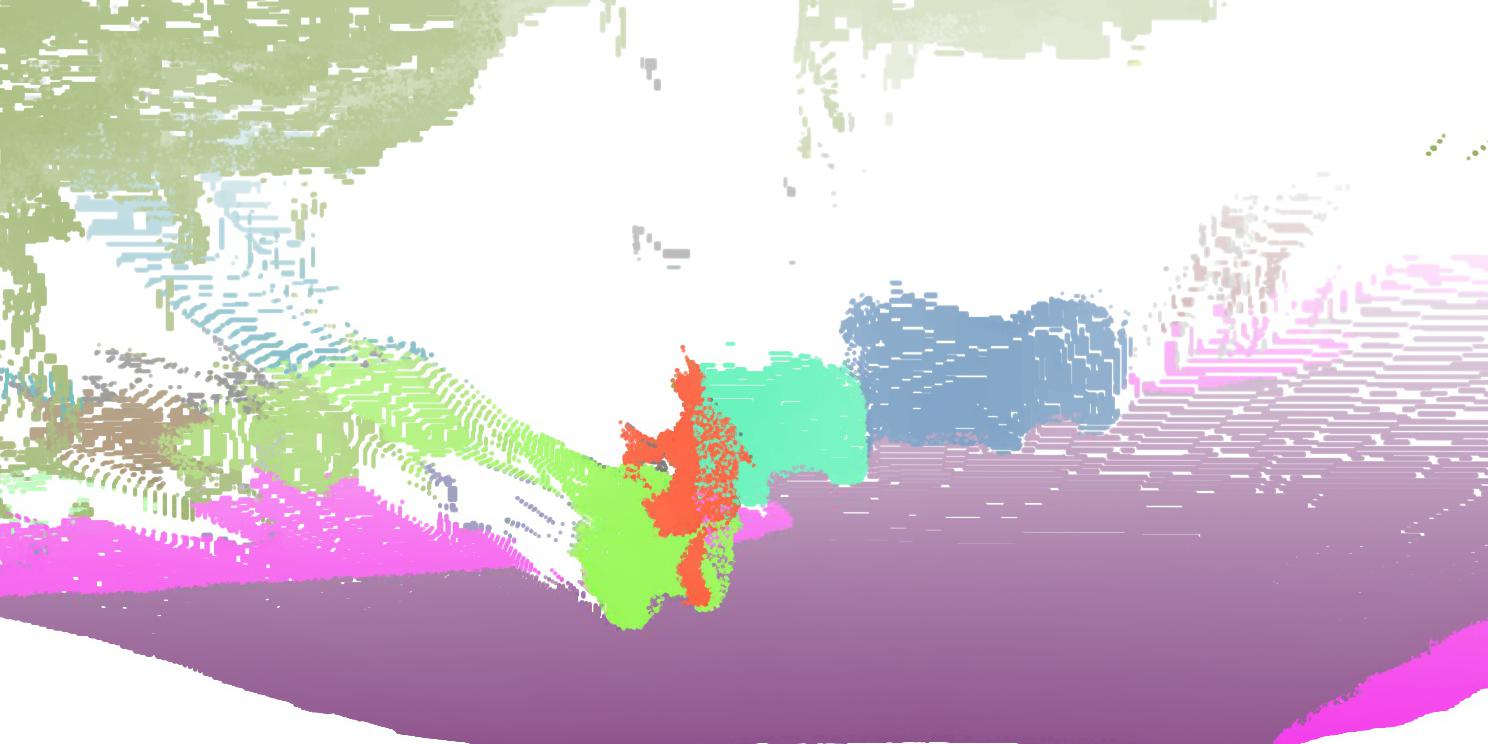}
\end{subfigure}\\\begin{subfigure}{.245\linewidth}
  \centering
  \includegraphics[trim={0 100 0 100},clip,width=\linewidth]{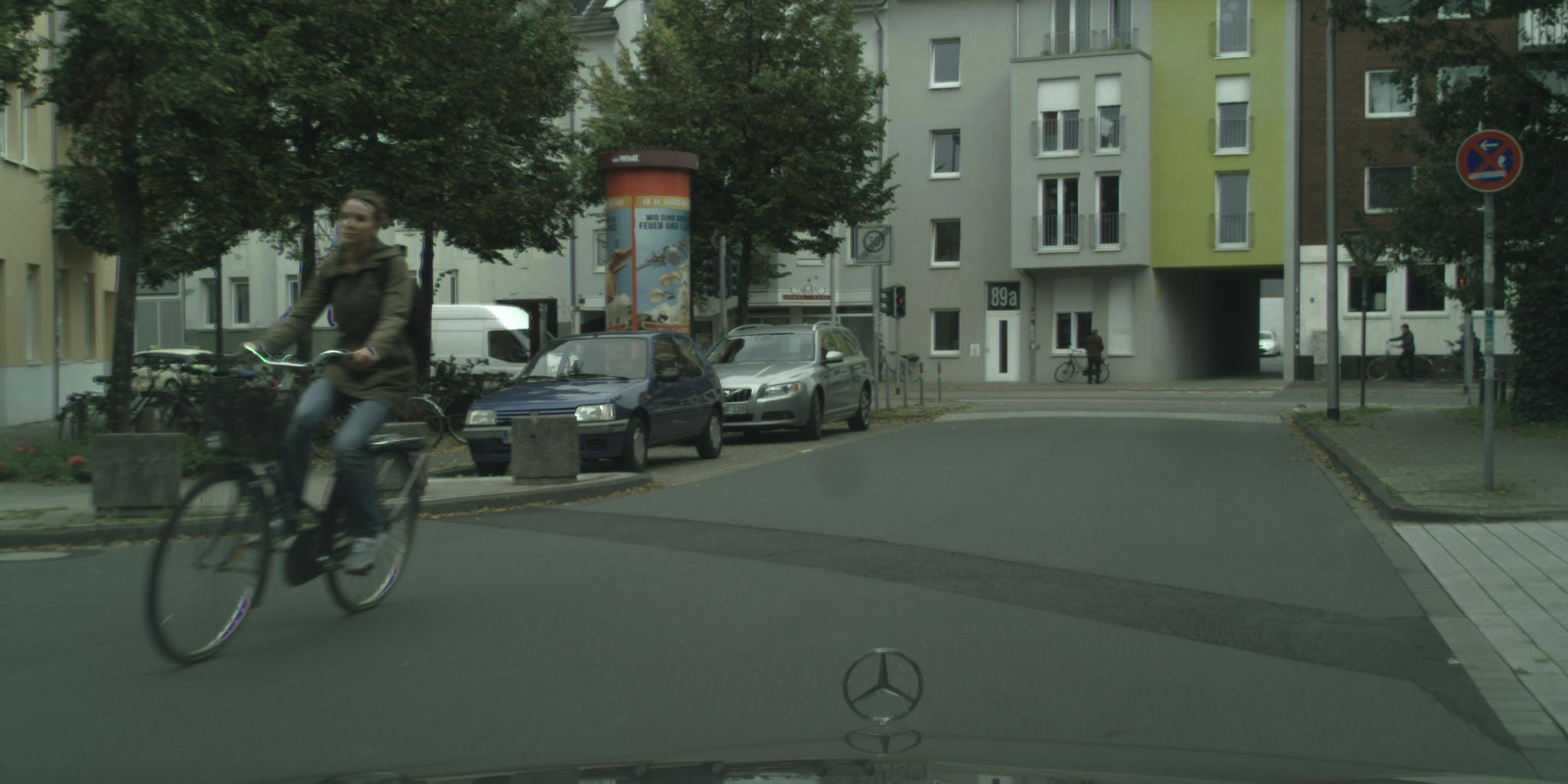}
\end{subfigure}\hfill\begin{subfigure}{.245\linewidth}
  \centering
  \includegraphics[trim={0 100 0 100},clip,width=\linewidth]{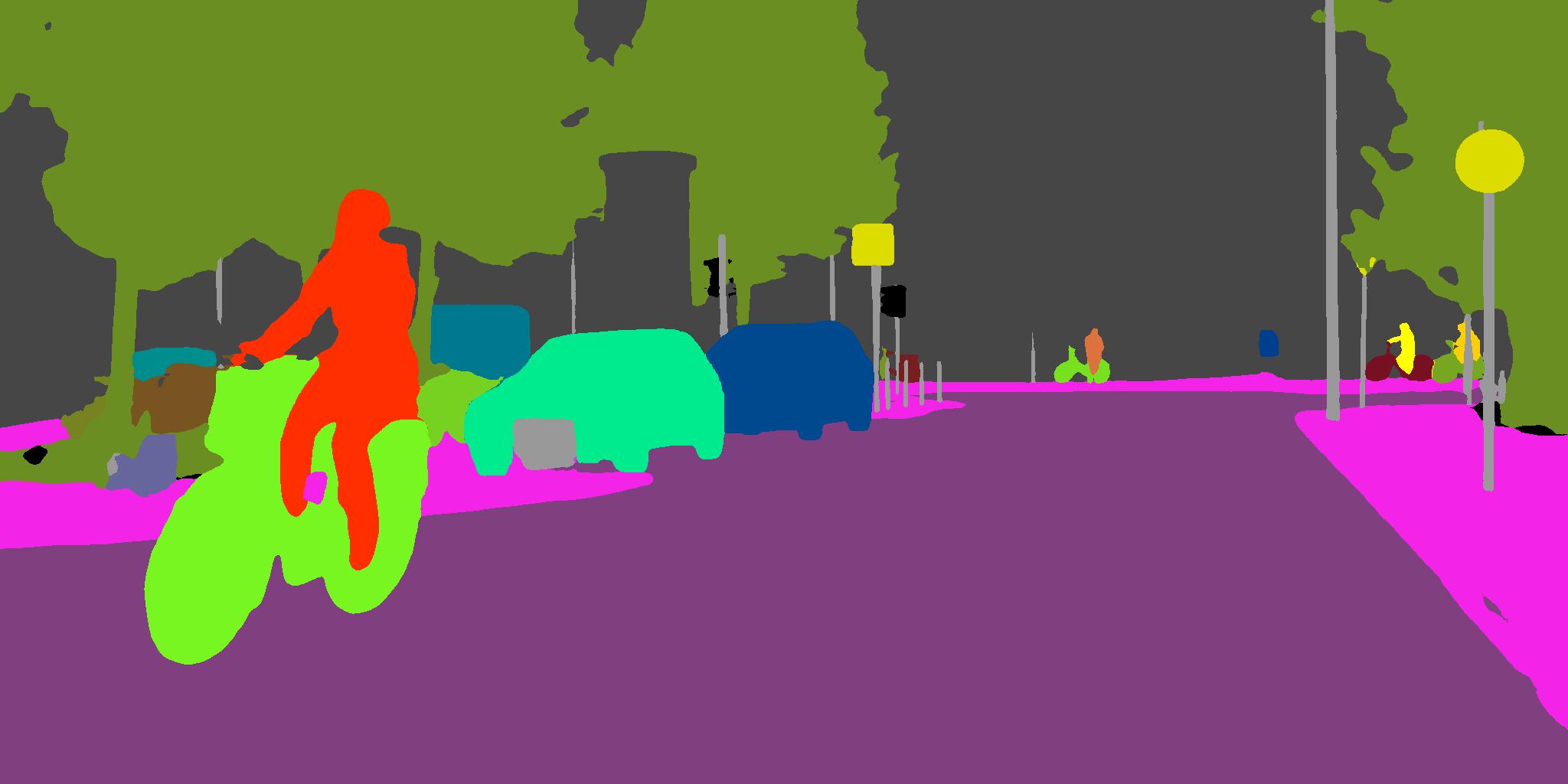}
\end{subfigure}\hfill\begin{subfigure}{.245\linewidth}
  \centering
  \includegraphics[trim={0 100 0 100},clip,width=\linewidth]{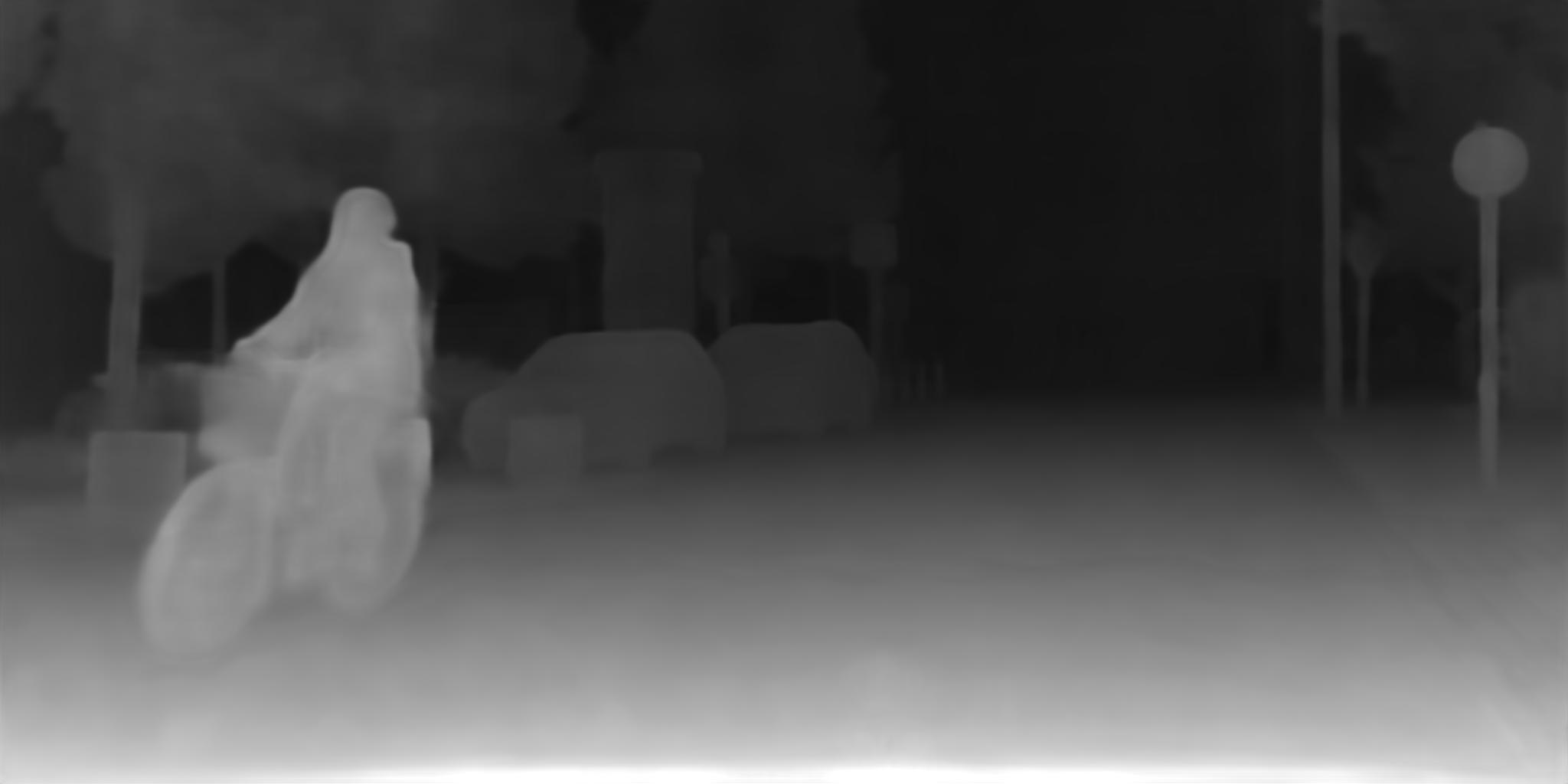}
\end{subfigure}\hfill\begin{subfigure}{.245\linewidth}
  \centering
  \includegraphics[trim={400 130 0 200},clip,width=\linewidth]{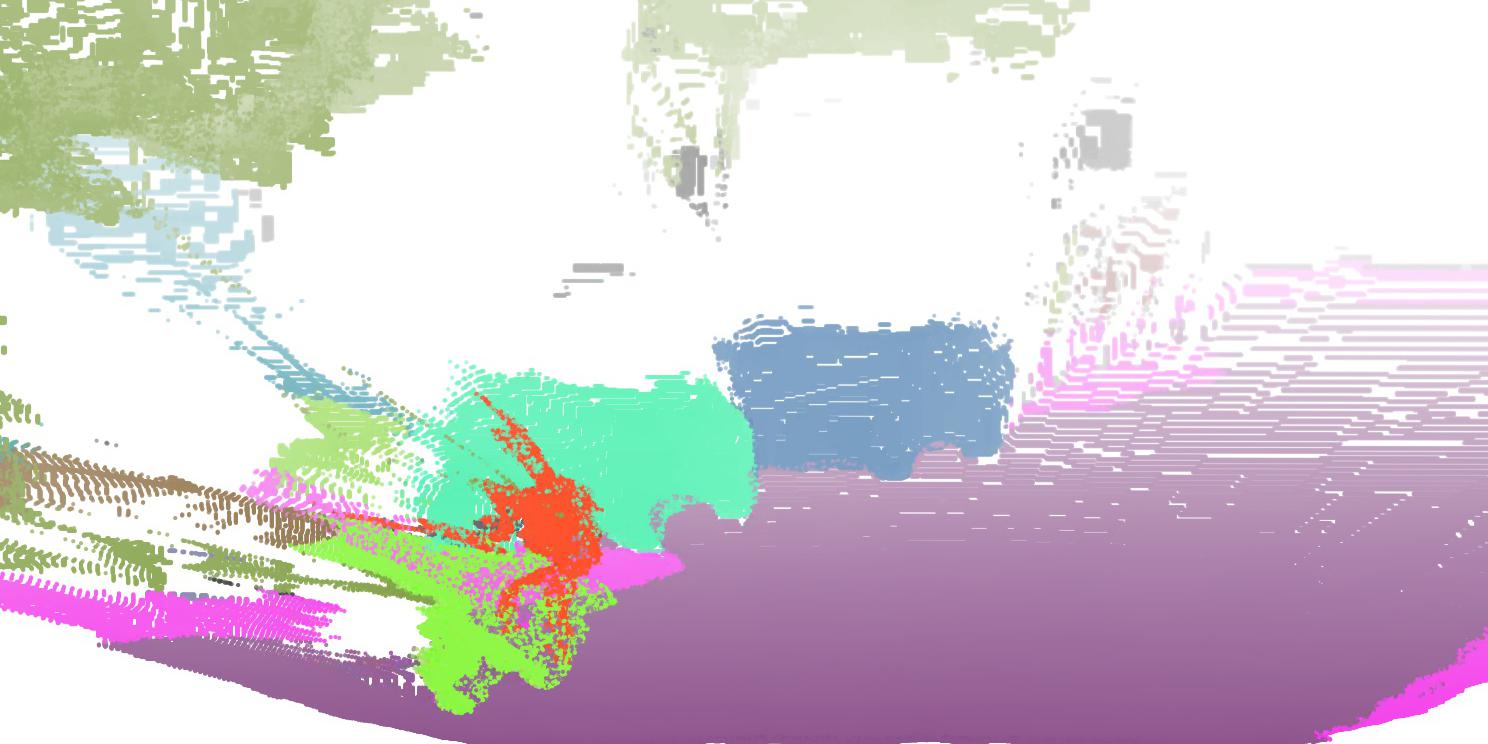}
\end{subfigure}\\\begin{subfigure}{.245\linewidth}
  \centering
  \includegraphics[trim={0 100 0 100},clip,width=\linewidth]{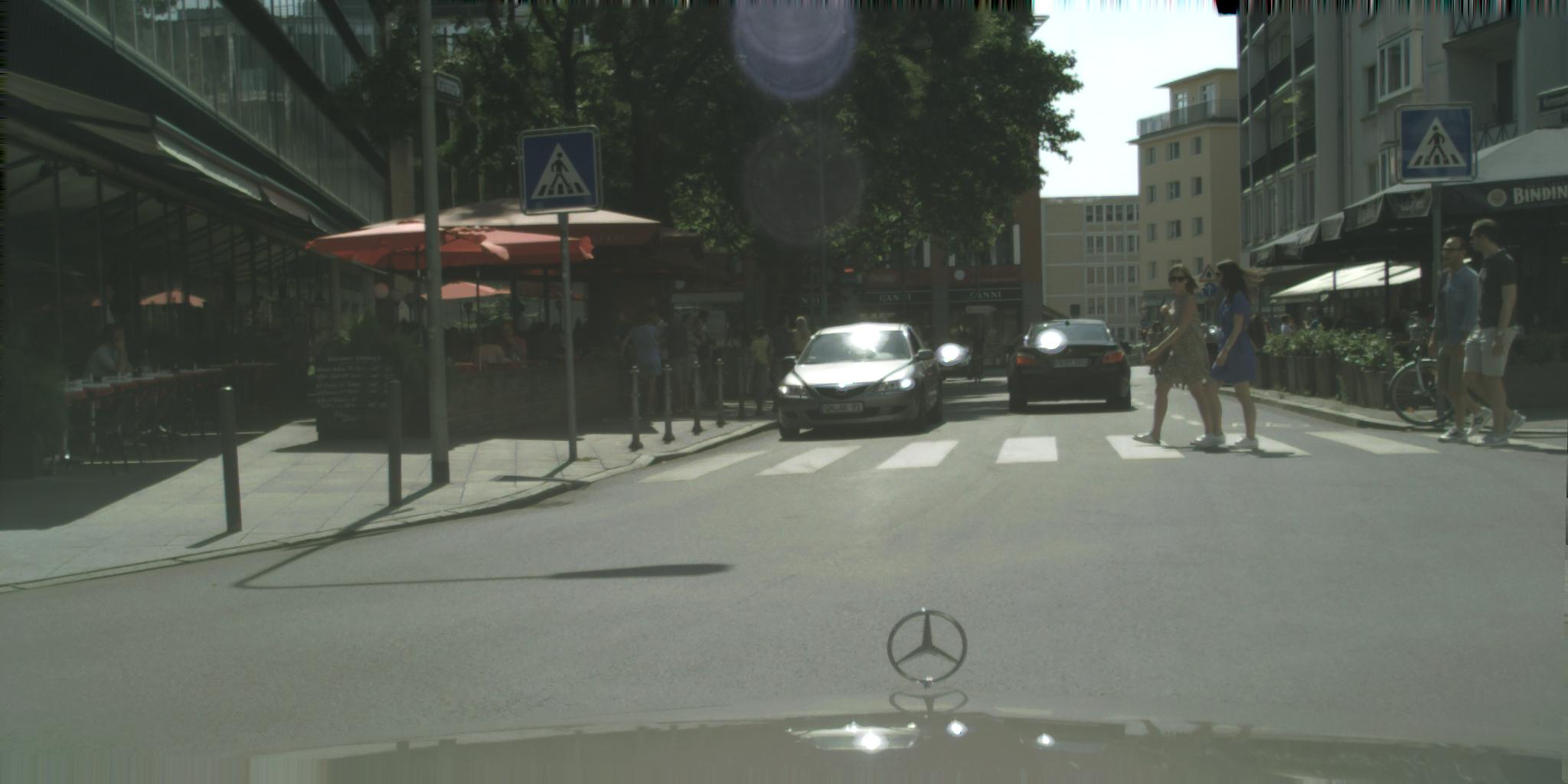}
\end{subfigure}\hfill\begin{subfigure}{.245\linewidth}
  \centering
  \includegraphics[trim={0 100 0 100},clip,width=\linewidth]{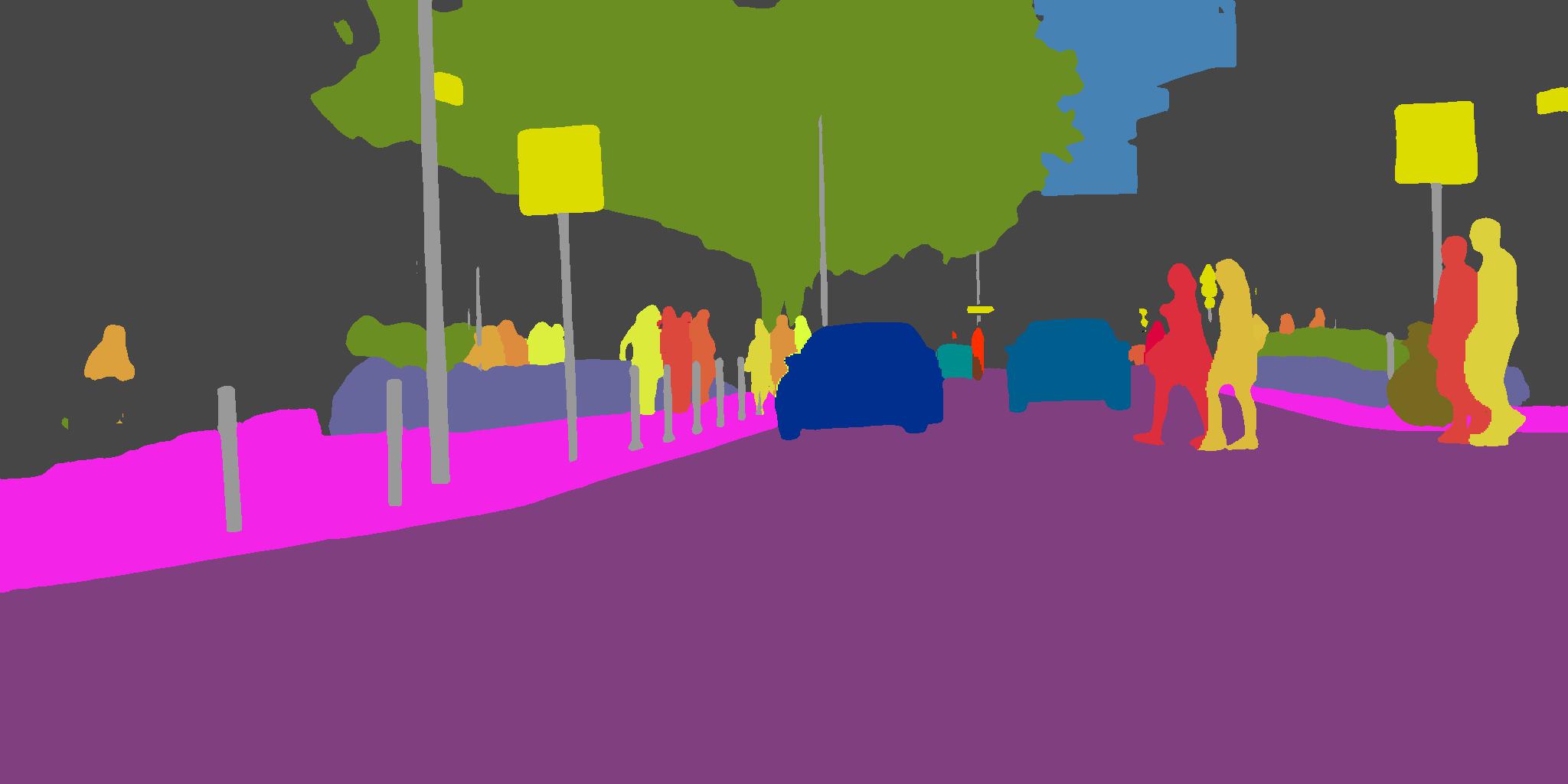}
\end{subfigure}\hfill\begin{subfigure}{.245\linewidth}
  \centering
  \includegraphics[trim={0 100 0 100},clip,width=\linewidth]{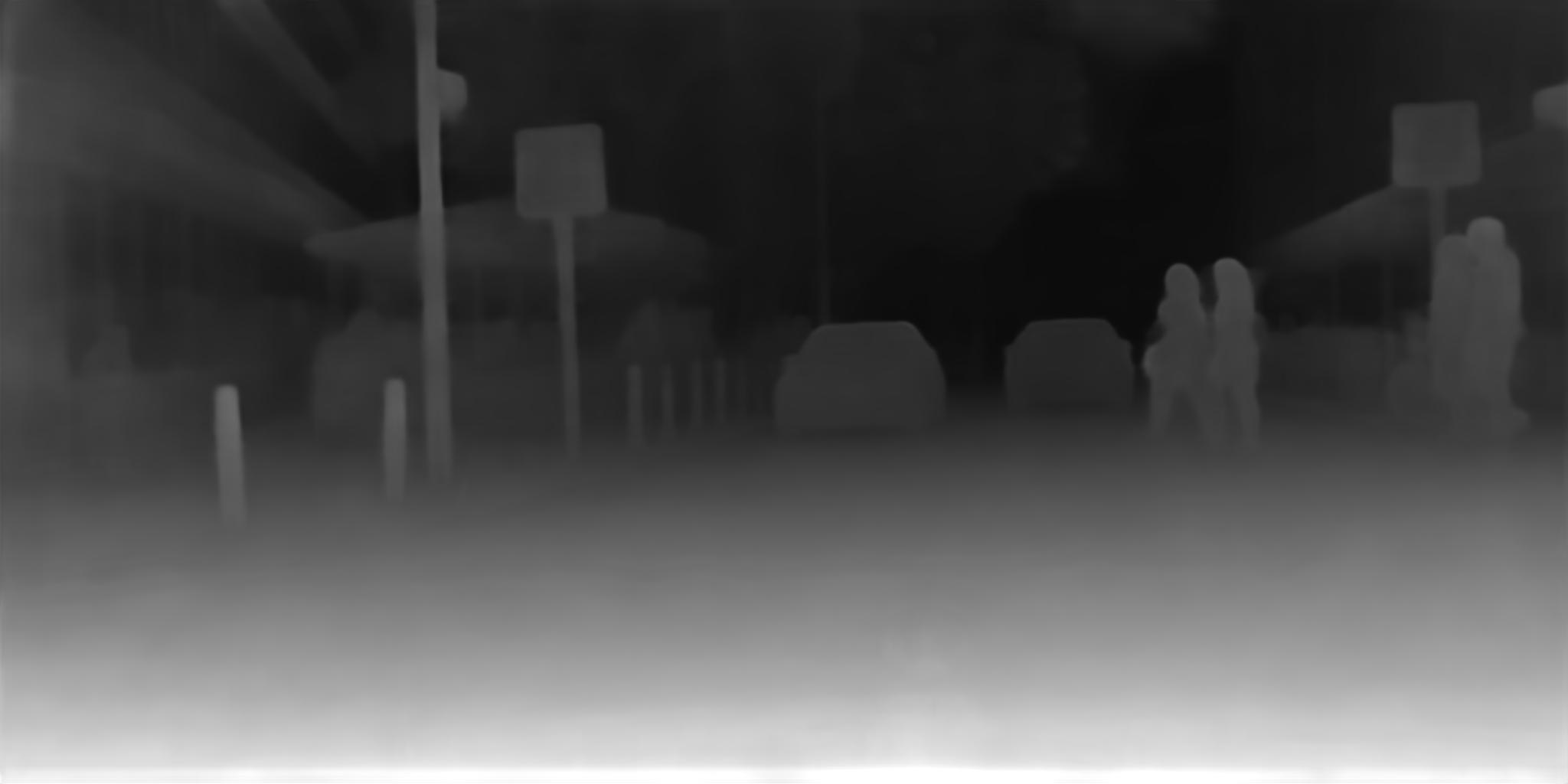}
\end{subfigure}\hfill\begin{subfigure}{.245\linewidth}
  \centering
  \includegraphics[trim={400 130 0 200},clip,width=\linewidth]{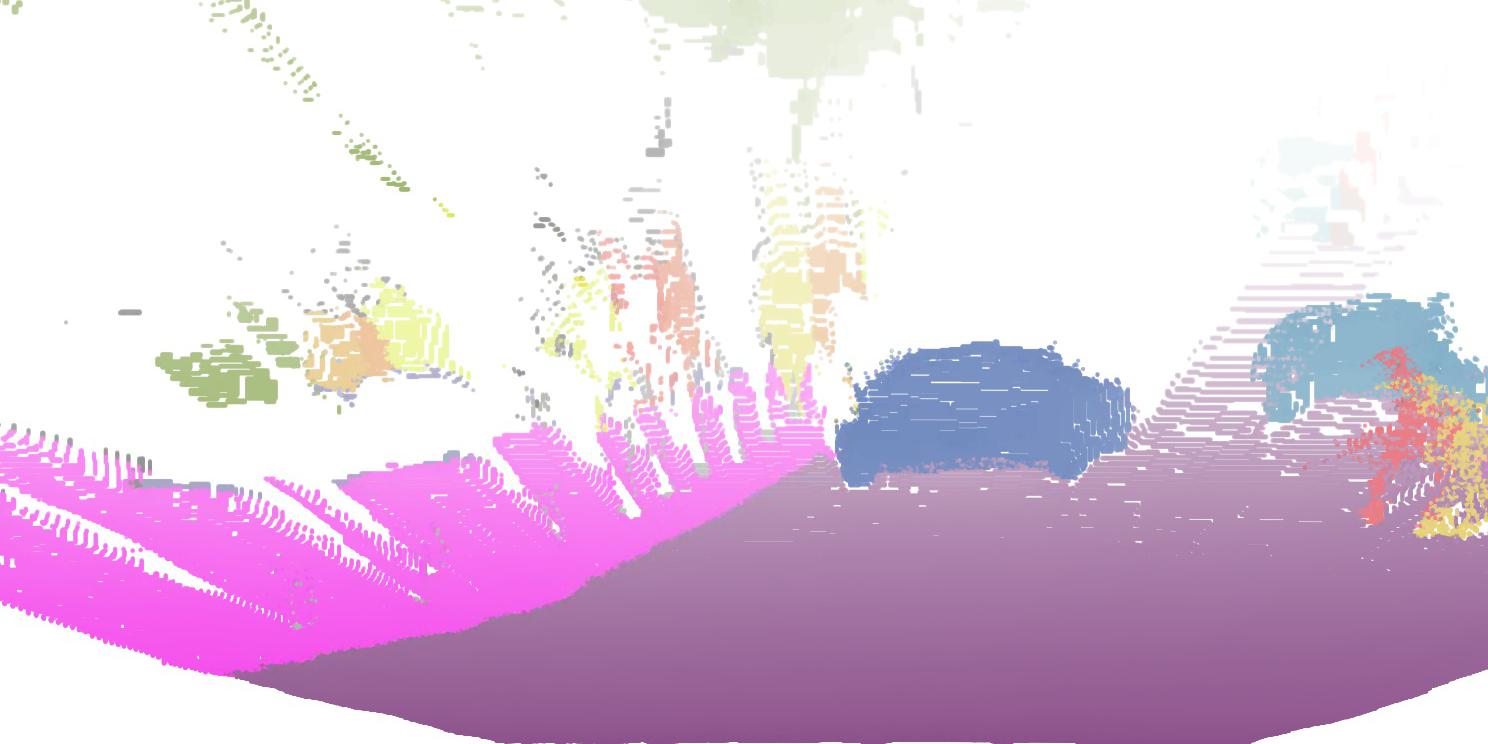}
\end{subfigure}\\\begin{subfigure}{.245\linewidth}
  \centering
  \includegraphics[trim={0 100 0 100},clip,width=\linewidth]{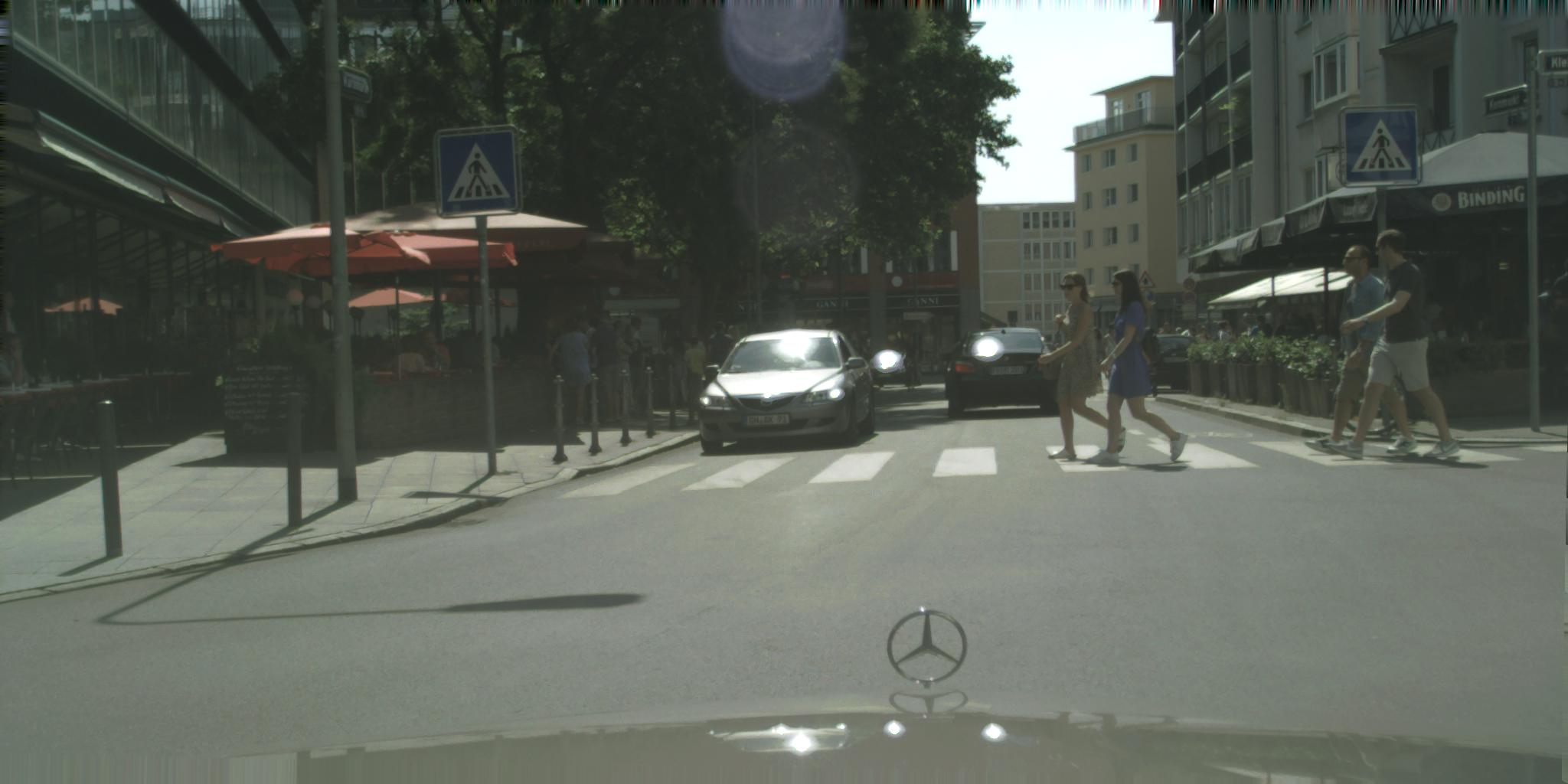}
\end{subfigure}\hfill\begin{subfigure}{.245\linewidth}
  \centering
  \includegraphics[trim={0 100 0 100},clip,width=\linewidth]{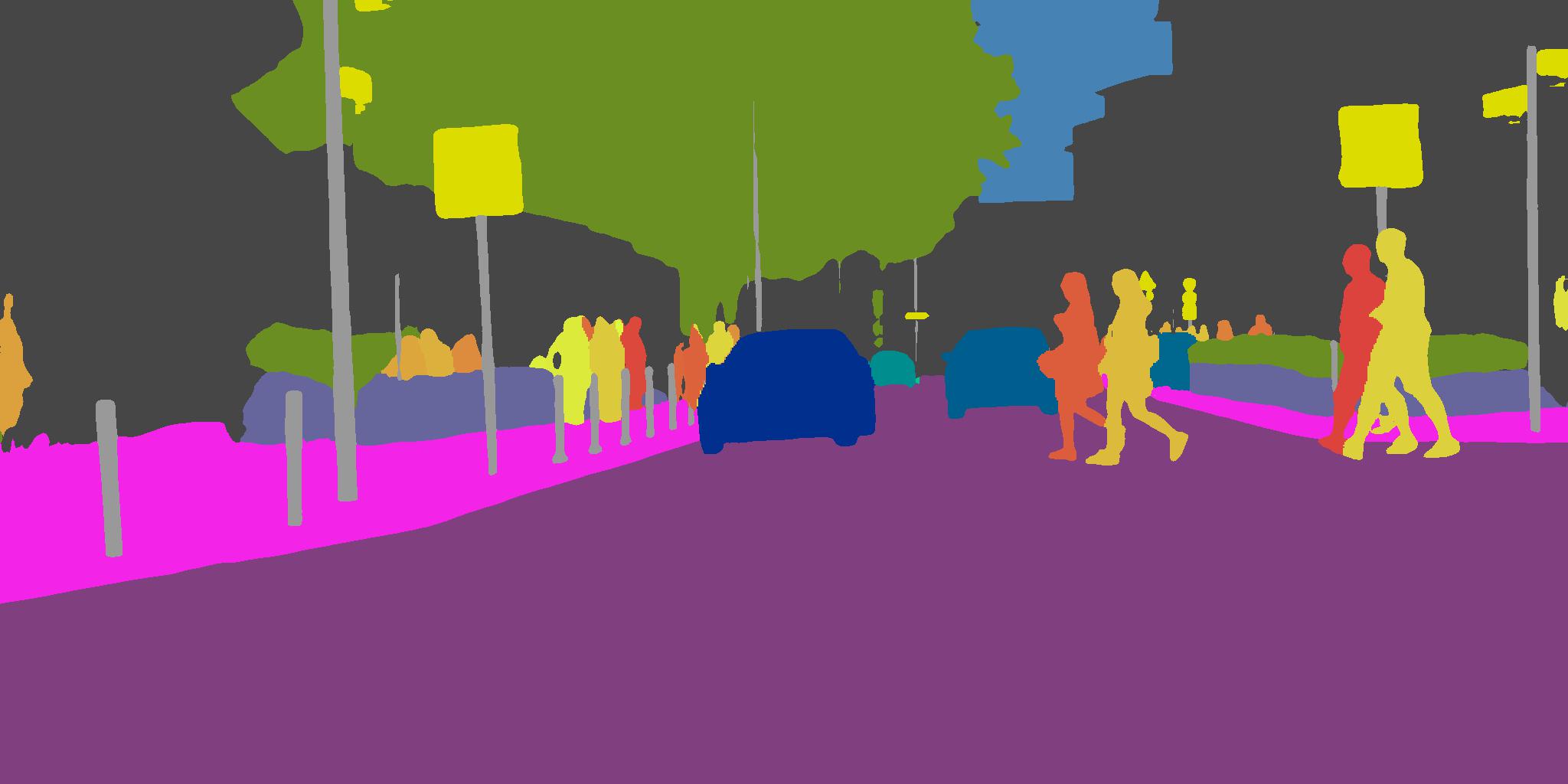}
\end{subfigure}\hfill\begin{subfigure}{.245\linewidth}
  \centering
  \includegraphics[trim={0 100 0 100},clip,width=\linewidth]{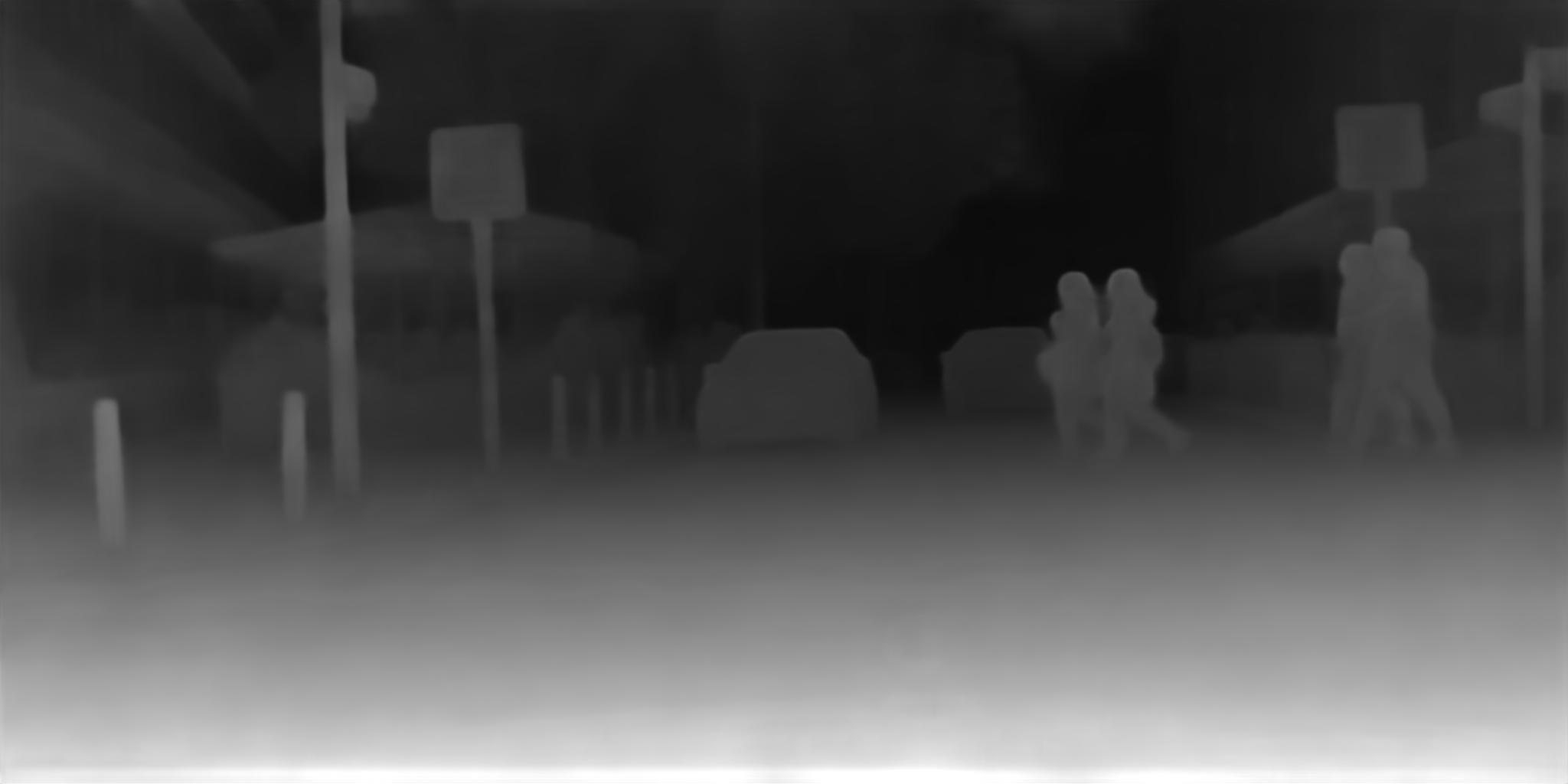}
\end{subfigure}\hfill\begin{subfigure}{.245\linewidth}
  \centering
  \includegraphics[trim={400 130 0 200},clip,width=\linewidth]{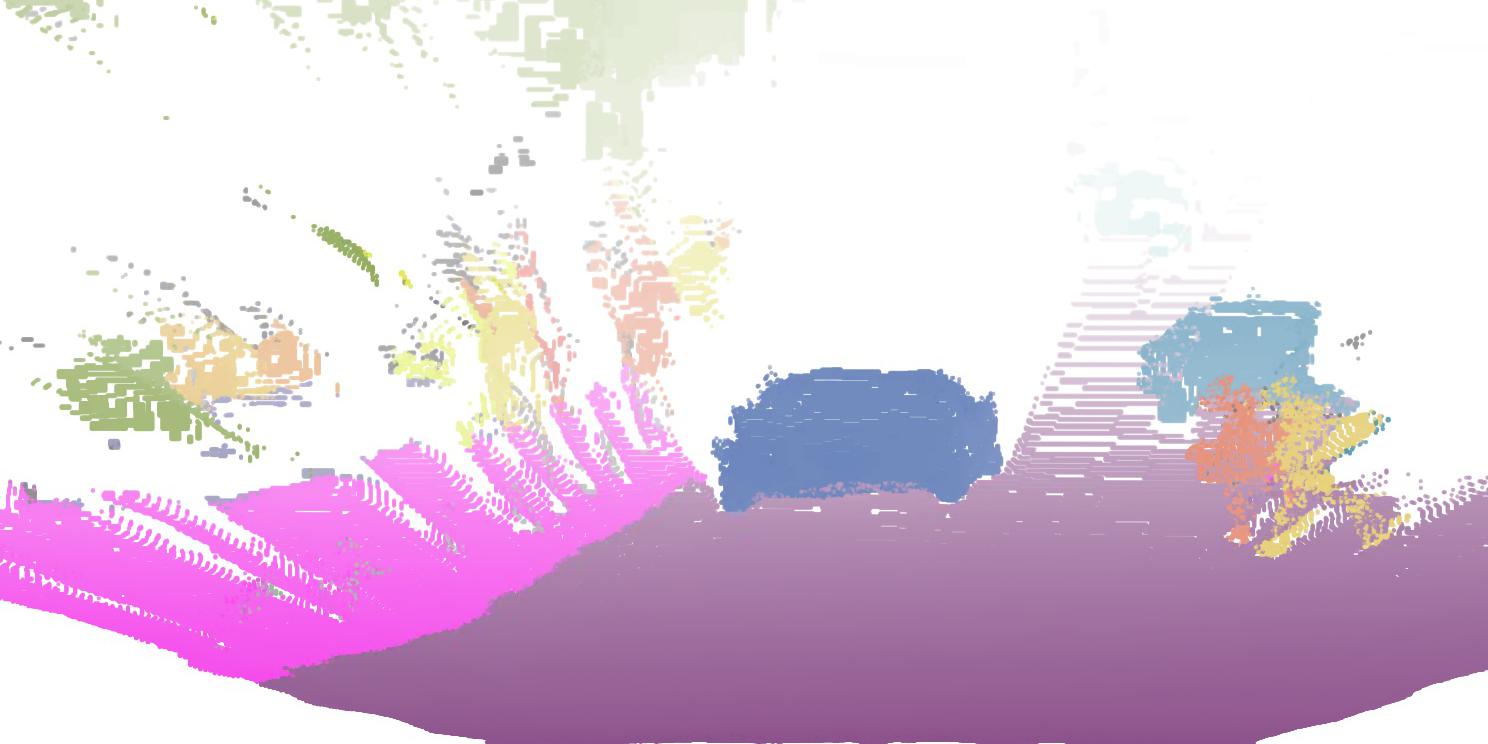}
\end{subfigure}\\\begin{subfigure}{.245\linewidth}
  \centering
  \includegraphics[trim={0 100 0 100},clip,width=\linewidth]{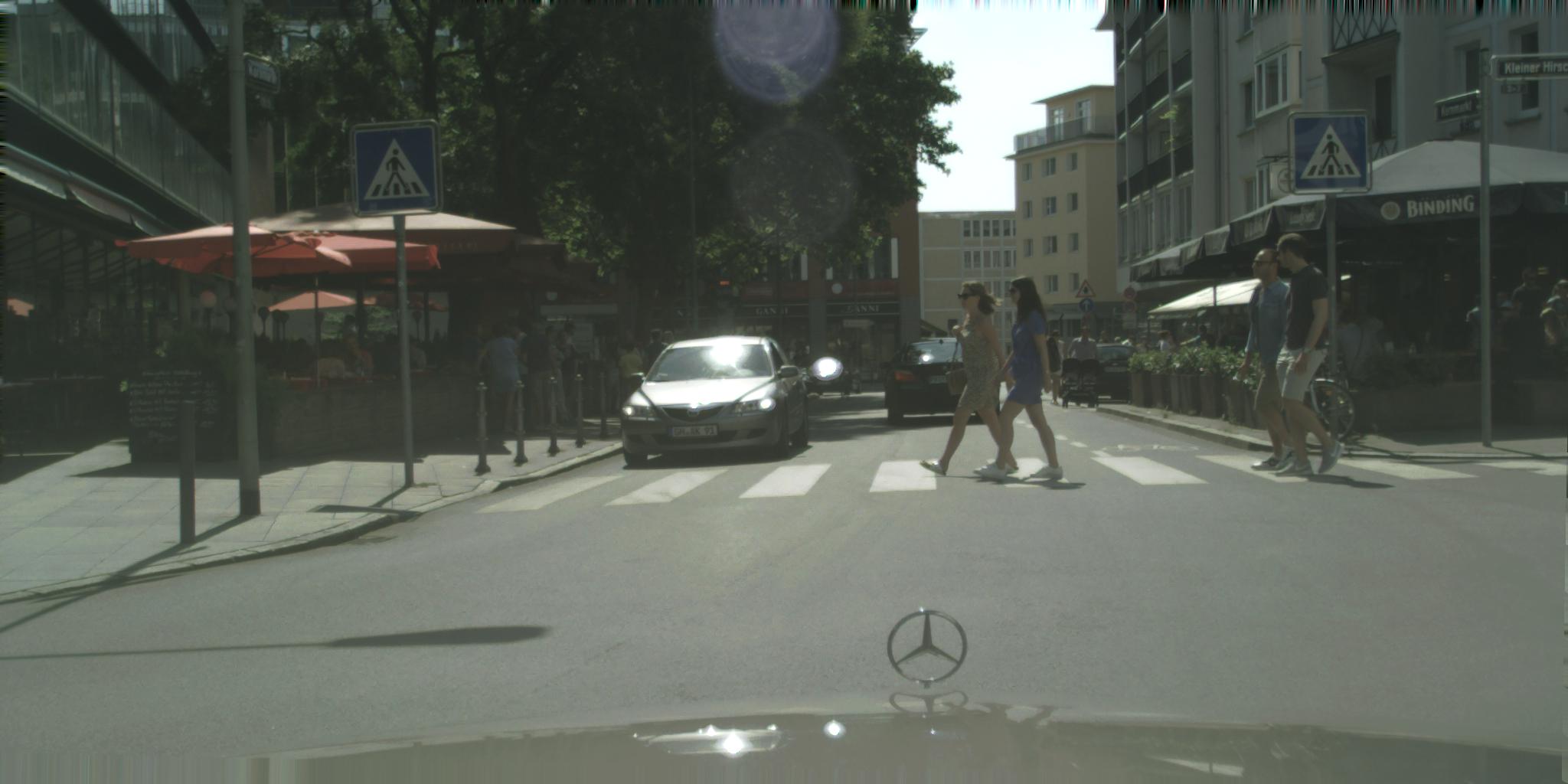}
\end{subfigure}\hfill\begin{subfigure}{.245\linewidth}
  \centering
  \includegraphics[trim={0 100 0 100},clip,width=\linewidth]{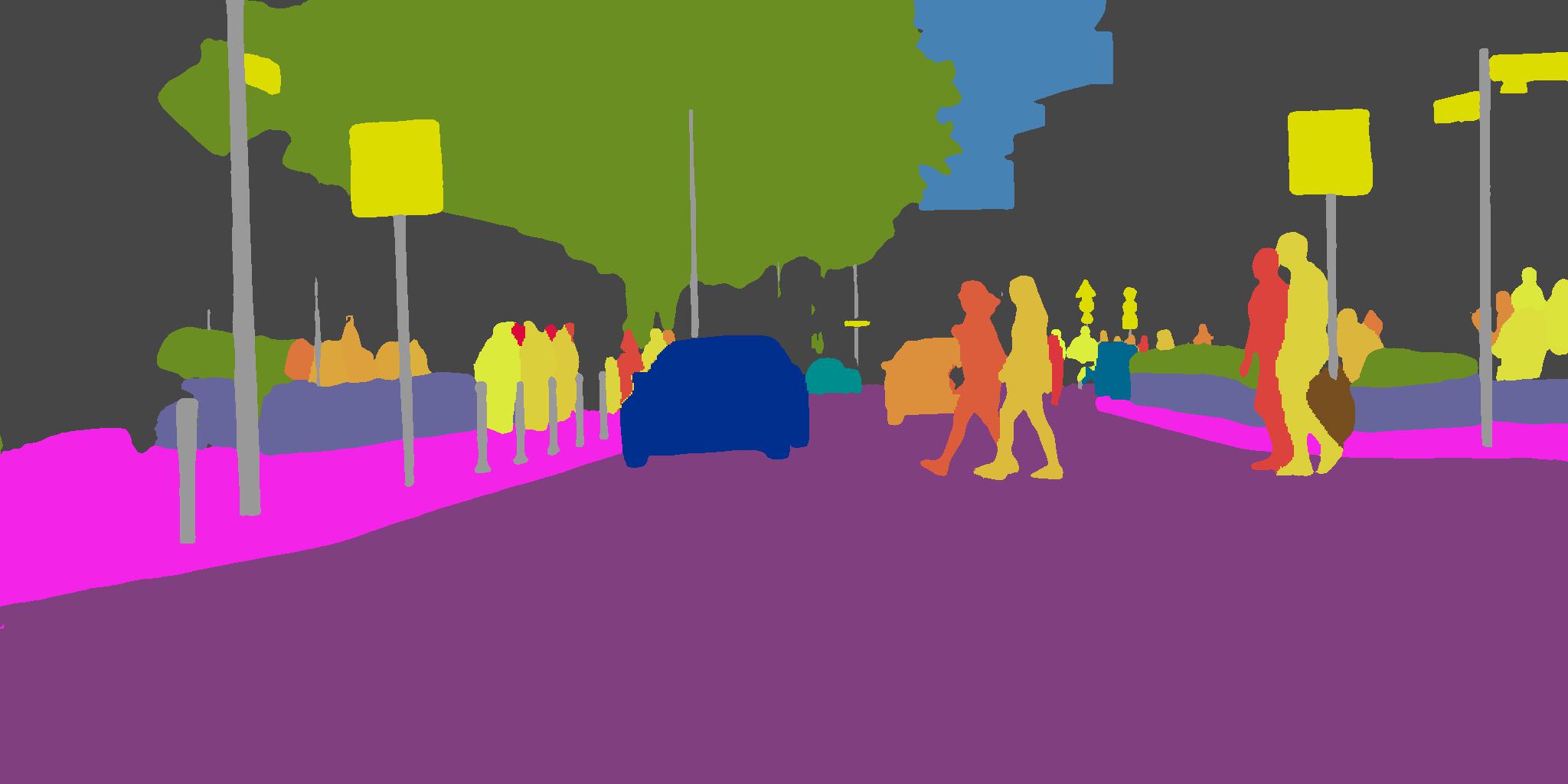}
\end{subfigure}\hfill\begin{subfigure}{.245\linewidth}
  \centering
  \includegraphics[trim={0 100 0 100},clip,width=\linewidth]{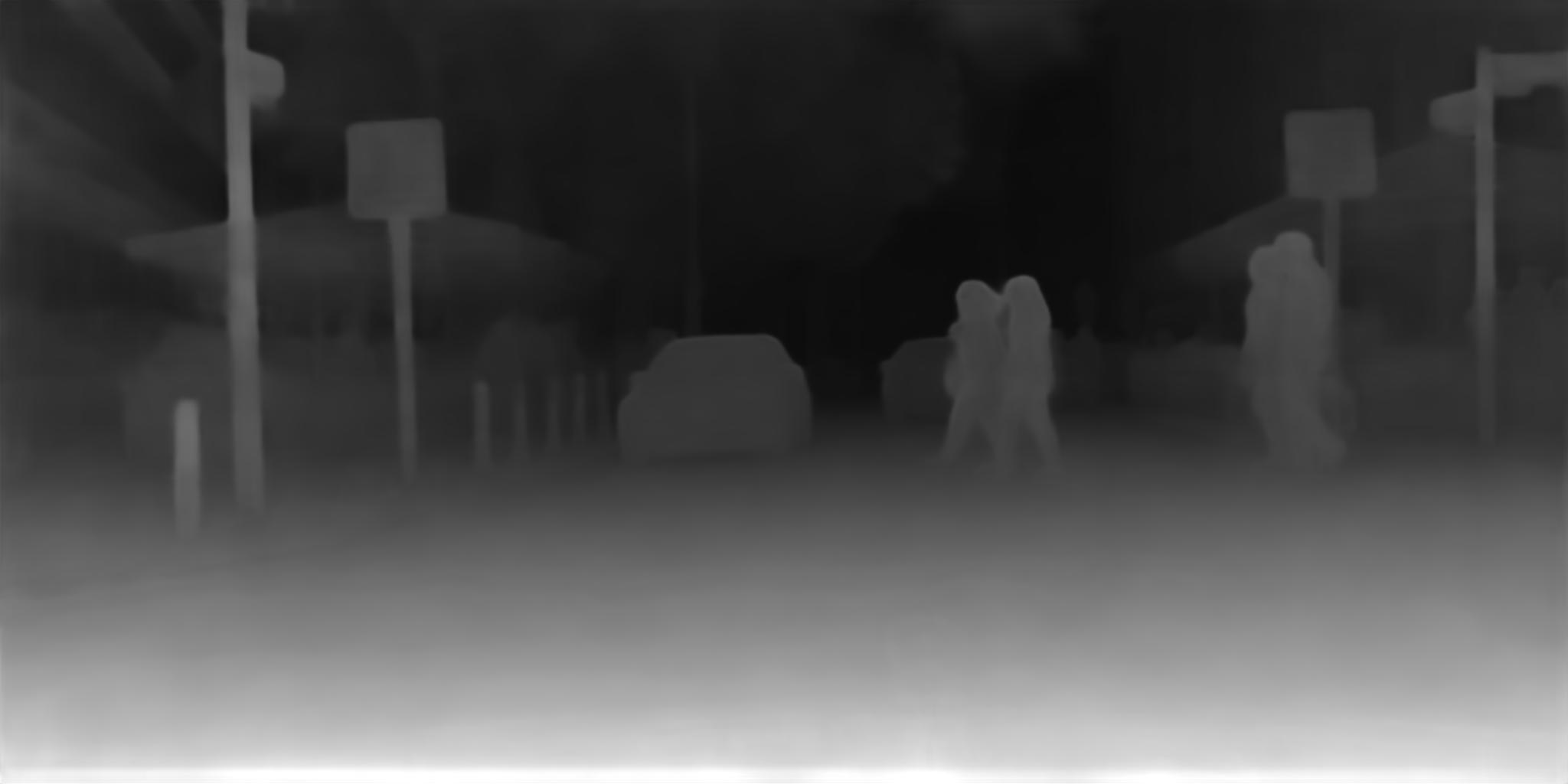}
\end{subfigure}\hfill\begin{subfigure}{.245\linewidth}
  \centering
  \includegraphics[trim={400 130 0 200},clip,width=\linewidth]{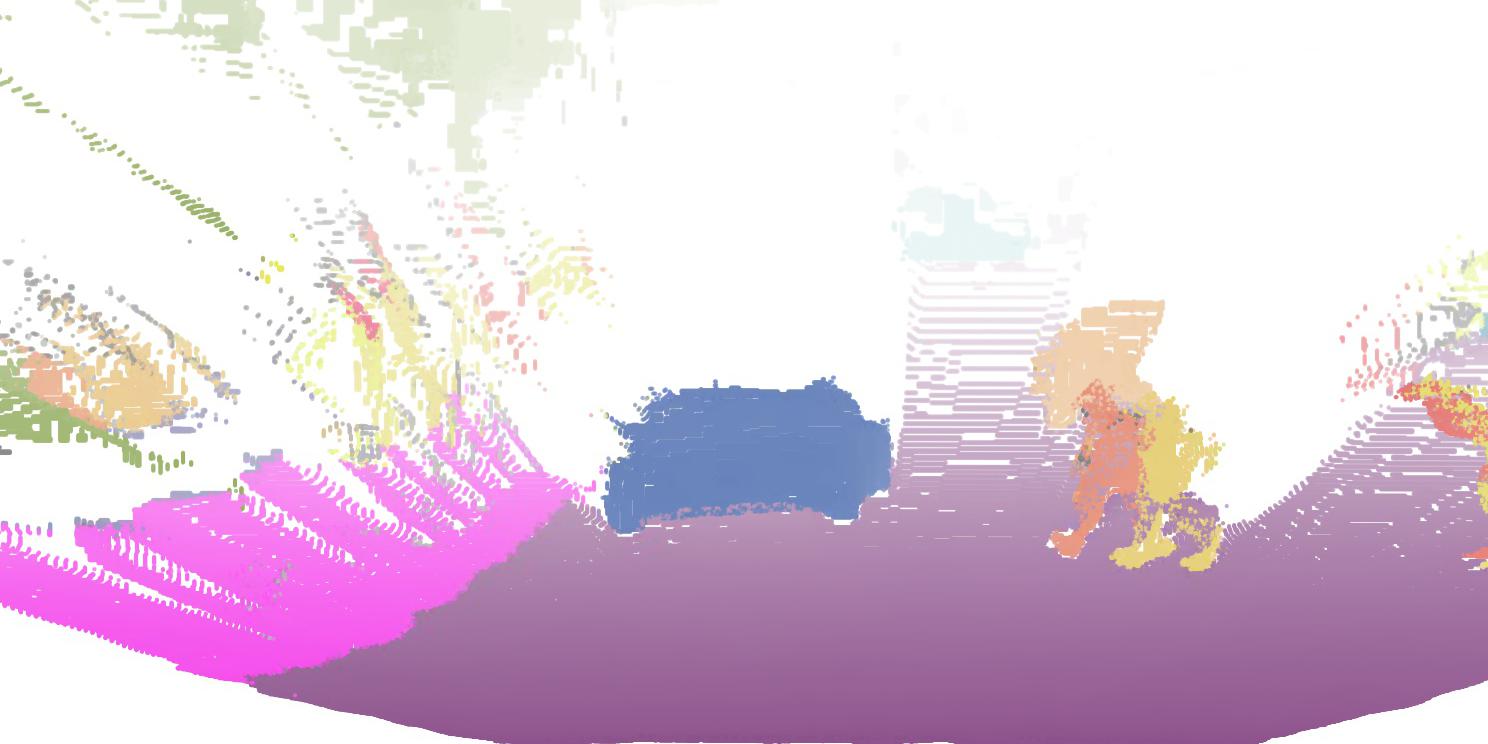}
\end{subfigure}\\\begin{subfigure}{.245\linewidth}
  \centering
  \includegraphics[trim={0 100 0 100},clip,width=\linewidth]{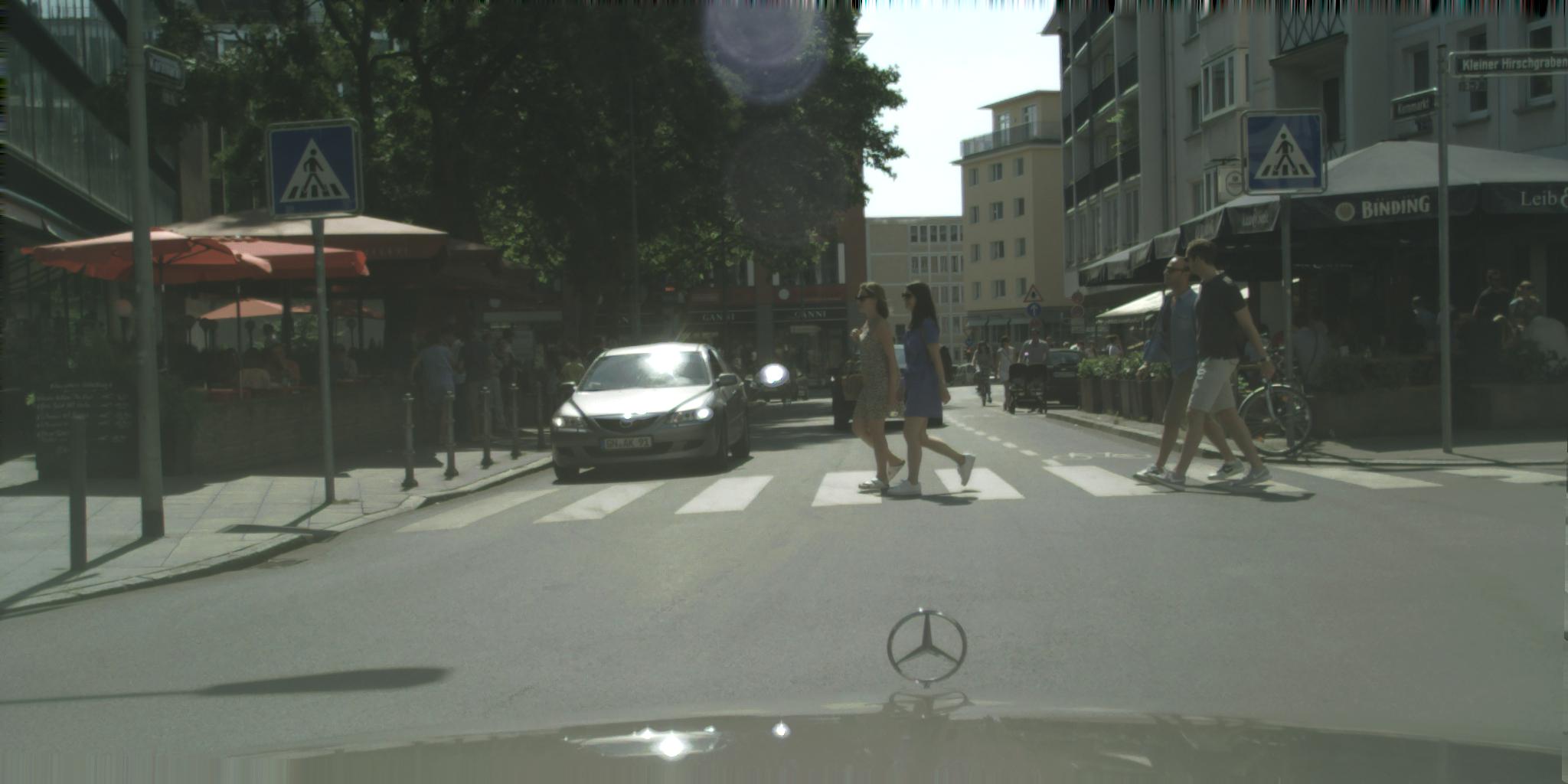}
\end{subfigure}\hfill\begin{subfigure}{.245\linewidth}
  \centering
  \includegraphics[trim={0 100 0 100},clip,width=\linewidth]{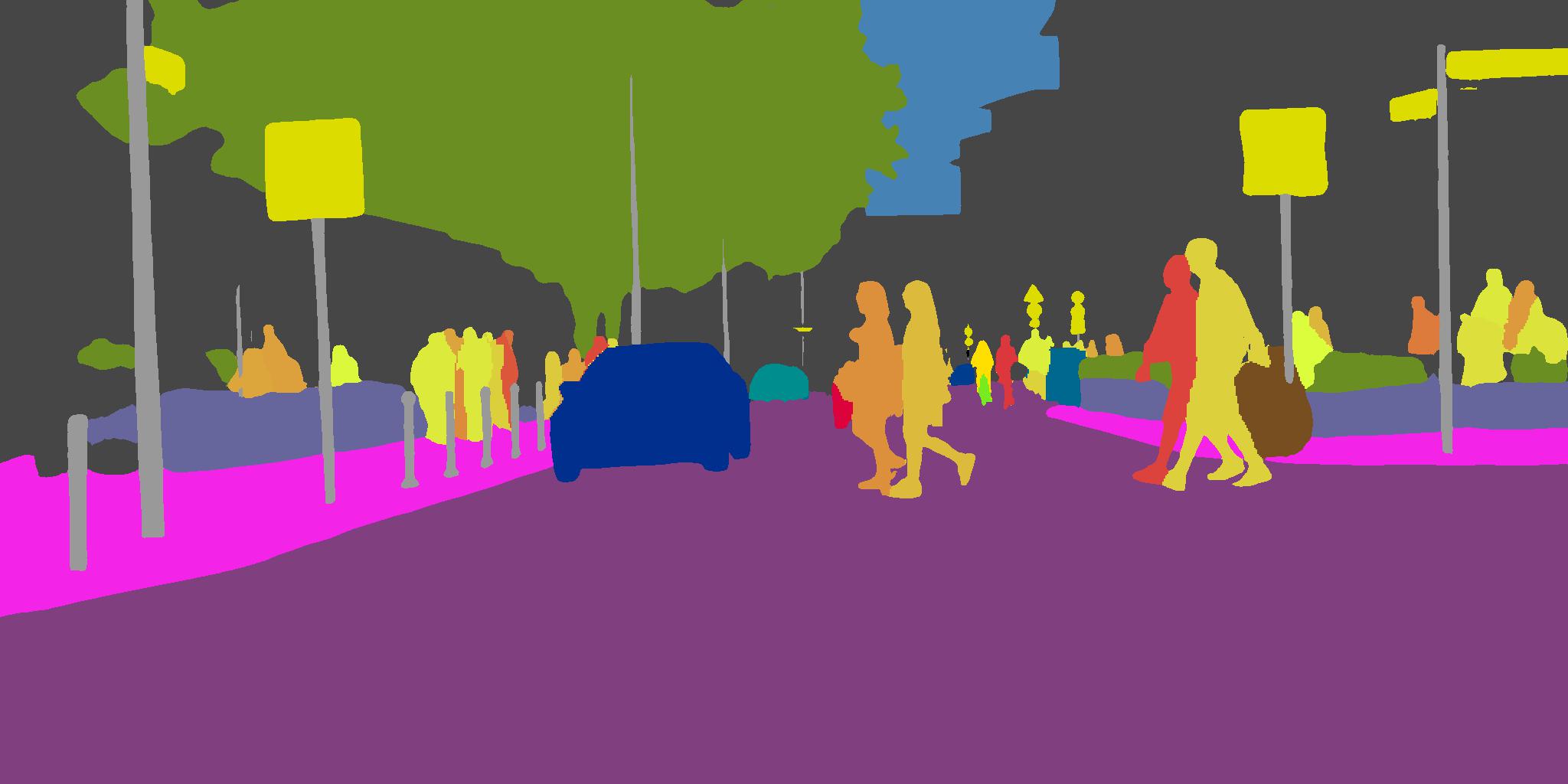}
\end{subfigure}\hfill\begin{subfigure}{.245\linewidth}
  \centering
  \includegraphics[trim={0 100 0 100},clip,width=\linewidth]{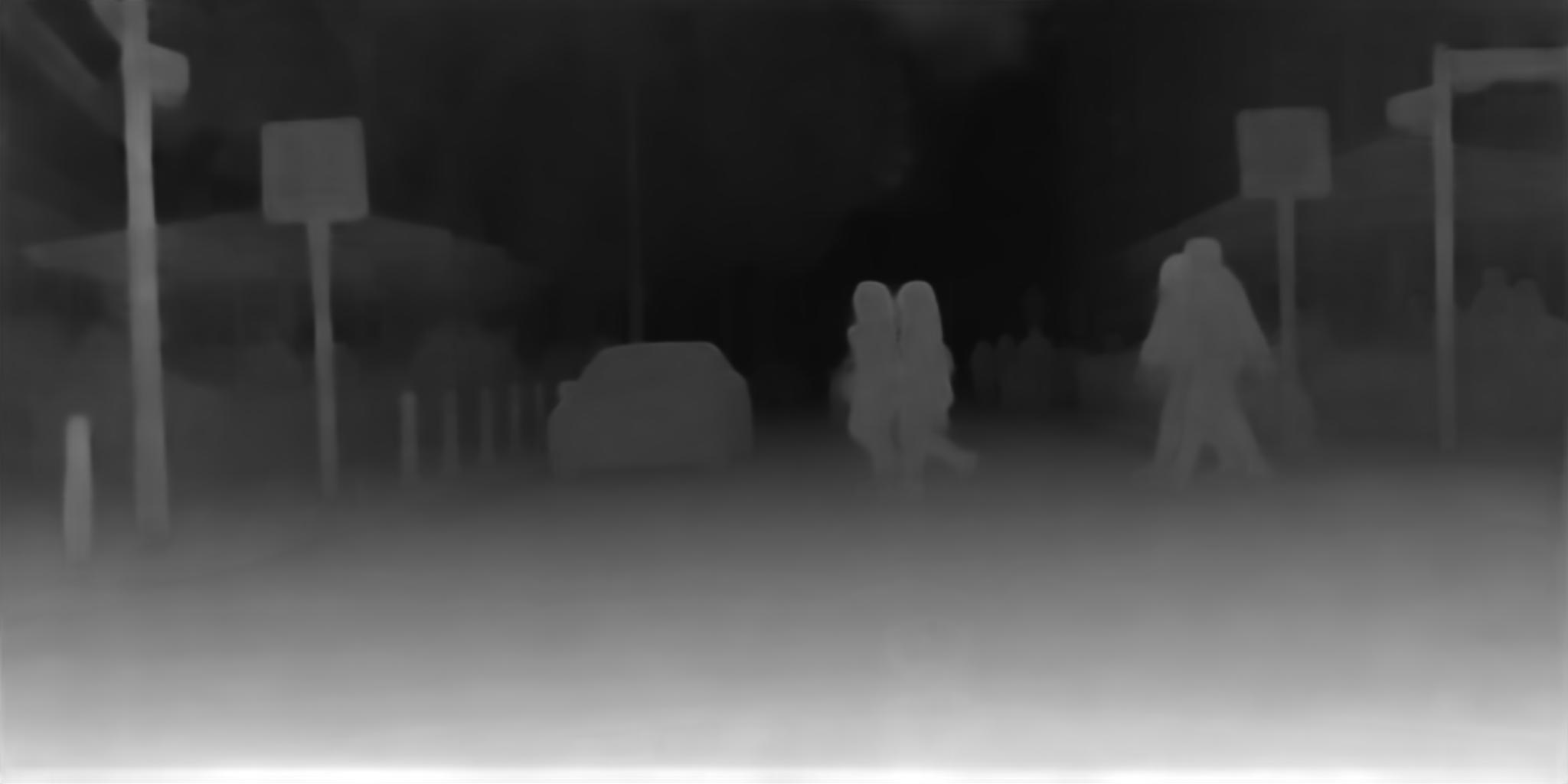}
\end{subfigure}\hfill\begin{subfigure}{.245\linewidth}
  \centering
  \includegraphics[trim={400 130 0 200},clip,width=\linewidth]{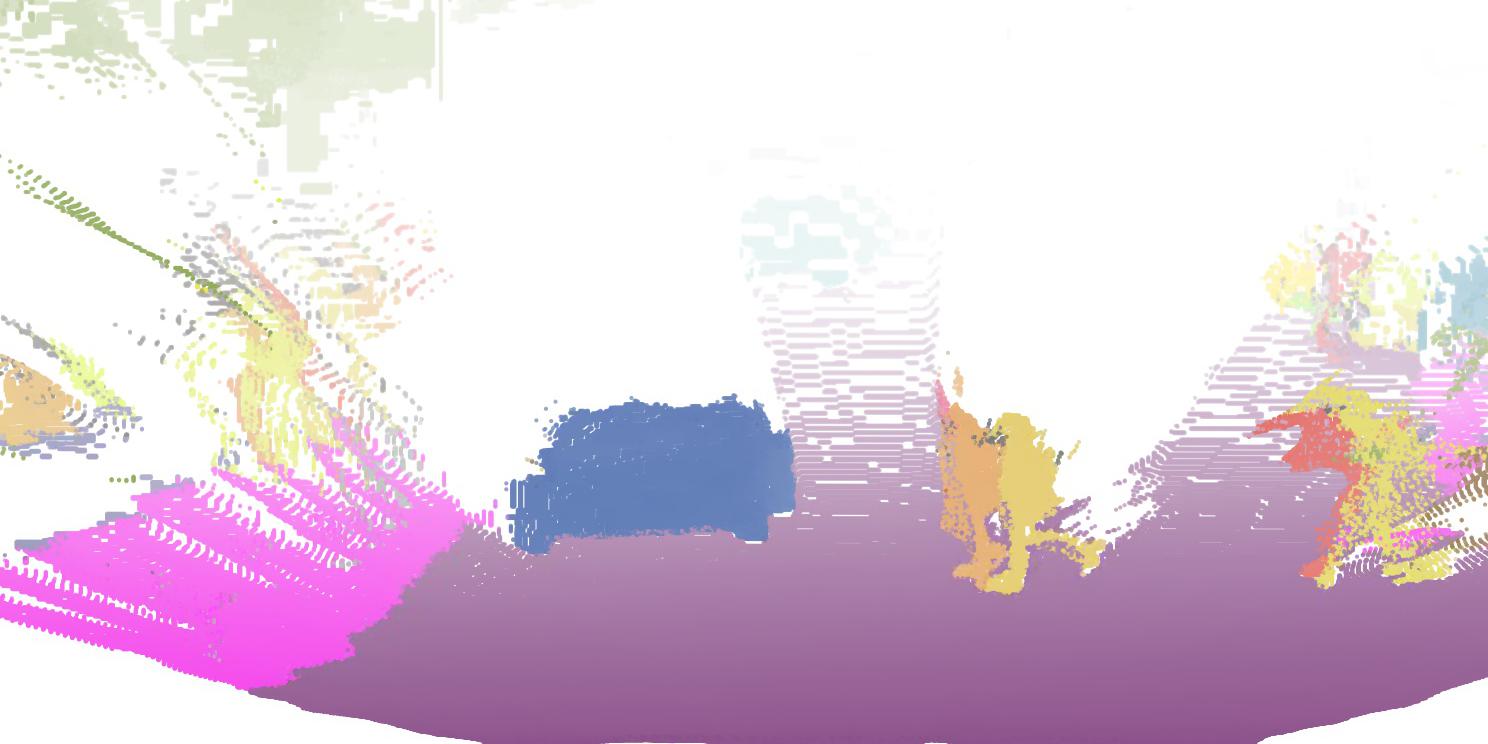}
\end{subfigure}\\
\begin{subfigure}{.245\linewidth}
  \centering
  \includegraphics[trim={0 100 0 100},clip,width=\linewidth]{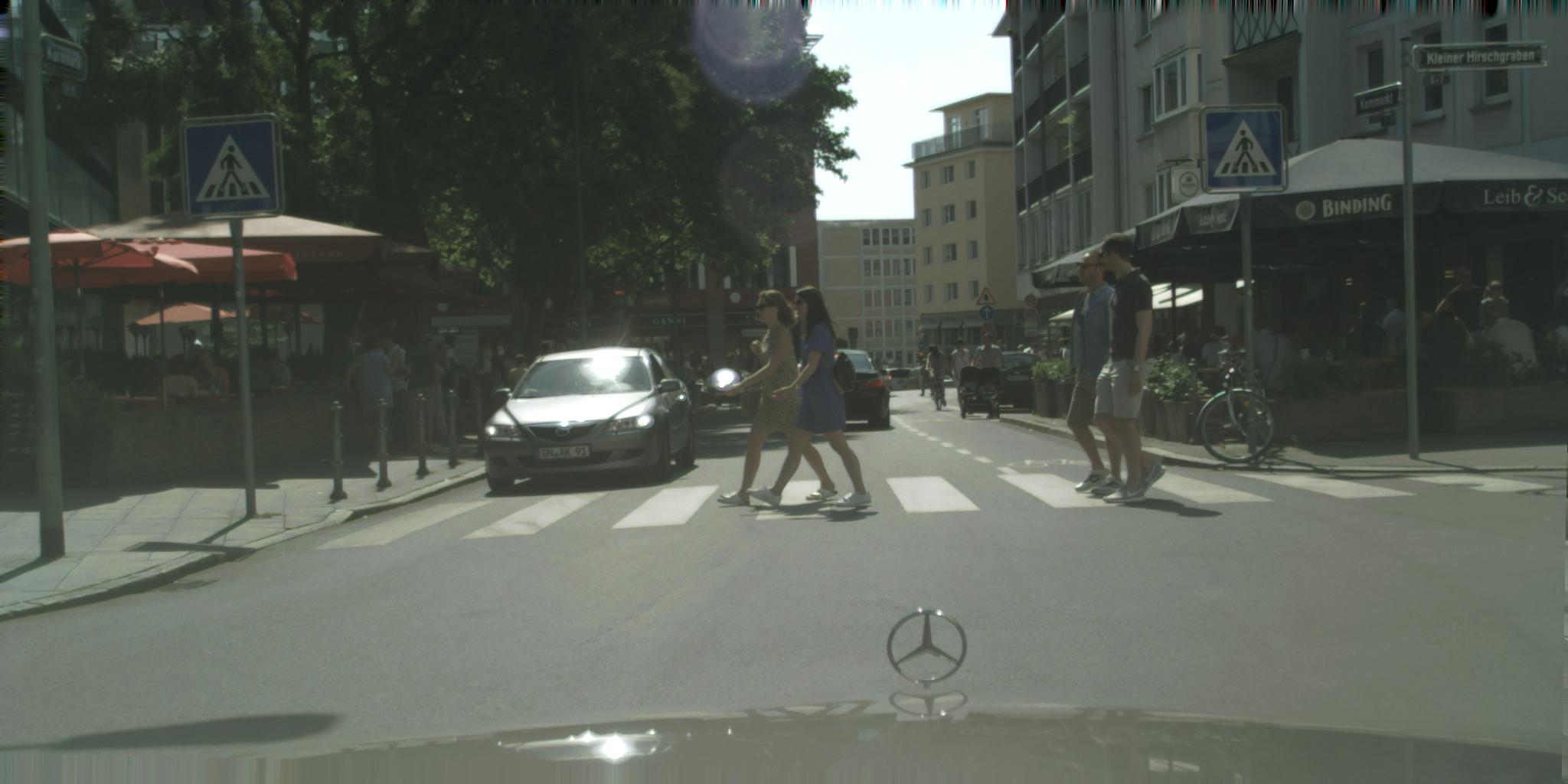}
\end{subfigure}\hfill\begin{subfigure}{.245\linewidth}
  \centering
  \includegraphics[trim={0 100 0 100},clip,width=\linewidth]{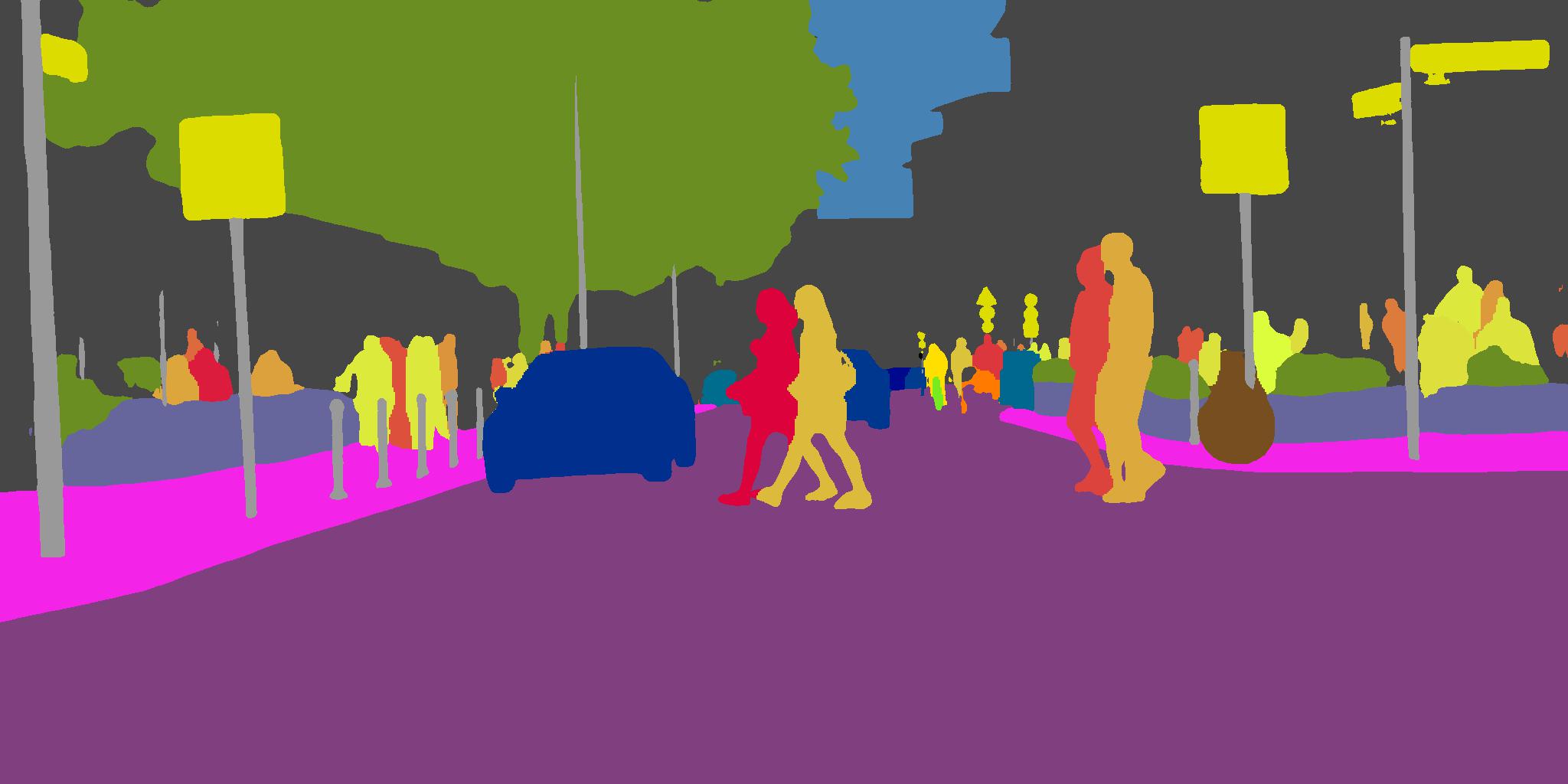}
\end{subfigure}\hfill\begin{subfigure}{.245\linewidth}
  \centering
  \includegraphics[trim={0 100 0 100},clip,width=\linewidth]{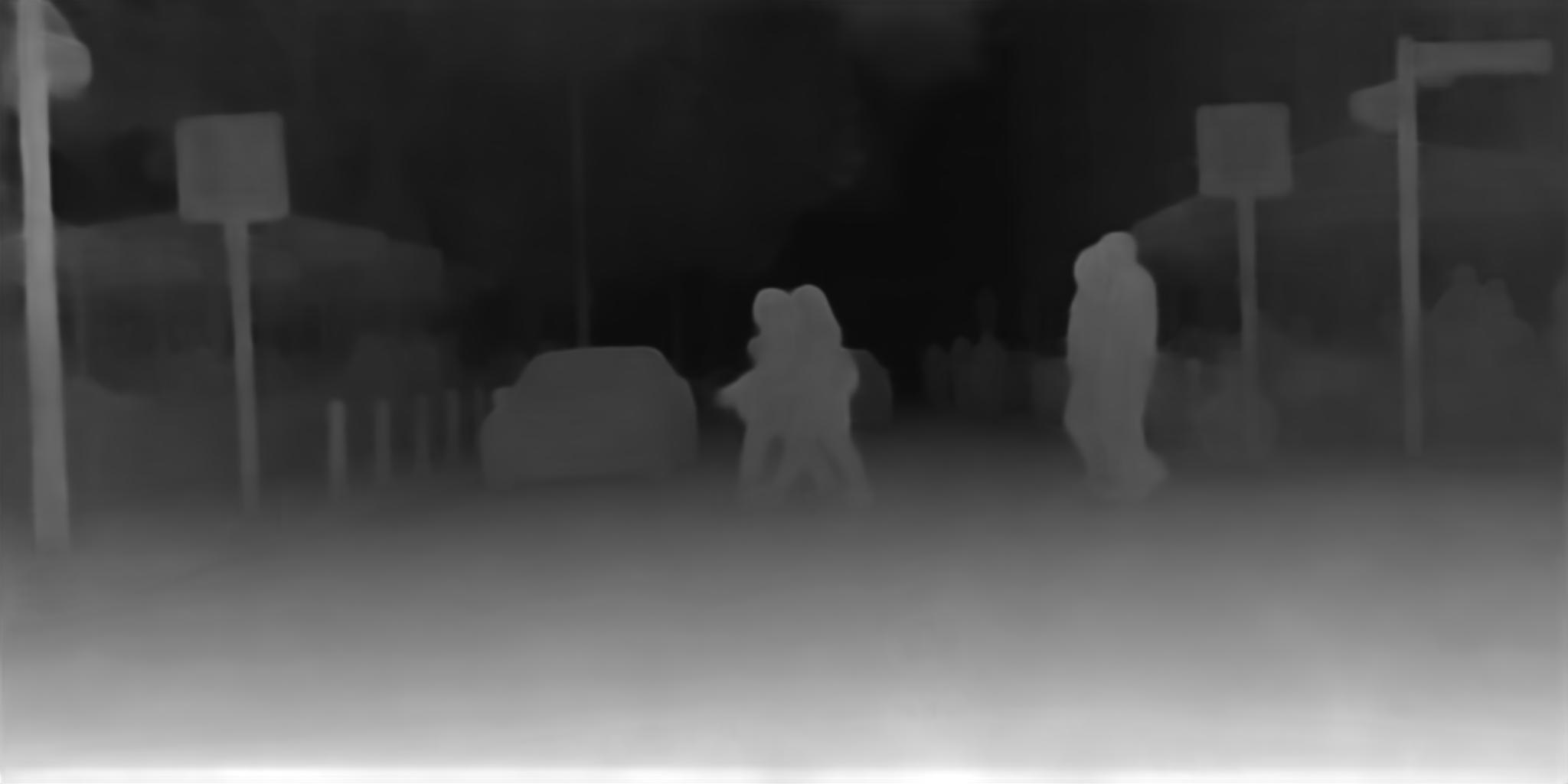}
\end{subfigure}\hfill\begin{subfigure}{.245\linewidth}
  \centering
  \includegraphics[trim={400 130 0 200},clip,width=\linewidth]{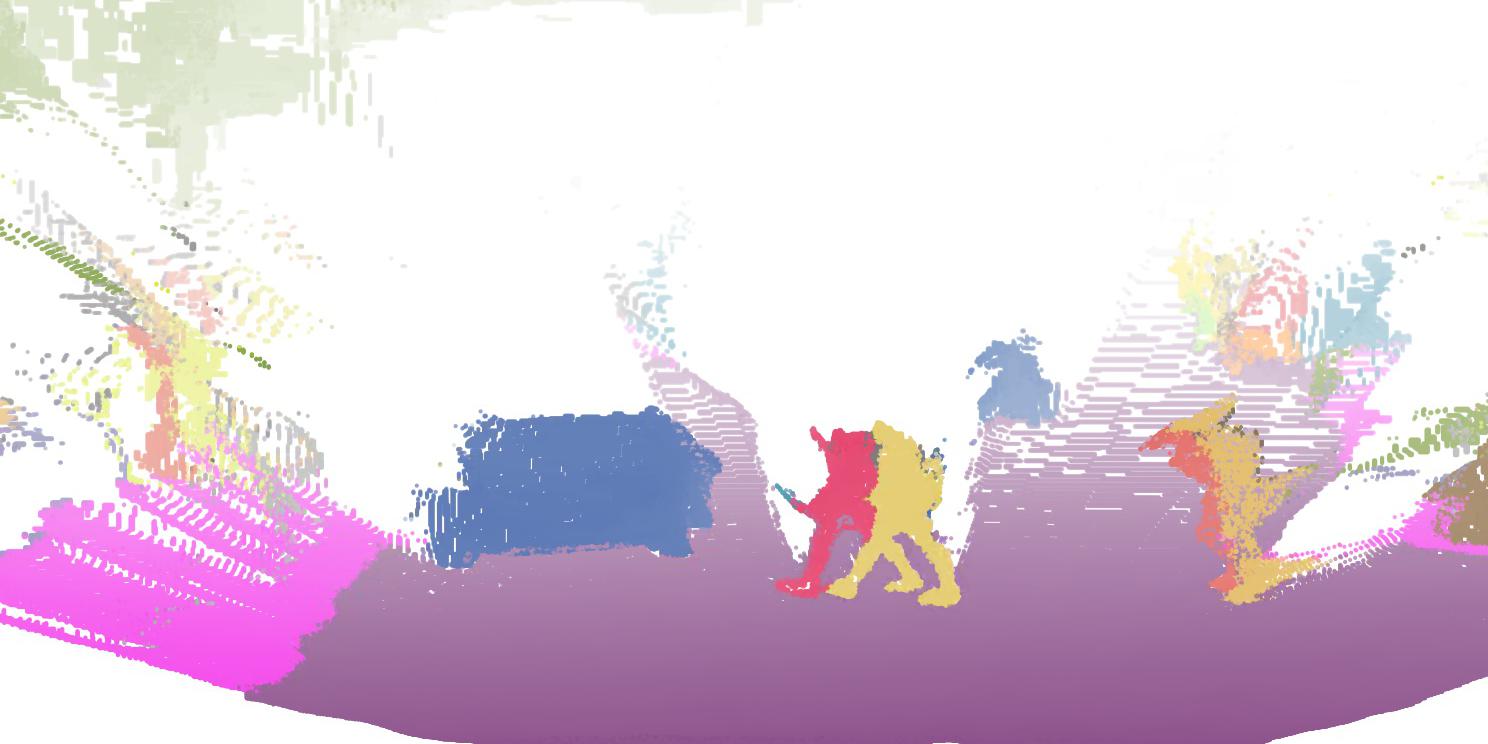}
\end{subfigure}\\\begin{subfigure}{.245\linewidth}
  \centering
  \includegraphics[trim={0 100 0 100},clip,width=\linewidth]{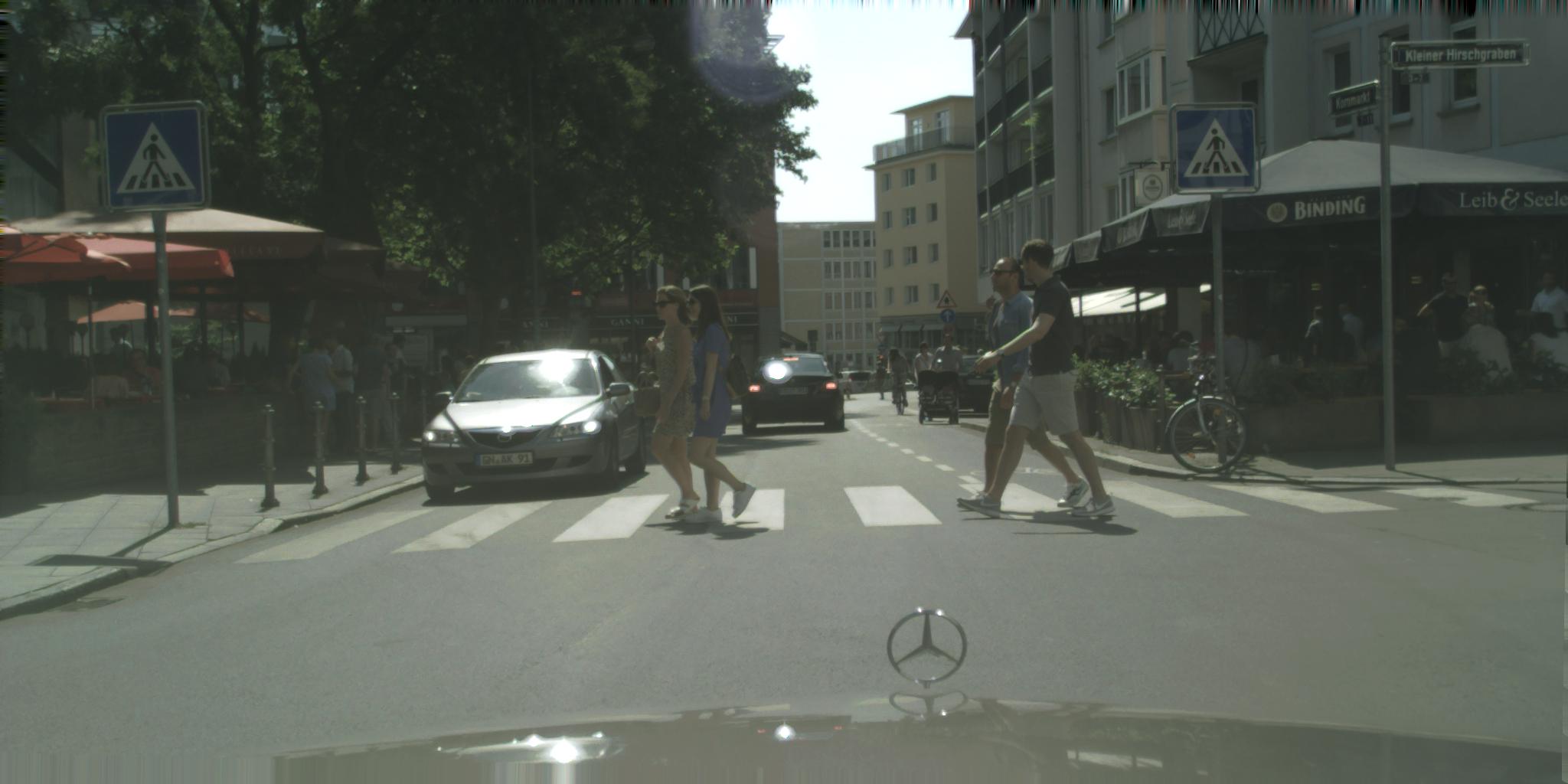}
\end{subfigure}\hfill\begin{subfigure}{.245\linewidth}
  \centering
  \includegraphics[trim={0 100 0 100},clip,width=\linewidth]{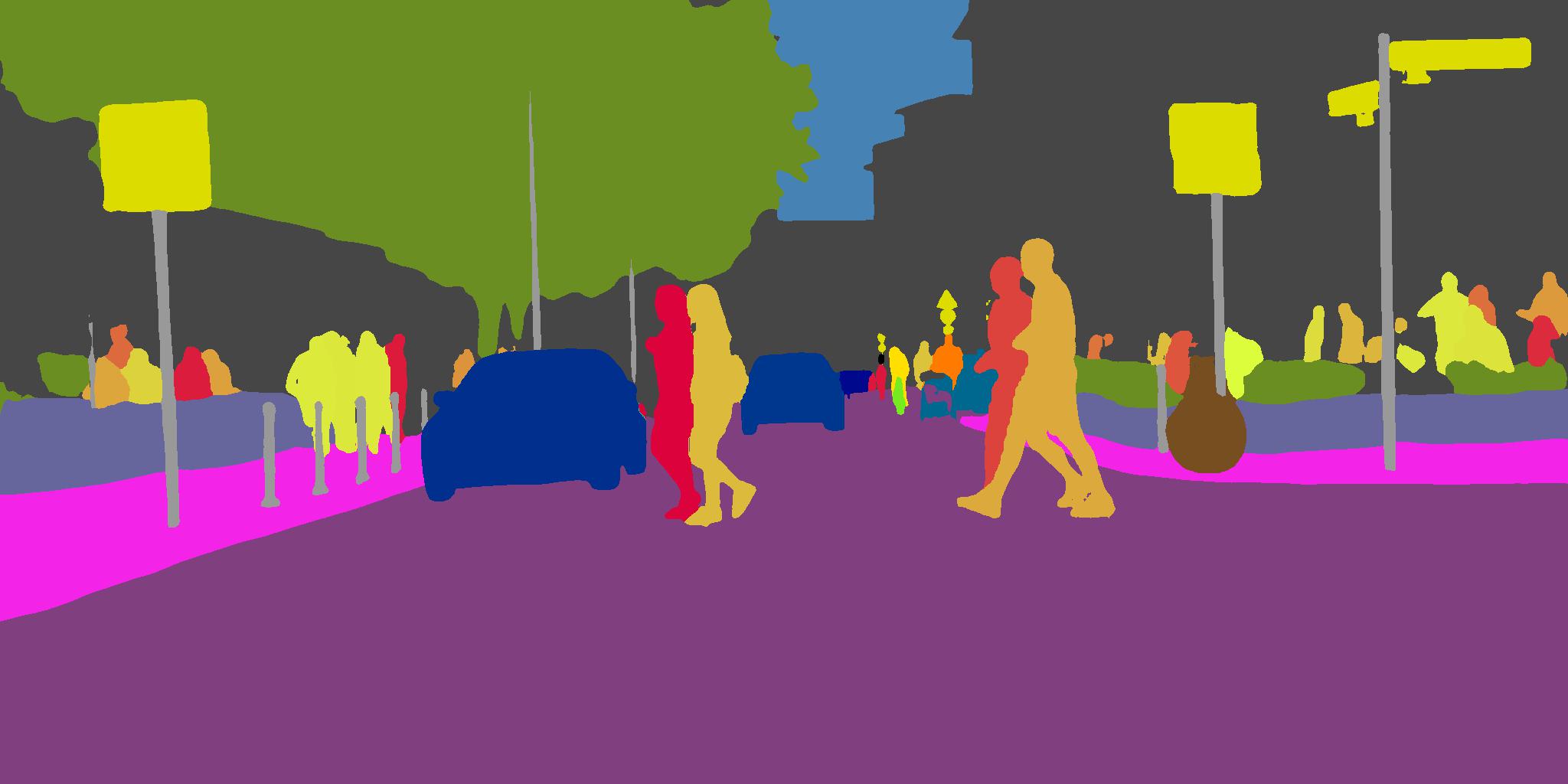}
\end{subfigure}\hfill\begin{subfigure}{.245\linewidth}
  \centering
  \includegraphics[trim={0 100 0 100},clip,width=\linewidth]{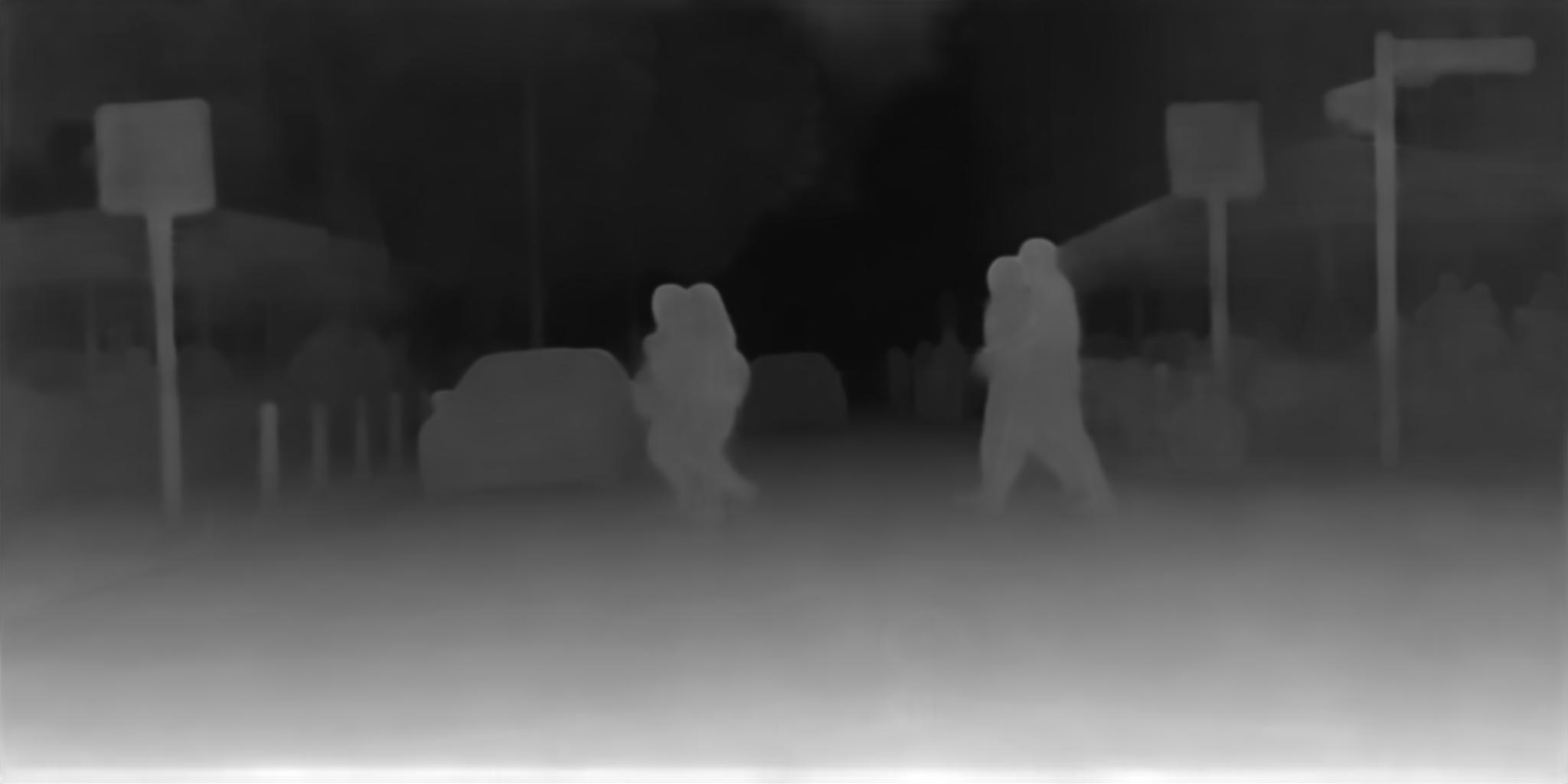}
\end{subfigure}\hfill\begin{subfigure}{.245\linewidth}
  \centering
  \includegraphics[trim={400 130 0 200},clip,width=\linewidth]{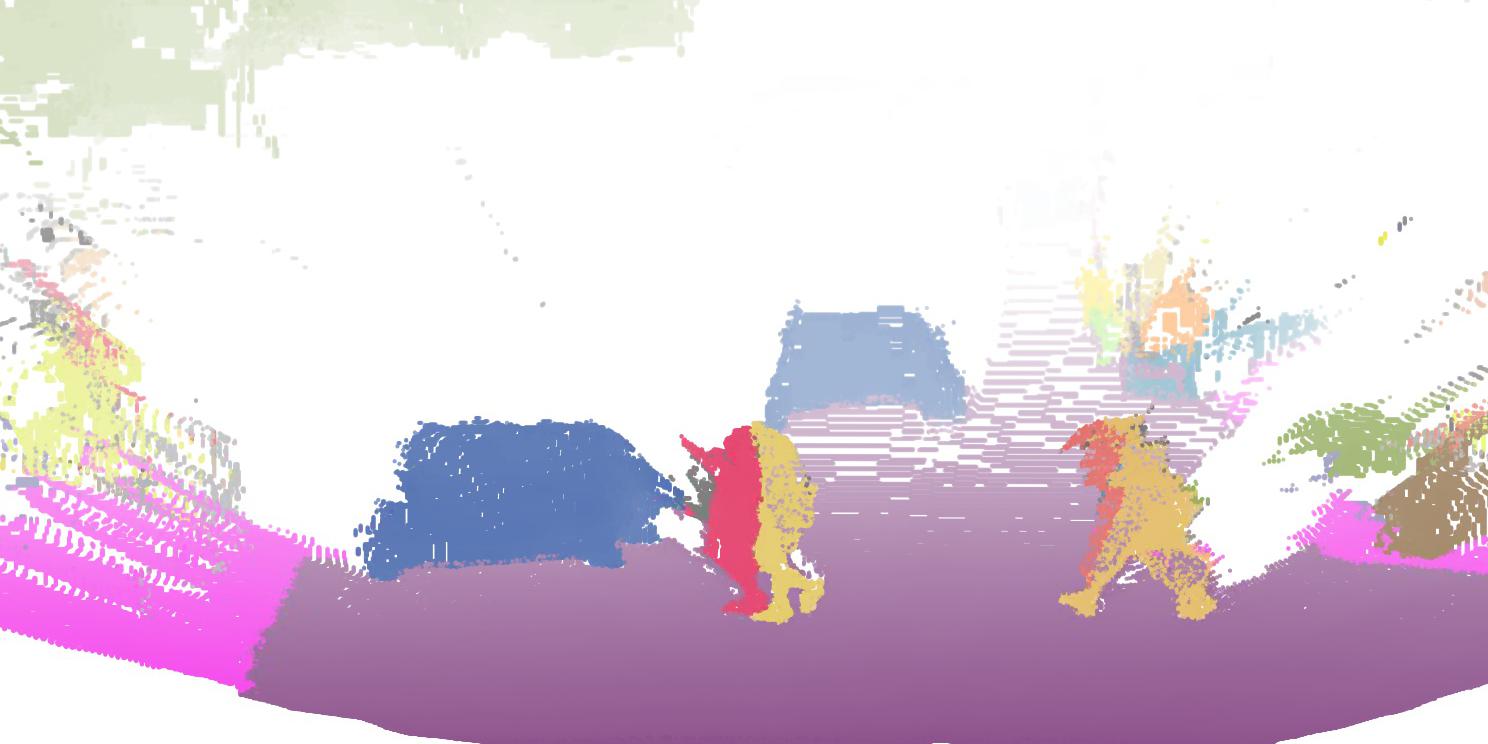}
\end{subfigure}\\
\caption{Prediction visualizations on Cityscapes-DVPS. From left to right: input image, temporally consistent panoptic segmentation prediction, monocular depth prediction, and point cloud visualization.}
\label{fig:cs_2}
\end{figure*}

\begin{figure*}[!t]
\centering
\begin{subfigure}{.245\linewidth}
  \centering
  \includegraphics[trim={150 0 150 100},clip,width=\linewidth]{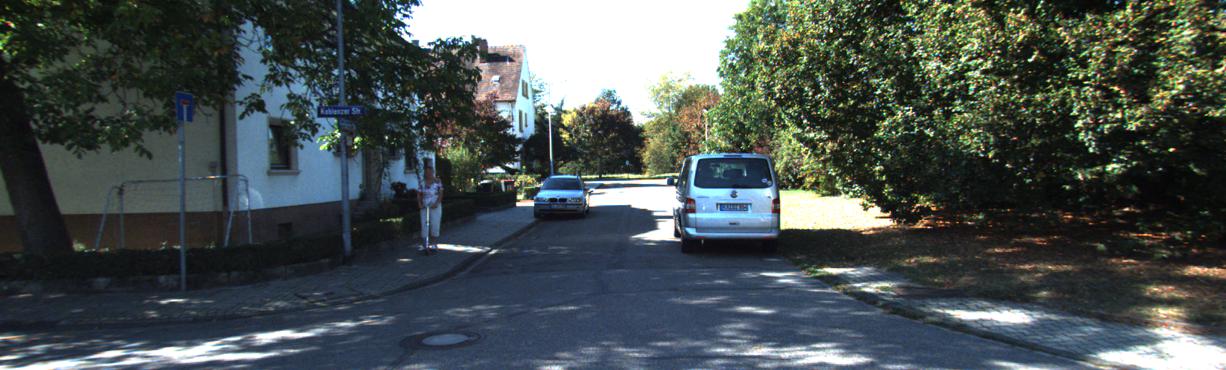}
\end{subfigure}\hfill
\begin{subfigure}{.245\linewidth}
  \centering
  \includegraphics[trim={150 0 150 100},clip,width=\linewidth]{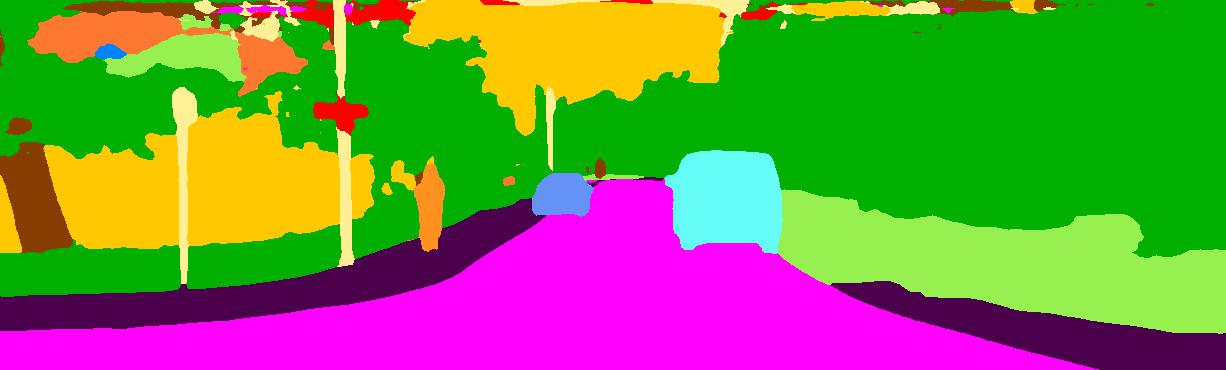}
\end{subfigure}\hfill
\begin{subfigure}{.245\linewidth}
  \centering
  \includegraphics[trim={150 0 150 100},clip,width=\linewidth]{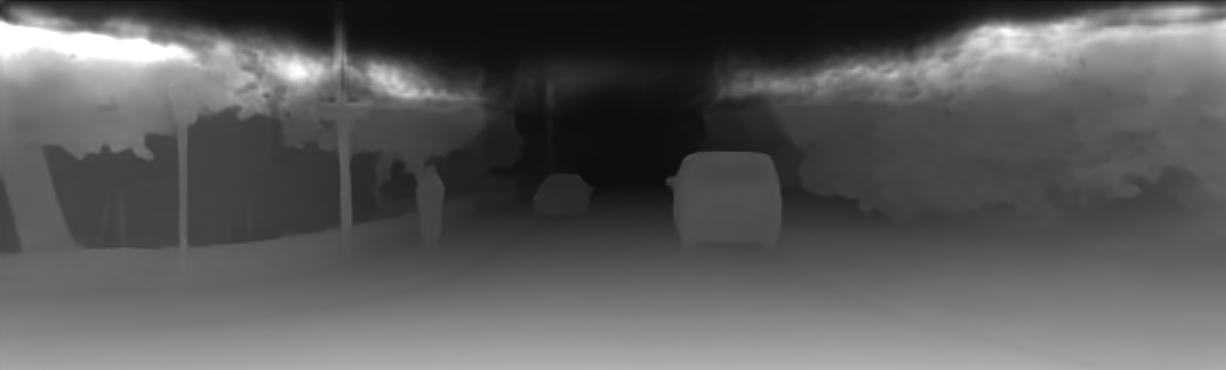}
\end{subfigure}\hfill
\begin{subfigure}{.245\linewidth}
  \centering
  \includegraphics[trim={0 0 0 300},clip,width=\linewidth]{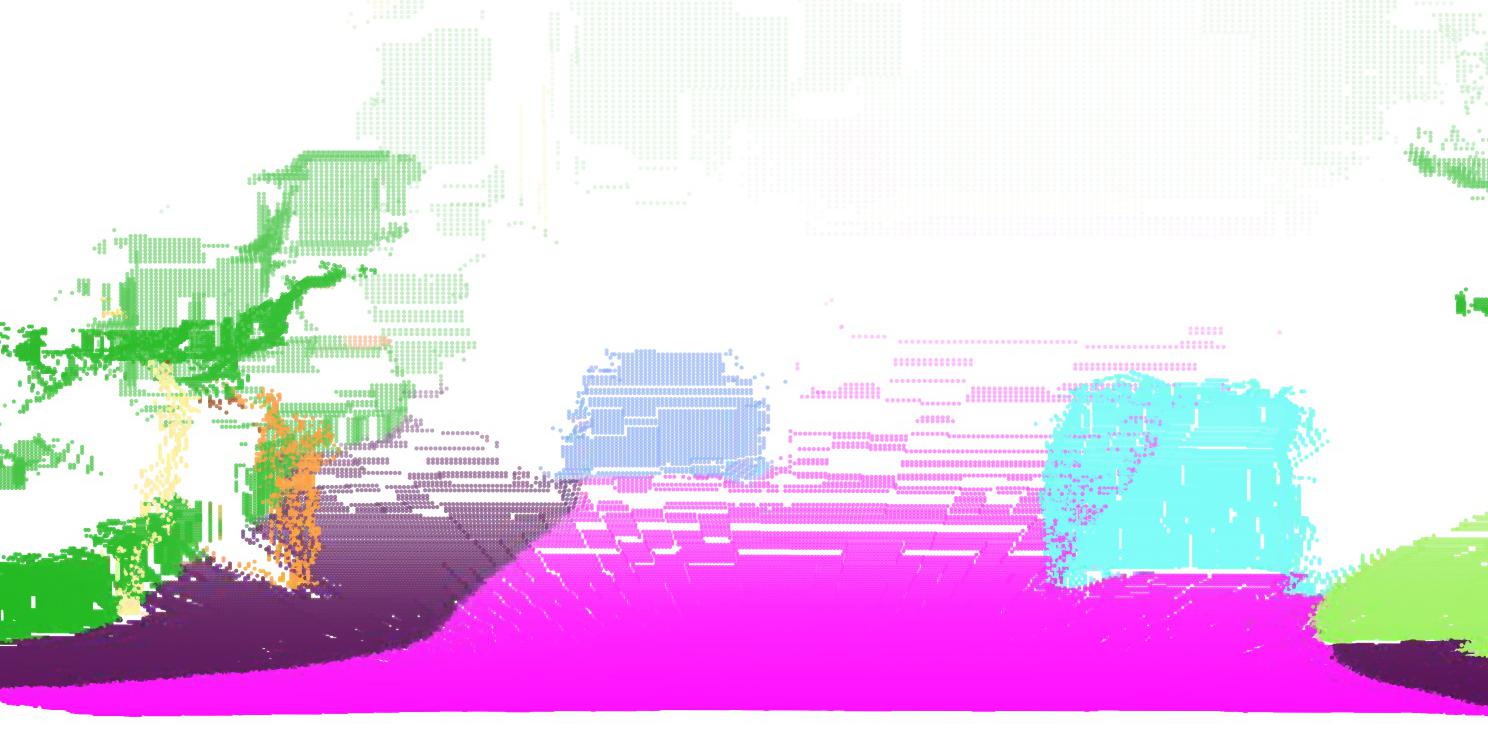}
\end{subfigure}\\
\begin{subfigure}{.245\linewidth}
  \centering
  \includegraphics[trim={150 0 150 100},clip,width=\linewidth]{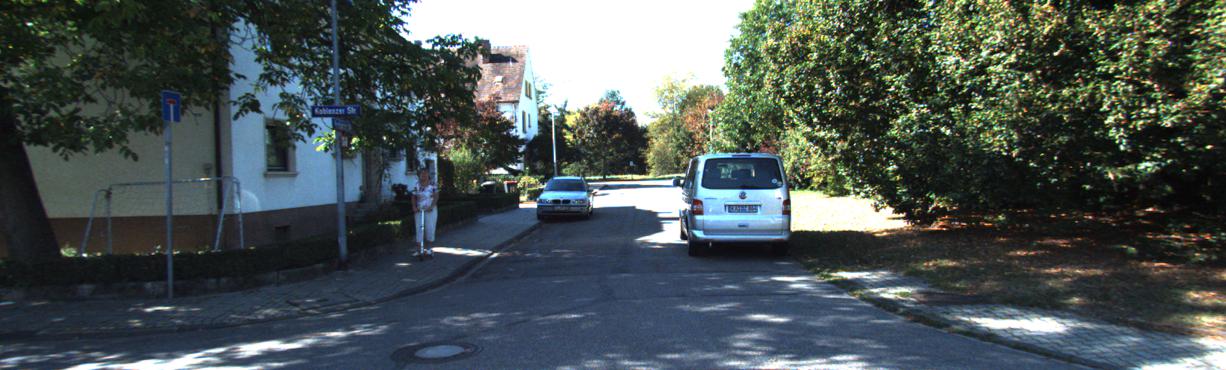}
\end{subfigure}\hfill
\begin{subfigure}{.245\linewidth}
  \centering
  \includegraphics[trim={150 0 150 100},clip,width=\linewidth]{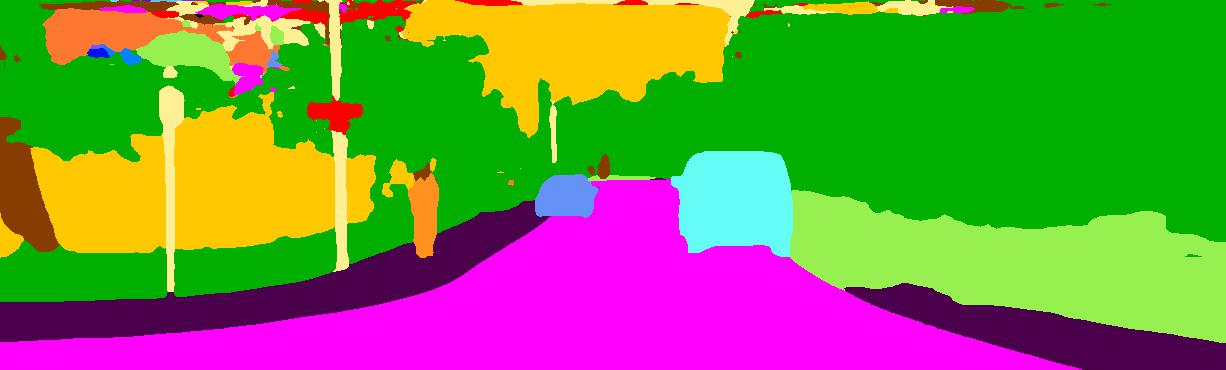}
\end{subfigure}\hfill
\begin{subfigure}{.245\linewidth}
  \centering
  \includegraphics[trim={150 0 150 100},clip,width=\linewidth]{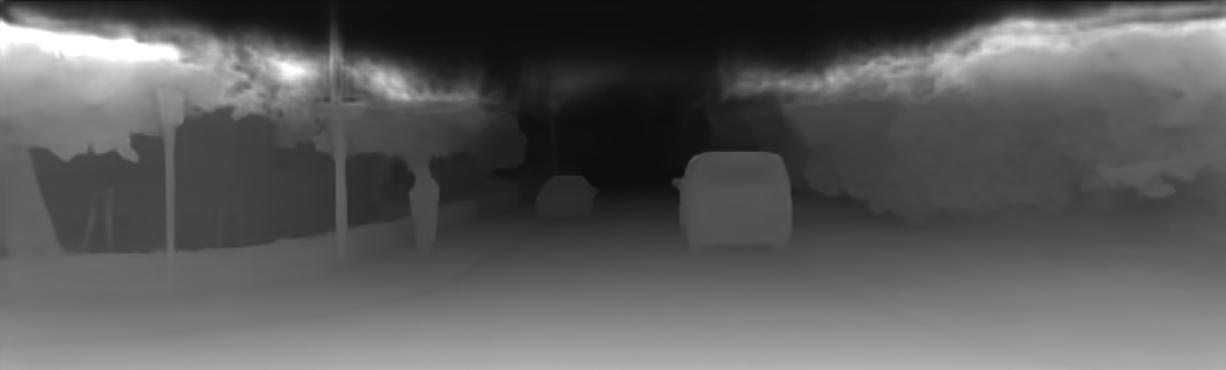}
\end{subfigure}\hfill
\begin{subfigure}{.245\linewidth}
  \centering
  \includegraphics[trim={0 0 0 300},clip,width=\linewidth]{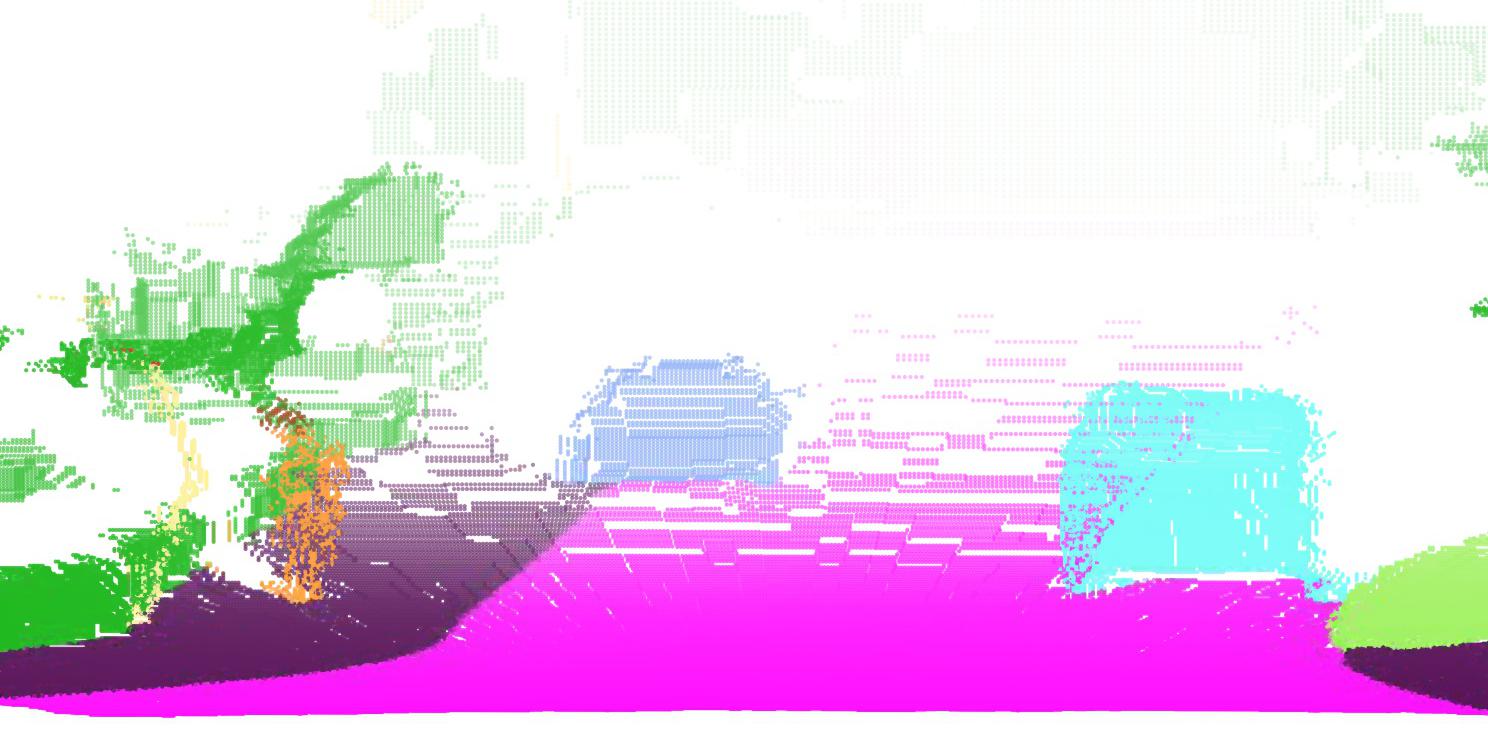}
\end{subfigure}\\
\begin{subfigure}{.245\linewidth}
  \centering
  \includegraphics[trim={150 0 150 100},clip,width=\linewidth]{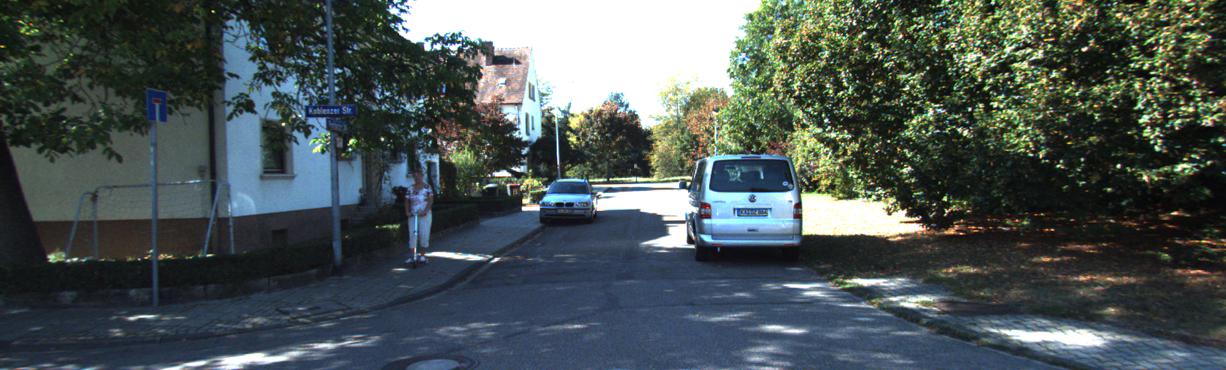}
\end{subfigure}\hfill
\begin{subfigure}{.245\linewidth}
  \centering
  \includegraphics[trim={150 0 150 100},clip,width=\linewidth]{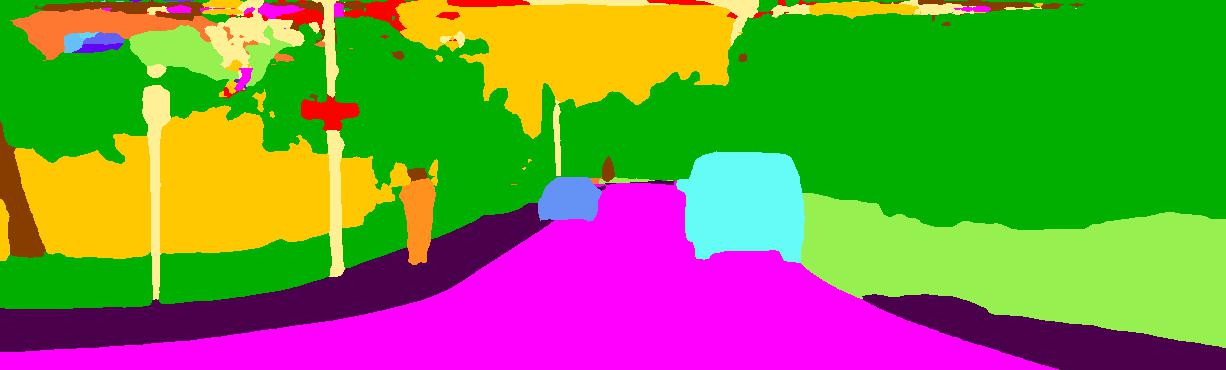}
\end{subfigure}\hfill
\begin{subfigure}{.245\linewidth}
  \centering
  \includegraphics[trim={150 0 150 100},clip,width=\linewidth]{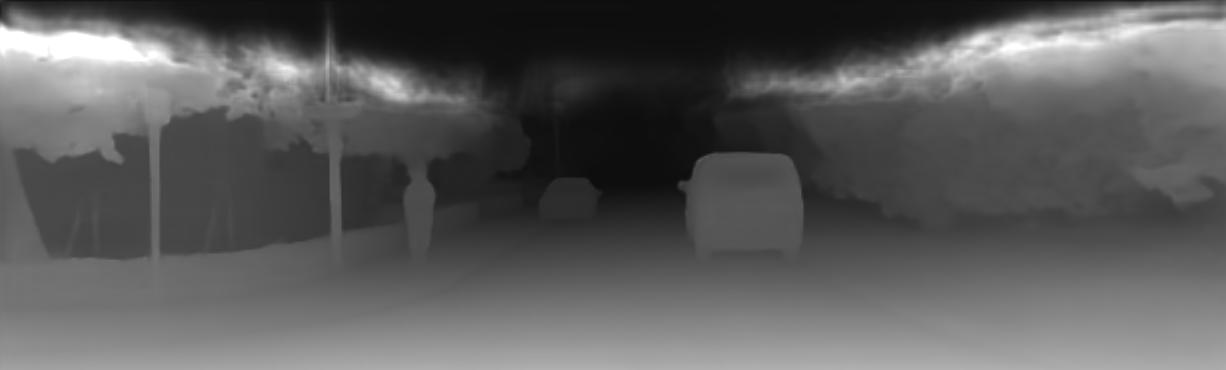}
\end{subfigure}\hfill
\begin{subfigure}{.245\linewidth}
  \centering
  \includegraphics[trim={0 0 0 300},clip,width=\linewidth]{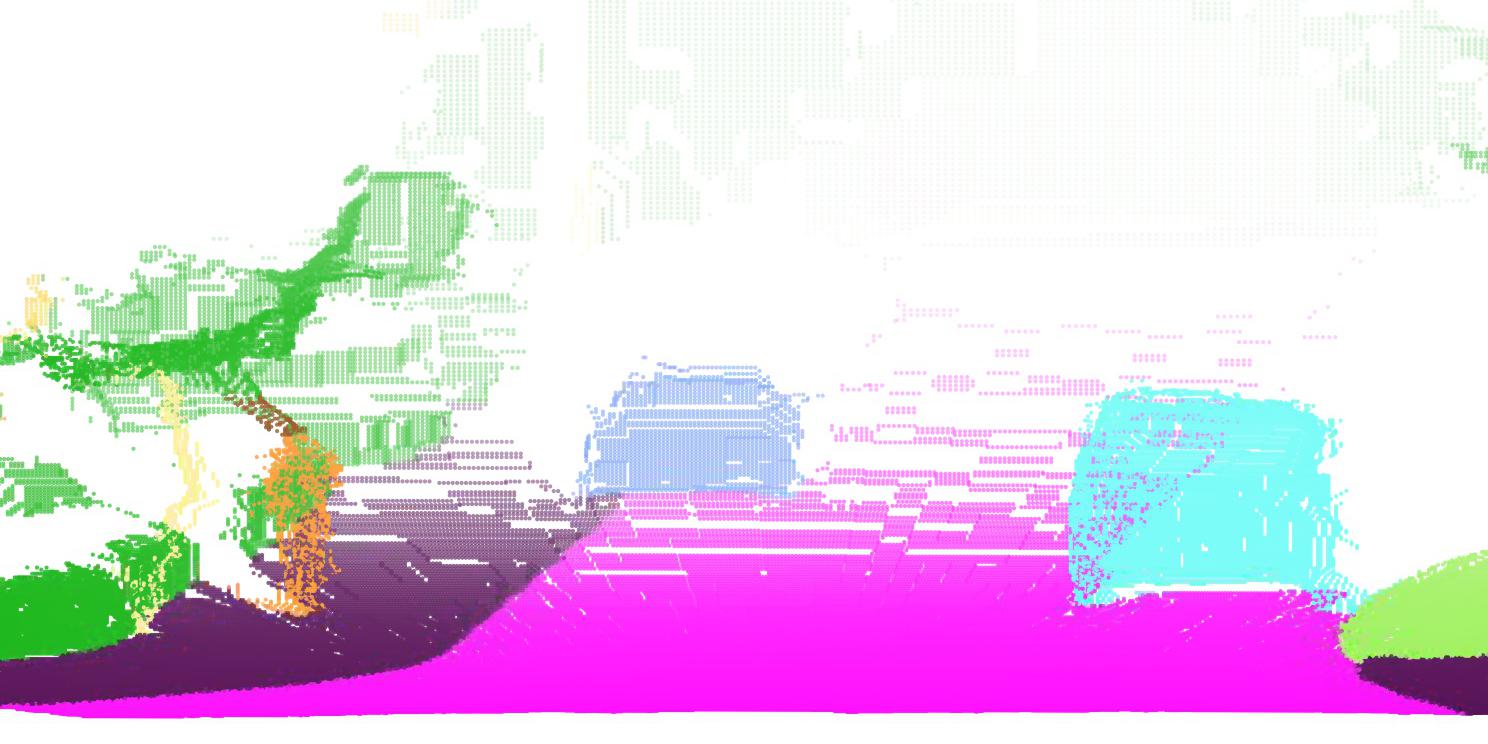}
\end{subfigure}\\
\begin{subfigure}{.245\linewidth}
  \centering
  \includegraphics[trim={150 0 150 100},clip,width=\linewidth]{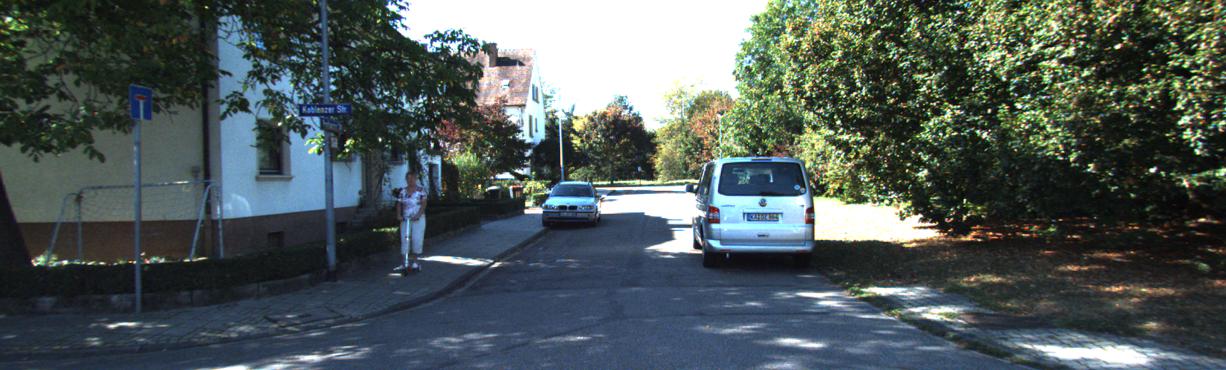}
\end{subfigure}\hfill
\begin{subfigure}{.245\linewidth}
  \centering
  \includegraphics[trim={150 0 150 100},clip,width=\linewidth]{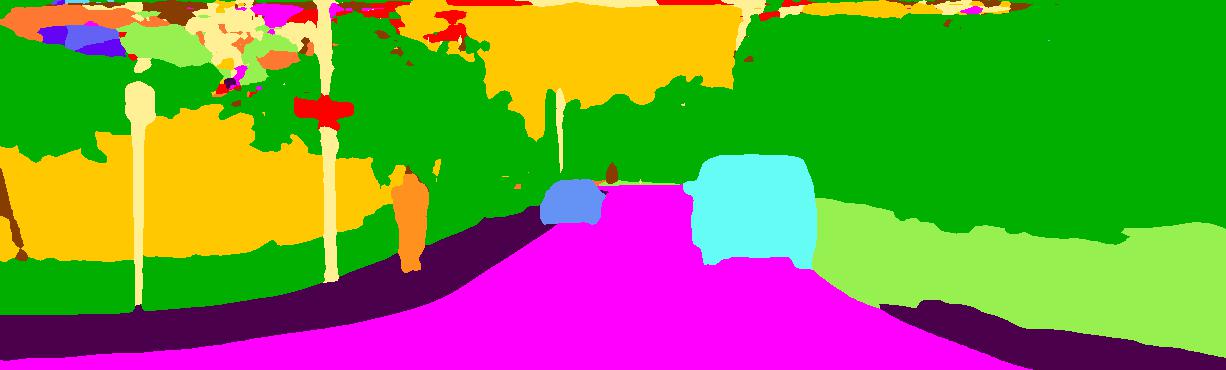}
\end{subfigure}\hfill
\begin{subfigure}{.245\linewidth}
  \centering
  \includegraphics[trim={150 0 150 100},clip,width=\linewidth]{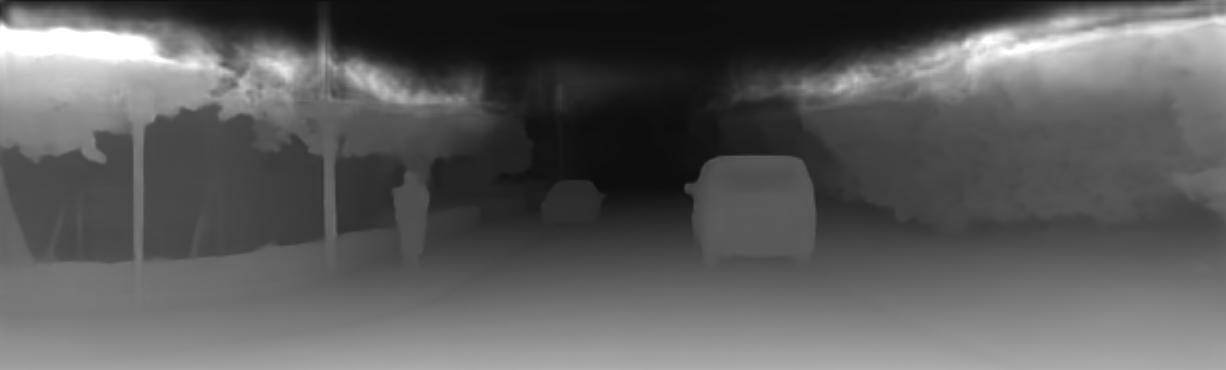}
\end{subfigure}\hfill
\begin{subfigure}{.245\linewidth}
  \centering
  \includegraphics[trim={0 0 0 300},clip,width=\linewidth]{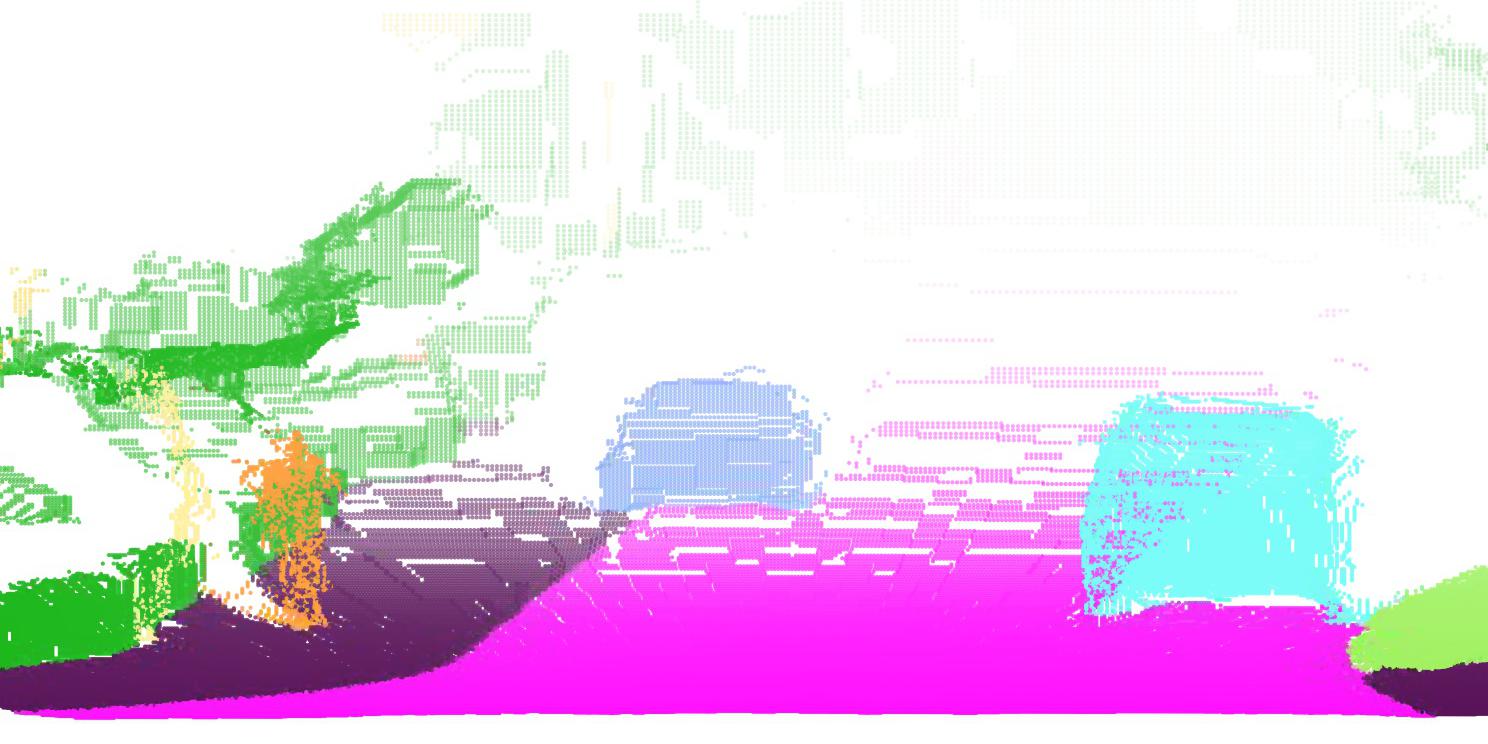}
\end{subfigure}\\
\begin{subfigure}{.245\linewidth}
  \centering
  \includegraphics[trim={150 0 150 100},clip,width=\linewidth]{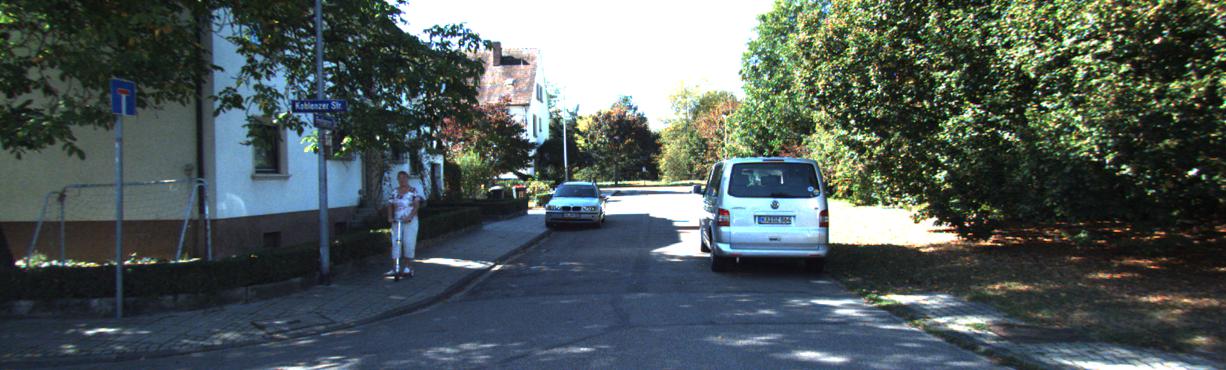}
\end{subfigure}\hfill
\begin{subfigure}{.245\linewidth}
  \centering
  \includegraphics[trim={150 0 150 100},clip,width=\linewidth]{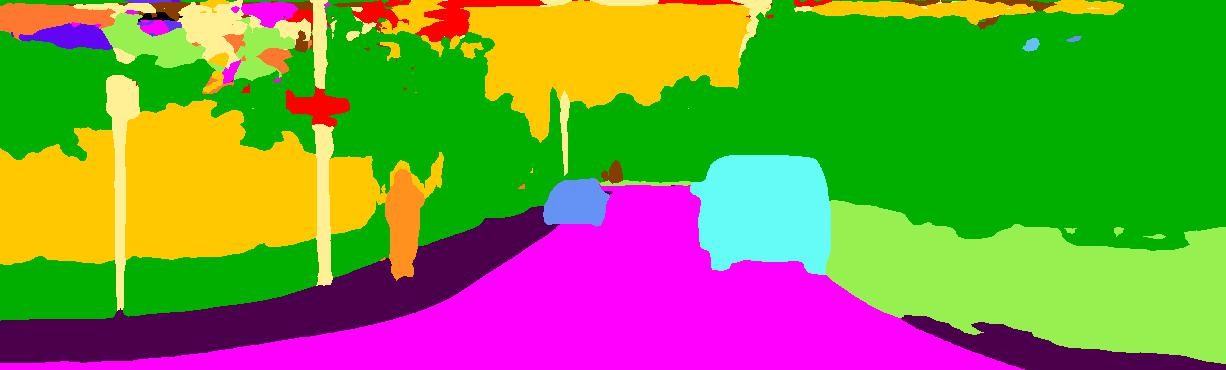}
\end{subfigure}\hfill
\begin{subfigure}{.245\linewidth}
  \centering
  \includegraphics[trim={150 0 150 100},clip,width=\linewidth]{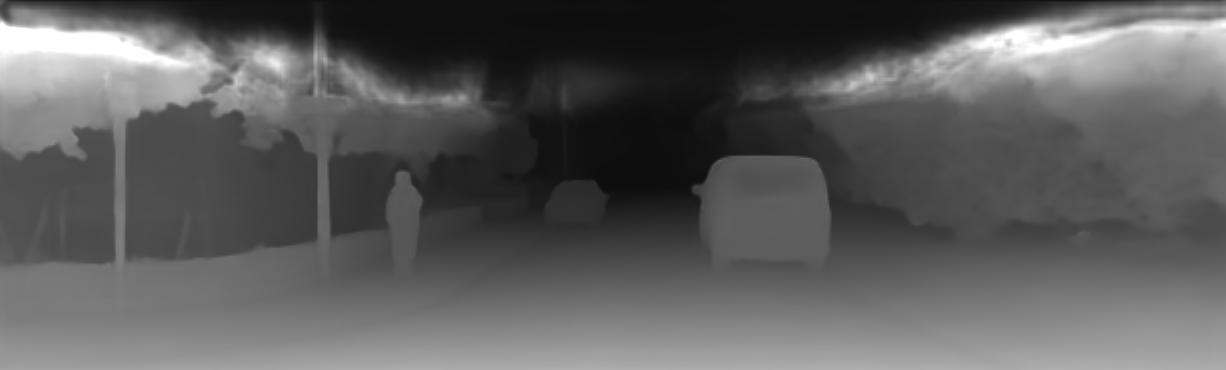}
\end{subfigure}\hfill
\begin{subfigure}{.245\linewidth}
  \centering
  \includegraphics[trim={0 0 0 300},clip,width=\linewidth]{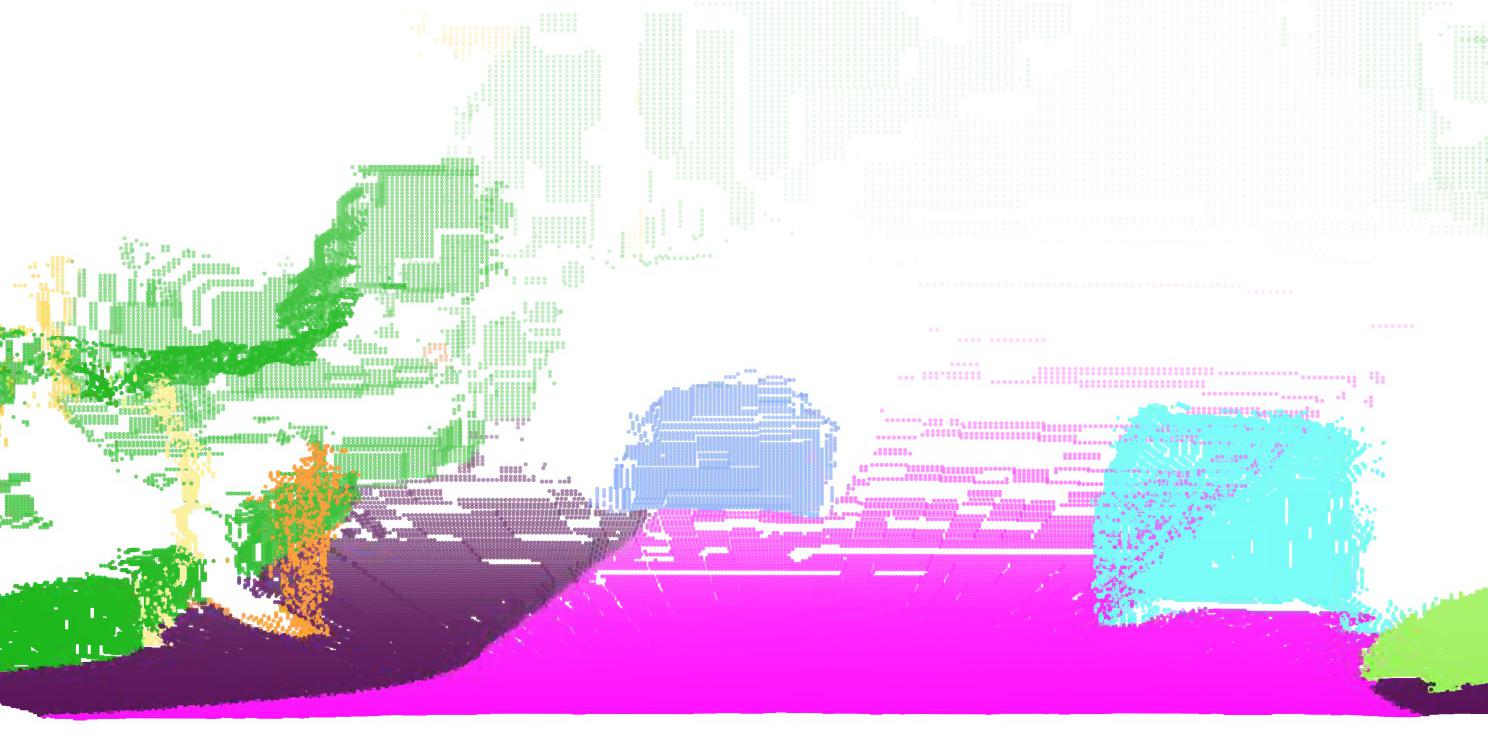}
\end{subfigure}\\
\begin{subfigure}{.245\linewidth}
  \centering
  \includegraphics[trim={150 0 150 100},clip,width=\linewidth]{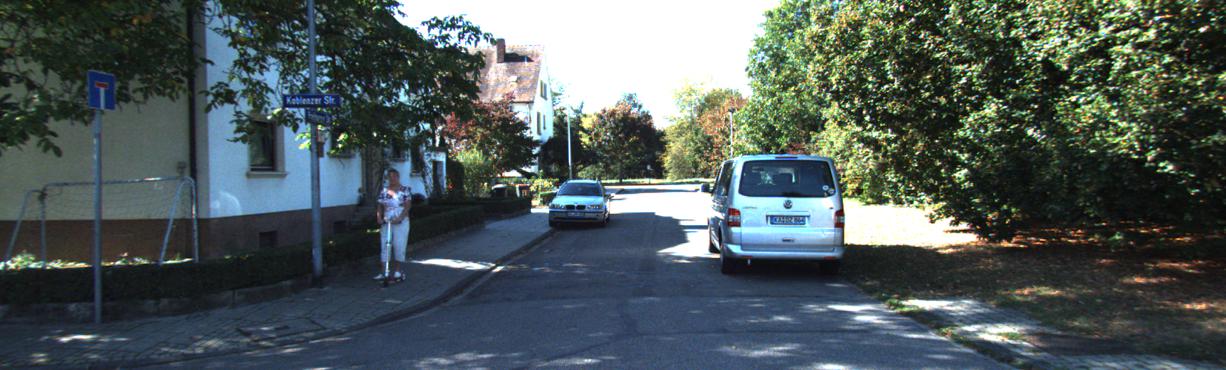}
\end{subfigure}\hfill
\begin{subfigure}{.245\linewidth}
  \centering
  \includegraphics[trim={150 0 150 100},clip,width=\linewidth]{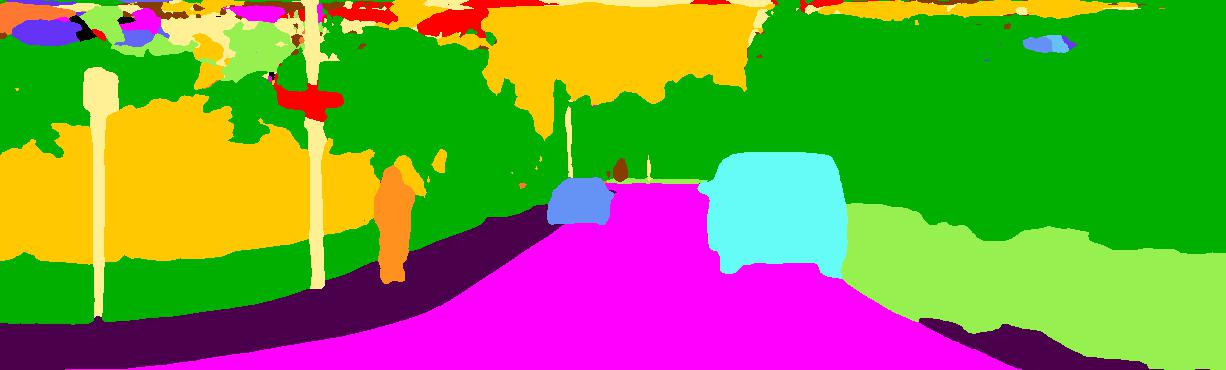}
\end{subfigure}\hfill
\begin{subfigure}{.245\linewidth}
  \centering
  \includegraphics[trim={150 0 150 100},clip,width=\linewidth]{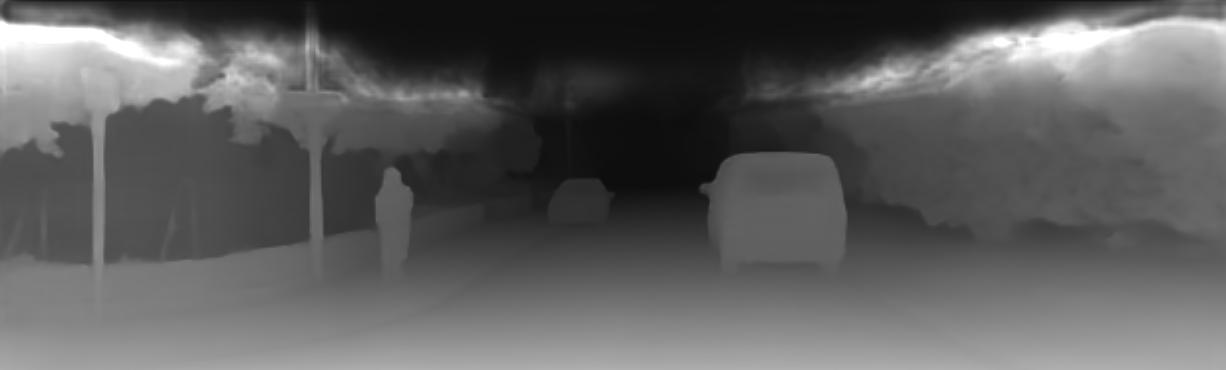}
\end{subfigure}\hfill
\begin{subfigure}{.245\linewidth}
  \centering
  \includegraphics[trim={0 0 0 300},clip,width=\linewidth]{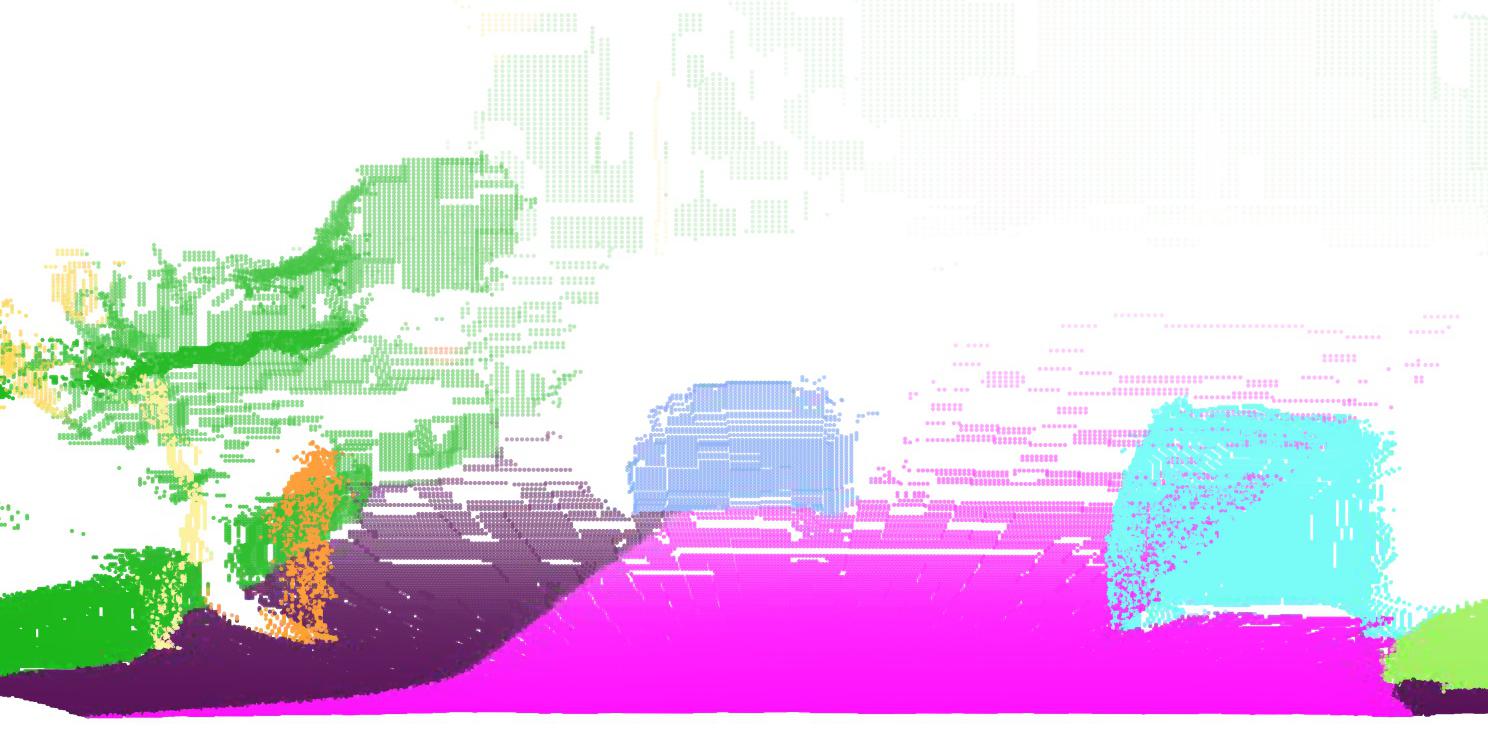}
\end{subfigure}\\
\begin{subfigure}{.245\linewidth}
  \centering
  \includegraphics[trim={150 0 150 100},clip,width=\linewidth]{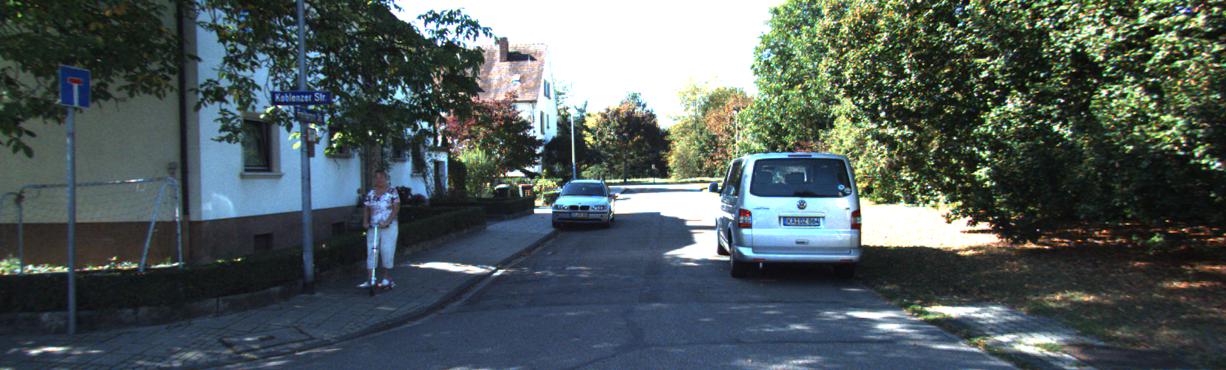}
\end{subfigure}\hfill
\begin{subfigure}{.245\linewidth}
  \centering
  \includegraphics[trim={150 0 150 100},clip,width=\linewidth]{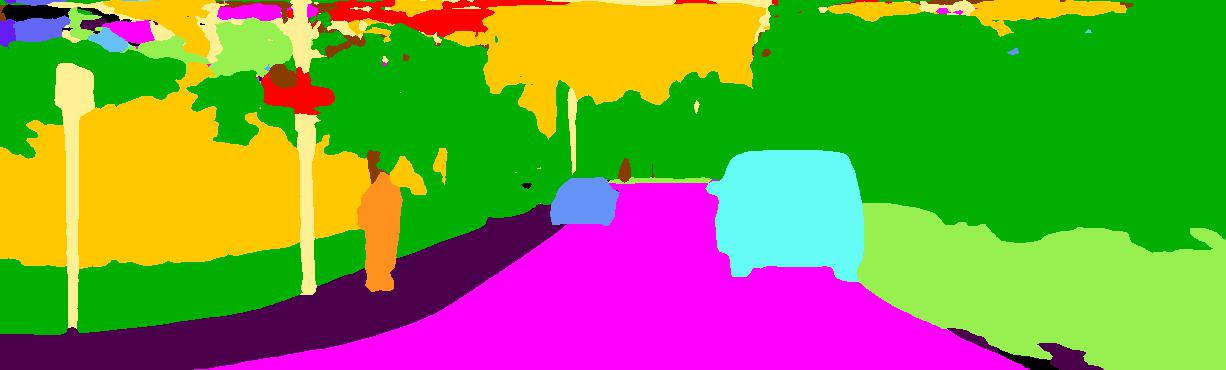}
\end{subfigure}\hfill
\begin{subfigure}{.245\linewidth}
  \centering
  \includegraphics[trim={150 0 150 100},clip,width=\linewidth]{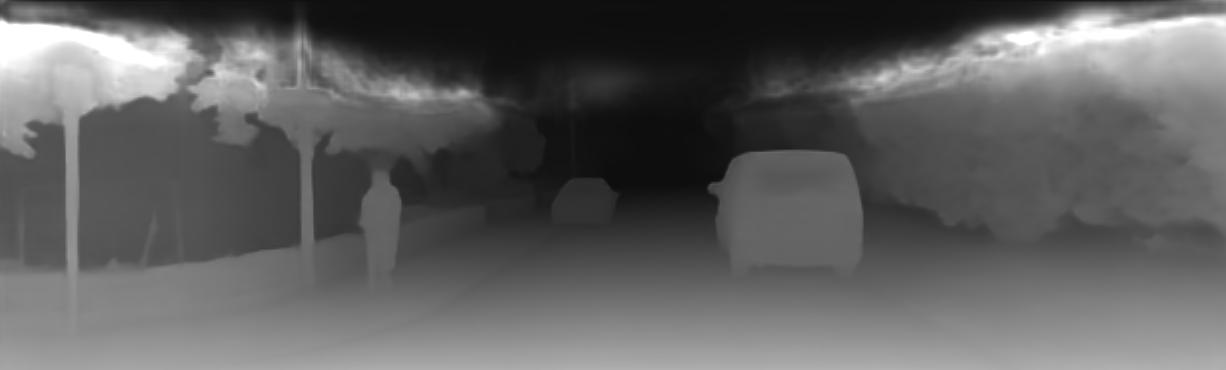}
\end{subfigure}\hfill
\begin{subfigure}{.245\linewidth}
  \centering
  \includegraphics[trim={0 0 0 300},clip,width=\linewidth]{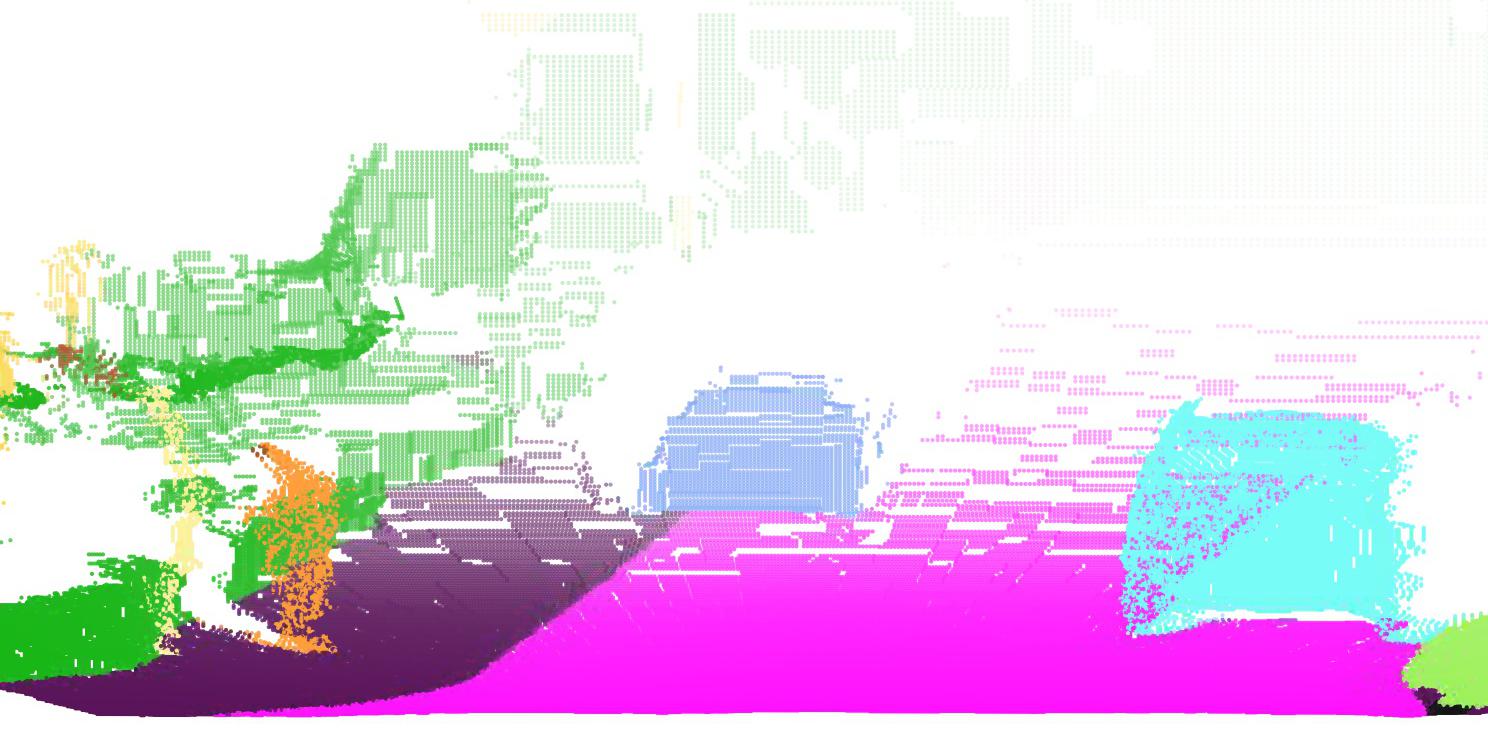}
\end{subfigure}\\
\begin{subfigure}{.245\linewidth}
  \centering
  \includegraphics[trim={150 0 150 100},clip,width=\linewidth]{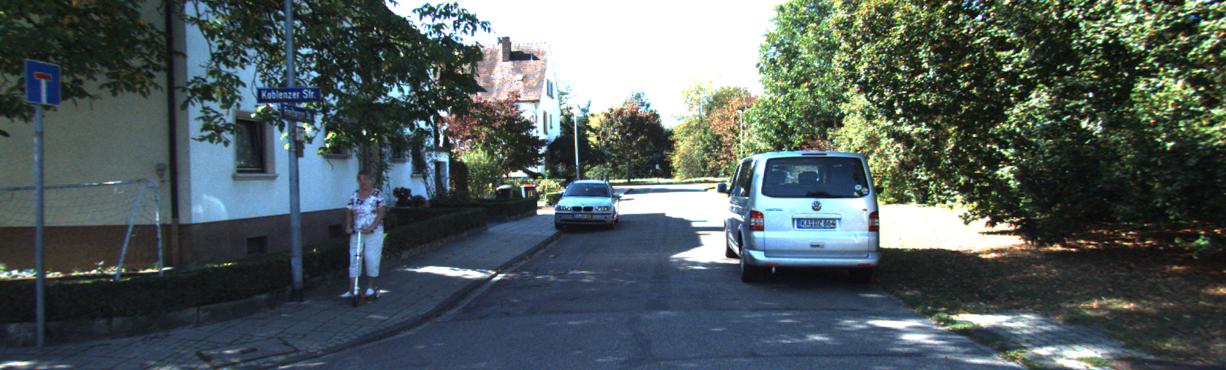}
\end{subfigure}\hfill
\begin{subfigure}{.245\linewidth}
  \centering
  \includegraphics[trim={150 0 150 100},clip,width=\linewidth]{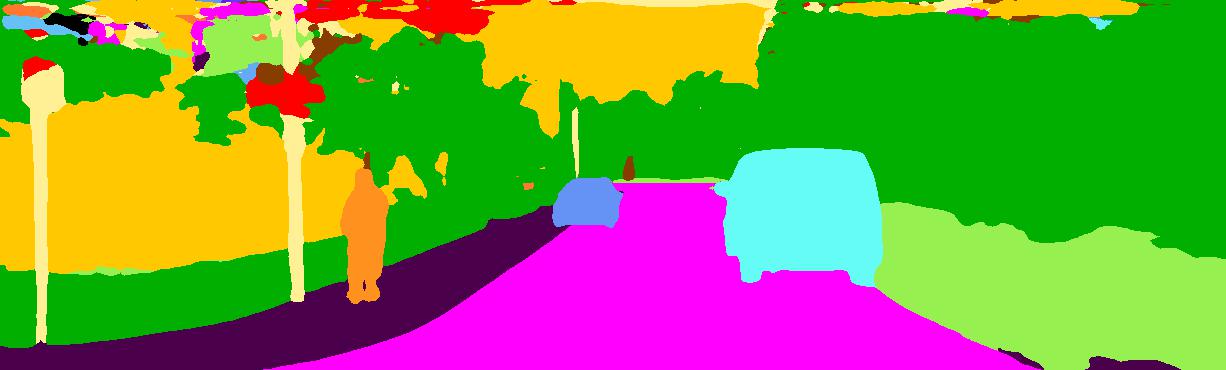}
\end{subfigure}\hfill
\begin{subfigure}{.245\linewidth}
  \centering
  \includegraphics[trim={150 0 150 100},clip,width=\linewidth]{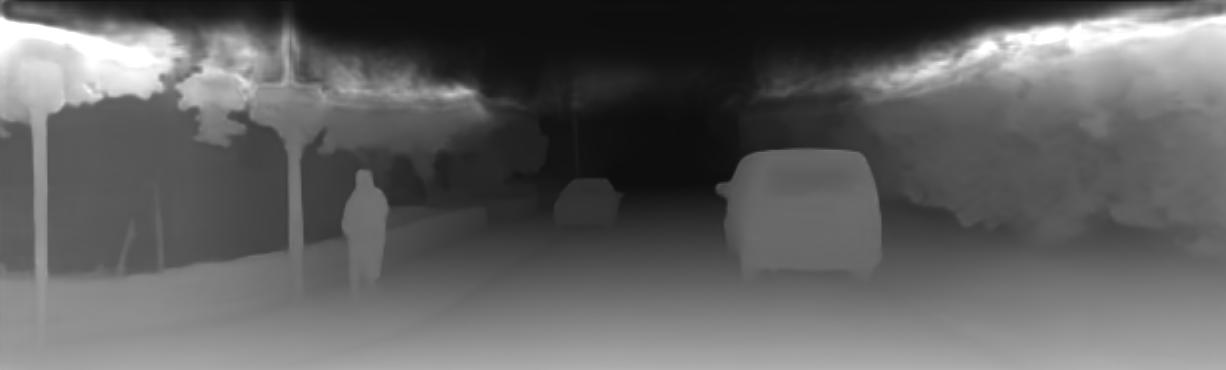}
\end{subfigure}\hfill
\begin{subfigure}{.245\linewidth}
  \centering
  \includegraphics[trim={0 0 0 300},clip,width=\linewidth]{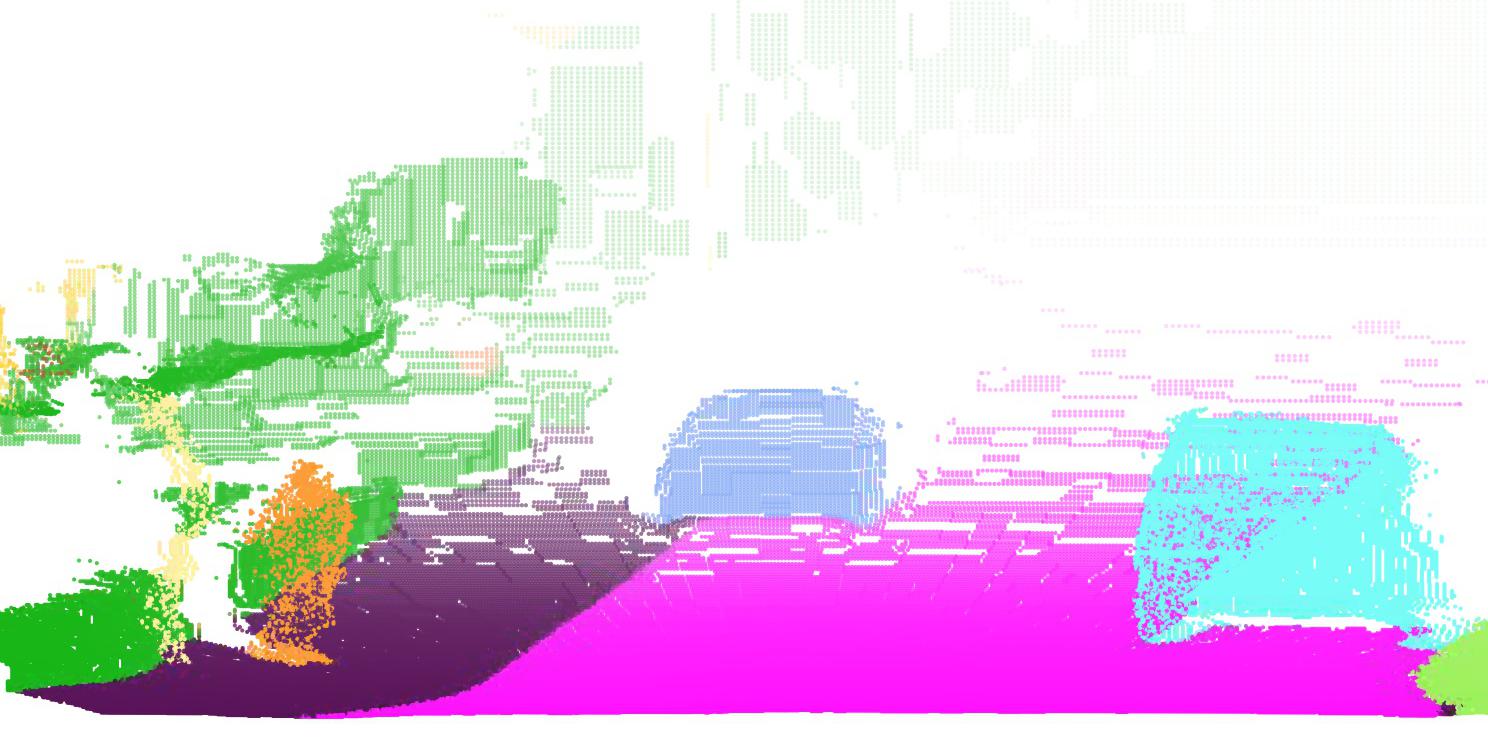}
\end{subfigure}\\
\begin{subfigure}{.245\linewidth}
  \centering
  \includegraphics[trim={150 0 150 100},clip,width=\linewidth]{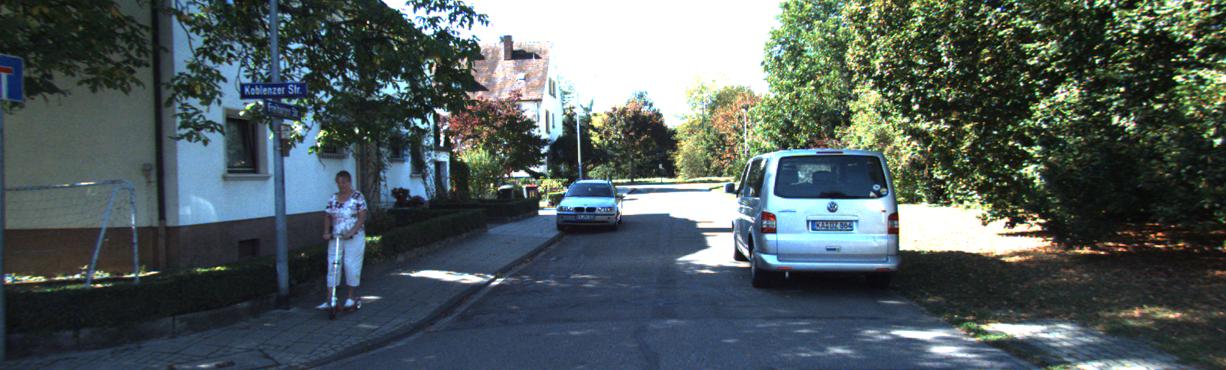}
\end{subfigure}\hfill
\begin{subfigure}{.245\linewidth}
  \centering
  \includegraphics[trim={150 0 150 100},clip,width=\linewidth]{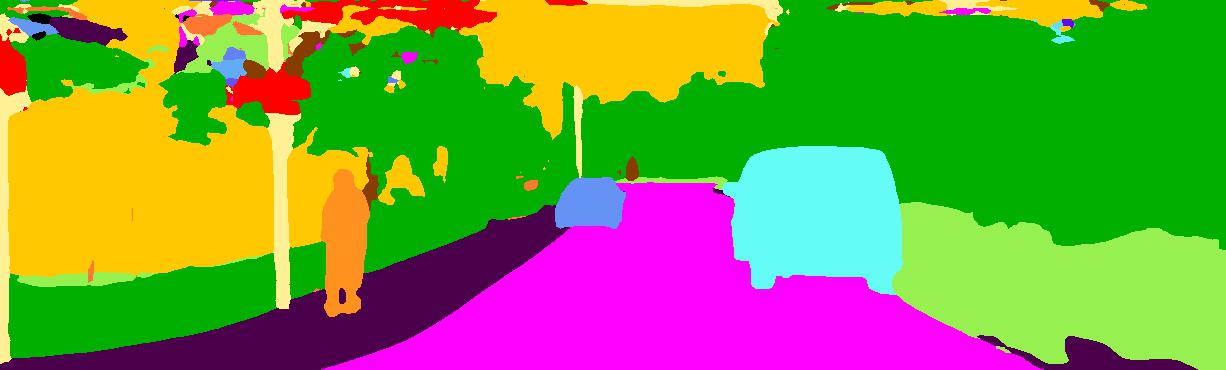}
\end{subfigure}\hfill
\begin{subfigure}{.245\linewidth}
  \centering
  \includegraphics[trim={150 0 150 100},clip,width=\linewidth]{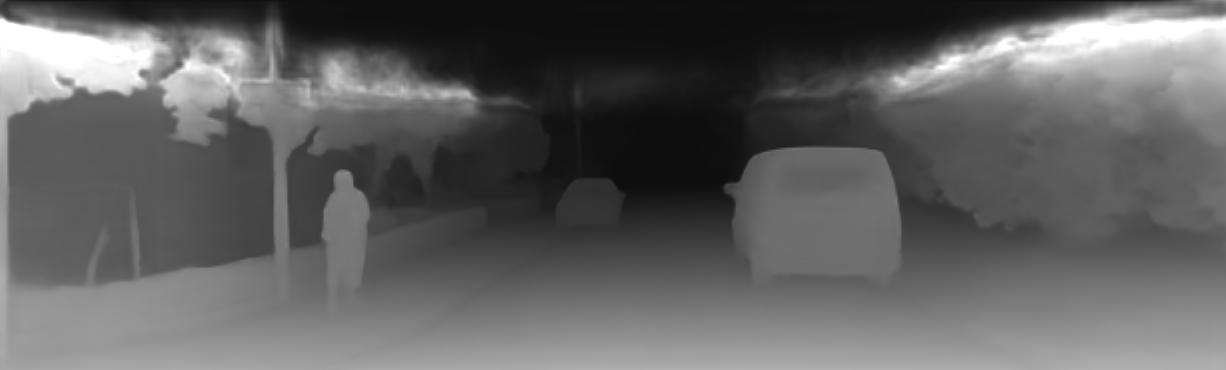}
\end{subfigure}\hfill
\begin{subfigure}{.245\linewidth}
  \centering
  \includegraphics[trim={0 0 0 300},clip,width=\linewidth]{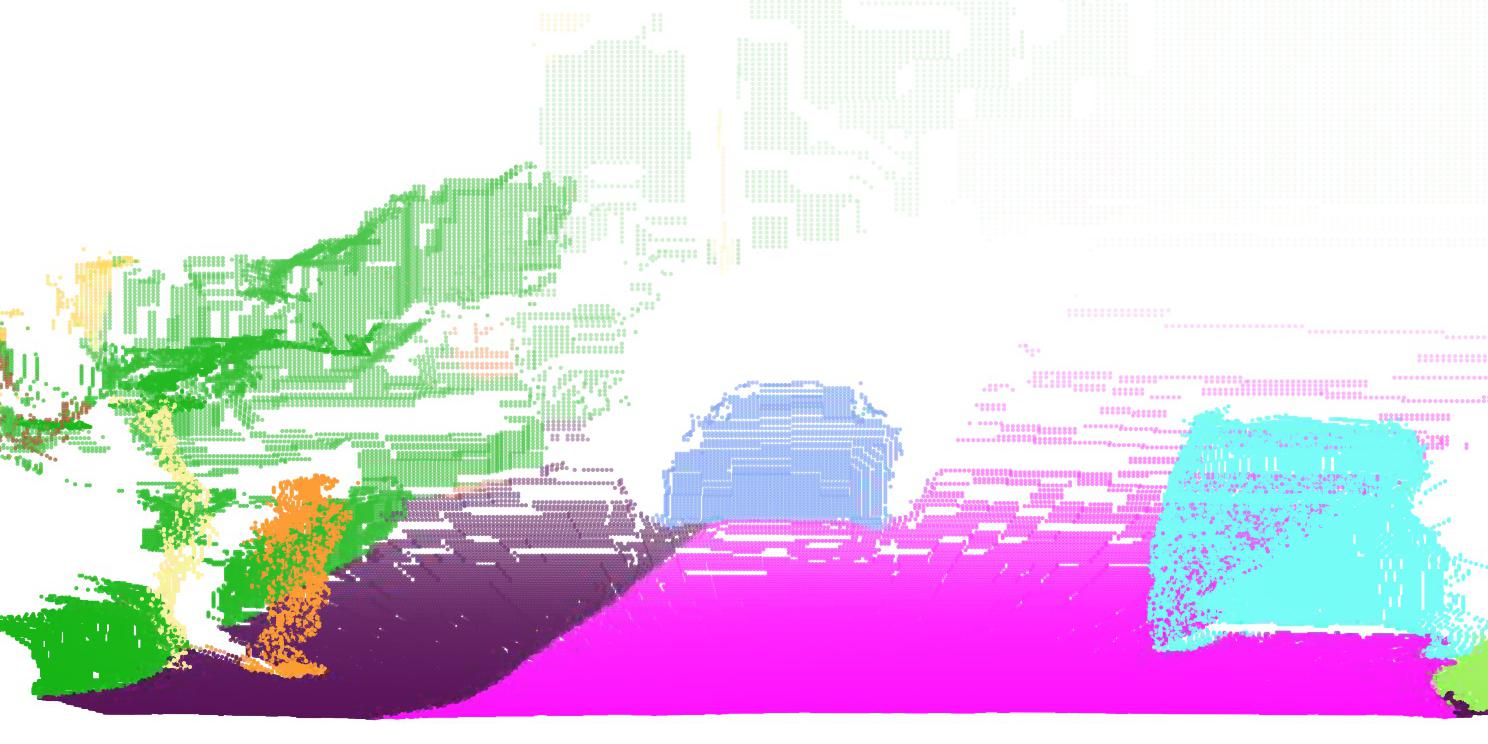}
\end{subfigure}\\
\begin{subfigure}{.245\linewidth}
  \centering
  \includegraphics[trim={150 0 150 100},clip,width=\linewidth]{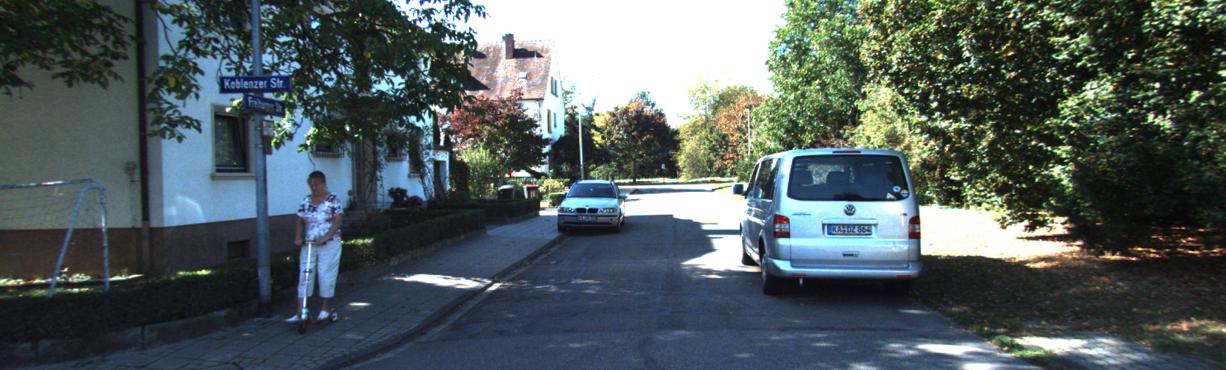}
\end{subfigure}\hfill
\begin{subfigure}{.245\linewidth}
  \centering
  \includegraphics[trim={150 0 150 100},clip,width=\linewidth]{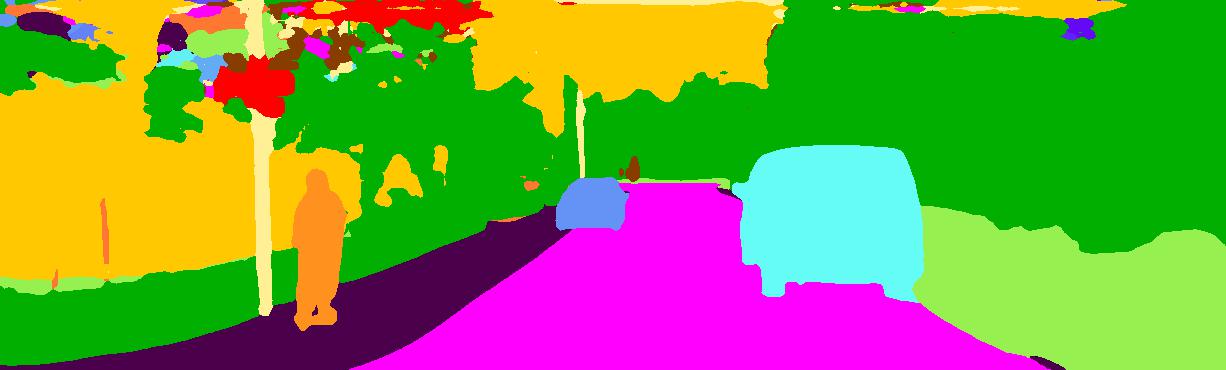}
\end{subfigure}\hfill
\begin{subfigure}{.245\linewidth}
  \centering
  \includegraphics[trim={150 0 150 100},clip,width=\linewidth]{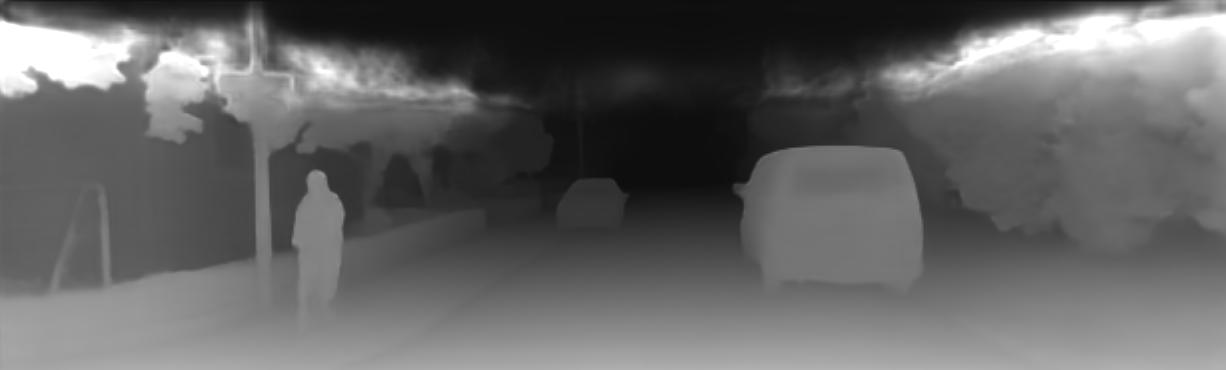}
\end{subfigure}\hfill
\begin{subfigure}{.245\linewidth}
  \centering
  \includegraphics[trim={0 0 0 300},clip,width=\linewidth]{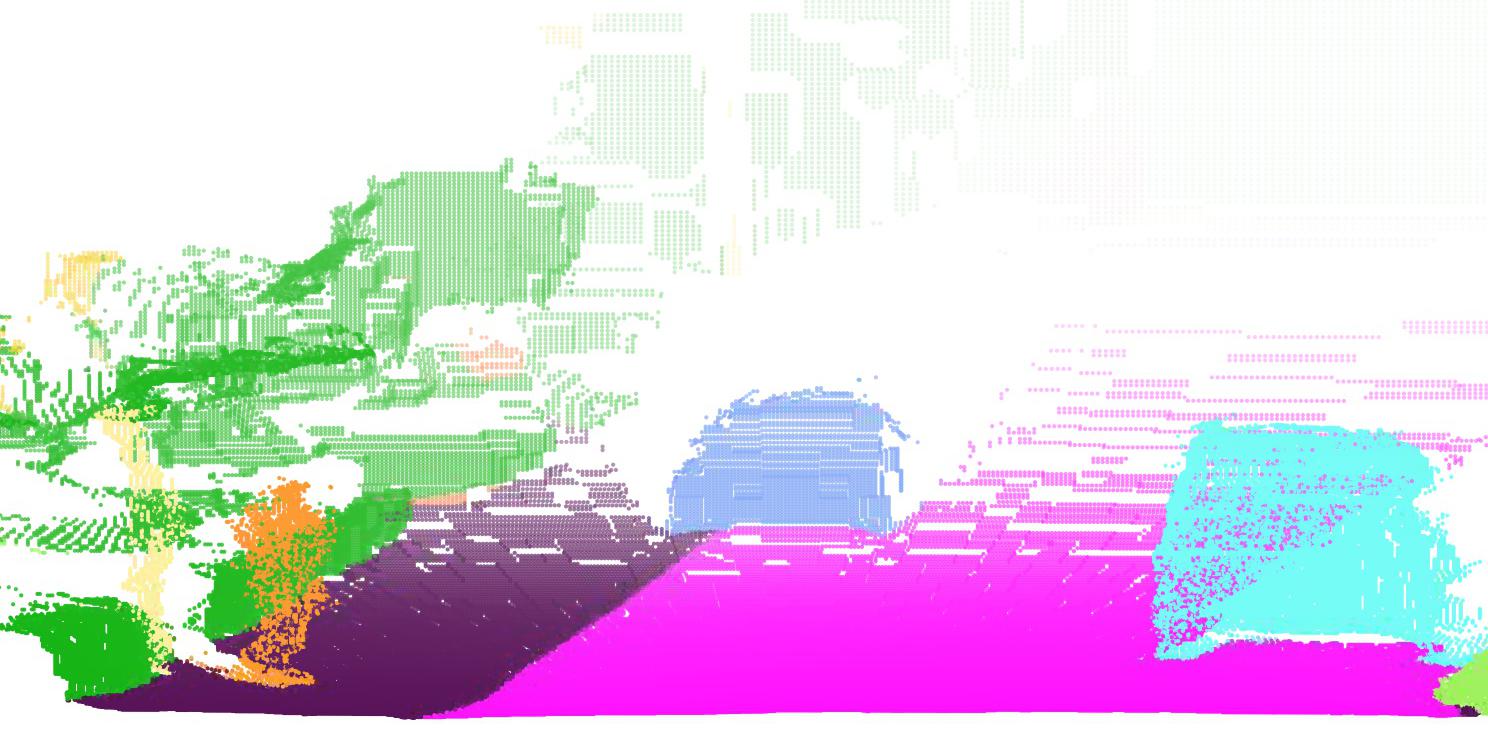}
\end{subfigure}\\
\begin{subfigure}{.245\linewidth}
  \centering
  \includegraphics[trim={150 0 150 100},clip,width=\linewidth]{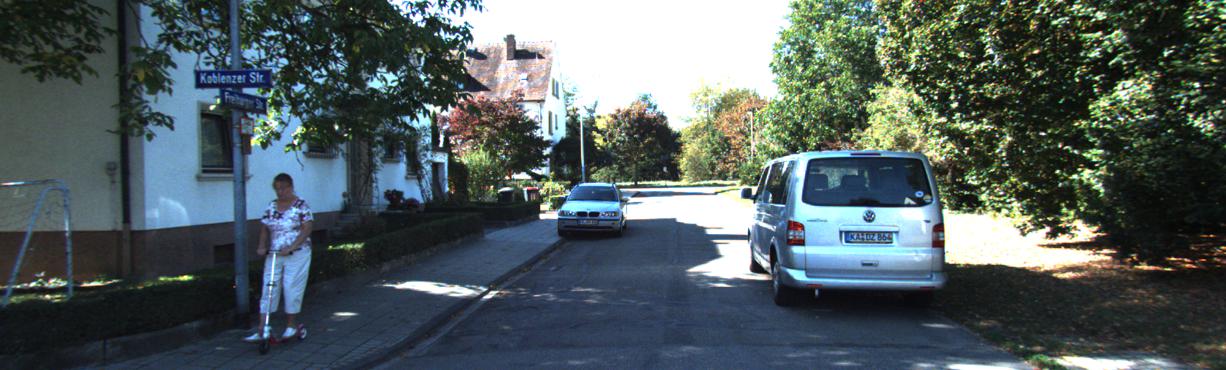}
\end{subfigure}\hfill
\begin{subfigure}{.245\linewidth}
  \centering
  \includegraphics[trim={150 0 150 100},clip,width=\linewidth]{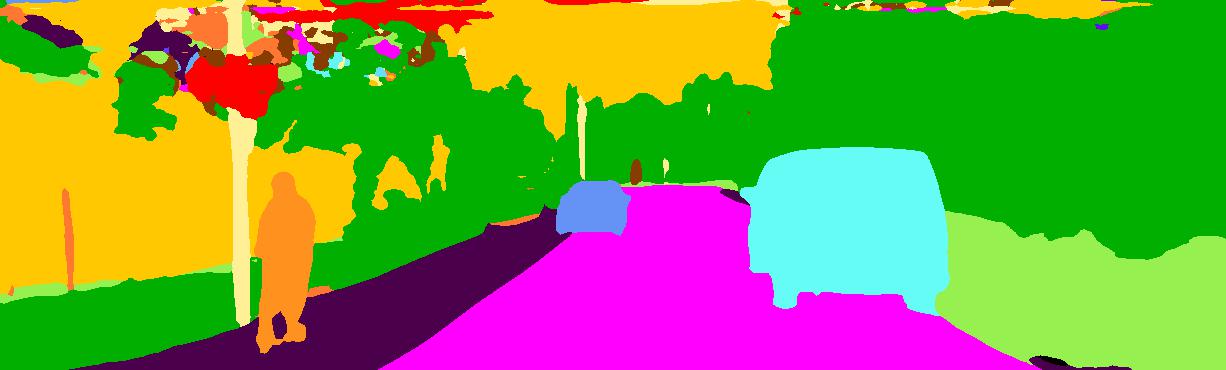}
\end{subfigure}\hfill
\begin{subfigure}{.245\linewidth}
  \centering
  \includegraphics[trim={150 0 150 100},clip,width=\linewidth]{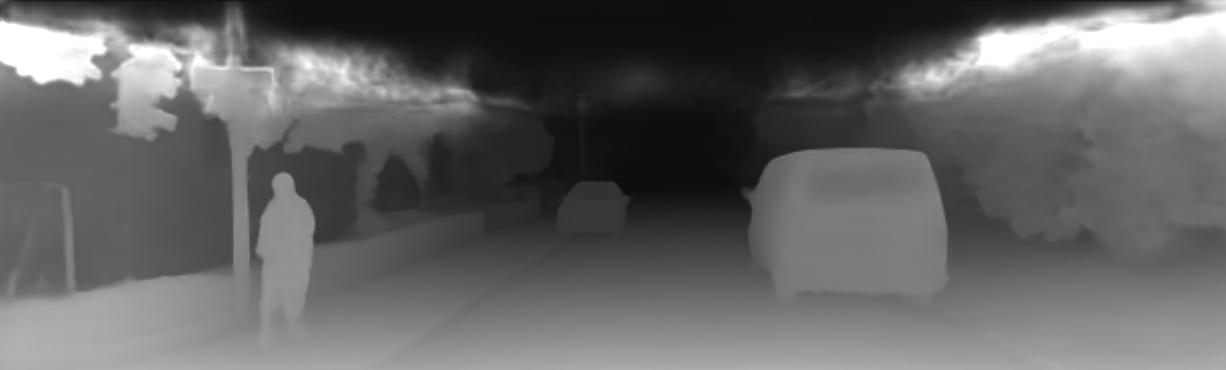}
\end{subfigure}\hfill
\begin{subfigure}{.245\linewidth}
  \centering
  \includegraphics[trim={0 0 0 300},clip,width=\linewidth]{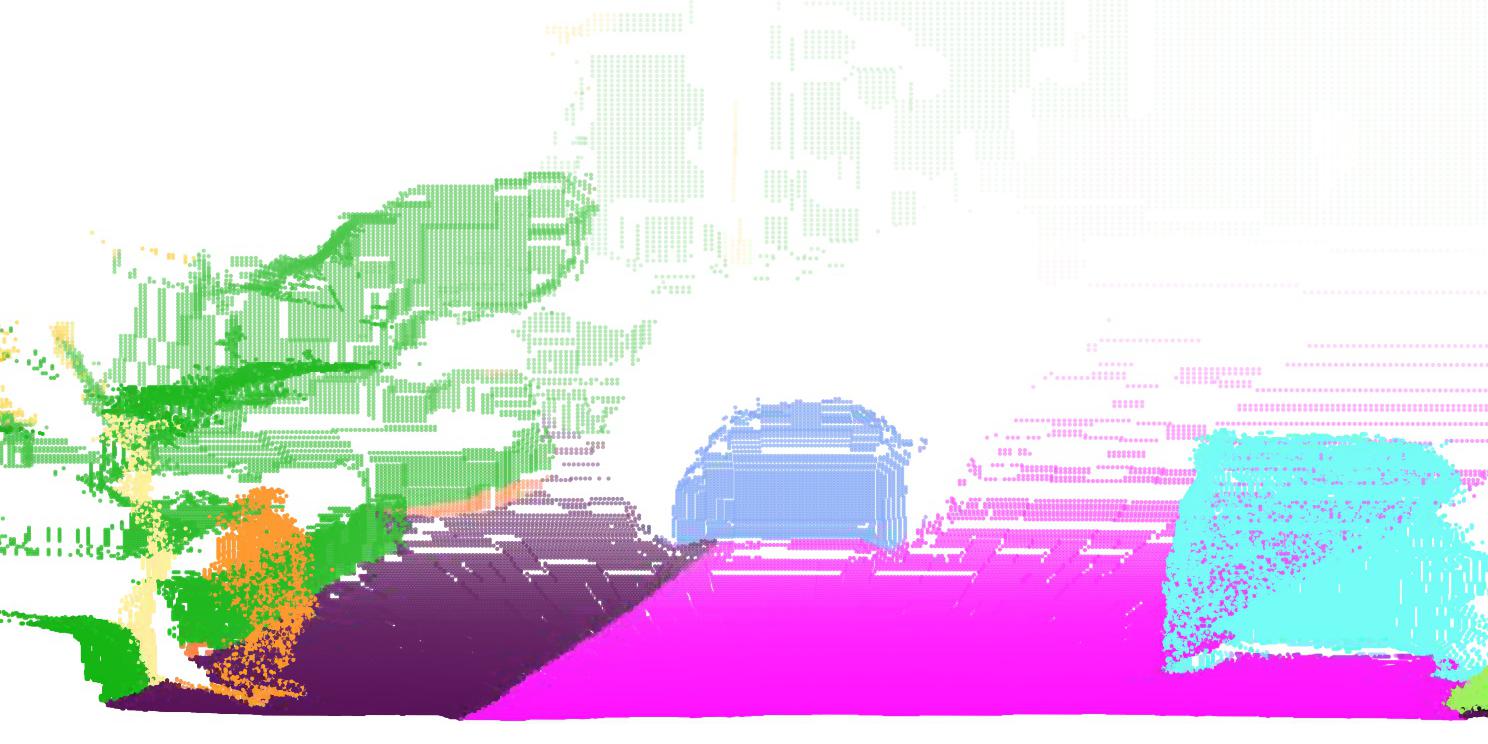}
\end{subfigure}\\
\begin{subfigure}{.245\linewidth}
  \centering
  \includegraphics[trim={150 0 150 100},clip,width=\linewidth]{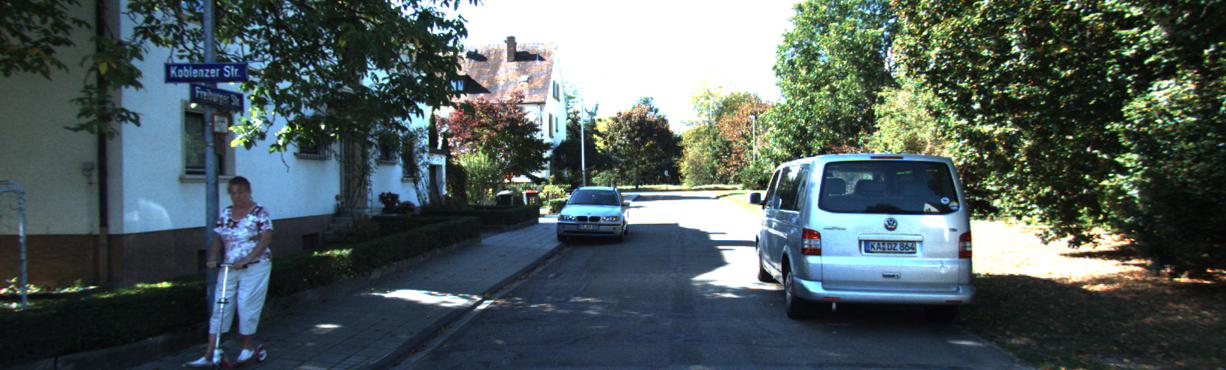}
\end{subfigure}\hfill
\begin{subfigure}{.245\linewidth}
  \centering
  \includegraphics[trim={150 0 150 100},clip,width=\linewidth]{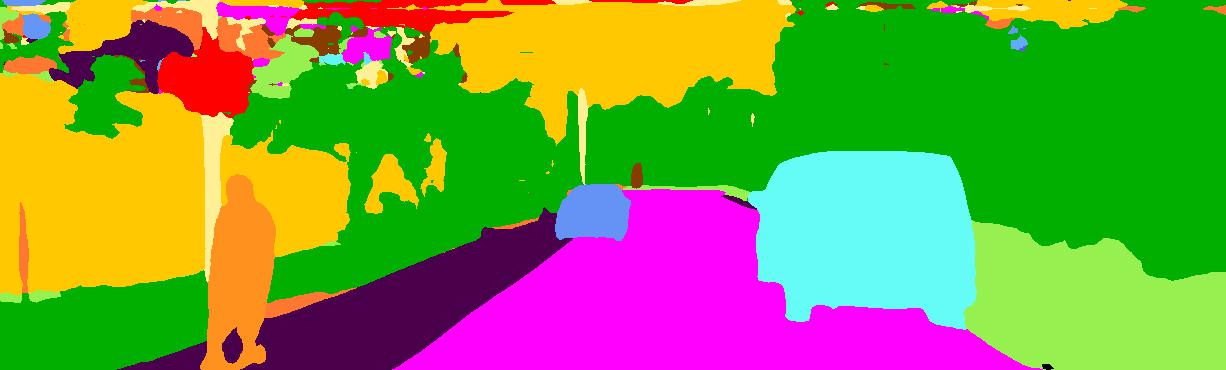}
\end{subfigure}\hfill
\begin{subfigure}{.245\linewidth}
  \centering
  \includegraphics[trim={150 0 150 100},clip,width=\linewidth]{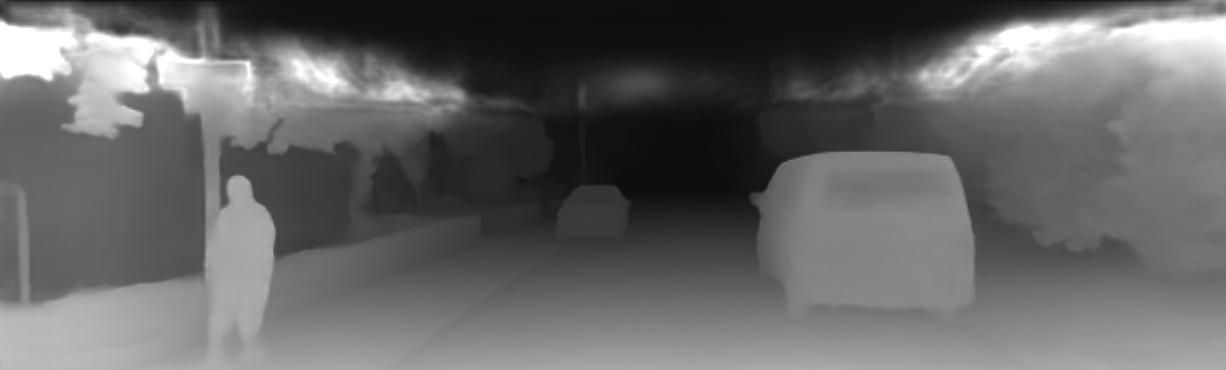}
\end{subfigure}\hfill
\begin{subfigure}{.245\linewidth}
  \centering
  \includegraphics[trim={0 0 0 300},clip,width=\linewidth]{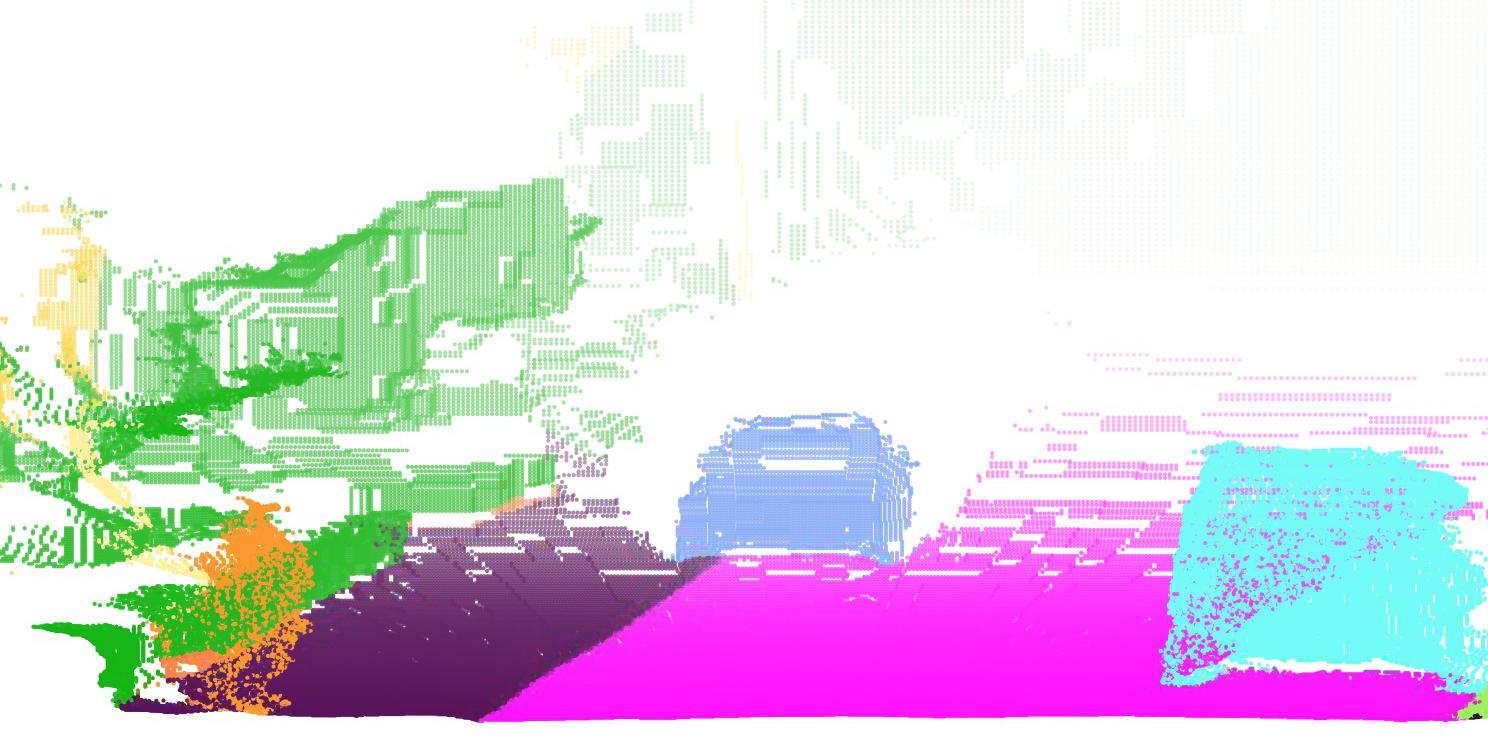}
\end{subfigure}\\
\begin{subfigure}{.245\linewidth}
  \centering
  \includegraphics[trim={150 0 150 100},clip,width=\linewidth]{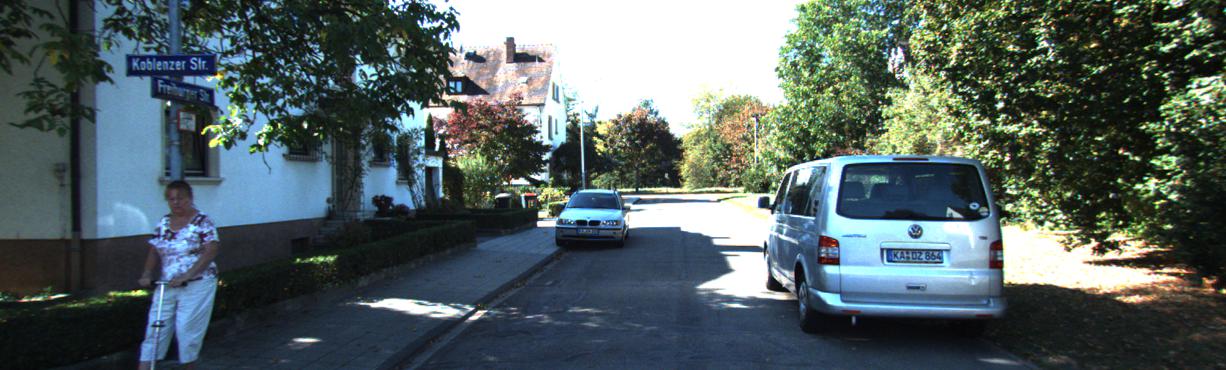}
\end{subfigure}\hfill
\begin{subfigure}{.245\linewidth}
  \centering
  \includegraphics[trim={150 0 150 100},clip,width=\linewidth]{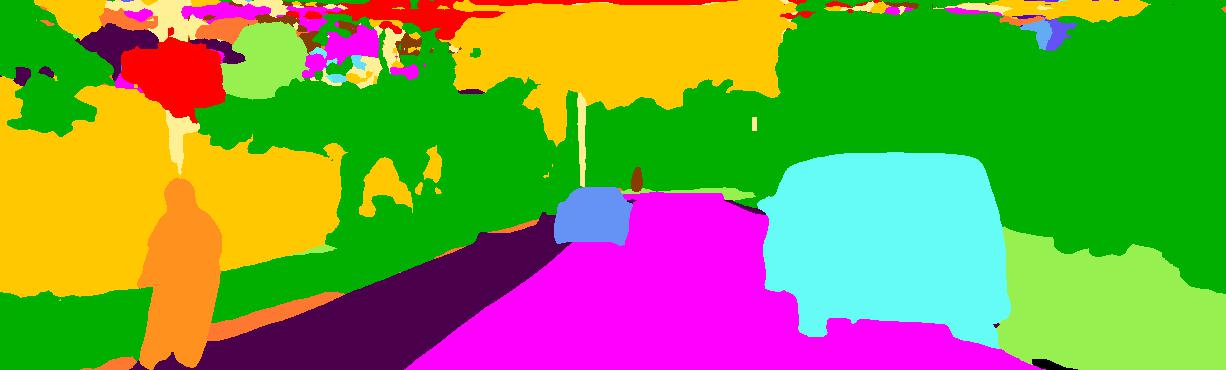}
\end{subfigure}\hfill
\begin{subfigure}{.245\linewidth}
  \centering
  \includegraphics[trim={150 0 150 100},clip,width=\linewidth]{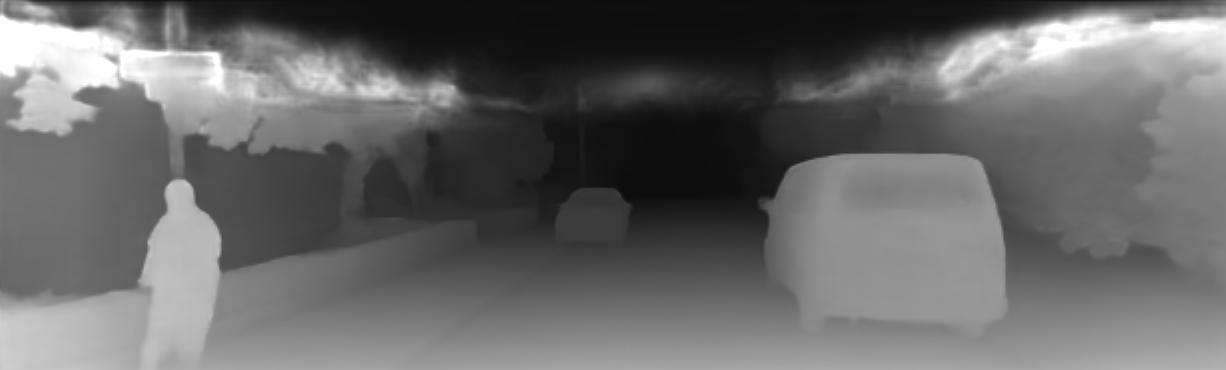}
\end{subfigure}\hfill
\begin{subfigure}{.245\linewidth}
  \centering
  \includegraphics[trim={0 0 0 300},clip,width=\linewidth]{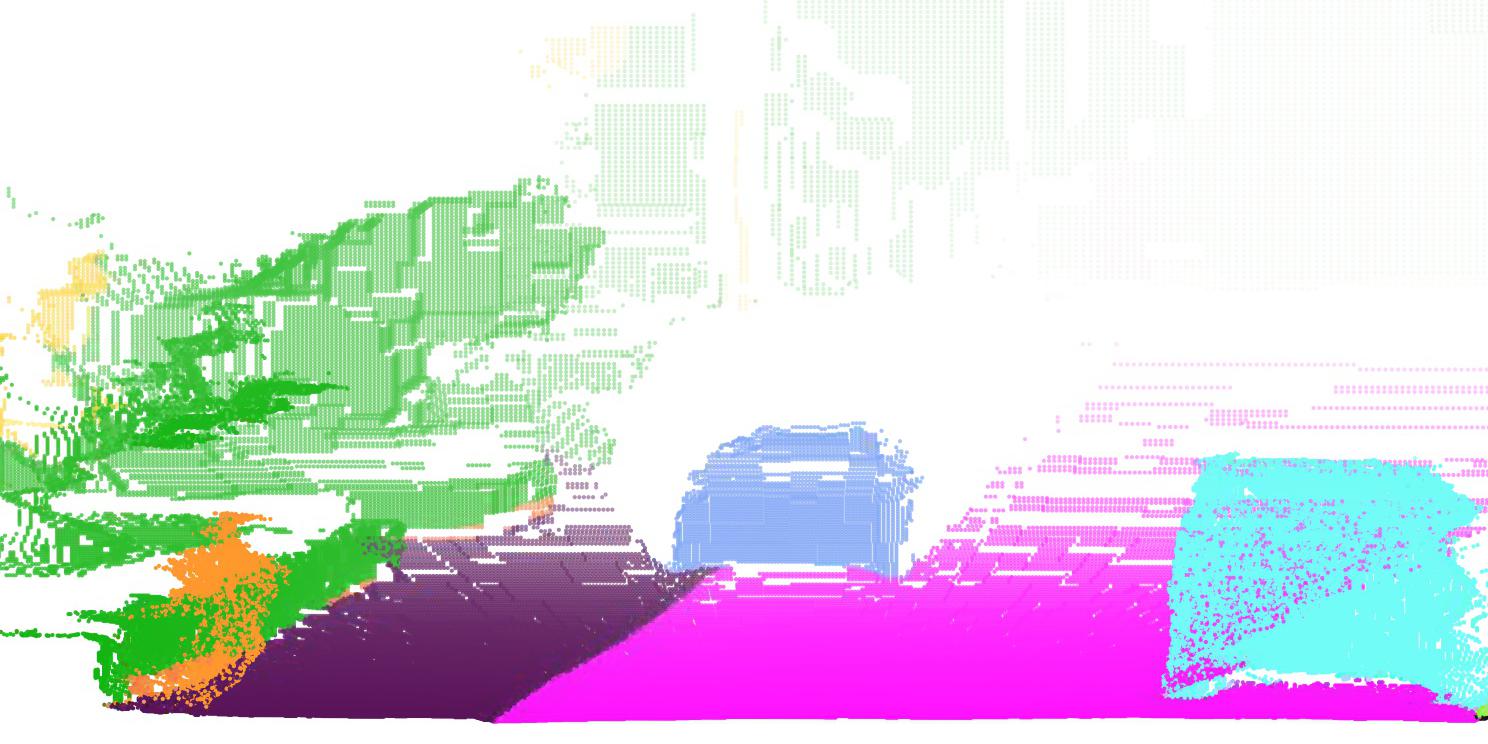}
\end{subfigure}\\
\begin{subfigure}{.245\linewidth}
  \centering
  \includegraphics[trim={150 0 150 100},clip,width=\linewidth]{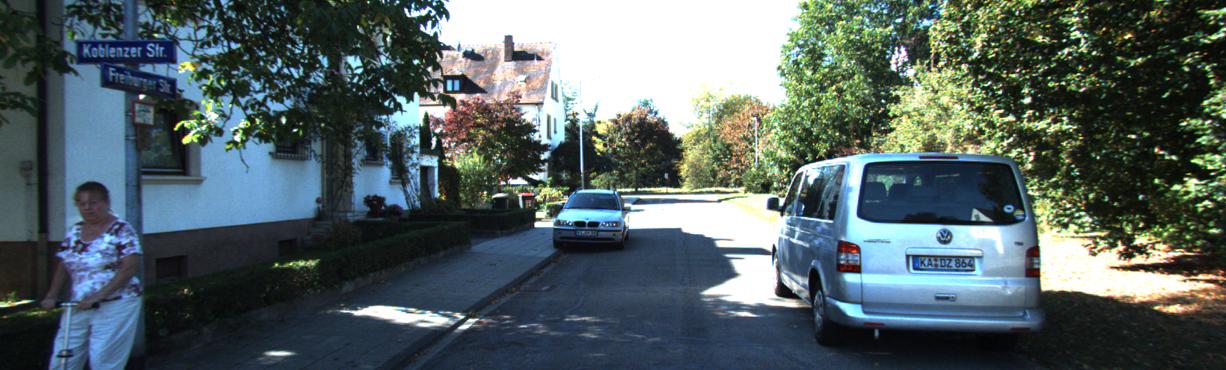}
\end{subfigure}\hfill
\begin{subfigure}{.245\linewidth}
  \centering
  \includegraphics[trim={150 0 150 100},clip,width=\linewidth]{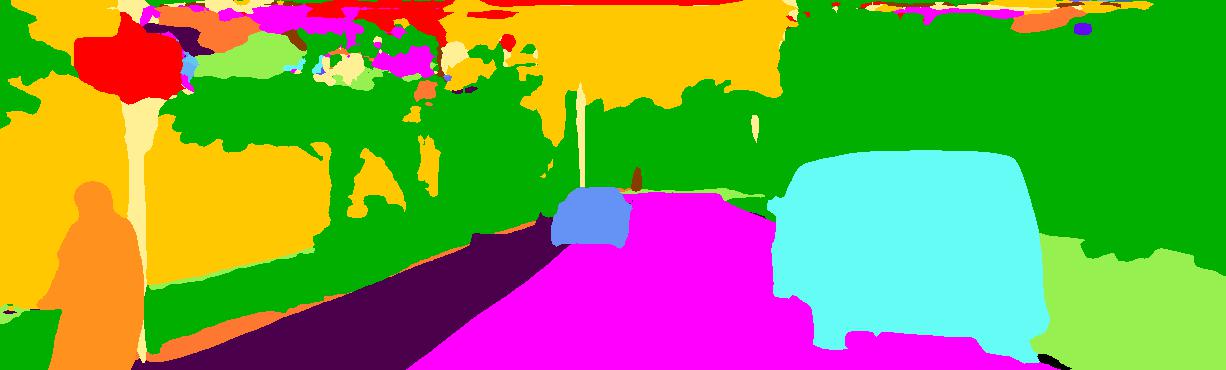}
\end{subfigure}\hfill
\begin{subfigure}{.245\linewidth}
  \centering
  \includegraphics[trim={150 0 150 100},clip,width=\linewidth]{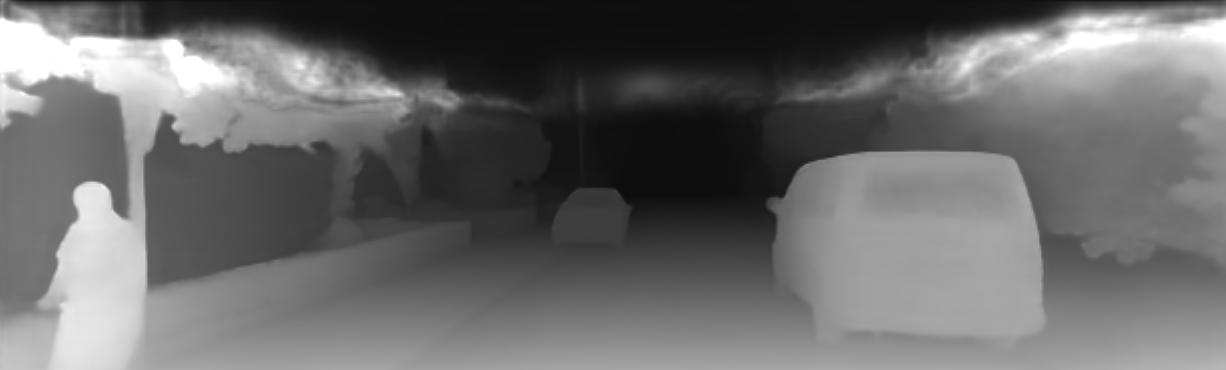}
\end{subfigure}\hfill
\begin{subfigure}{.245\linewidth}
  \centering
  \includegraphics[trim={0 0 0 300},clip,width=\linewidth]{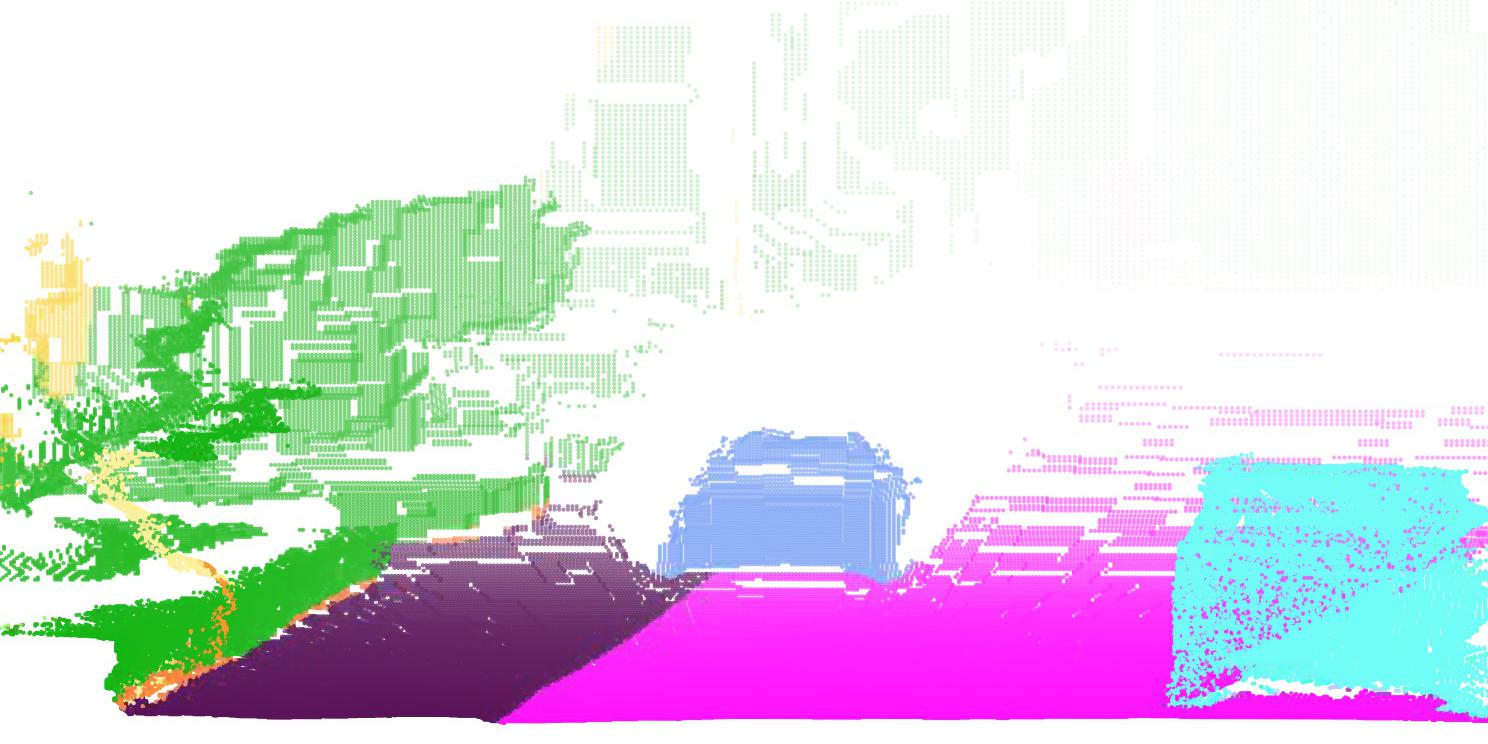}
\end{subfigure}\\
\begin{subfigure}{.245\linewidth}
  \centering
  \includegraphics[trim={150 0 150 100},clip,width=\linewidth]{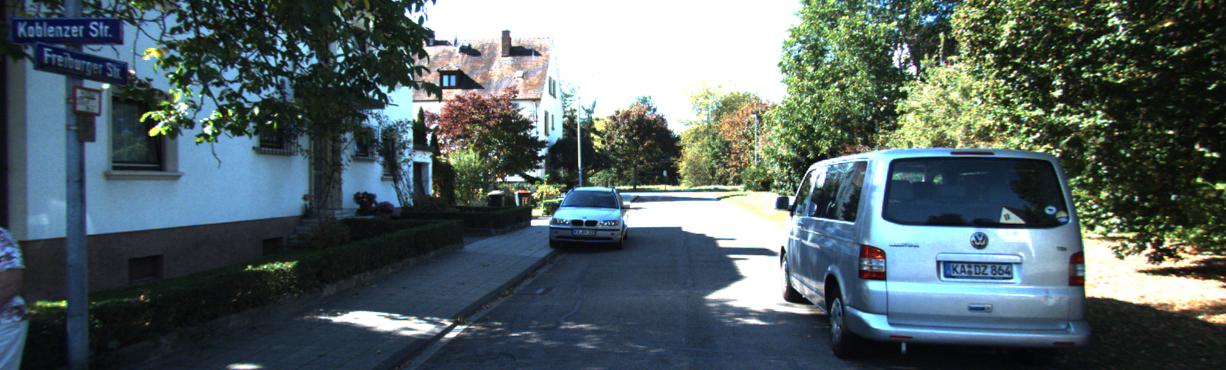}
\end{subfigure}\hfill
\begin{subfigure}{.245\linewidth}
  \centering
  \includegraphics[trim={150 0 150 100},clip,width=\linewidth]{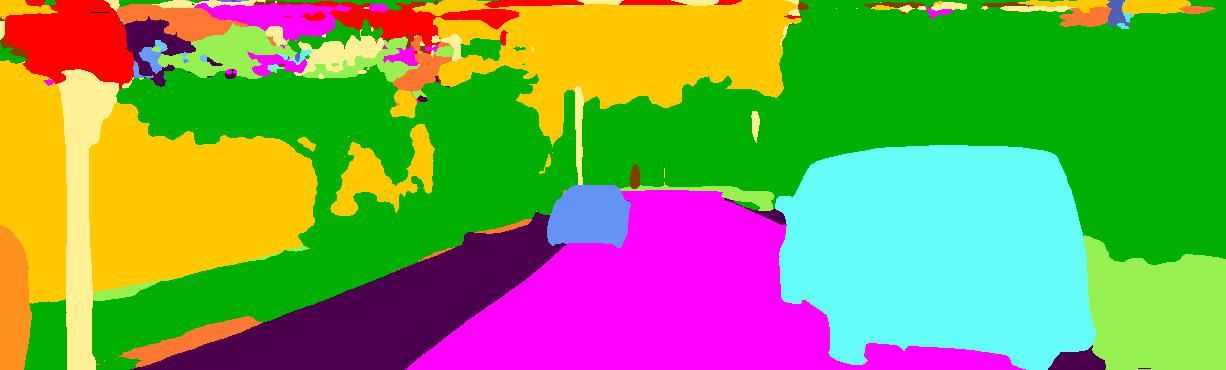}
\end{subfigure}\hfill
\begin{subfigure}{.245\linewidth}
  \centering
  \includegraphics[trim={150 0 150 100},clip,width=\linewidth]{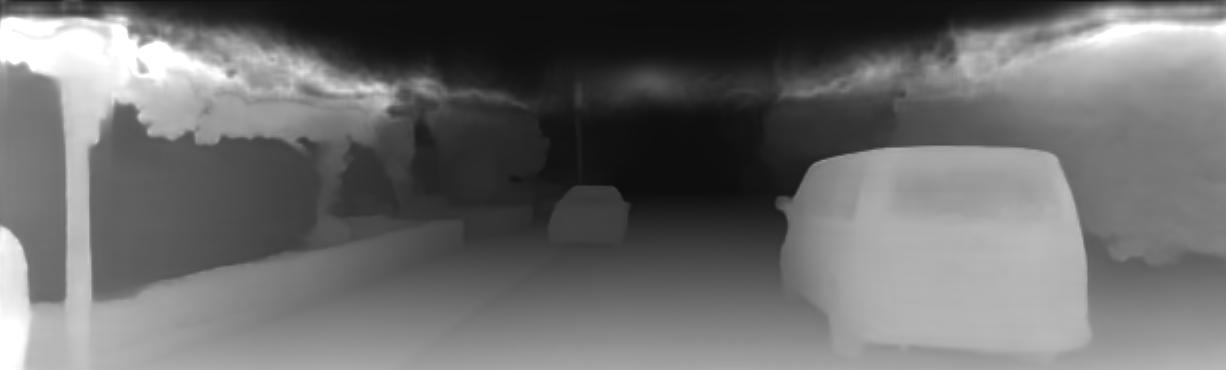}
\end{subfigure}\hfill
\begin{subfigure}{.245\linewidth}
  \centering
  \includegraphics[trim={0 0 0 300},clip,width=\linewidth]{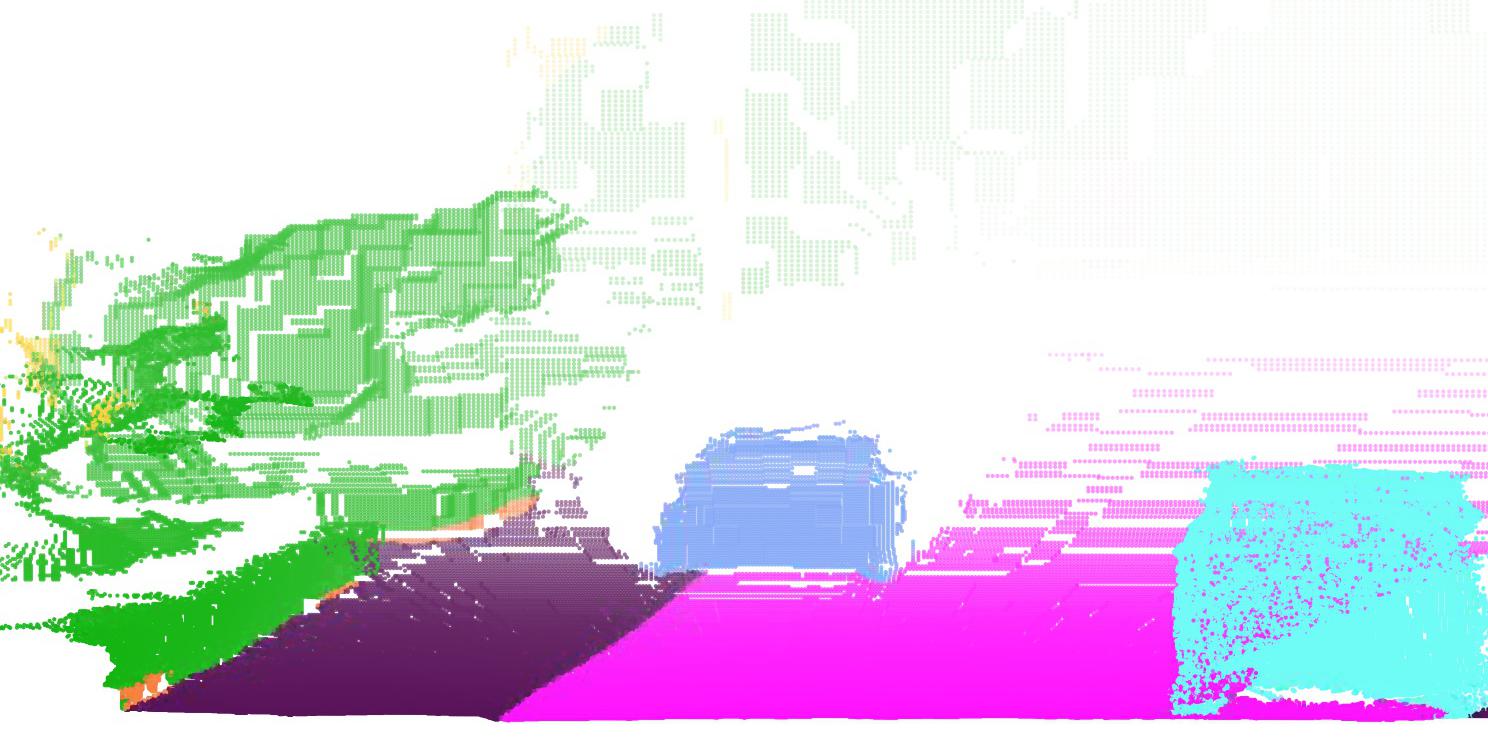}
\end{subfigure}\\
\begin{subfigure}{.245\linewidth}
  \centering
  \includegraphics[trim={150 0 150 100},clip,width=\linewidth]{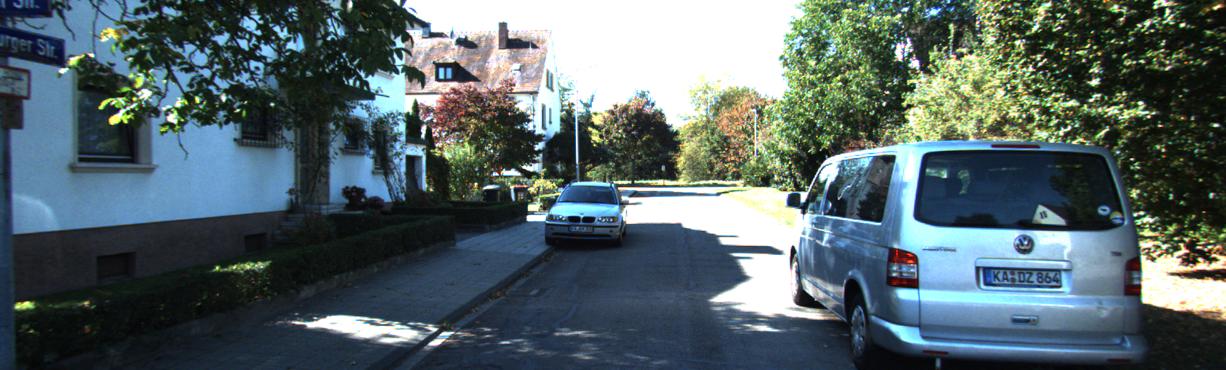}
\end{subfigure}\hfill
\begin{subfigure}{.245\linewidth}
  \centering
  \includegraphics[trim={150 0 150 100},clip,width=\linewidth]{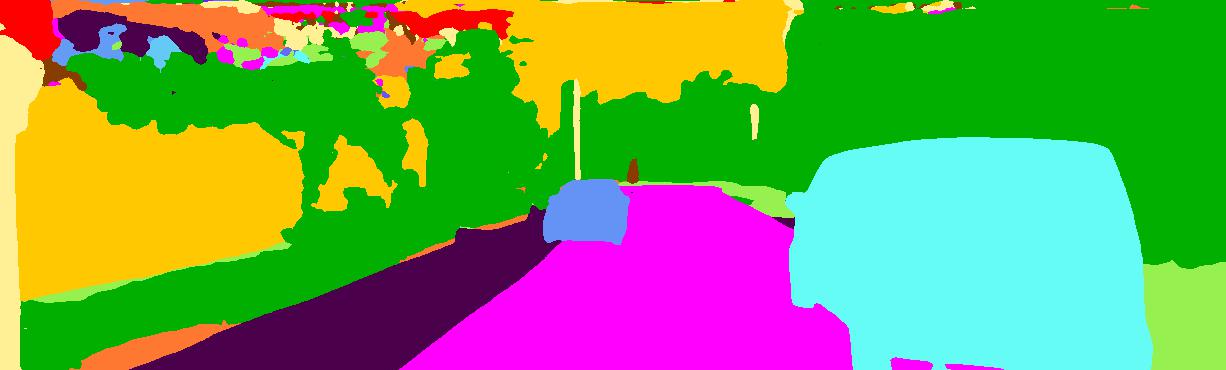}
\end{subfigure}\hfill
\begin{subfigure}{.245\linewidth}
  \centering
  \includegraphics[trim={150 0 150 100},clip,width=\linewidth]{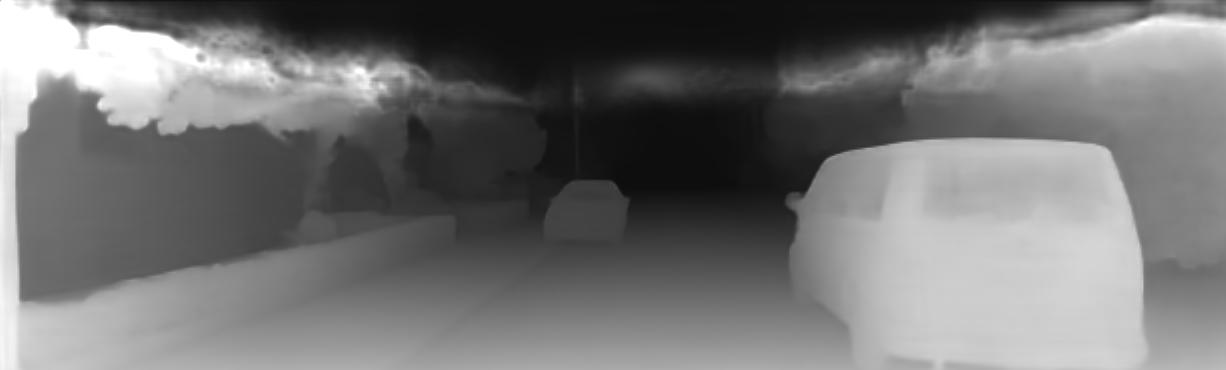}
\end{subfigure}\hfill
\begin{subfigure}{.245\linewidth}
  \centering
  \includegraphics[trim={0 0 0 300},clip,width=\linewidth]{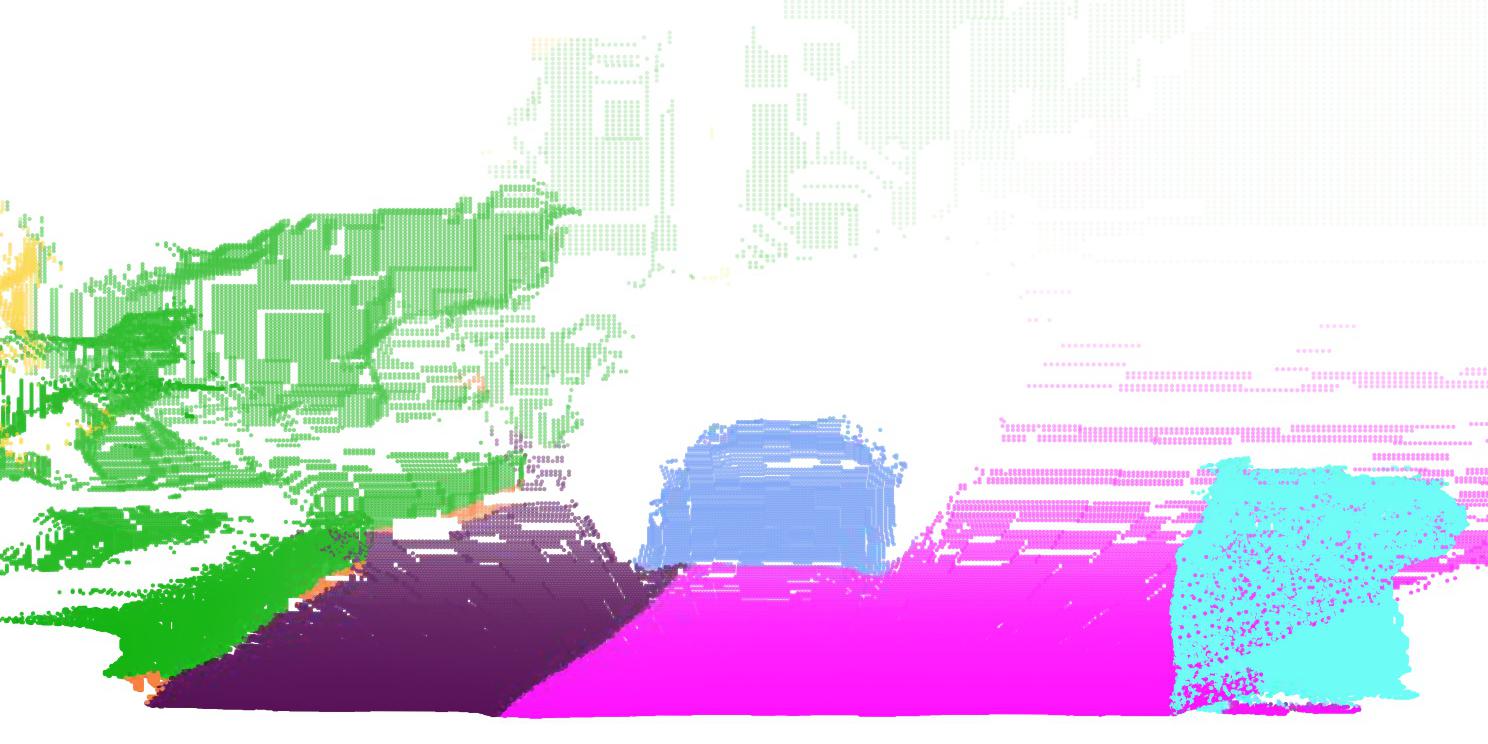}
\end{subfigure}\\
\caption{Prediction visualizations on SemKITTI-DVPS. From left to right: input image, temporally consistent panoptic segmentation prediction, monocular depth prediction, and point cloud visualization.}
\label{fig:sk_1}
\end{figure*}

\begin{figure*}
\centering
\begin{subfigure}{.245\linewidth}
  \centering
  \includegraphics[trim={150 0 150 100},clip,width=\linewidth]{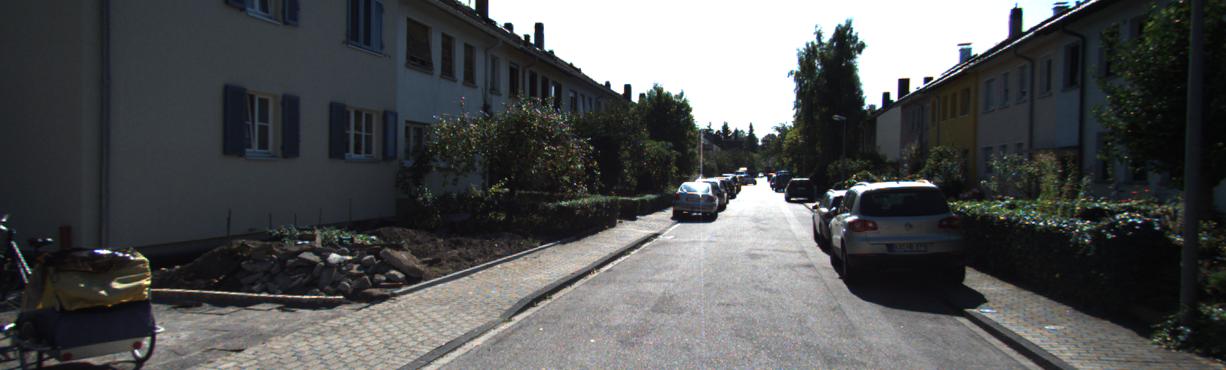}
\end{subfigure}\hfill
\begin{subfigure}{.245\linewidth}
  \centering
  \includegraphics[trim={150 0 150 100},clip,width=\linewidth]{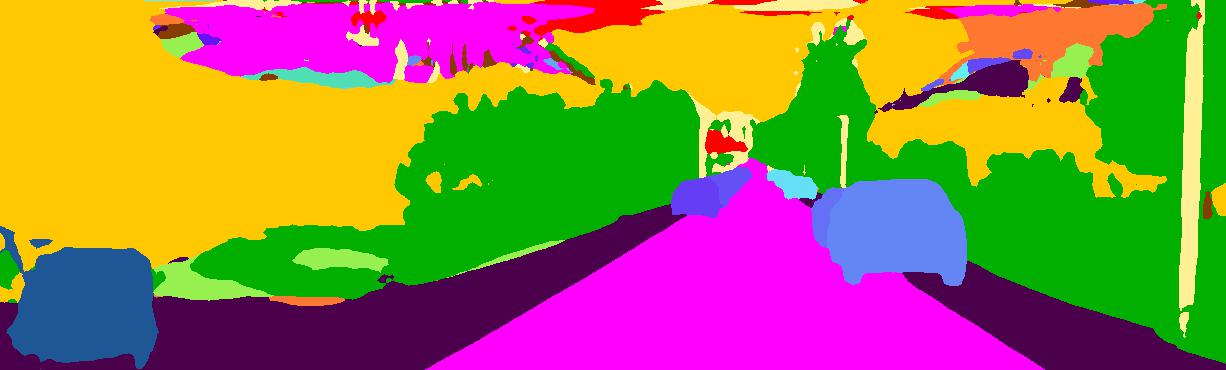}
\end{subfigure}\hfill
\begin{subfigure}{.245\linewidth}
  \centering
  \includegraphics[trim={150 0 150 100},clip,width=\linewidth]{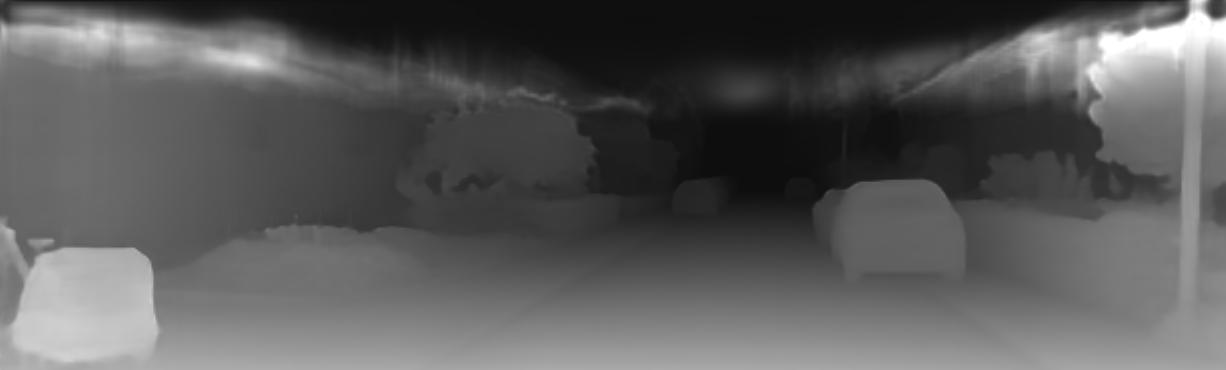}
\end{subfigure}\hfill
\begin{subfigure}{.245\linewidth}
  \centering
  \includegraphics[trim={0 0 0 300},clip,width=\linewidth]{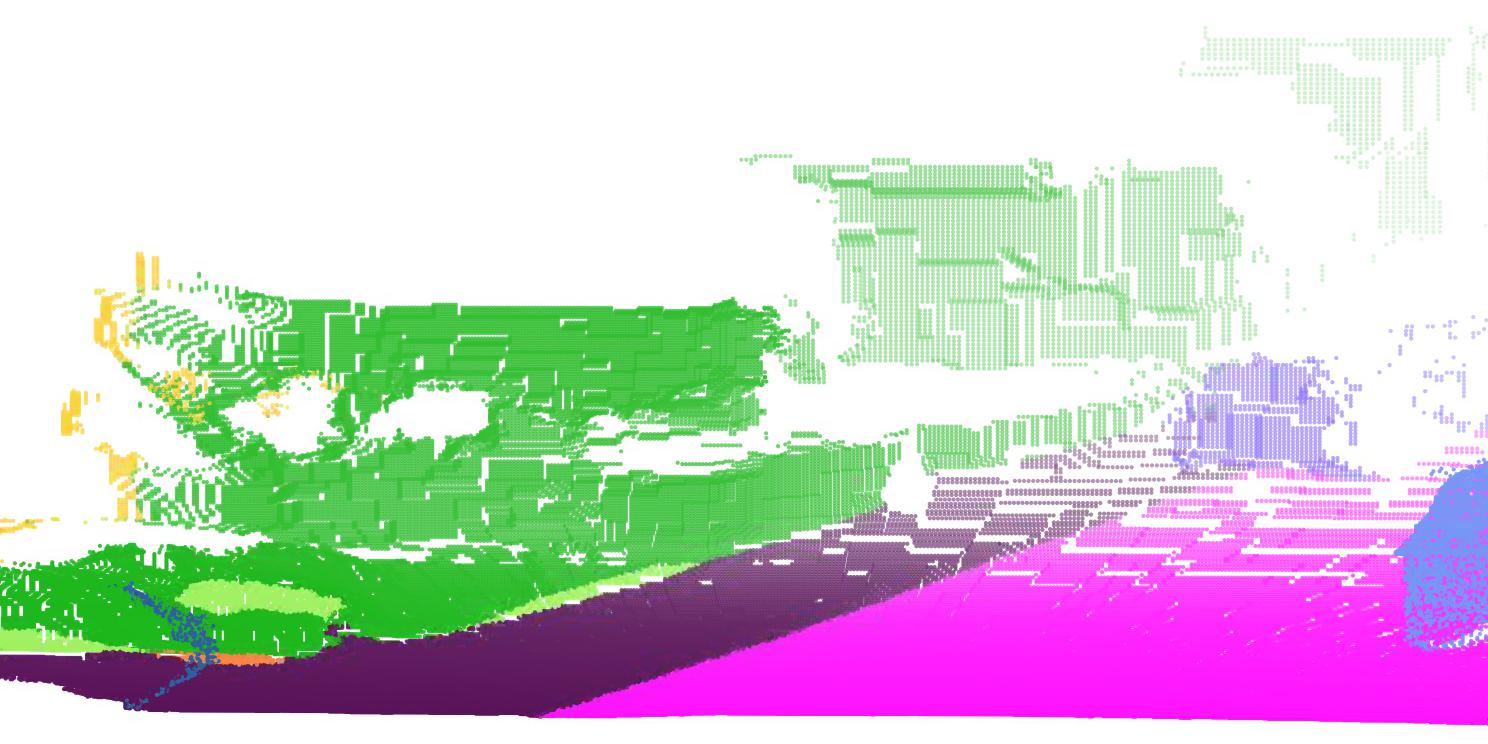}
\end{subfigure}\\
\begin{subfigure}{.245\linewidth}
  \centering
  \includegraphics[trim={150 0 150 100},clip,width=\linewidth]{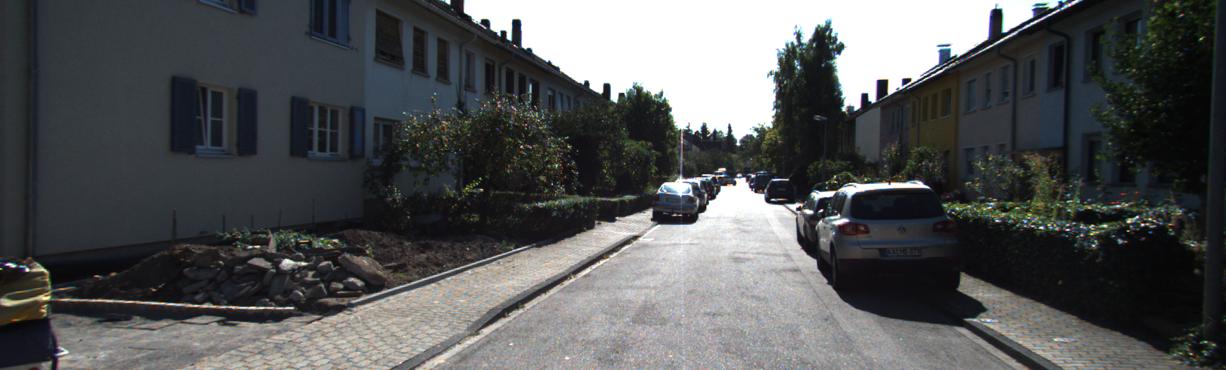}
\end{subfigure}\hfill
\begin{subfigure}{.245\linewidth}
  \centering
  \includegraphics[trim={150 0 150 100},clip,width=\linewidth]{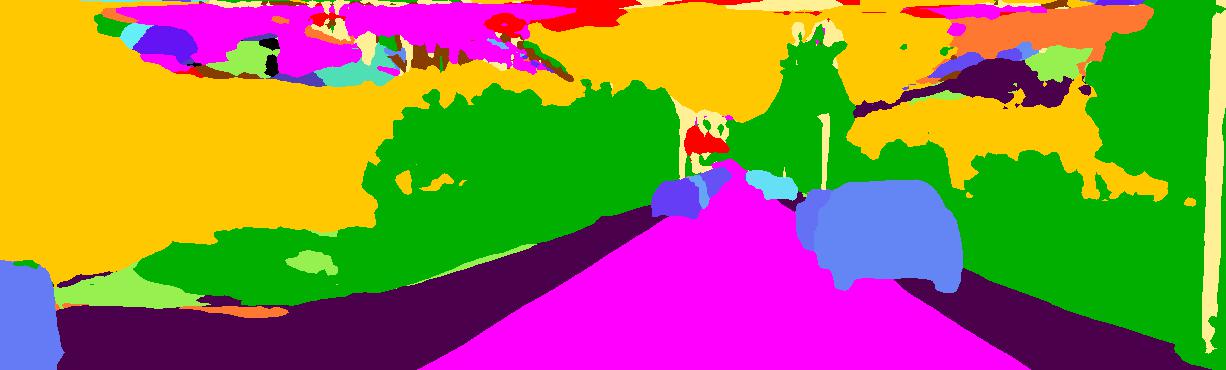}
\end{subfigure}\hfill
\begin{subfigure}{.245\linewidth}
  \centering
  \includegraphics[trim={150 0 150 100},clip,width=\linewidth]{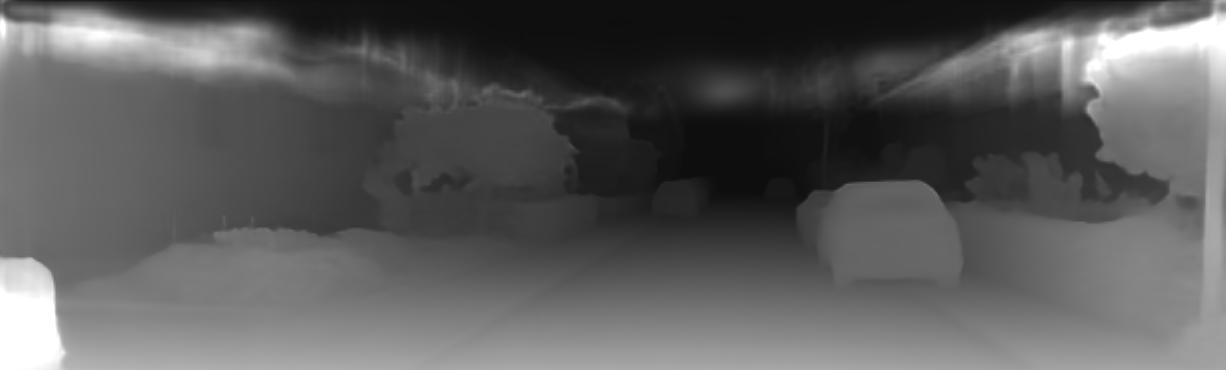}
\end{subfigure}\hfill
\begin{subfigure}{.245\linewidth}
  \centering
  \includegraphics[trim={0 0 0 300},clip,width=\linewidth]{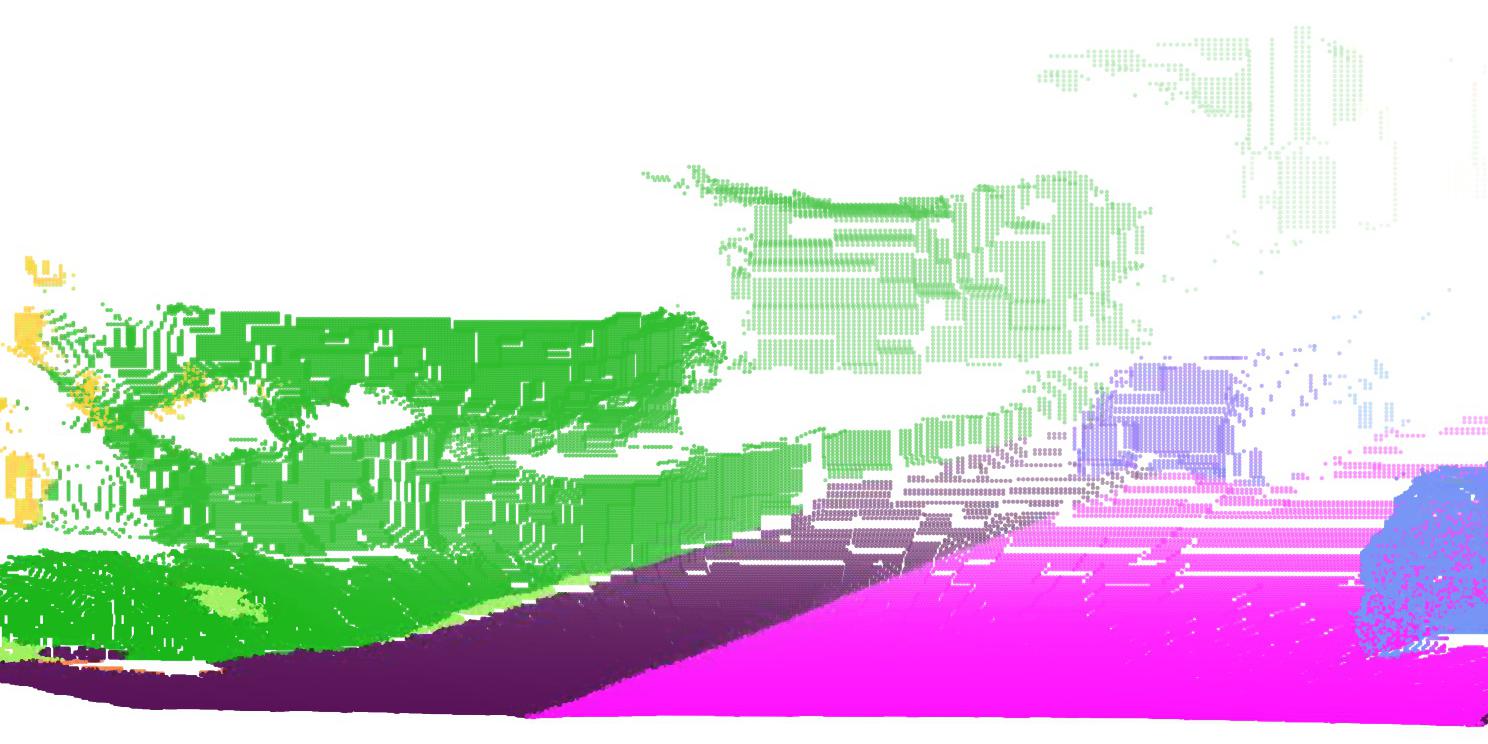}
\end{subfigure}\\
\begin{subfigure}{.245\linewidth}
  \centering
  \includegraphics[trim={150 0 150 100},clip,width=\linewidth]{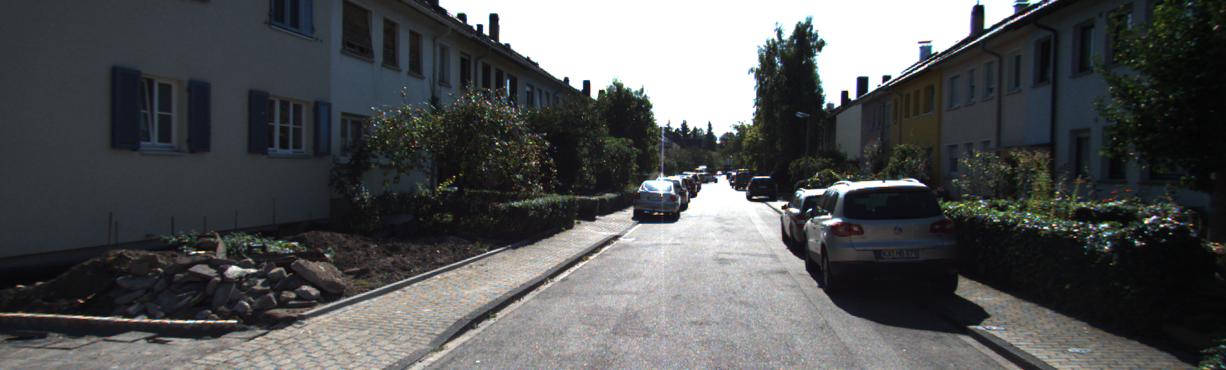}
\end{subfigure}\hfill
\begin{subfigure}{.245\linewidth}
  \centering
  \includegraphics[trim={150 0 150 100},clip,width=\linewidth]{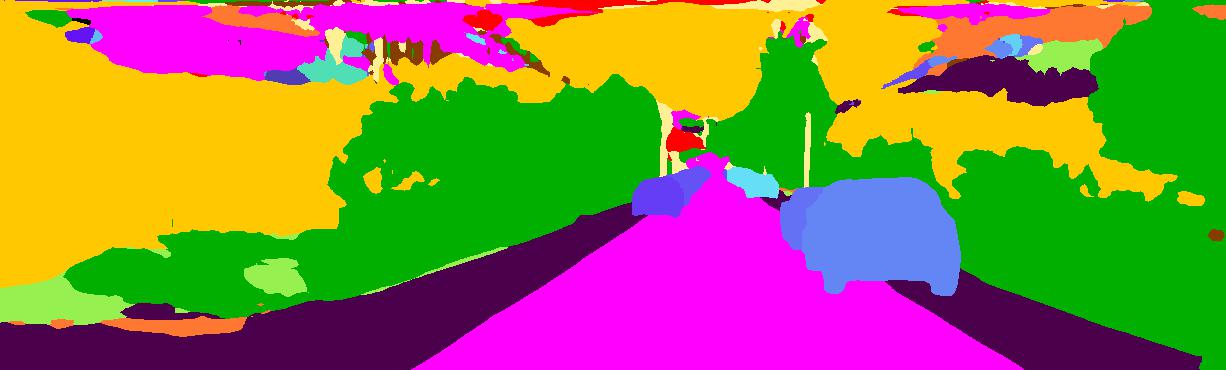}
\end{subfigure}\hfill
\begin{subfigure}{.245\linewidth}
  \centering
  \includegraphics[trim={150 0 150 100},clip,width=\linewidth]{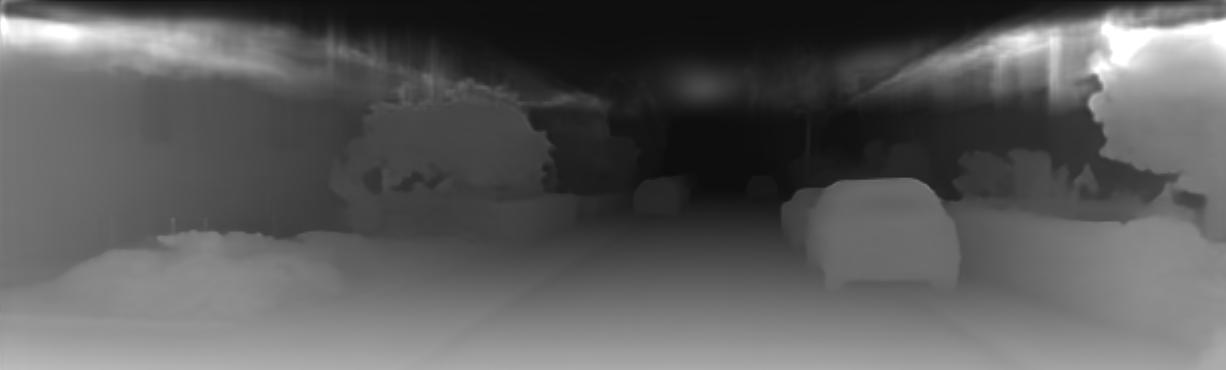}
\end{subfigure}\hfill
\begin{subfigure}{.245\linewidth}
  \centering
  \includegraphics[trim={0 0 0 300},clip,width=\linewidth]{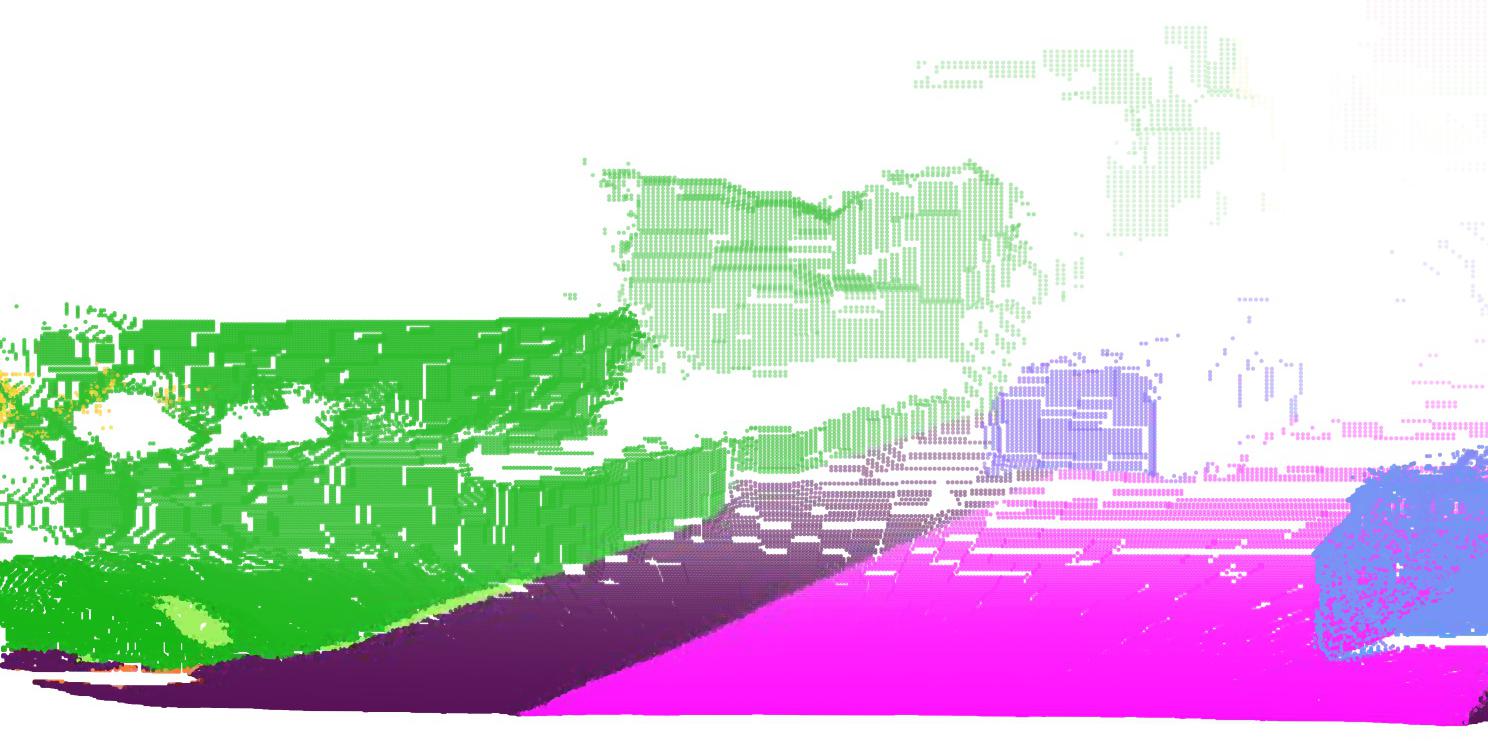}
\end{subfigure}\\
\begin{subfigure}{.245\linewidth}
  \centering
  \includegraphics[trim={150 0 150 100},clip,width=\linewidth]{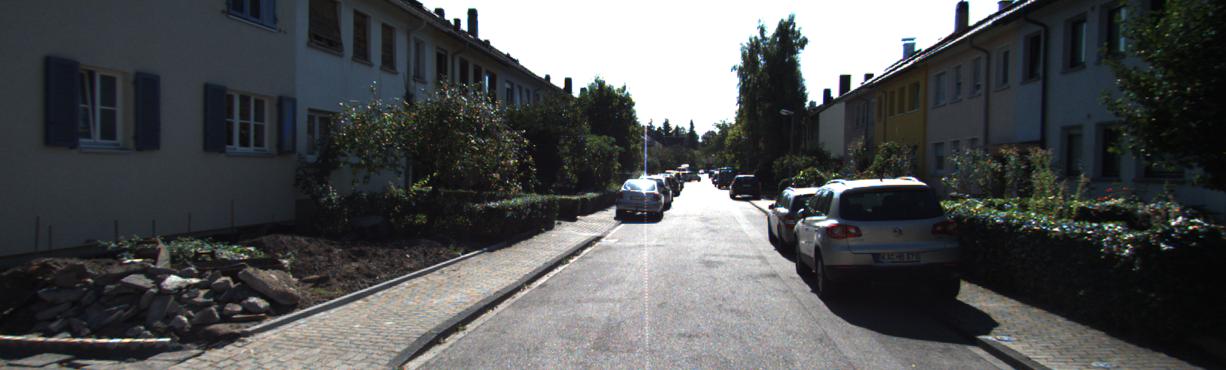}
\end{subfigure}\hfill
\begin{subfigure}{.245\linewidth}
  \centering
  \includegraphics[trim={150 0 150 100},clip,width=\linewidth]{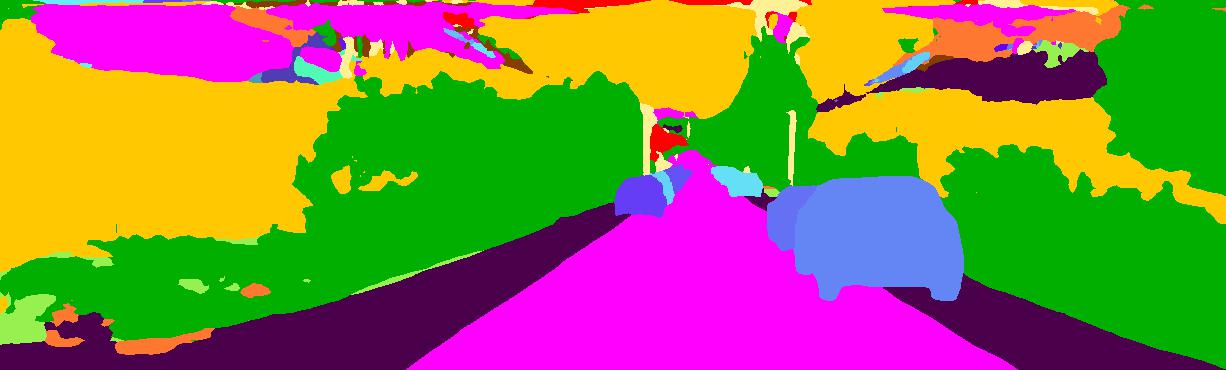}
\end{subfigure}\hfill
\begin{subfigure}{.245\linewidth}
  \centering
  \includegraphics[trim={150 0 150 100},clip,width=\linewidth]{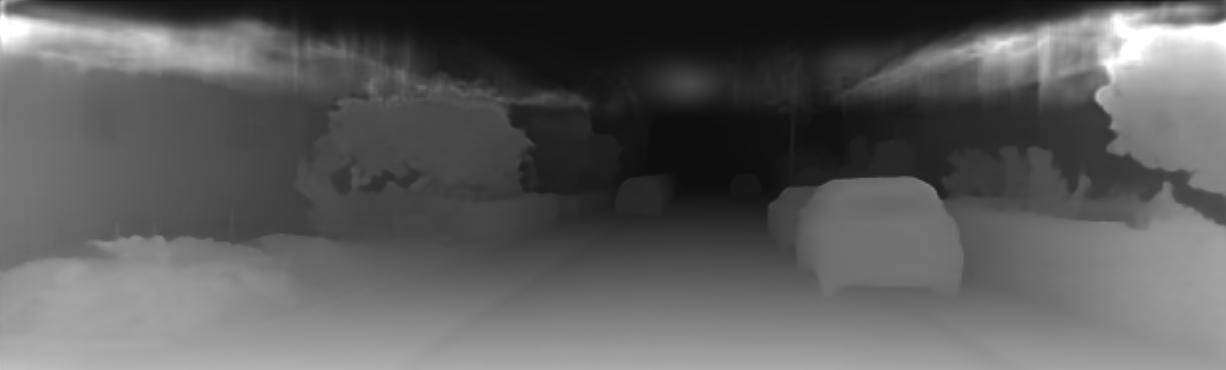}
\end{subfigure}\hfill
\begin{subfigure}{.245\linewidth}
  \centering
  \includegraphics[trim={0 0 0 300},clip,width=\linewidth]{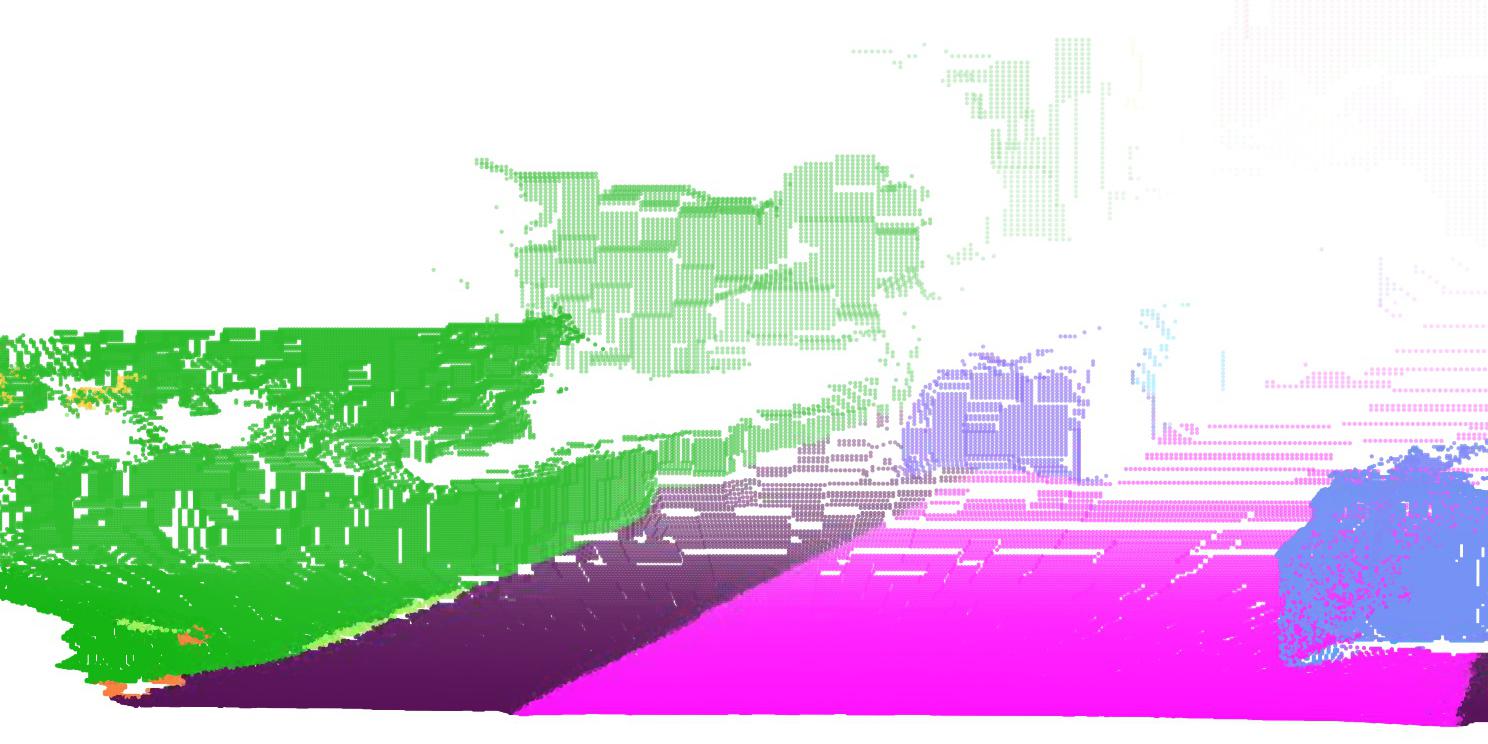}
\end{subfigure}\\
\begin{subfigure}{.245\linewidth}
  \centering
  \includegraphics[trim={150 0 150 100},clip,width=\linewidth]{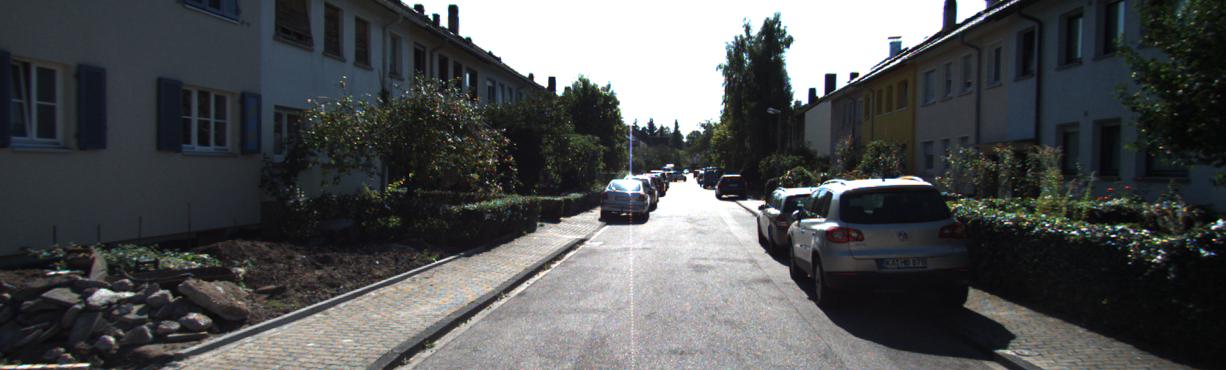}
\end{subfigure}\hfill
\begin{subfigure}{.245\linewidth}
  \centering
  \includegraphics[trim={150 0 150 100},clip,width=\linewidth]{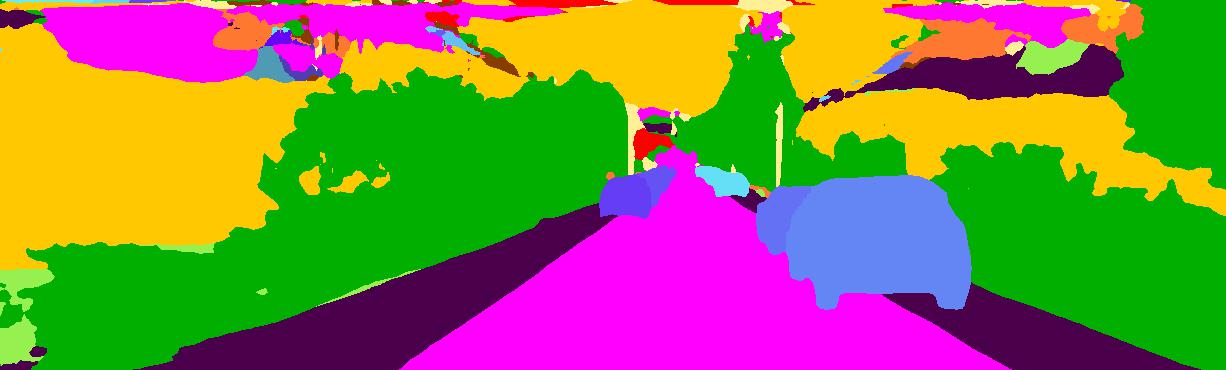}
\end{subfigure}\hfill
\begin{subfigure}{.245\linewidth}
  \centering
  \includegraphics[trim={150 0 150 100},clip,width=\linewidth]{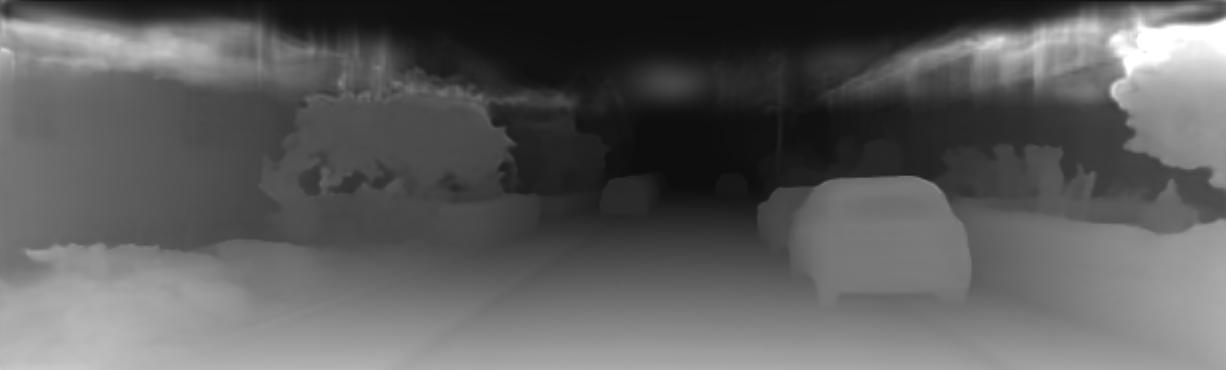}
\end{subfigure}\hfill
\begin{subfigure}{.245\linewidth}
  \centering
  \includegraphics[trim={0 0 0 300},clip,width=\linewidth]{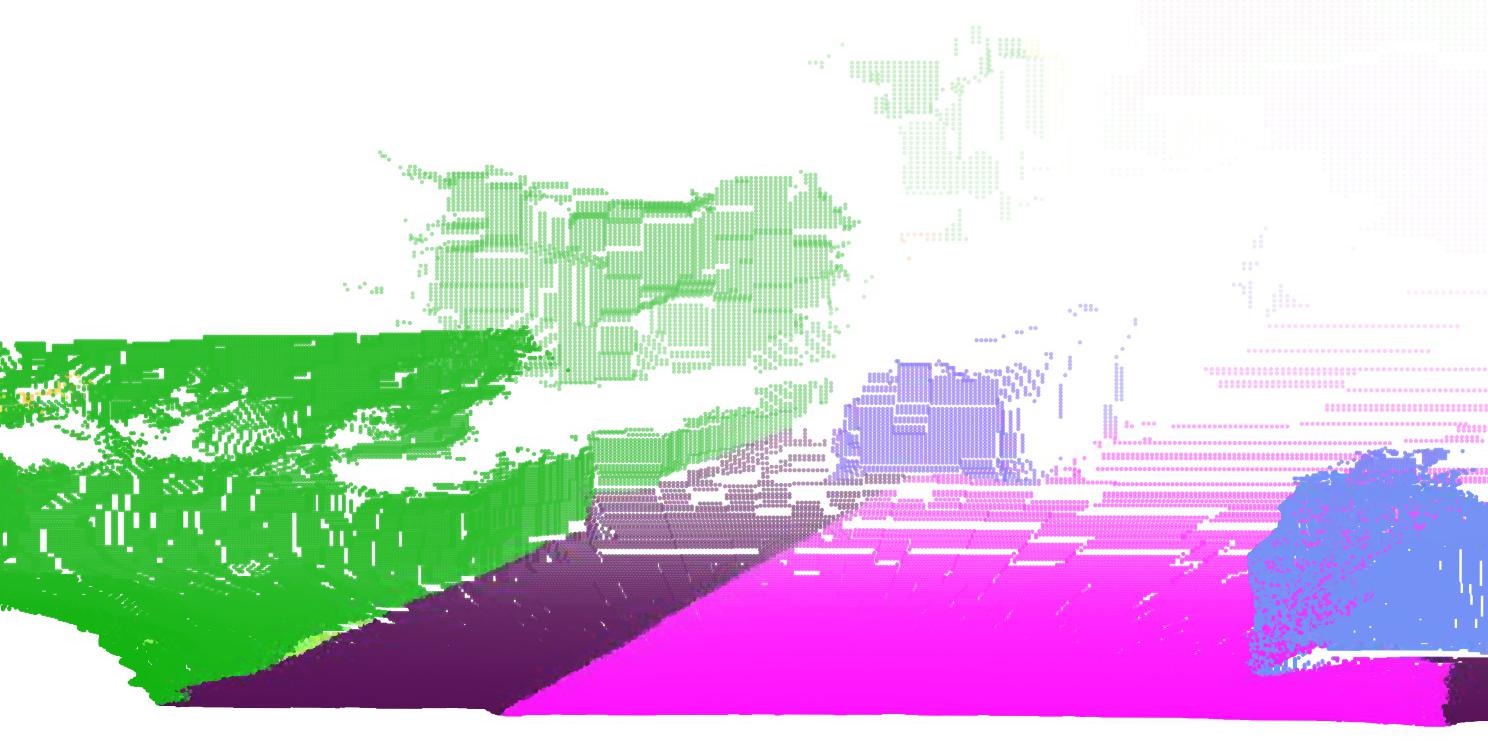}
\end{subfigure}\\
\begin{subfigure}{.245\linewidth}
  \centering
  \includegraphics[trim={150 0 150 100},clip,width=\linewidth]{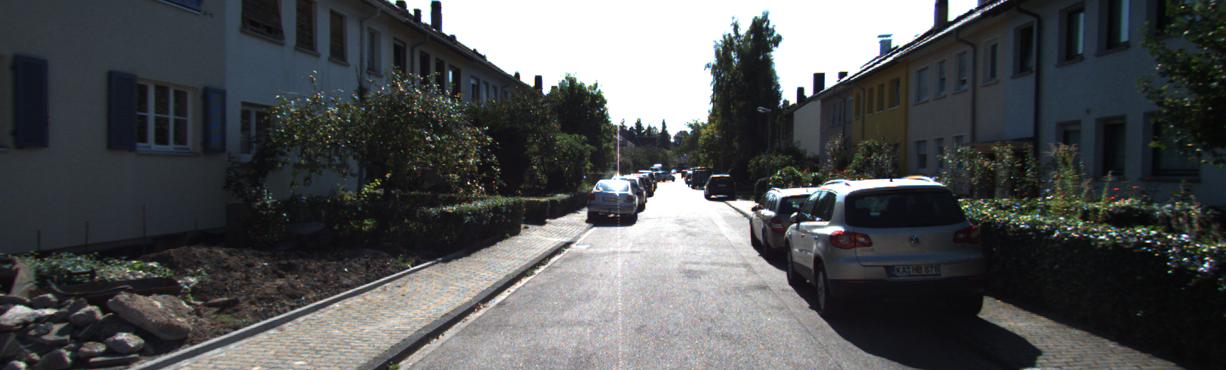}
\end{subfigure}\hfill
\begin{subfigure}{.245\linewidth}
  \centering
  \includegraphics[trim={150 0 150 100},clip,width=\linewidth]{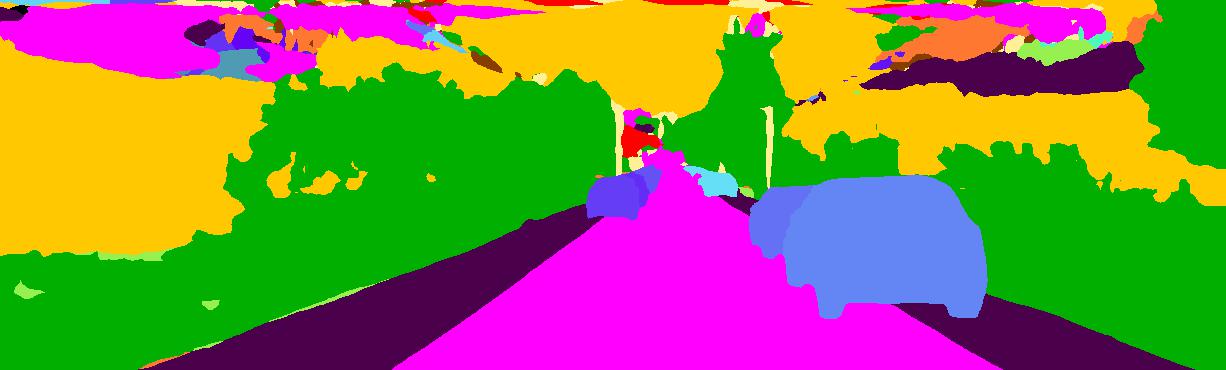}
\end{subfigure}\hfill
\begin{subfigure}{.245\linewidth}
  \centering
  \includegraphics[trim={150 0 150 100},clip,width=\linewidth]{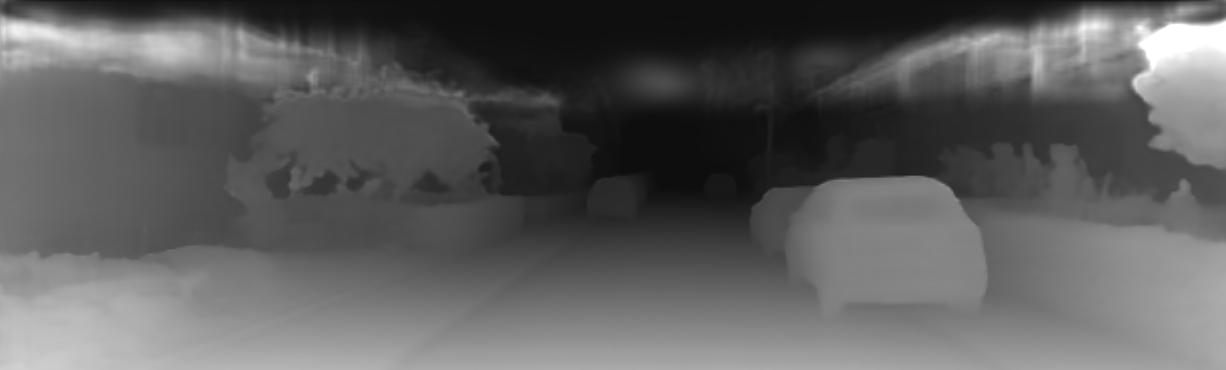}
\end{subfigure}\hfill
\begin{subfigure}{.245\linewidth}
  \centering
  \includegraphics[trim={0 0 0 300},clip,width=\linewidth]{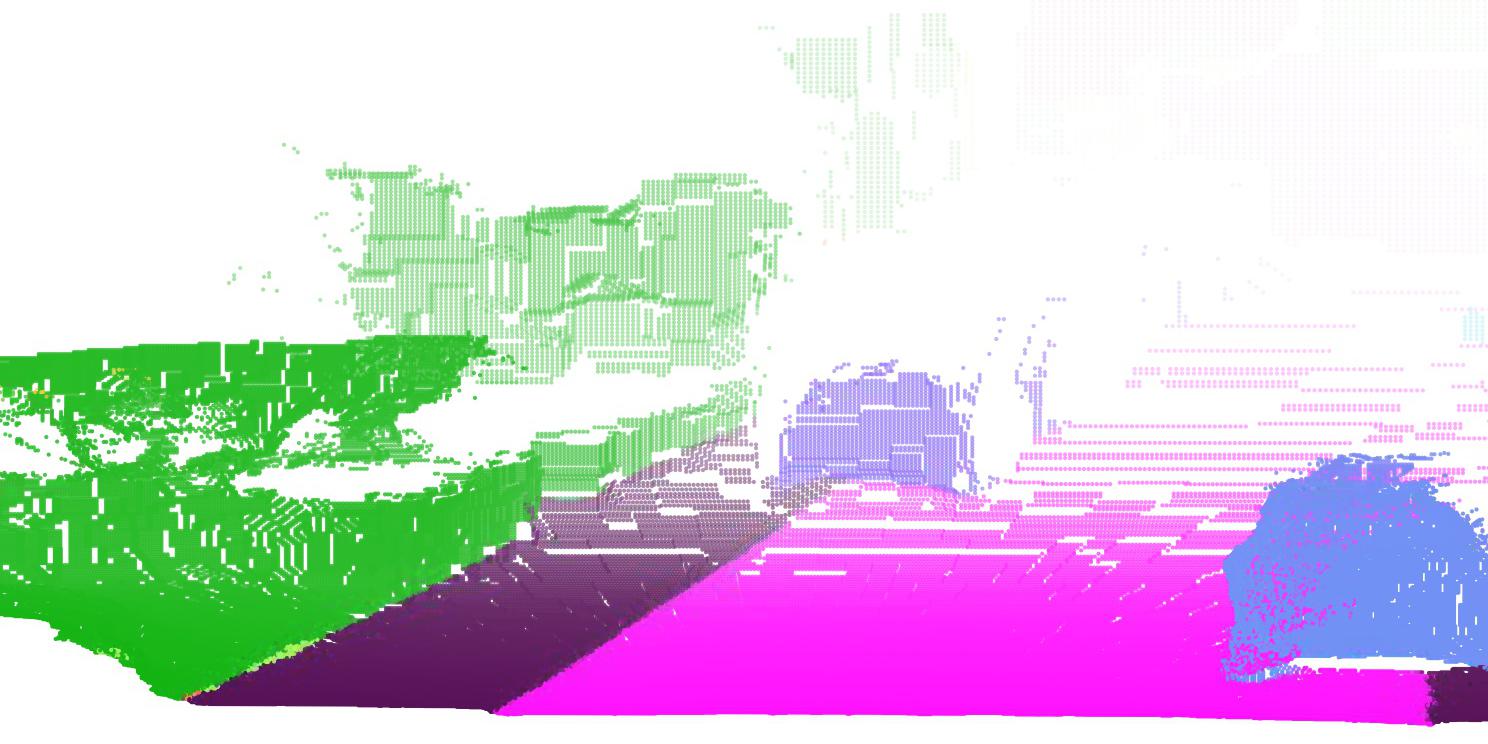}
\end{subfigure}\\
\begin{subfigure}{.245\linewidth}
  \centering
  \includegraphics[trim={150 0 150 100},clip,width=\linewidth]{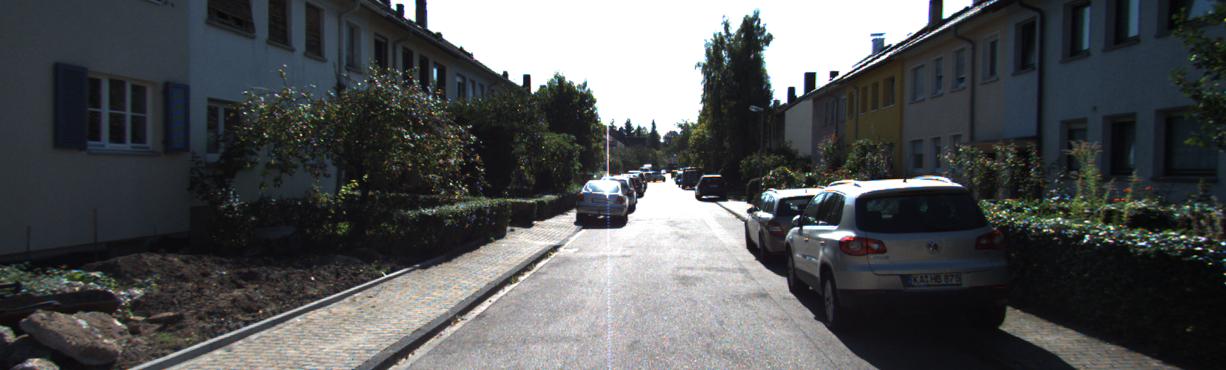}
\end{subfigure}\hfill
\begin{subfigure}{.245\linewidth}
  \centering
  \includegraphics[trim={150 0 150 100},clip,width=\linewidth]{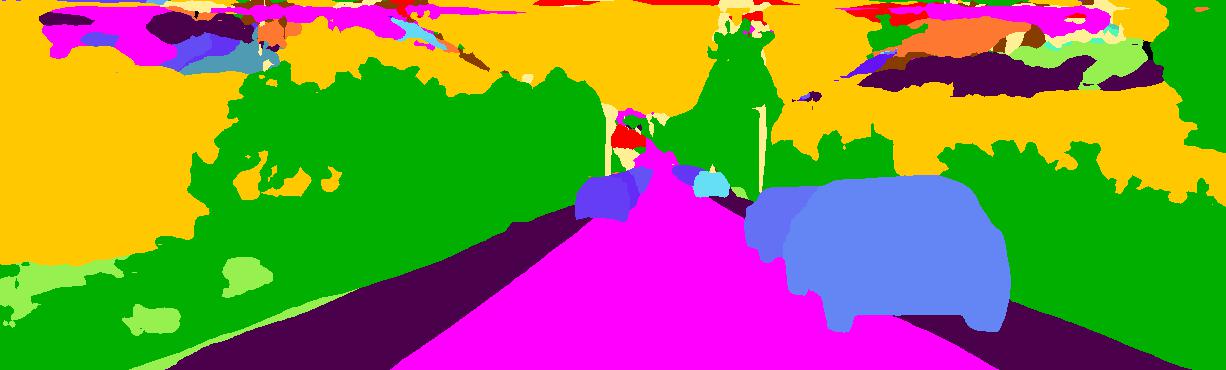}
\end{subfigure}\hfill
\begin{subfigure}{.245\linewidth}
  \centering
  \includegraphics[trim={150 0 150 100},clip,width=\linewidth]{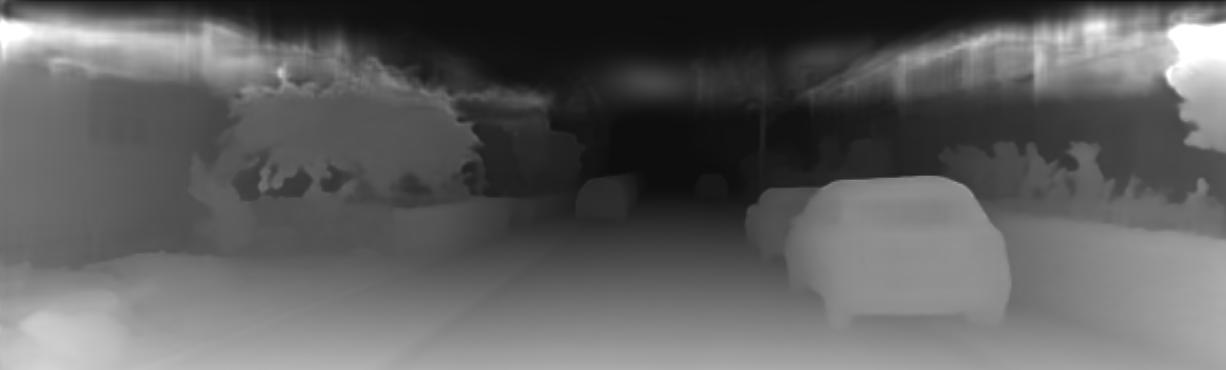}
\end{subfigure}\hfill
\begin{subfigure}{.245\linewidth}
  \centering
  \includegraphics[trim={0 0 0 300},clip,width=\linewidth]{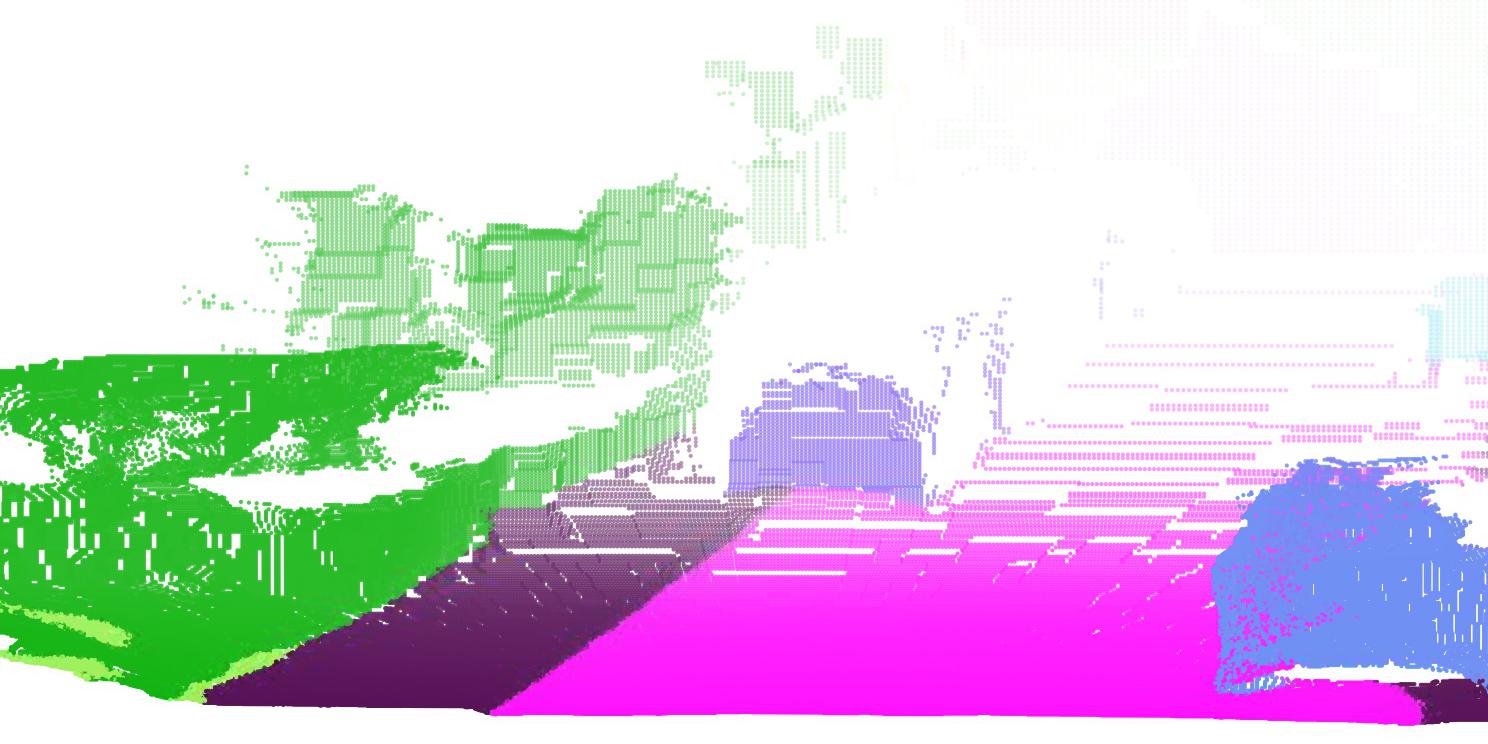}
\end{subfigure}\\
\begin{subfigure}{.245\linewidth}
  \centering
  \includegraphics[trim={150 0 150 100},clip,width=\linewidth]{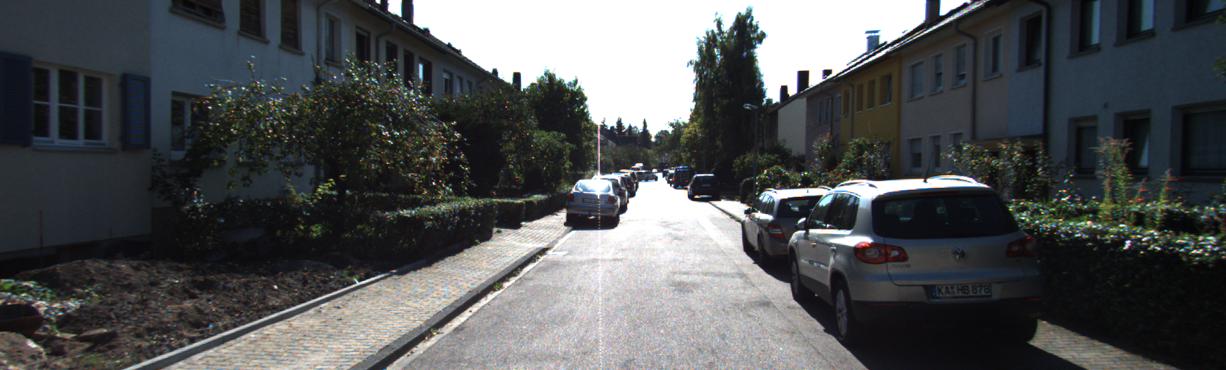}
\end{subfigure}\hfill
\begin{subfigure}{.245\linewidth}
  \centering
  \includegraphics[trim={150 0 150 100},clip,width=\linewidth]{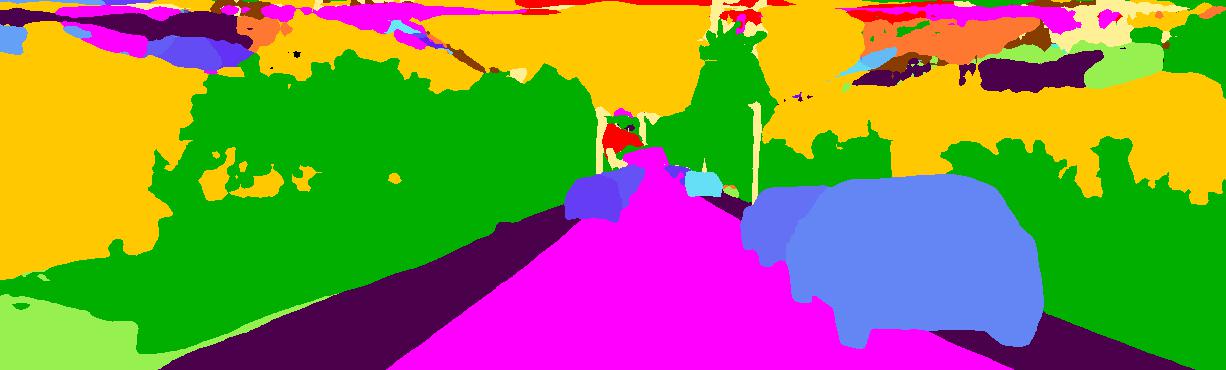}
\end{subfigure}\hfill
\begin{subfigure}{.245\linewidth}
  \centering
  \includegraphics[trim={150 0 150 100},clip,width=\linewidth]{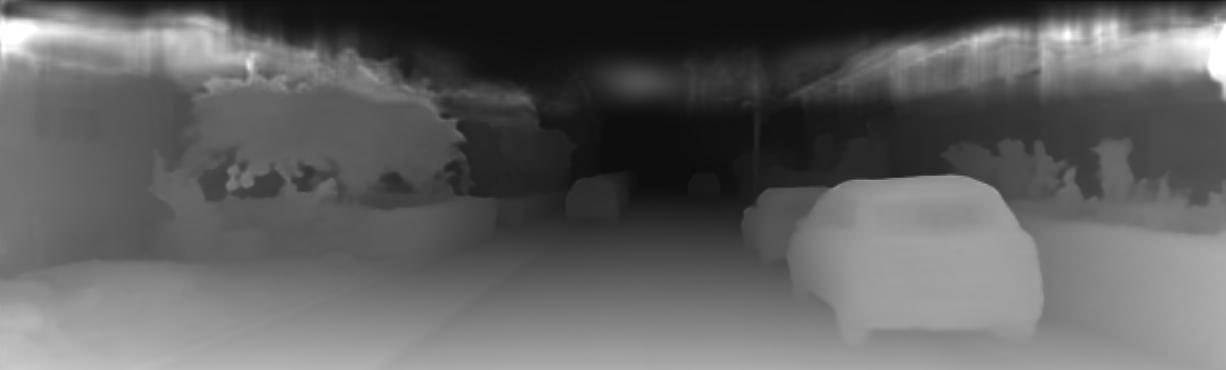}
\end{subfigure}\hfill
\begin{subfigure}{.245\linewidth}
  \centering
  \includegraphics[trim={0 0 0 300},clip,width=\linewidth]{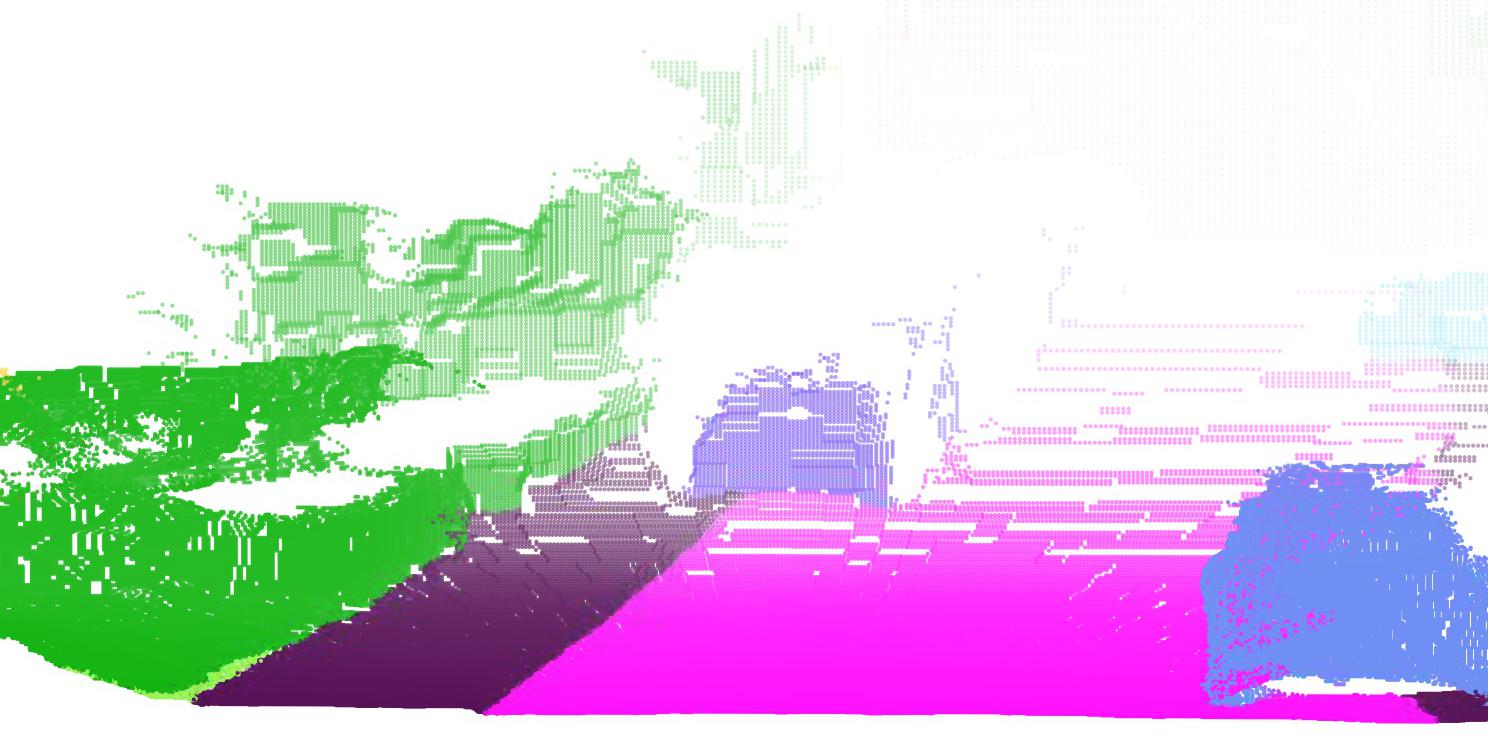}
\end{subfigure}\\
\begin{subfigure}{.245\linewidth}
  \centering
  \includegraphics[trim={150 0 150 100},clip,width=\linewidth]{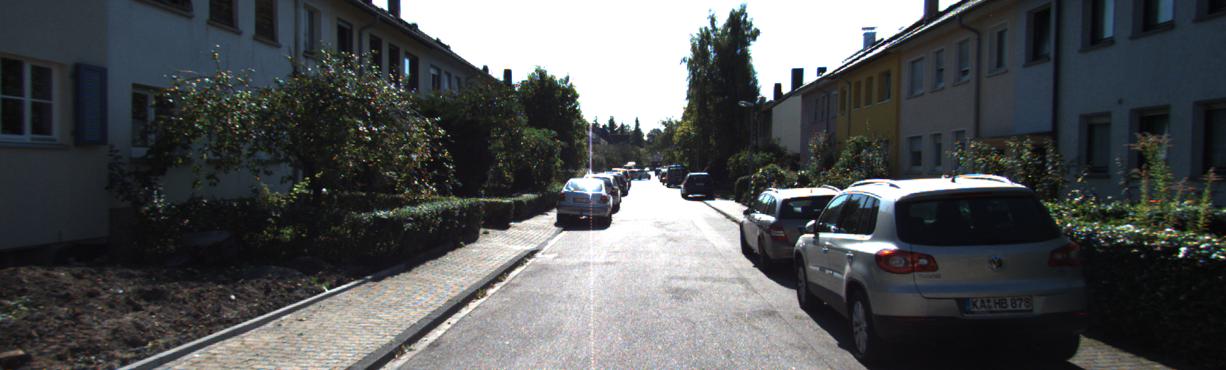}
\end{subfigure}\hfill
\begin{subfigure}{.245\linewidth}
  \centering
  \includegraphics[trim={150 0 150 100},clip,width=\linewidth]{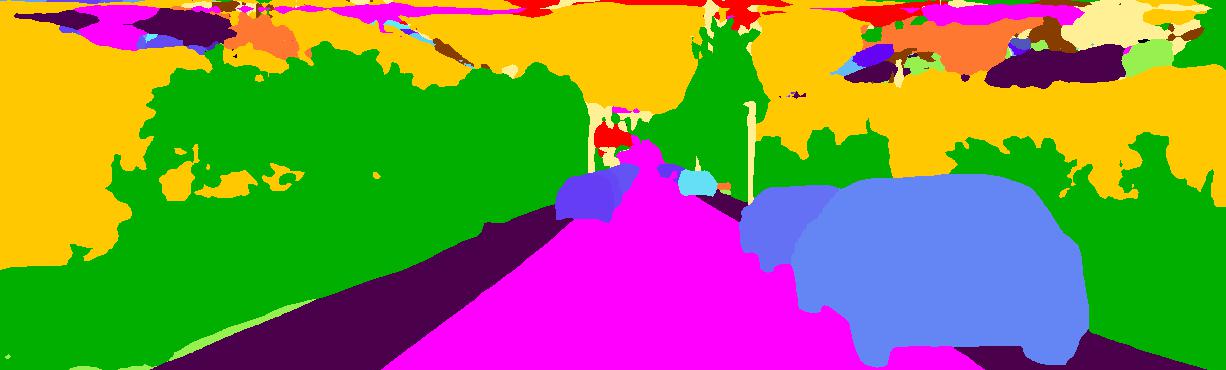}
\end{subfigure}\hfill
\begin{subfigure}{.245\linewidth}
  \centering
  \includegraphics[trim={150 0 150 100},clip,width=\linewidth]{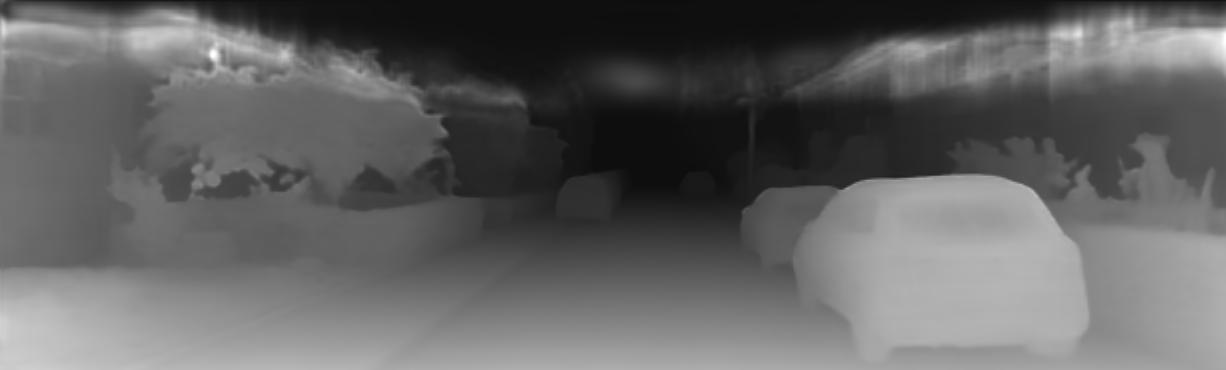}
\end{subfigure}\hfill
\begin{subfigure}{.245\linewidth}
  \centering
  \includegraphics[trim={0 0 0 300},clip,width=\linewidth]{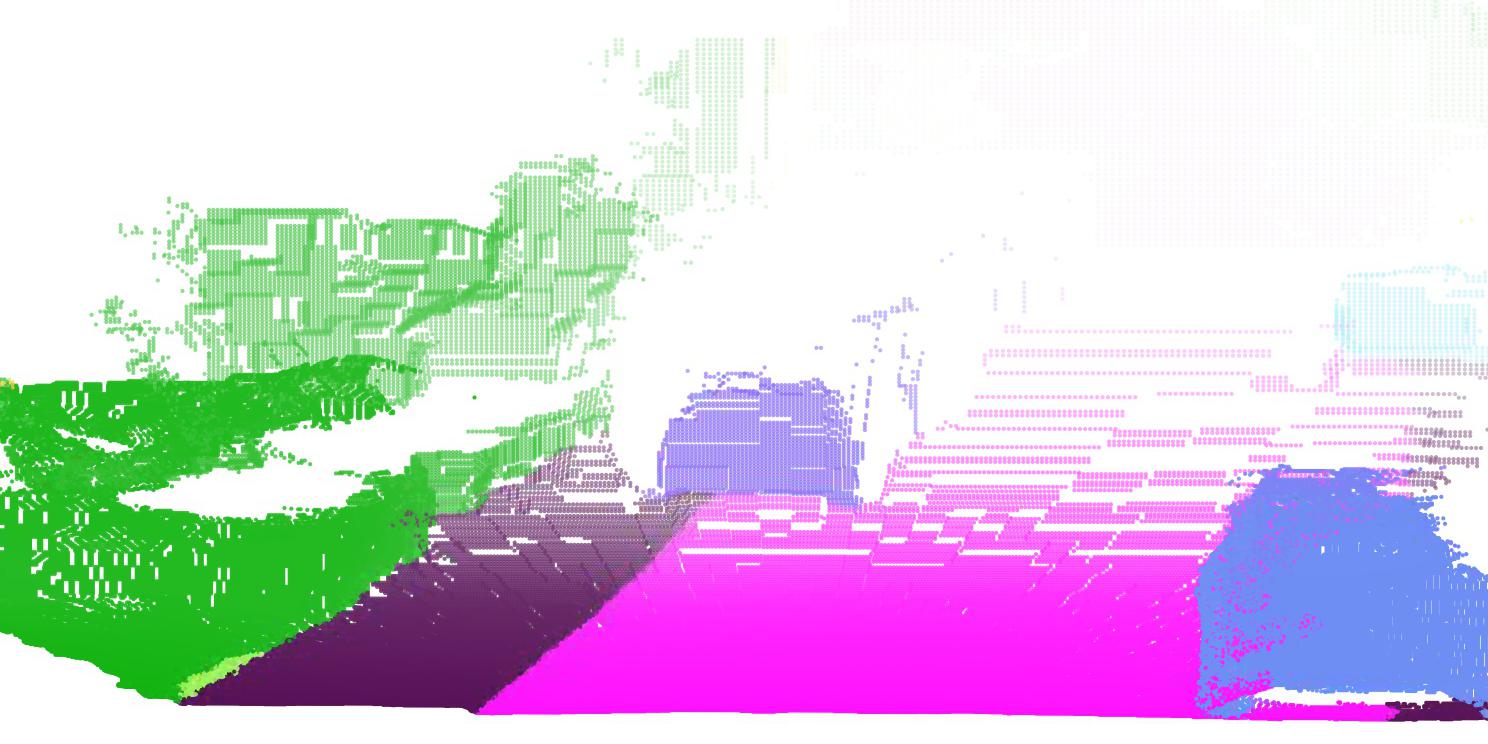}
\end{subfigure}\\
\begin{subfigure}{.245\linewidth}
  \centering
  \includegraphics[trim={150 0 150 100},clip,width=\linewidth]{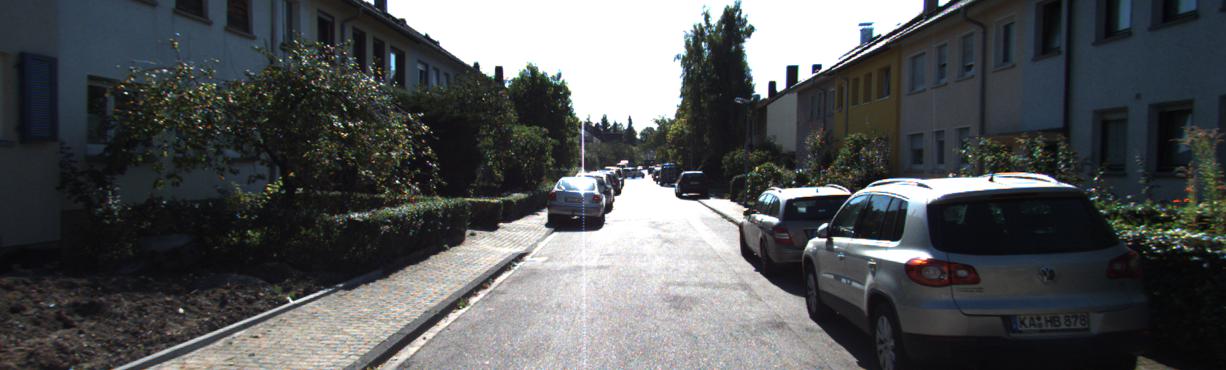}
\end{subfigure}\hfill
\begin{subfigure}{.245\linewidth}
  \centering
  \includegraphics[trim={150 0 150 100},clip,width=\linewidth]{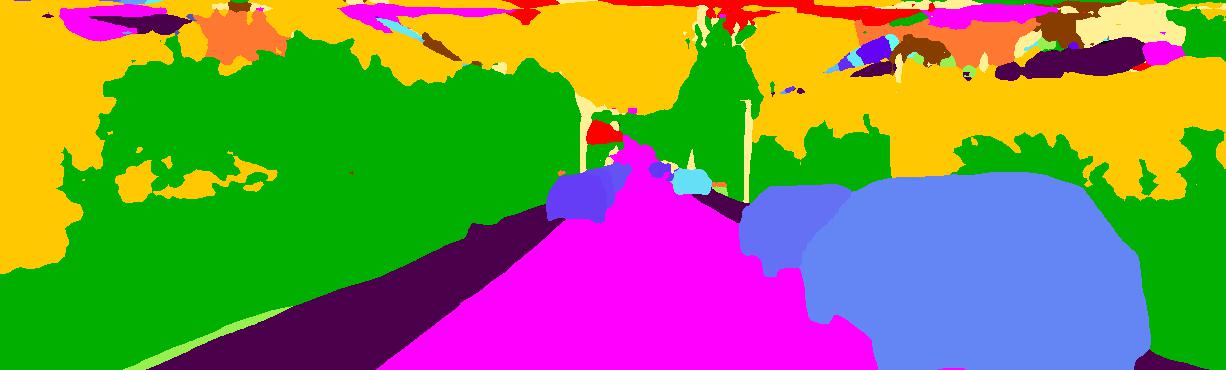}
\end{subfigure}\hfill
\begin{subfigure}{.245\linewidth}
  \centering
  \includegraphics[trim={150 0 150 100},clip,width=\linewidth]{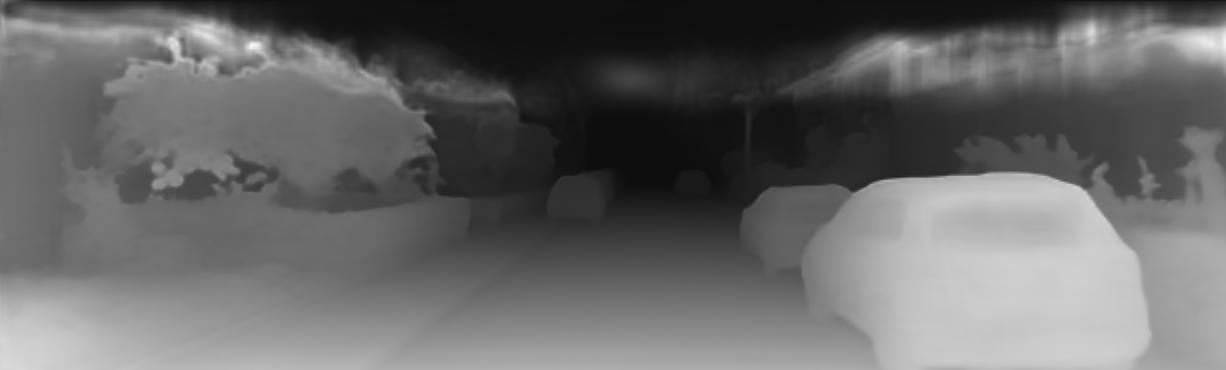}
\end{subfigure}\hfill
\begin{subfigure}{.245\linewidth}
  \centering
  \includegraphics[trim={0 0 0 300},clip,width=\linewidth]{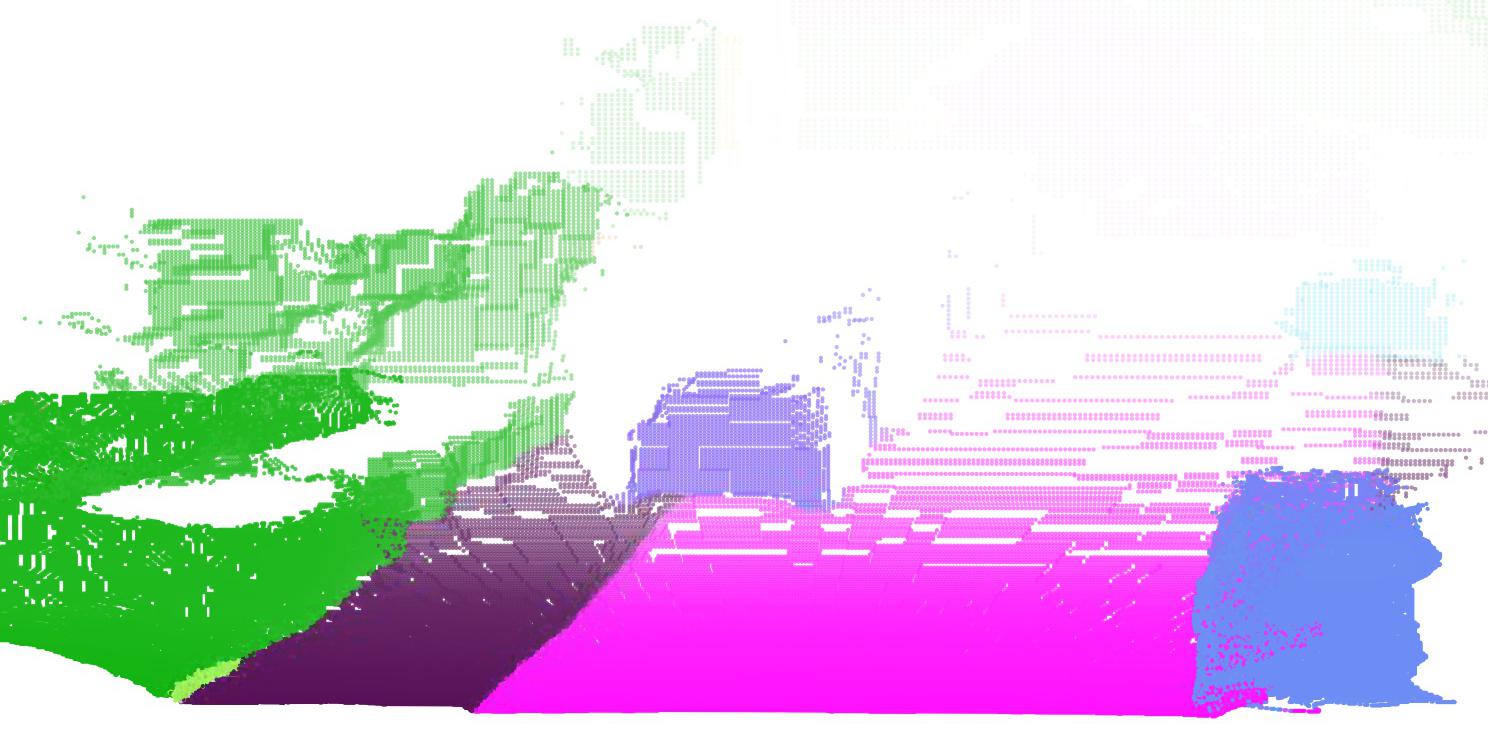}
\end{subfigure}\\
\begin{subfigure}{.245\linewidth}
  \centering
  \includegraphics[trim={150 0 150 100},clip,width=\linewidth]{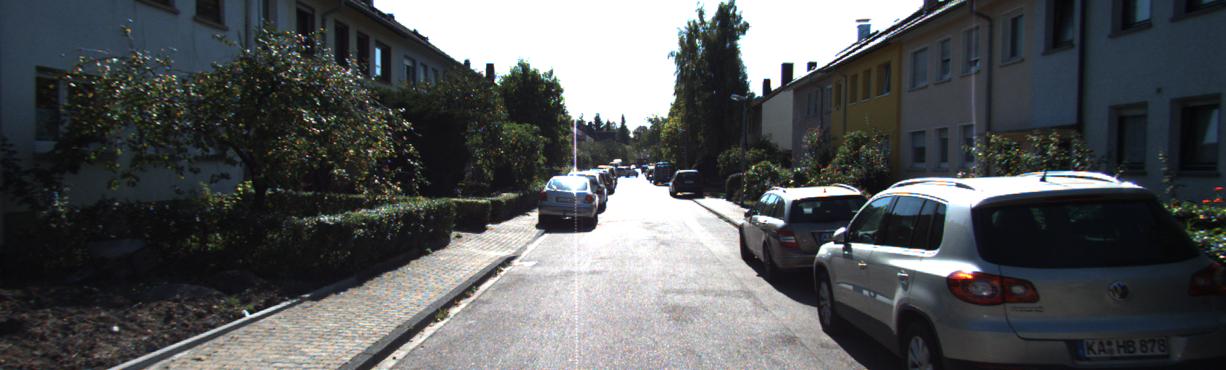}
\end{subfigure}\hfill
\begin{subfigure}{.245\linewidth}
  \centering
  \includegraphics[trim={150 0 150 100},clip,width=\linewidth]{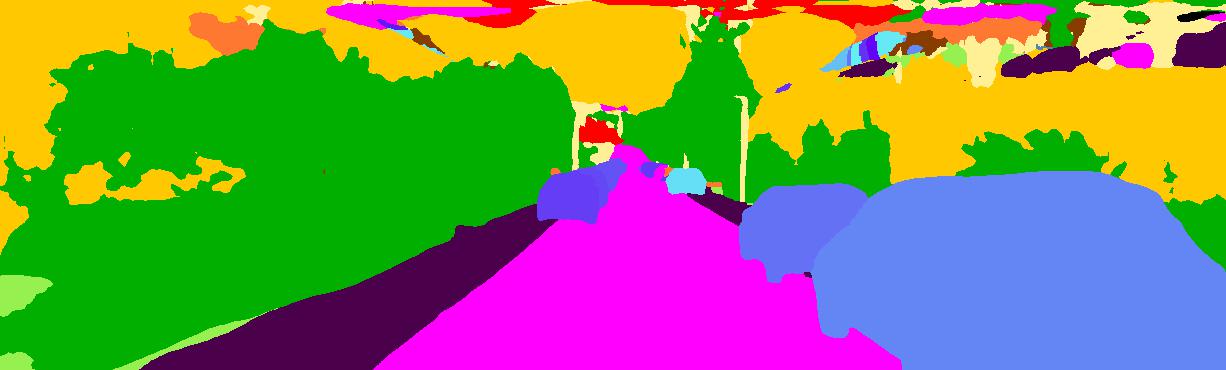}
\end{subfigure}\hfill
\begin{subfigure}{.245\linewidth}
  \centering
  \includegraphics[trim={150 0 150 100},clip,width=\linewidth]{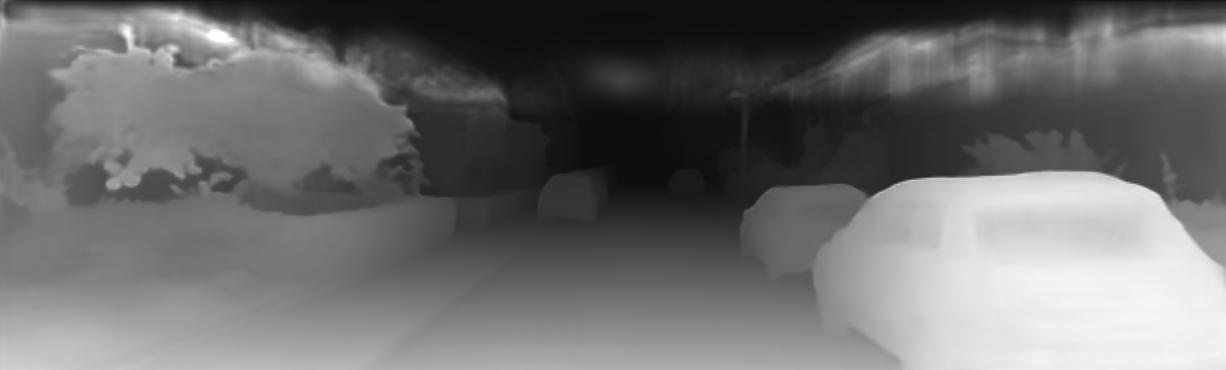}
\end{subfigure}\hfill
\begin{subfigure}{.245\linewidth}
  \centering
  \includegraphics[trim={0 0 0 300},clip,width=\linewidth]{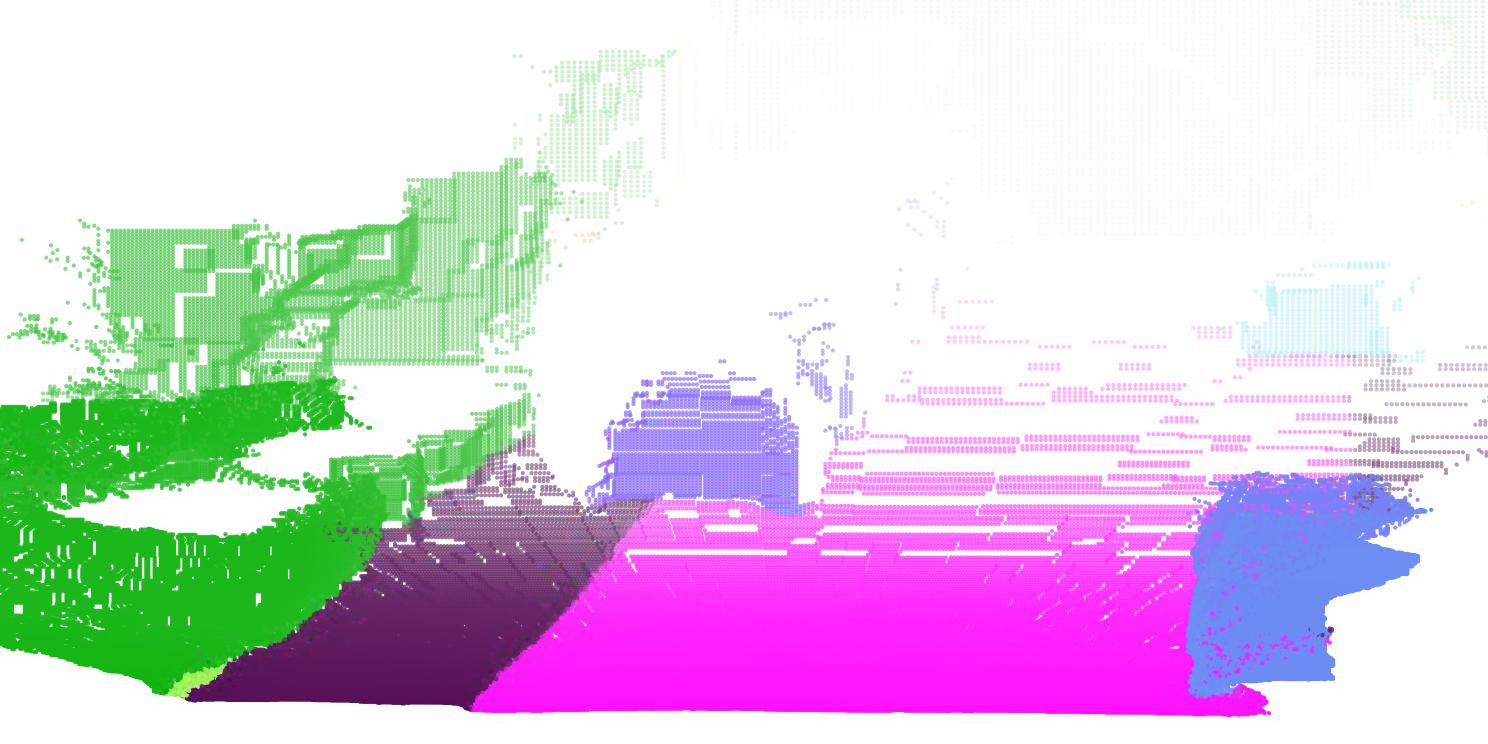}
\end{subfigure}\\
\begin{subfigure}{.245\linewidth}
  \centering
  \includegraphics[trim={150 0 150 100},clip,width=\linewidth]{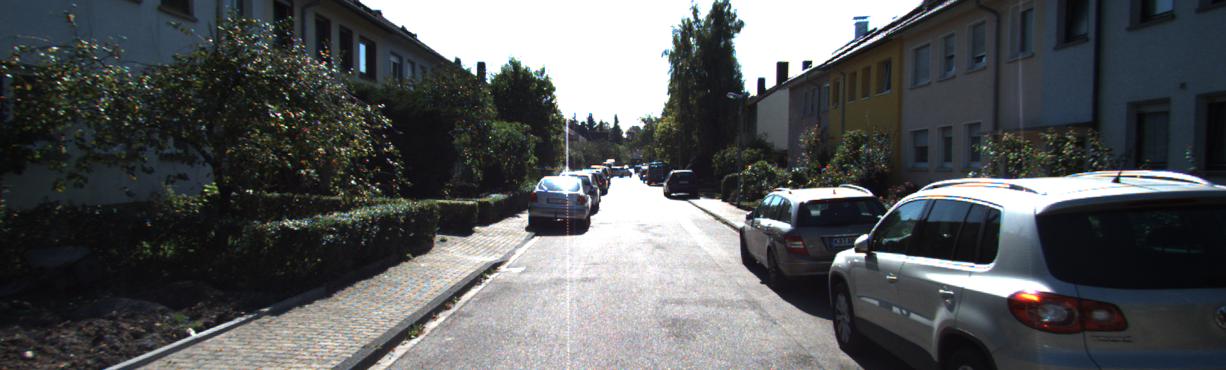}
\end{subfigure}\hfill
\begin{subfigure}{.245\linewidth}
  \centering
  \includegraphics[trim={150 0 150 100},clip,width=\linewidth]{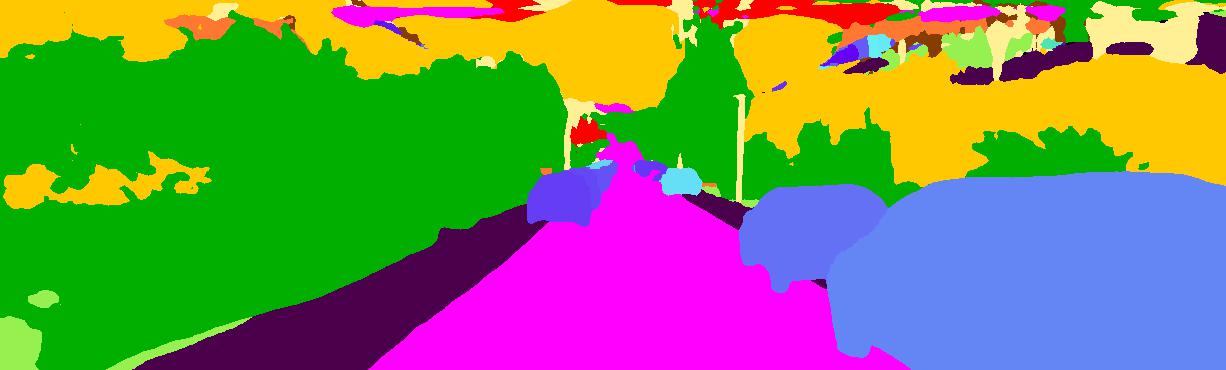}
\end{subfigure}\hfill
\begin{subfigure}{.245\linewidth}
  \centering
  \includegraphics[trim={150 0 150 100},clip,width=\linewidth]{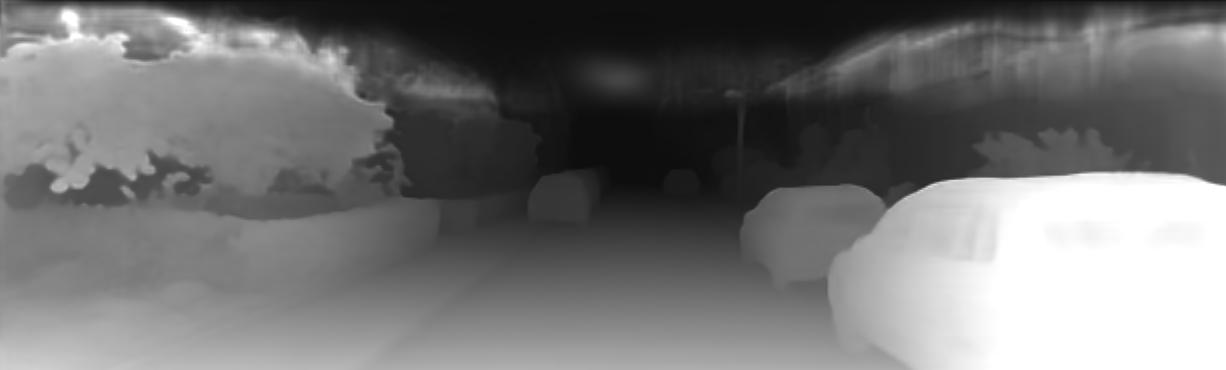}
\end{subfigure}\hfill
\begin{subfigure}{.245\linewidth}
  \centering
  \includegraphics[trim={0 0 0 300},clip,width=\linewidth]{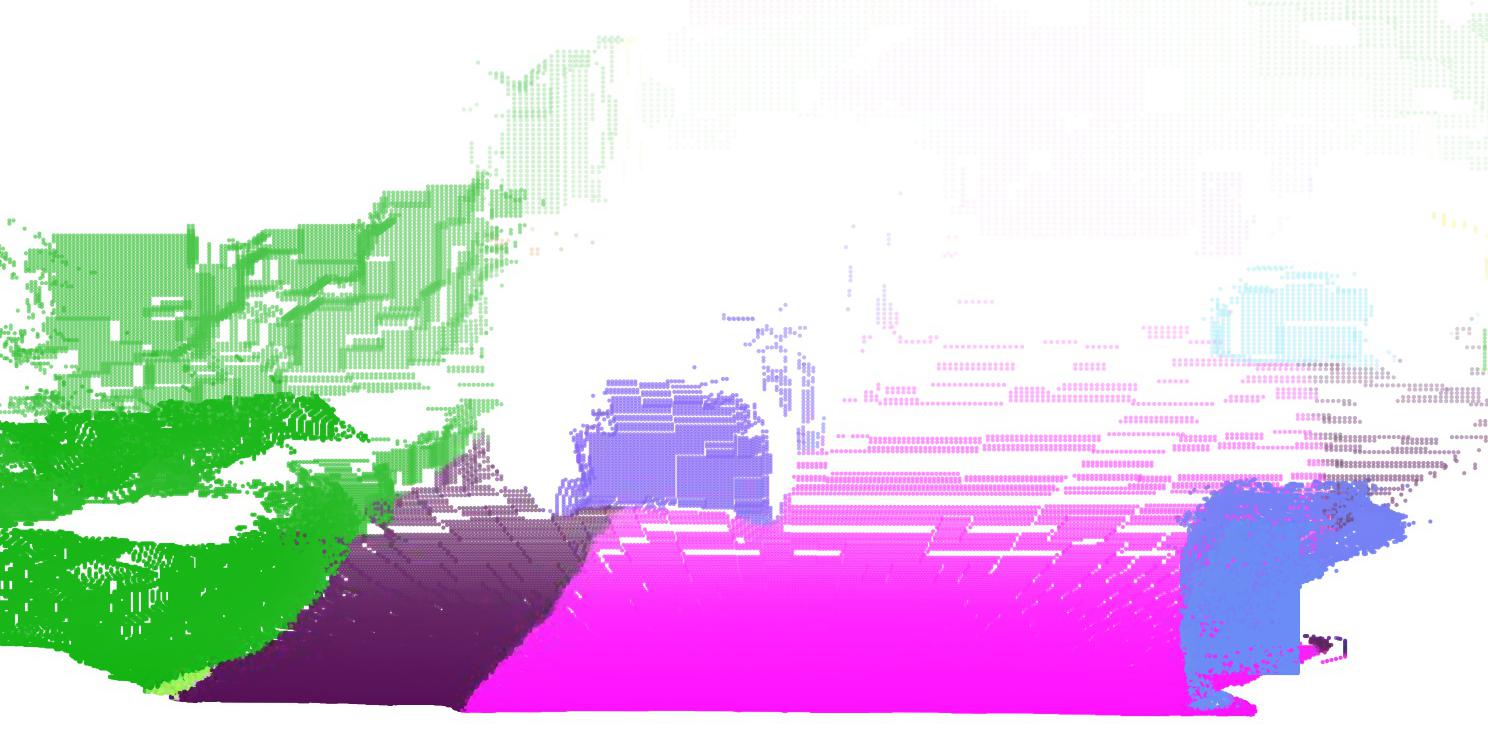}
\end{subfigure}\\
\begin{subfigure}{.245\linewidth}
  \centering
  \includegraphics[trim={150 0 150 100},clip,width=\linewidth]{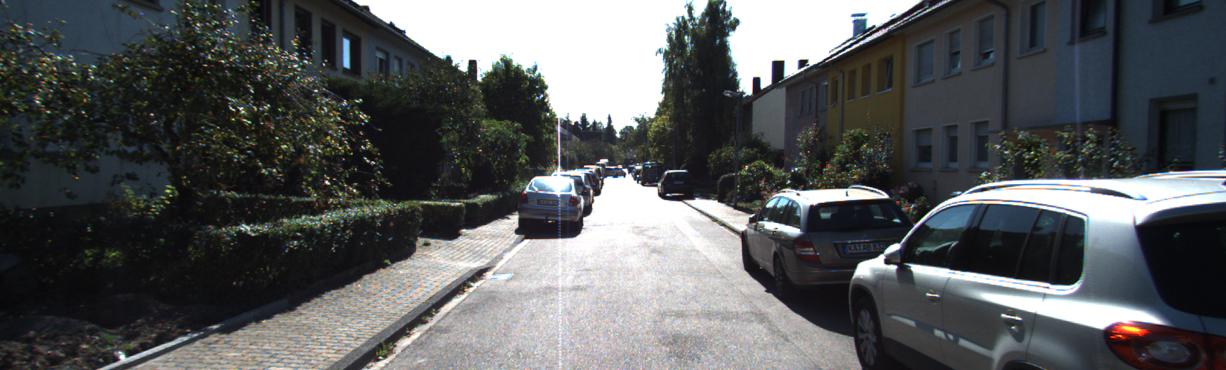}
\end{subfigure}\hfill
\begin{subfigure}{.245\linewidth}
  \centering
  \includegraphics[trim={150 0 150 100},clip,width=\linewidth]{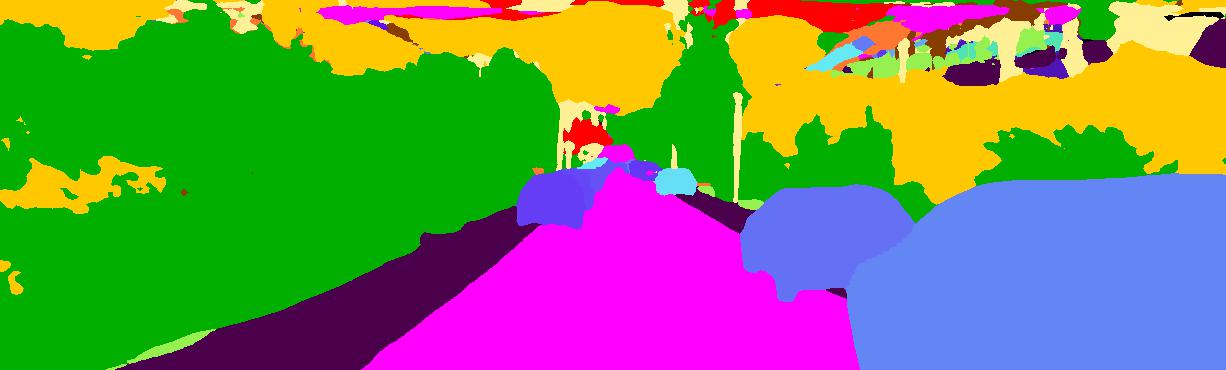}
\end{subfigure}\hfill
\begin{subfigure}{.245\linewidth}
  \centering
  \includegraphics[trim={150 0 150 100},clip,width=\linewidth]{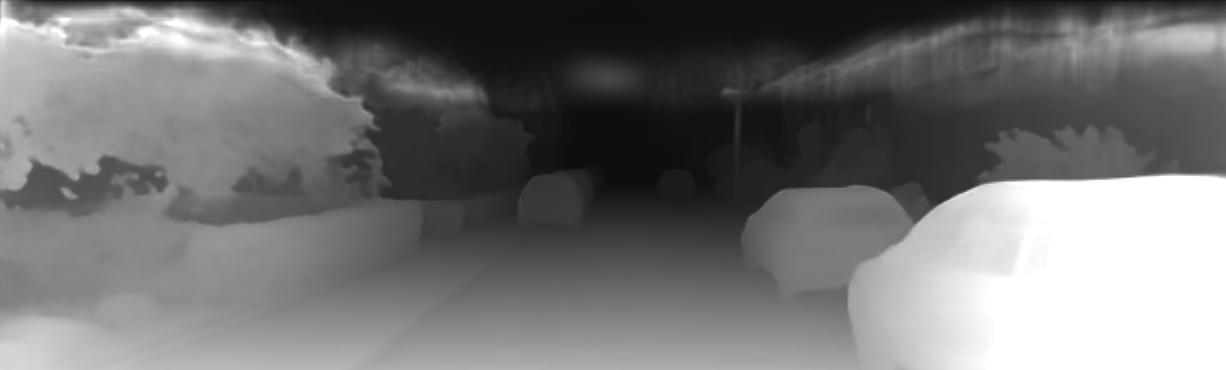}
\end{subfigure}\hfill
\begin{subfigure}{.245\linewidth}
  \centering
  \includegraphics[trim={0 0 0 300},clip,width=\linewidth]{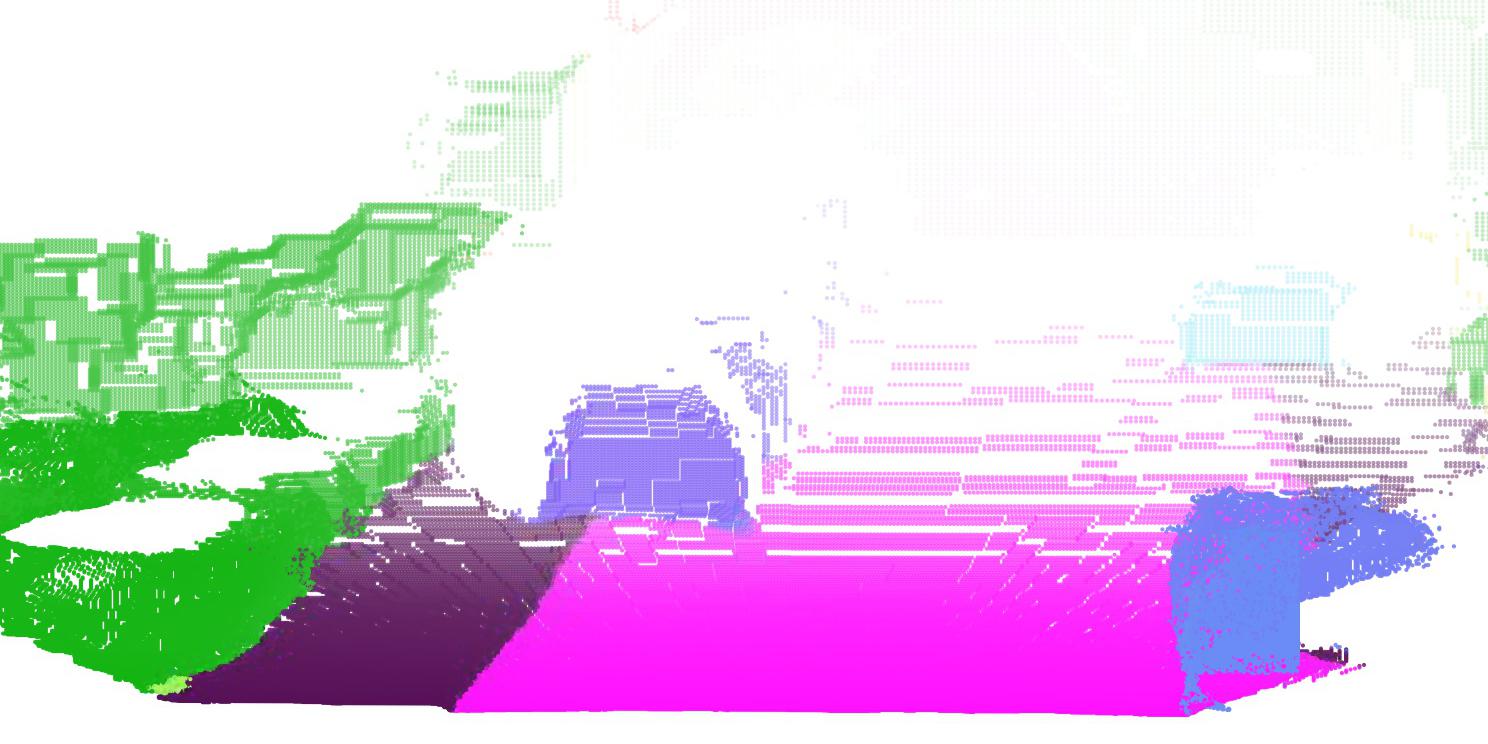}
\end{subfigure}\\
\begin{subfigure}{.245\linewidth}
  \centering
  \includegraphics[trim={150 0 150 100},clip,width=\linewidth]{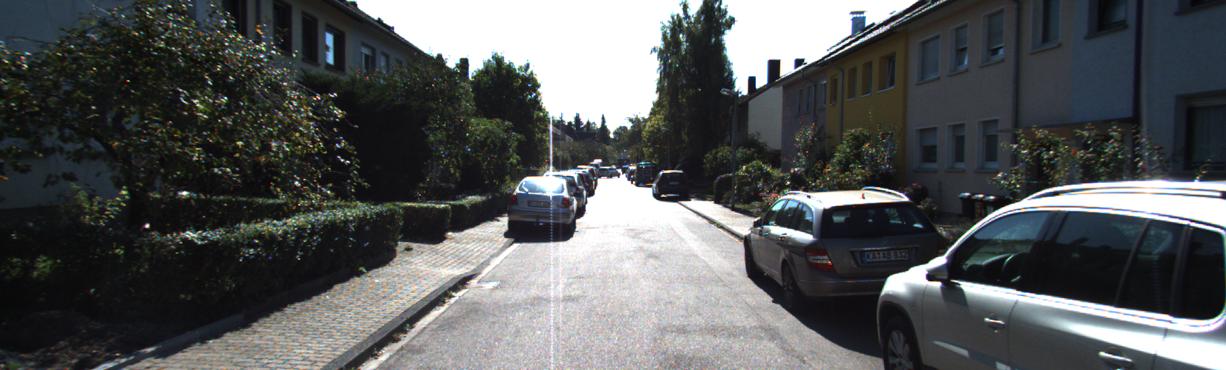}
\end{subfigure}\hfill
\begin{subfigure}{.245\linewidth}
  \centering
  \includegraphics[trim={150 0 150 100},clip,width=\linewidth]{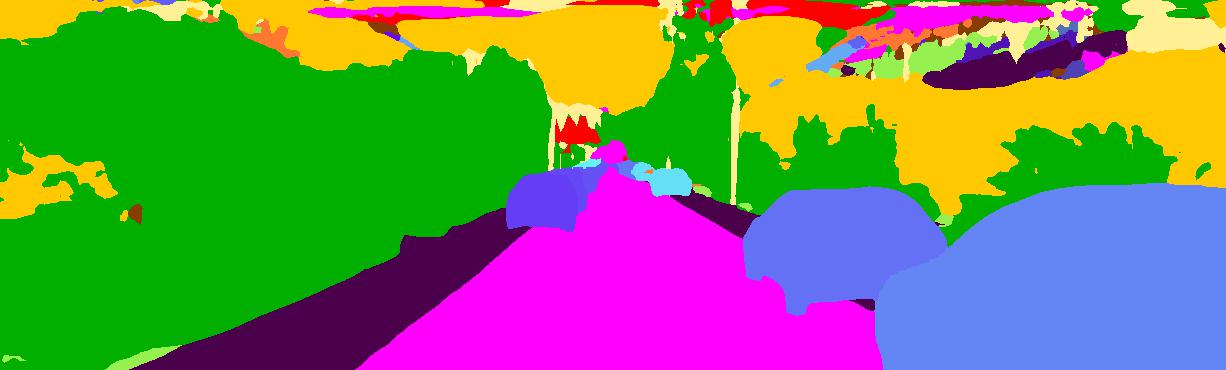}
\end{subfigure}\hfill
\begin{subfigure}{.245\linewidth}
  \centering
  \includegraphics[trim={150 0 150 100},clip,width=\linewidth]{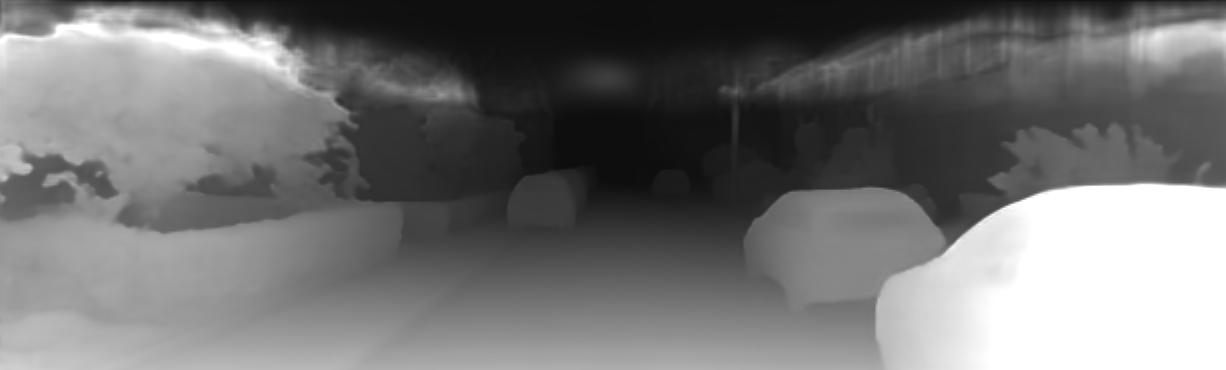}
\end{subfigure}\hfill
\begin{subfigure}{.245\linewidth}
  \centering
  \includegraphics[trim={0 0 0 300},clip,width=\linewidth]{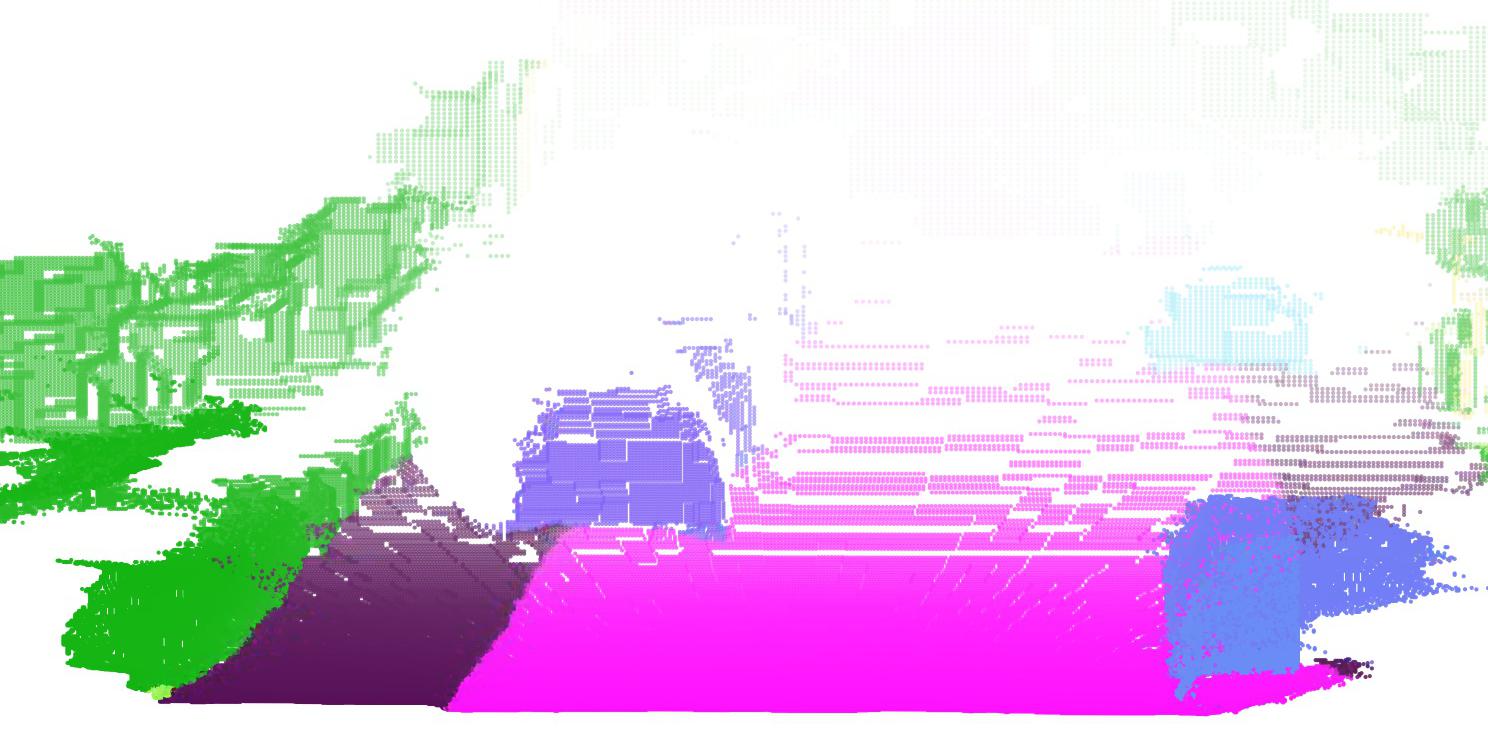}
\end{subfigure}\\
\begin{subfigure}{.245\linewidth}
  \centering
  \includegraphics[trim={150 0 150 100},clip,width=\linewidth]{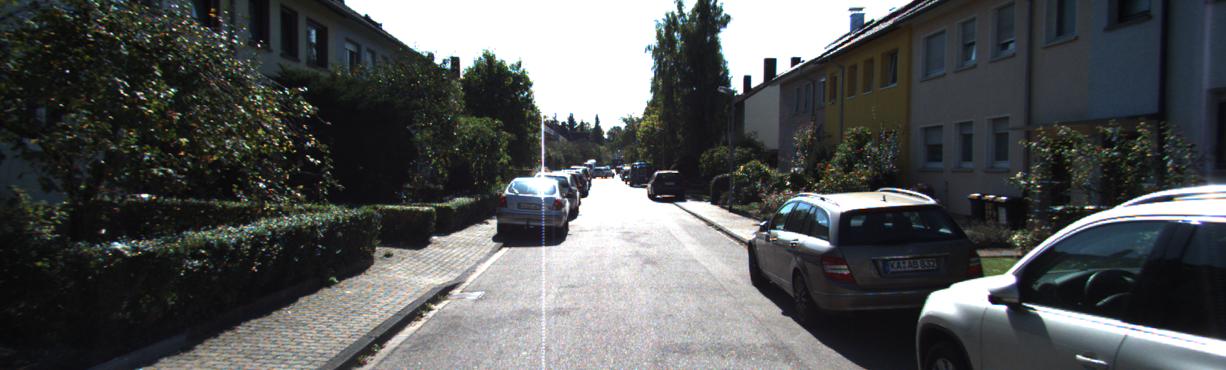}
\end{subfigure}\hfill
\begin{subfigure}{.245\linewidth}
  \centering
  \includegraphics[trim={150 0 150 100},clip,width=\linewidth]{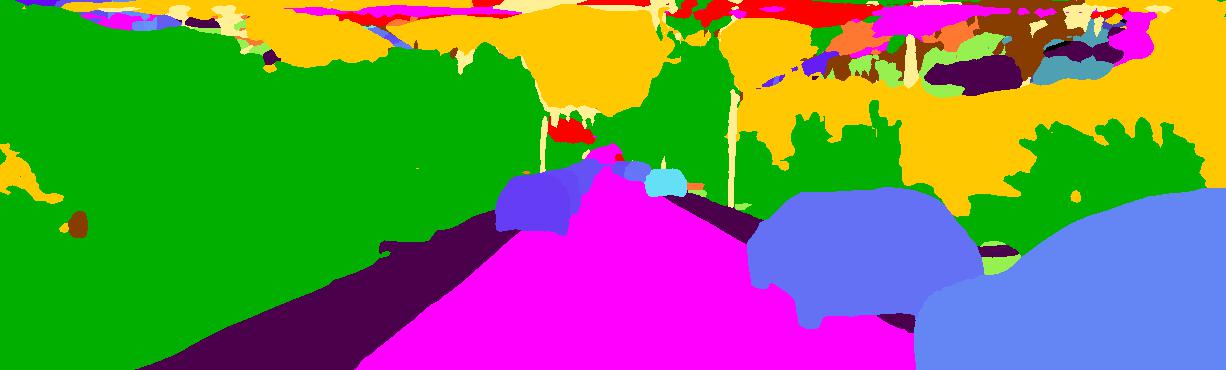}
\end{subfigure}\hfill
\begin{subfigure}{.245\linewidth}
  \centering
  \includegraphics[trim={150 0 150 100},clip,width=\linewidth]{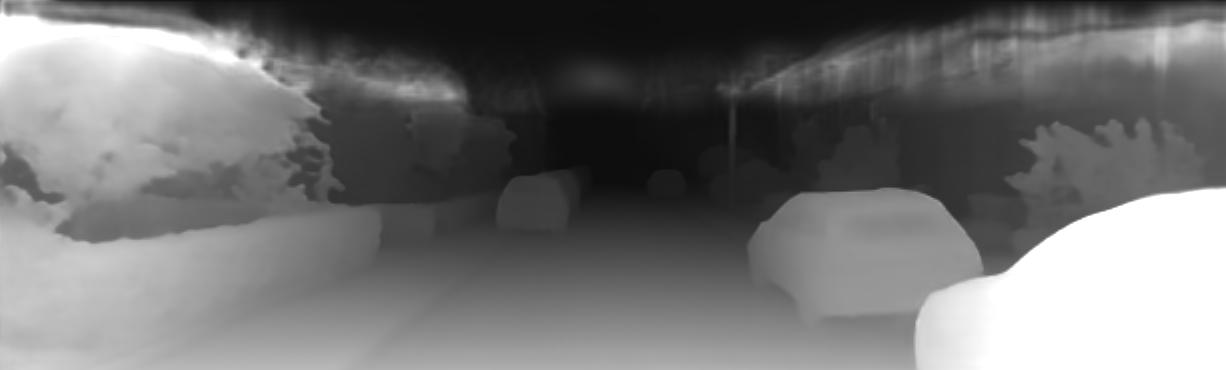}
\end{subfigure}\hfill
\begin{subfigure}{.245\linewidth}
  \centering
  \includegraphics[trim={0 0 0 300},clip,width=\linewidth]{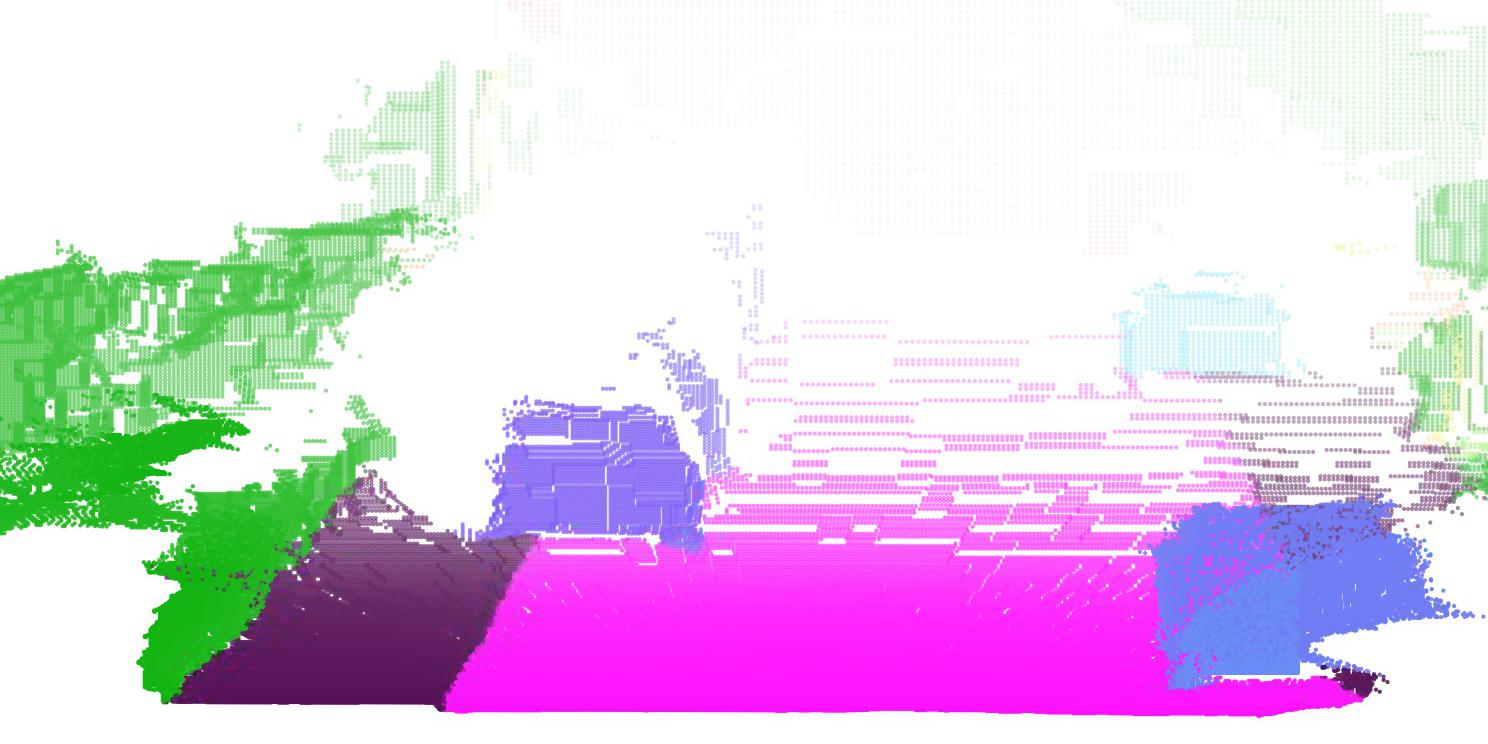}
\end{subfigure}\\
\begin{subfigure}{.245\linewidth}
  \centering
  \includegraphics[trim={150 0 150 100},clip,width=\linewidth]{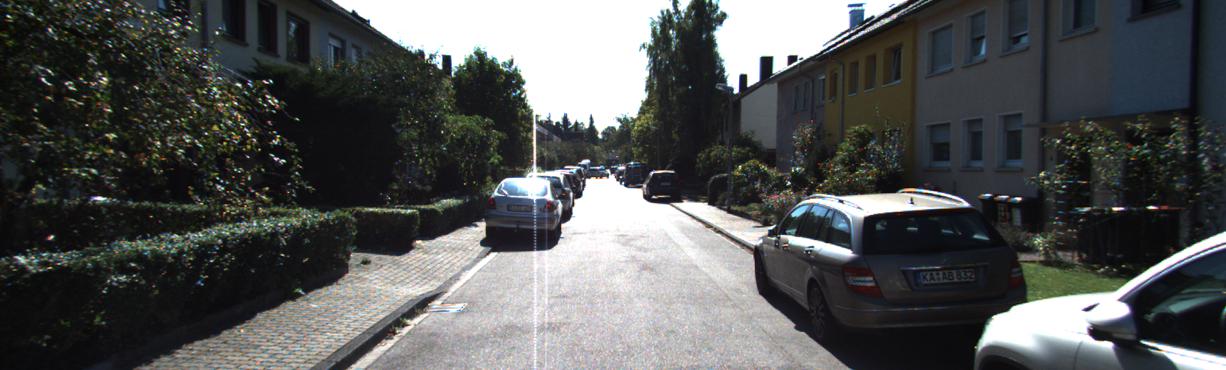}
\end{subfigure}\hfill
\begin{subfigure}{.245\linewidth}
  \centering
  \includegraphics[trim={150 0 150 100},clip,width=\linewidth]{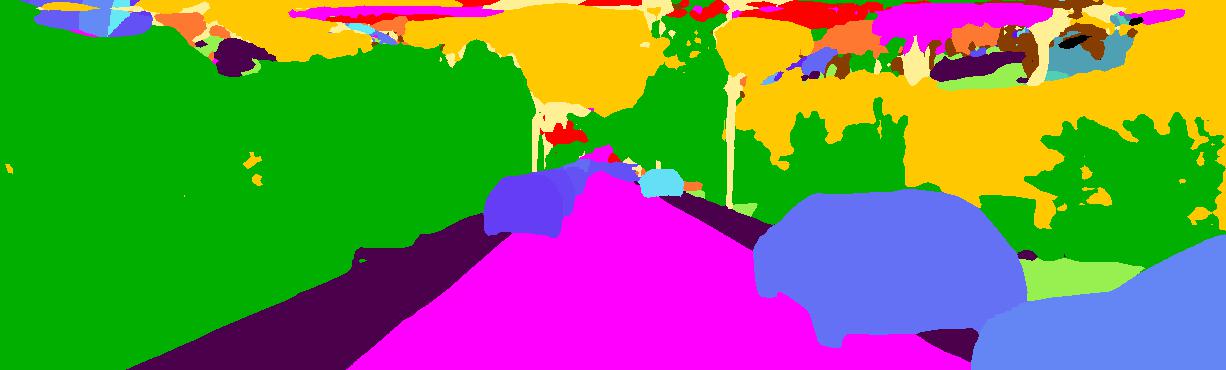}
\end{subfigure}\hfill
\begin{subfigure}{.245\linewidth}
  \centering
  \includegraphics[trim={150 0 150 100},clip,width=\linewidth]{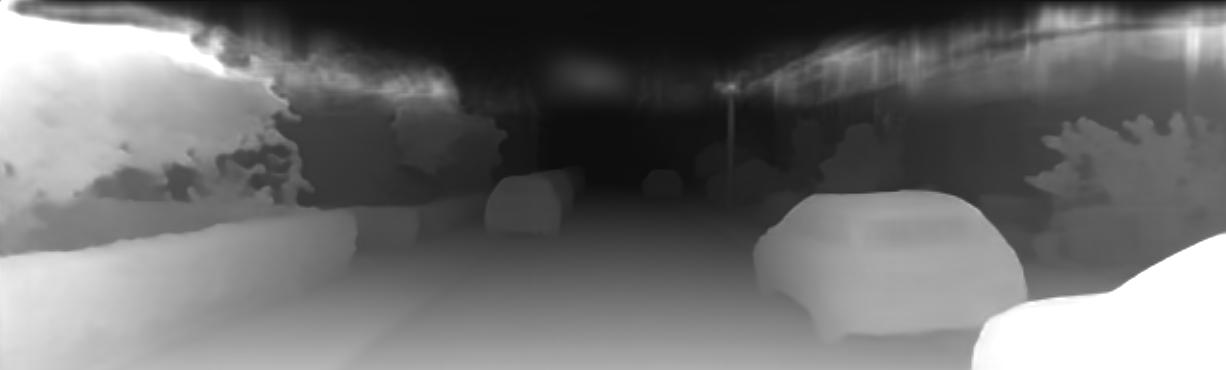}
\end{subfigure}\hfill
\begin{subfigure}{.245\linewidth}
  \centering
  \includegraphics[trim={0 0 0 300},clip,width=\linewidth]{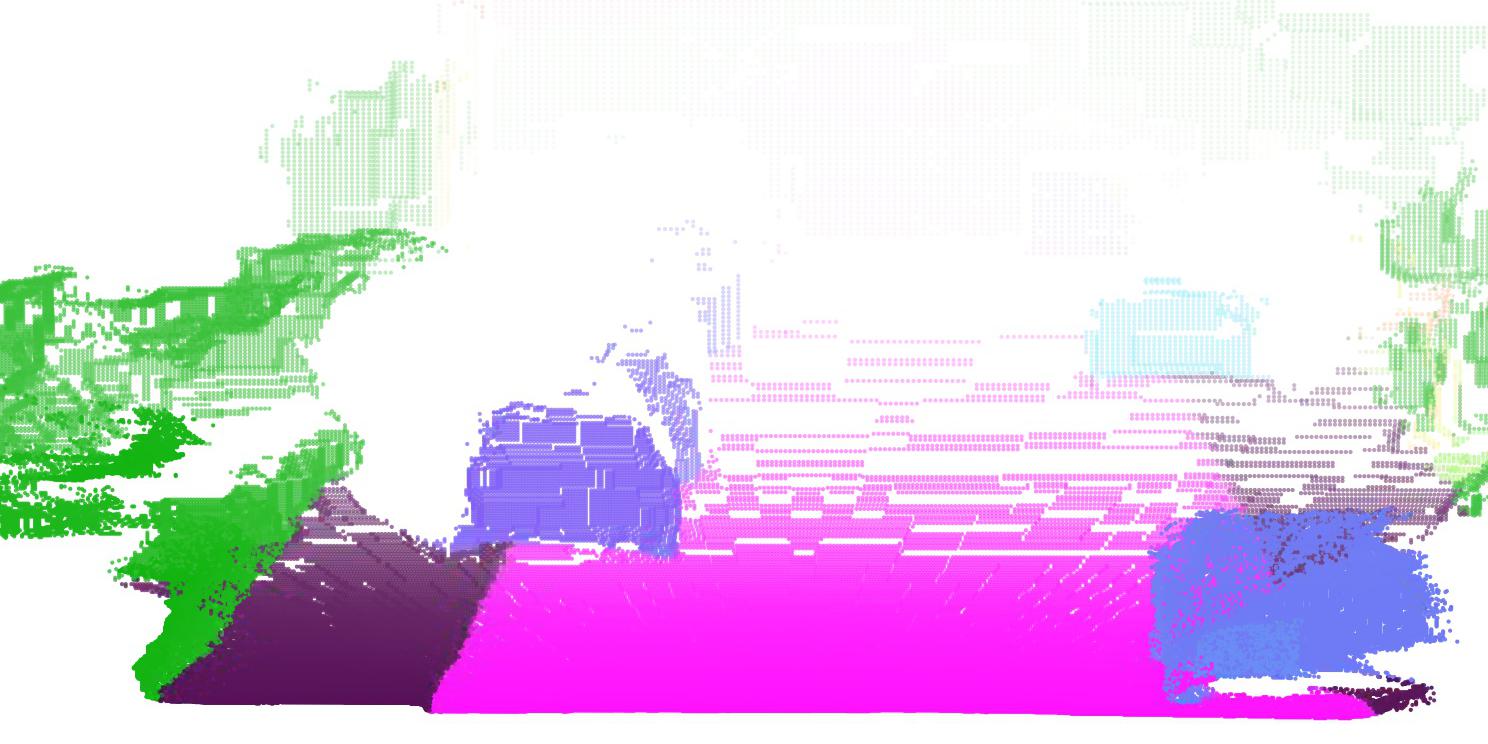}
\end{subfigure}\\

\caption{Prediction visualizations on SemKITTI-DVPS. From left to right: input image, temporally consistent panoptic segmentation prediction, monocular depth prediction, and point cloud visualization.}
\label{fig:sk_2}
\end{figure*}

We show more prediction visualizations in \figref{fig:cs_1}, \figref{fig:cs_2}, \figref{fig:sk_1}, and \figref{fig:sk_2}.
We choose four sequences from 50 validation sequences of Cityscapes-DVPS, and the results are shown in \figref{fig:cs_1} and \figref{fig:cs_2}.
As each sequence contains only 6 frames, the figures show all the frames of the four sequences.
Here, the video panoptic predictions demonstrate the results after the stitching algorithm, so each instance has the same instance ID in all the frames.
SemKITTI-DVPS results are shown in \figref{fig:sk_1} and \figref{fig:sk_2}.
We present the results on two 16-frame video clips from the validation sequence.
From the visualizations we can see that ViP-DeepLab is capable of outputting accurate video panoptic predictions and high quality depth predictions.

\end{document}